\documentclass[runningheads]{llncs}
\usepackage{graphicx}
\usepackage{comment}
\usepackage{amsmath,amssymb} % define this before the line numbering.
\usepackage{color}

\usepackage{float}

\usepackage[pagebackref=true,breaklinks=true,letterpaper=true,colorlinks,bookmarks=false]{hyperref}

\usepackage[shortlabels]{enumitem} 

\usepackage{graphicx}
\usepackage{booktabs}       % professional-quality tables
\usepackage{bm}       % blackboard math symbols
\usepackage[noend,ruled,linesnumbered]{algorithm2e}
\graphicspath{{figures/}}

\newcommand{\cD}{\mathcal{D}}

\newcommand{\cC}{\mathcal{C}}
\newcommand{\cG}{\mathcal{G}}
\newcommand{\cI}{\mathcal{I}}
\newcommand{\cV}{\mathcal{V}}
\newcommand{\cN}{\mathcal{N}}
\newcommand{\cE}{\mathcal{E}}
\newcommand{\cF}{\mathcal{F}}
\newcommand{\cP}{\mathcal{P}}
\newcommand{\cR}{\mathcal{R}}

\newcommand{\cS}{\mathcal{S}}
\newcommand{\cX}{\mathcal{X}}

\newcommand{\cL}{\mathcal{L}}
\newcommand{\indep}{\raisebox{0.05em}{\rotatebox[origin=c]{90}{$\models$}}} 
\DeclareMathOperator*{\argmax}{arg\,max}

\newcommand{\etal}{\textit{et al}.}
\newcommand{\ie}{\textit{i}.\textit{e}.}
\newcommand{\eg}{\textit{e}.\textit{g}.}

\def\causal{\hbox{$\circ$}\kern-1.5pt\hbox{$\rightarrow$}}
\def\reversecausal{\hbox{$\leftarrow$}\kern-1.5pt\hbox{$\circ$}}

\begin{document}
\pagestyle{headings}
\mainmatter
\def\ECCVSubNumber{4535}  % Insert your submission number here
%%%%%%%%% TITLE
\title{SG-VAE: Scene Grammar Variational Autoencoder to generate new indoor scenes}

\begin{comment}
\titlerunning{ECCV-20 submission ID \ECCVSubNumber} 
\authorrunning{ECCV-20 submission ID \ECCVSubNumber} 
\author{Anonymous ECCV submission}
\institute{Paper ID \ECCVSubNumber}
\end{comment}
%******************

%\begin{comment}
\titlerunning{Scene Grammar Variational Autoencoder}

\author{Pulak Purkait\inst{1} \orcidID{0000-0003-0684-1209} \and
Christopher Zach\inst{2} \orcidID{0000-0003-2840-6187} \and
Ian Reid\inst{1} \orcidID{0000-0001-7790-6423}}
\authorrunning{P. Purkait et al.}
\institute{Australian Institute of Machine Learning and School of Computer Science, \\ The University of Adelaide, Adelaide SA 5005, Australia 
%\email{\{pulak.purkait,ian.reid\}@adelaide.edu.au} 
\and Chalmers University of Technology, Goteborg 41296, Sweden 
}
%\end{comment}
%******************

\maketitle
%\thispagestyle{empty}
%%%%%%%%% ABSTRACT
\begin{abstract}
Deep generative models have been used in recent years to learn coherent latent representations in order to synthesize high-quality images. In this work, we propose a neural network to learn a generative model for sampling consistent indoor scene layouts. Our method learns the co-occurrences, and appearance parameters such as shape and pose, for different objects categories through a grammar-based auto-encoder, resulting in a compact and accurate representation for scene layouts. In contrast to existing grammar-based methods with a user-specified grammar, we construct the grammar automatically by extracting a set of production rules on reasoning about object co-occurrences in training data. The extracted grammar is able to represent a scene by an augmented parse tree. The proposed auto-encoder encodes these parse trees to a latent code, and decodes the latent code to a parse tree, thereby ensuring the generated scene is always valid. We experimentally demonstrate that the proposed auto-encoder learns not only to generate valid scenes (i.e. the arrangements and appearances of objects), but it also learns coherent latent representations where nearby latent samples decode to similar scene outputs.
The obtained generative model is applicable to several computer vision tasks such as 3D pose and layout estimation from RGB-D data. 
 %The prior knowledge of object appearances encoded in the latent space can be readily adapted for other computer vision tasks.  
\end{abstract}

%%%%%%%%% BODY TEXT
\section{Introduction} 
Recently proposed approaches for deep generative models have seen great success in producing high quality RGB images~\cite{goodfellow2014generative,li2019grains,liu2016coupled,ritchie2019fast} and continuous latent representations from images~\cite{kingma2013auto}. Our work aims to learn coherent latent representations for generating natural indoor scenes comprising different object categories and their respective appearances (\ie~pose and shape). %%\iancomment{appearances?}. 
Such a learned representation has direct use for various computer vision and scene understanding tasks, including ({\bf i}) 3D scene-layout estimation~\cite{song2016deep}, ({\bf ii}) 3D visual grounding~\cite{deng2018visual,xiao2017weakly}, ({\bf iii})  Visual Question Answering~\cite{anderson2018bottom,lu2016hierarchical}, and ({\bf iv}) robot navigation~\cite{meyer2003map}.  % However, these models are limited producing meaningful scene in a way that one has to employ  another set of computer vision tools to understand the generated scene~\cite{chen20153d,song2016deep}.  

\begin{figure}[H]
\centering
\includegraphics[width=1\textwidth]{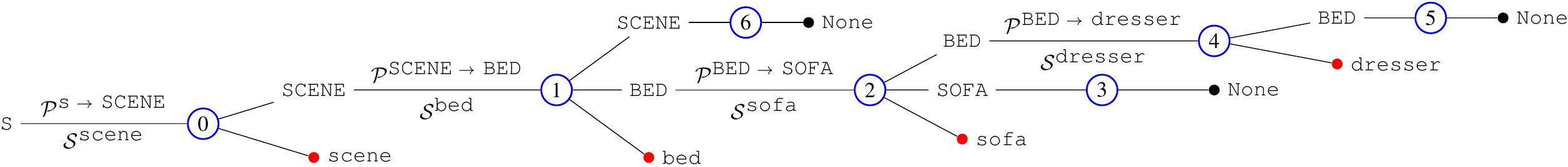}
\caption{An example of parse tree obtained by applying the CFG to a scene comprising \texttt{bed, sofa, dresser}. The sequence of production rules {\color{blue}\textcircled{\raisebox{-0.5pt}{\scriptsize 0}}-\color{blue}\textcircled{\raisebox{-0.5pt}{\scriptsize 6}}} are marked in order. The  attributes of production rules are displayed above and below of the rules. %Note that attributes corresponding to \texttt{None} are zero and not shown above.
} \label{fig:parsetree} %\vspace{-1em}
\end{figure} 
%\end{rcenter} 
%The current work incorporates the co-occurrences and appearances of different objects, \ie 3D bounding boxes of the scene, while generating an indoor scene. 
Developing generative models for such discrete domains has been explored in a limited number of works~\cite{huang2018holistic,qi2018human,yu2011make}. These works utilize prior knowledge of indoor scenes by manually defining attributed grammars. 
However, the number of rules in such grammars can be prohibitively large for real indoor environments, and consequently, these methods are evaluated only on synthetic data with a small number of objects. 
%Further, as the inference method is based on MCMC sampling, due to the large number of parameters the inference time is prohibitively slow.
Further, the Monte Carlo based inference method can be intractably slow: up to 40~minutes~\cite{qi2018human} or one hour~\cite{huang2018holistic} to estimate a single layout.
%e.g., the method propsed in~\cite{qi2018human} takes up to 40~minutes and the one described in~\cite{huang2018holistic} requires approximately one hour to estimate a single layout. 
%We proceed along a fast alternative of MCMC based sampling techniques, \ie deep generative models in discrete domain~\cite{gomez2018automatic}.
Deep generative models for discrete domains have been proposed in~\cite{gomez2018automatic} (employing sequential representations) and in~\cite{kusner2017grammar} (based on formal grammars). 
%That work was followed by a grammar-based approach  in that has inspired our work in representing scene layouts.
In our work, we extend \cite{kusner2017grammar} by integrating object attributes, such as pose and shape of objects in a scene.  %The sequential representation proposed in~\cite{gomez2018automatic} is later replaced by a grammar-based representation in \cite{kusner2017grammar}.
%Our work is inspired by the later.
Further, the underlying grammar is often defined manually~\cite{kusner2017grammar,qi2018human},  %\iancomment{"often" is weak here.  is it manual in [12]?}, 
but we propose to extract suitable grammar rules from training data automatically.
%Note that one could define the grammar manually, but it would be time consuming and may not be accurate.   In section~\ref{sec:structurediscovery} and \ref{sec:grammar} we propose a data-driven method to discover the grammar.
The main components of our approach are thus: 
\begin{itemize}[itemsep=0.5ex,leftmargin=4.5mm,topsep=0.5ex,partopsep=0.5ex]
\item a scene grammar variational autoencoder (SG-VAE) that captures the appearances (\ie~pose and shape) of objects in the same 3D spatial configurations in a compact latent code (Sec~\ref{sec:ourmethod}); % the quality of the smooth latent code is illustrated by interpolating different scenes in Figure~\ref{fig:token}; 
\item a context free grammar that explains causal relationships among objects which frequently co-occur, automatically extracted from training data (Sec~\ref{sec:structurediscovery});
\item the practicality of the learned latent space is also demonstrated for a computer vision task. 
\end{itemize}
Our SG-VAE is fast and has the ability to represent the scene in a coherent latent space as shown in Sec \ref{sec:experiments}.  %\vspace{-0.5em}

% Being a rule-based VAE, the proposed method is interpretable compared to the traditional VAEs~\cite{doersch2016tutorial}\cite{kingma2013auto}\cite{blei2017variational}. 

\section{Deep generative model for scene generation}\label{sec:ourmethod}

The proposed method is influenced by the Grammar Variational Autoencoder~\cite{kusner2017grammar}, so we begin with a brief description of that prior art. 

The Grammar VAE takes a valid string (in their case a chemical formula) and begins by parsing it into a set of production rules. These rules are represented as 1-hot binary vectors and encoded compactly to a latent code by the VAE. Latent codes can then be sampled and decoded to production rules and corresponding valid strings. 
%which takes a valid string and parses it into a sequence of production rules generating the input string.
%As an instance of a VAE latent variables can be sampled according to the prior, which are subsequently decoded into production rules and corresponding valid strings.
%The sampled latent vector is then decoded to reconstruct the same sentence. 
%The benefit of the method is that as the decoder produces a sequence of production rules that always yields a valid sentence defined by the grammar. 
More specifically, each production rule is represented by a 1-hot vector of size $N$, where $N$ is the total number of rules, \ie~$N = |\cR|$ (where $\cR$ is the set of rules). The maximum size $T$ of the sequence is fixed in advance. Thus the scene is represented by a sequence $\cX\in \{0, 1\}^{N\times T}$ of 1-hot vectors (note that when fewer than $T$ rules are needed, a dummy/null rule is used to pad the sequence up to length $T$ ensuring that the input to the autoencoder is always the same size). $\cX$ is then encoded to a continuous (low)-dimensional latent posterior distribution
$\cN(\bm{\mu}(\cX), \bm{\Sigma}(\cX))$. 
%% CZ: this is the posterior and should read as $\cN(\bm{\mu}(x), \bm{\Sigma}(x))$, where x is the input. Pulak: x is a matrix here, shall we make it \cX?  
The decoding network, which is a recurrent network, maps latent vectors to a set of unnormalized log probability vectors (logits) corresponding to the production rules. 
%samples the latent
%$\bm{z}\sim \cN(\bm{0}, \mathtt{I})$ 
%CZ: this is the prior, e.g. $\bm{z}\sim \cN(\bm{0}, \bm{I})$! Pulak: I am still bit confused here, since we are describing the whole autoencoder pipeline, arn't we referring z as a sample from $\bm{z}\sim \cN(\bm{0}, \mathtt{I})$ 
%from the prior distribution 
%%\iancomment{what do you mean "from the prior"} : Pulak-- Its the prior latent distribution 
To convert from the output logits to a valid sequence of production rules, each logit vector is considered in turn. The max output in the logit vector gives a 1-hot encoding of a production rule, but only some sequences of rules are valid. To avoid generating a rule that is inconsistent with the rules that have preceded it, invalid rules are masked out of the logit and the max is taken over only unmasked elements.  This ensures that the Grammar VAE only ever generates valid outputs. 
Further details of the Grammar VAE can be found in~\cite{kusner2017grammar}. 
%\iancoment{the text here remains confusing/confused between the role of the grammar and the role of the VAE.  which are you describing here and why?}

Adapting this idea to the case of generating scenes requires that we incorporate not only valid co-occurrences of objects, but also valid attributes such as absolute pose (3D location and orientation) and shape (3D bounding boxes) of the objects in the scene.  More specifically, our proposed SG-VAE is adapted from the Grammar VAE in the following ways: 
\begin{itemize}[itemsep=0.5ex,leftmargin=4.5mm,topsep=0.5ex,partopsep=0.5ex]
\item The object attributes, \ie~absolute pose and shape of the objects are estimated while inferring the production rules. %The relative pose is computed with respect to the parent non-terminal. 
\item The SG-VAE is moreover designed to generate valid 3D scenes which adhere not only to the rules of grammar, but also generate valid poses. 
\end{itemize} 

\subsection{Scene-Grammar Variational Autoencoder} 
%The proposed SG-VAE generates similar object appearances and co-occurrences that we observe in the training data. 

We represent the objects in indoor scenes explicitly by a set of production rules, so that the entire  arrangement---\ie~the occurrences and appearances (\ie~pose and shape) of the objects in a scene---is guaranteed to be consistent during inference. %During the generation of a new scene, objects therein must form a valid structure and therefore must follow the scene grammar. 
Nevertheless we also aim to capture the advantages of deep generative models in admitting a compact representation that can be rapidly decoded. While a standard VAE would implicitly {\em encourage} decoded outputs to be scene-like, our proposed solution extends the Grammar VAE~\cite{kusner2017grammar} to explicitly enforce an underlying grammar, while still possessing the aforementioned advantages of deep generative models.  % Each object in a scene consists of a set of internal and external attributes (shape and pose parameters respectively). Our generative model is a sequential one, \ie it generates the scene one object at a time.
% In indoor environments the appearances of different objects have correlations between them. 
For example, given an appearance of an object \emph{bed}, the model finds strong evidence for co-occurrence of another indoor object, \eg~\emph{dresser}. Furthermore, given the attributes (3D pose and bounding boxes) of one object (\emph{bed}), the attributes of the latter (\emph{dresser}) can be inferred. 
%Our scene grammar aims to represent these relations. 
%% \iancomment{This para has its argument topsy-turvy and repetitive} : Pulak : I have tried removing a couple of sentences 

The model comprises two parts: {\bf (i)} a context free grammar (CFG) that represents valid configurations of objects; {\bf (ii)} a Variational  Autoencoder (VAE) that maps a sequence of production rules (\ie~a valid scene) to a low dimensional latent space, and decodes a latent vector to a sequence of production rules which in turn define a valid scene. % We describe each of these two parts in the sub-sections below. 

\subsection{CFG of indoor scenes}
A context-free grammar can be defined by a 4-tuple of sets $G = (S, \Sigma, \cV, \cR)$ where $S$ is a distinct non-terminal symbol known as start symbol; $\Sigma$ is the finite set of non-terminal symbols; 
$\cV$ is the set of terminal symbols; and $\cR$ is the set of production rules. 
% The rules are formally described as $\alpha \rightarrow \beta$ where $\alpha \in \Sigma$ and $\beta \in (\Sigma\cup\cV)^*$, with $*$ denoting the Kleene closure, each of which represents a generating process from a parent node $\alpha$ to its child nodes $\beta$. 
Note that in a CFG, the left hand side is always a non-terminal symbol. 
A set of all valid configurations $\cC$ derived from the production rules defined by the CFG $G$ is called a language. In contrast to~\cite{qi2018human} 
%% IAN \comment{This ref needs to come earlier as well, prob in introduction} : Pulak -- placed the reference in the introduction 
where the grammar is pre-specified, we propose a data-driven algorithm to generate a set of production rules that constitutes a CFG. 

We select a few objects and associate a number of non-terminals. Only those objects that lead to co-occurrence of other objects also exist as non-terminals (described in detail in Sec~\ref{sec:prune}). %% \iancomment{What do you mean "we select"?}%% CZ: How? Why? Reasoning?
A valid production rule is thus \emph{``an object category,  corresponding to a non-terminal, generates another object category''.} 
For clarity, non-terminals %% CZ: what is an anchor node? Not defined so far
are denoted in upper-case with the object name. % and the terminal symbols (objects) are placed inside \texttt{'\;'} symbol. 
For example \texttt{BED} and \texttt{bed} are the non-terminal and the terminal symbols corresponding to the object category \emph{bed}. 
%% \iancomment{why are the commas in inverted commas?} Pulak : These are dummy terminals incorporated to separate out objects, no use in our formulation though 
Thus occurrence of a non-terminal \texttt{\small BED} leads to occurrence of the immediate terminal symbol \texttt{bed} and possibly further occurrences of other terminal symbols that \emph{bed} co-occurs with, \eg~\texttt{dresser}. Thus, a set of rules \texttt{\scriptsize \{S $\rightarrow$ scene SCENE; SCENE $\rightarrow$ bed BED SCENE; BED $\rightarrow$ bed BED; BED $\rightarrow$ dresser  BED; BED $\rightarrow$ None; SCENE $\rightarrow$ None\}} \normalsize can be defined accordingly. % \texttt{\scriptsize{','}} is a dummy terminal symbol. 
Note that an additional object category \emph{scene} is incorporated to represent the shape and size of the room. 
The learned scene grammar is composed of following rules: 
\begin{enumerate}[(R1),itemsep=0.5ex,leftmargin=9.5mm,topsep=0.5ex,partopsep=0.5ex]
\item \emph{involving start symbol} \texttt{S}: generates the terminal \texttt{scene} and non-terminal \texttt{\small SCENE} that represents the indoor scene layout with attributes as the room size and room orientation, \eg~\texttt{\small S $\rightarrow$ scene SCENE;}. This rule ensures generating a room first. 
\item \emph{involving non-terminal} \texttt{\small SCENE}: generates a terminal and a non-terminal corresponding to an object category, \eg~{\small \texttt{SCENE $\rightarrow$ bed BED SCENE;}}. 
\item \emph{generating a terminal object category}: a non-terminal generates a terminal corresponding to another object category, \eg~{\small~\texttt{BED $\rightarrow$ dresser  BED;}}.  %$x_k\in \cV$, \eg \\ ${X_j} \rightarrow x_k,~ X_j $ % or ${X_j}\text{\texttt{inc}} \rightarrow X_k,~ X_k\text{\texttt{inc}},~ X_j\text{\texttt{inc}}  $  
\item \emph{involving }\texttt{None}: non-terminal symbols assigned to \texttt{None}, \eg~\texttt{\small BED $\rightarrow$ None;}. %\eg ${X_j}\text{\texttt{inc}} \rightarrow \text{\texttt{'None'}}$
\end{enumerate} 
% Where \texttt{S} is a non-terminal known as start symbol in CFG and 
% where $X_{j}$\texttt{inc} is a non-terminal corresponding to an object $X_j$.  
\texttt{None} is an empty object and corresponding rule is a dummy rule indicating that the generation of the non-terminal is complete and the parser is now ready to handle the next non-terminal in the stack. The proposed method to deduce a CFG from data is described in detail in Sec~\ref{sec:structurediscovery}. 

Note that the above CFG creates a necessary but not sufficient description. For example, a million dresser and a bed in a bedroom is a valid configuration by the grammar. Likewise, the relative orientation and shape are not included in the grammar, therefore a scene consisting of couple of small beds on a huge pillow is also a valid scene under the grammar. However, these issues are handled further by the co-occurrence distributions learned by the autoencoder.

\subsection{The VAE network} 

Let $\cD$ be a set of scenes comprising multiple objects. Let $\cS_i^j$ be the (bounding box) shape parameters and $\cP_i^j = (T_i^j;\;\gamma_i^j)$ be the (absolute) pose parameters of $j$th object in the $i$th scene where $T_i^j$ is the center and $\gamma_i^j$ is the (yaw) angle corresponding to the direction of the object in the horizontal plane, respectively.  Note that an object bounding box is aligned with gravity, thus there is only one degree of freedom in its orientation. The world co-ordinates are aligned with the camera co-ordinate frame. %Further, a scene  $\cI_i$ is consists of only a fraction of objects $j \in \cV$. 

The pose and shape attributes $\Theta^{j \rightarrow k} = ({\cP}_i^{j \rightarrow k}, \cS_i^{k})$ are associated with a production rule in which a non-terminal $X_j$ yields a terminal $X_k$. 
%In particular, $\Theta^{j \rightarrow k}$ is the concatenation of 3D pose (relative) and shape (bounding box) parameters of the object $X_k$ in the $i$-th scene. 
The pose parameters ${\cP}_i^{j \rightarrow k}$ of the terminal object $X_k$ are computed w.r.t.\ the non-terminal object $X_j$ on the left of the production rule. \ie~${\cP}_i^{j \rightarrow k} = {(\cP_i^{j})}^{-1}\cP_i^k$. 
%%The pose of the non-terminals are computed with respect to the start symbol \texttt{S} (scene) and 
%The pose of \texttt{S} is chosen as the pose of the camera. \ie 
%\begin{equation} 
%{\cP}_i^{j \rightarrow k} = \begin{cases}
% \cP_i^j, ~~~~~~~~~~~~~~~~~~\text{{If} \texttt{'S'} is on the left and the terminal $X_j$ is on the right}, \\
% 
% {(\cP_i^{j})}^{-1}\cP_i^k, ~~~~~\text{{If} a non-terminal $X_j$\texttt{inc} is on the left and a terminal $X_k$ is on right,} \\ 
% \mathbf{0} ~~~~~~~~~~~~~~~~~~~~~\text{{ Otherwise}} \\
%\end{cases}
%\end{equation}
%where $\mathbf{0}$ is the zero vector of size $4$. The above assignments are different values corresponding to rules (R1)-(R3) as described in Sec~\ref{sec:grammar}.  Note that ${(\cP_i^j)}^{-1}\cP_i^k$ corresponds to relative pose of the terminal object $X_k$ correspond to the non-terminal object $X_j$\texttt{inc}. 
%%The updated input data representation is passed through the encoder of VAE and the decoder also estimates the attributes. %Thus proposed VAE not only tries to yield the co-occurrences of valid objects, also tries to infer the object pose and shape. 
%\ie at every terminal node of the parse tree, the relative pose is computed w.r.t. its parent node. 
%The pose of the root node is chosen as the camera co-ordinates. 
The absolute poses of the objects are determined by chaining the relative poses on the path from the root node to the terminal node in the parse tree (see Figure~\ref{fig:parsetree}). Note that pose and shape attributes of the production rules corresponding to \texttt{None} object are fixed to zero. % vector. 

The VAE must encode and decode both production rules (1-hot vectors) and the corresponding pose and shape parameters. We achieve this by having separate initial branches of the encoder into which the attributes $\Theta^{j \rightarrow k}$, and the 1-hot vectors are passed. Features from the 1-hot encoding branch and the pose-shape branch are then concatenated after a number of 1D convolutional layers. These concatenated features undergo further 1D convolutional layers before being flattened and mapped to the latent space (thereby predicting $\bm{\mu}$ and $\bm{\Sigma}$ of $\cN(\bm{\mu}, \bm{\Sigma})$). The decoding network is a recurrent network consisting of a stack of GRUs, that takes samples $\bm{z}\sim \cN(\bm{\mu}, \bm{\Sigma})$ (employing reparameterization trick~\cite{kingma2013auto}) and outputs logits (corresponding to the production rules) and corresponding attributes $\Theta^{j \rightarrow k}$.    
Logits corresponding to invalid production rules are masked out. % (as in Grammar VAE).
%Note that masking out of the components is performed only on the logits for a valid scene generation.
%In the experimental Sec~\ref{sec:experiments} we investigate alternative network architectures. 
 
The reconstruction loss of our SG-VAE consists of two parts: {\bf (i)} a cross entropy loss corresponding to the 1-hot encoding of the production rules---note that \textit{soft-max} is computed only on the components after mask-out---and {\bf (ii)} a mean squared error loss corresponding to the production rule attributes (but omitting the terms of \texttt{None} objects).
%corresponding to only a non-zero vector is considered, \ie attributes of \texttt{'None'} object is omitted in the loss.
%% CZ: I didn't understand that part : Pulak - hope I made it clear now 
Thus, the loss is given as follows:
\begin{align}
\cL_{total}(\phi, \theta; \cX, \Theta) =  \cL_{vae}(\phi, \theta; \cX) + \lambda_1 \Big(\cL_{pose}(\phi, \theta; \cP)  + \lambda_2 \cL_{shape}(\phi, \theta; \cS)\Big)
\end{align}
where $\cL_{vae}$ is the autoencoder loss~\cite{kusner2017grammar}, and $\cL_{pose}$ and $\cL_{shape}$ are mean squared error loss corresponding to pose and shape parameters, respectively; $\phi$, and $\theta$ are the encoder and decoder parameters of the autoencoder that we optimize; $(\cX, \Theta)$ are the set of training examples comprising 1-hot encoders and rule attributes. Instead of directly regressing the orientation parameter, the respective \textit{sines} and \textit{cosines} are regressed. Our choice is $\lambda_1=10$ and $\lambda_2=1$ in all experiments. 
% The details of the model architecture are described in the supplementary material. 

\section{Discovery of the scene grammar} \label{sec:structurediscovery}
% In this work, an indoor environment is considered  

In much previous work a grammar is manually specified. However in this work we aim to discover a suitable grammar for scene layouts in a data-driven manner.  It comprises two parts.  First we generate a causal graph of all pairwise relationships discovered in the training data, as described in more detail in Sec~\ref{sec:causal}. Second we prune this causal graph by removing all but the dominant discovered relationships, as described in Sec~\ref{sec:prune}.  

\subsection{Data-driven relationship discovery}\label{sec:causal} 

We aim to discover causal relationships of different objects that reflects the influence of the appearance (\ie~pose and shape) of one object to another. We learn the relationship using hypothesis testing, with each successful hypothesis added to a causal graph (directed) $\cG:(\cV,\;\cE)$ where the vertex set $\cV = \{X_1,\ldots, X_n\}$ is the set of different object categories, and edge set $\cE$ is the set of causal relationships. An edge $(X_j \causal X_{j^\prime})\in \cE$ corresponds to a direct causal influence on occurrence of the object $X_j$ to the object $X_{j^\prime}$. %In the causal graph, occurrence of an object influences occurrences of another adjacent object. 
We conduct separate {\bf (i)} appearance based and {\bf (ii)} co-occurrence based testing for causal relationships between a pair of object categories as set out below. \\ %\vspace{-1em}
%\\ \vspace{-0.5em}
\begin{algorithm}[H]%\captionsetup{labelfont={sc,bf}, labelsep=newline}
\scriptsize 
 \caption{$\chi^2$-test for conditional independence check}\label{chisquare_conditional_algo}
 
    \KwIn { Co-occurrences $O$ of the objects $X_j, X_{j^\prime}, X_k$ }
    \KwOut{ \emph{\bf True} if $X_j \indep X_{j^\prime} \; |\; X_k$ ~~and~~ \emph{\bf False}~~ Otherwise }
%	Discard the rows of the indices $O_i$ corresponding to the $i$ scenes where none of the objects occur. 
 \vspace{-0.27cm}
   \begin{minipage}{0.40\textwidth}
   \[
   \chi^2 = \sum_{j, j^\prime, k \in (\{0, 1\})^3} \frac{\left(O_{j, j^\prime, k} - \frac{O_{j,k}O_{j^\prime, k}}{O_k}\right)^2}{\frac{O_{j,k}O_{j^\prime, k}}{O_k}}, 
 %  \chi^2 = \!\!\!\!\sum_{j, j', k \in \{0, 1\} }\!\!\!\!
  % \big( O_{j, j', k} - O_{j,k} O_{j',k}/O_k \big)^2/\big(O_{j,k} O_{j',k}/O_k\big), 
   \]
   \end{minipage} ~~~~~~~~
   %\hfill\vline\hfill 
%   \hspace{0.8cm}
   \begin{minipage}{0.44\textwidth}
    $O_{j, j^\prime, k}$: frequency of occurrences of $(j, {j^\prime}, k)$, \\
    ~~$O_{j, k}$~~~: frequency of occurrences of $(j, k)$, \\
    ~~~$O_{k}$~~~~~: frequency of occurrences of $k$, \\
     ~~$N$~~~~~~: number of scenes 
   
   \end{minipage} \\ 
%   The value $0$ correspond to absence of an object instance and $1$ otherwise. \\ 
   Compute $p$-value from cumul. $\chi^2$ distrib. with above $\chi^2$ value and d.o.f. \tcc*{D.o.f is $2$} 
%    \If {$p$-value is less than a predefined threshold $\tau$ ($0.05$) }{
%	     \Return False \tcc*{The hypothesis is rejected} 
%    } {
%    \Return True
%    }
    \Return $p$-value $< \tau$ (we choose $\tau=0.05$)
    \end{algorithm} %\vspace{-1em}
%\DecMargin{1.5em} % \setlength{\floatsep}{0.1cm} 
%\fi

\noindent {\bf (i) Testing for dependency based on co-occurrences}
We seek to capture loose associations 
(\eg\ sofa and TV) and determine if these associations have a potentially causal nature. 
To do so, for each pair of object categories we then consider whether these categories are dependent, given a third category. This is performed using the Chi-squared ($\chi^2$) test described below. If the dependence persists across all possible choices of the third category, we conclude that the dependence is not induced by another object, therefore a potentially causal link should exist between them. 
This exhaustive series of tests is $O(Nm^3)$, where $N$ is the number of scenes and $m$ is the number of categories (in our case, $84$). However it is performed offline and only once. This procedure creates an undirected graph with links between pairs where a causal relationship is hypothesized to exist. To establish the direction of causation---\ie\ turn the undirected graph into a directed one, we use Pearl's Inductive Causation algorithm~\cite{pearl1995theory} and the procedure is summarized in Algorithm~2 of the supplementary.

In more detail the $\chi^2$-test checks for conditional independence of a pair of object categories $\{X_j, X_{j^\prime}\}$ given an additional object category $X_k \in \cV\setminus \{X_j, X_{j^\prime}\}$. The probabilities required for the test are obtained from the relative frequencies of the objects and their co-occurrences in the dataset.   Algorithm~\ref{chisquare_conditional_algo} describes this in detail. By way of example, \emph{pillow} and \emph{blanket} might co-occur in a substantial number of scenes, however, their co-occurrences are influenced by a third object category \emph{bed}. In this case, the pairwise relationship between \emph{pillow} and \emph{blanket} is determined to be independent, given the presence of \emph{bed}, so no link between \emph{pillow} and \emph{blanket} is created. %\\ \vspace{-2.5em}

%\noindent Our hypothesis is \emph{``if occurrence of an object category influences occurrence of another object category, there is a causal relationship from former to later''}. 
%Further, a pair of objects categories $(A, B)$ can co-occur for a triplet orderings with 

%an additional  object category $C$, \eg $A\causal C \causal B$  or $A \reversecausal C \causal B$, and thus the above 
% hypothesis might wrongly add a causal relationship of $A$ and $B$~\cite{pearl1995theory}.  

 % and the triplet \emph{pillow} $\reversecausal$ \emph{bed} $\causal$ \emph{blanket} will be established.  % In other words, given location and orientation of \emph{bed}, location and orientation of \emph{pillow} does not give us any additional information. 

%\def\skipalgoone{1}
%\ifx\skipalgoone\undefined
%The null hypothesis is 
%\begin{equation}
%\cH_0^{jj^\prime|k} : X_j \indep X_{j^\prime} | X_k \text{~~~\ie Objects $X_j$ and %$X_{j^\prime}$ are independent given $X_k$} \tag{$\cH_0$} \label{eq:hypo} 
%\end{equation} 
%\IncMargin{1.5em} \setlength{\textfloatsep}{0.1cm}  \setlength{\floatsep}{0.1cm} 
\noindent {\bf (ii) Testing for dependency based on shape and pose}\label{sec:poseprior}
In addition to the conditional co-occurrence captured above, we also seek to capture covering/enclosing and supporting relationships---which are defined by the shape and pose of the objects as well as the categories---in the causal graph. More precisely we hypothesize a causal
 relationship between object categories $A$ and $B$ if: 
\begin{figure}[H]%{l}{0.660\textwidth}
\centering 
%\vspace{-10pt}
\includegraphics[width=1\textwidth]{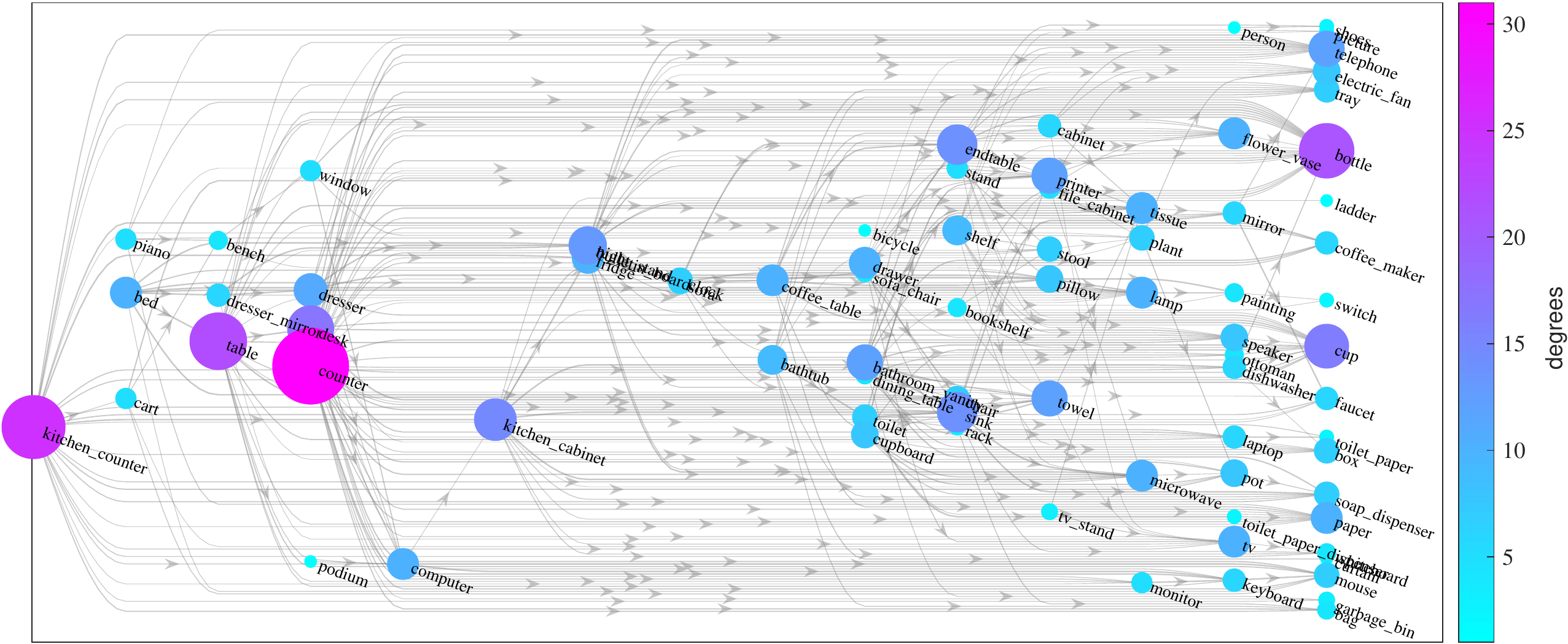}
\caption{The above graph is generated by the modified IC algorithm on SUNRGBD-3D Dataset. An arrowhead of an edge indicates the direction of causation and the color of a  node indicates its degree.  } %\vspace{-1em}
%\vspace{-10pt}
\label{fig:scene_graph} 
\end{figure} 
% \emph{``if an object is supported or covered by another object---pose of the former can be inferred given the pose of the later.} \ie 
%The causal graph $\cG$ is thus initialized with following physical (prior) knowledge: 
\begin{itemize}[itemsep=0.0ex,leftmargin=4.5mm,topsep=0.0ex,partopsep=0.0ex]
\item object category $A$ is found to support object category $B$ (\ie\ their relative poses and shapes are such that, within a threshold, one is above and touching the other), or,   
%---a causal relationship is instigated from former to later. 
%\eg \emph{tv$\_$stand} $\causal$ \emph{tv}.  
%This will encourage no flying objects in the generated scene. 
\item object category $B$ is enclosed/covered by another larger object category $A$ (again this is determined using a threshold on the objects' relative shapes and poses). 
%---a causal relationship is instigated from later to former.  
% For example, if a \emph{dishwasher} is covered by a \emph{kitchen$\_$counter}. %, \eg \emph{kitchen$\_$counter} \causal \emph{dishwasher}. % This will encourage occurrence of of a \emph{dishwasher} inside a \emph{kitchen$\_$counter}.  
\end{itemize}
We accept a hypothesis and establish the causal relationship (by entering a suitable edge into the causal graph) if at least $30\%$ of the co-occurrences of these object categories in the dataset agree with the hypothesis.  %However, 
%The dependencies constitute a graph $\cG = (\cV, \cE_0)$ which is the initialization of the final causal graph.   
%Note that the above is the only prior knowledge employed in this work. %In the following section we discover the dependencies from the co-occurrences to construct the full causal graph. 
The final directed causal relational graph $\cG$ is the union of the causal graphs generated by the above tests. Note that we do not consider any dependencies that would lead to a cycle in the graph~\cite{dor1992simple}. The result of the procedure is also displayed in Figure~\ref{fig:scene_graph}. 

\subsection{Creating a CFG from the causal graph}\label{sec:prune} 
%The causal graph $\cG$ generated in the above section is utilized for this matter. 
% For example, a rule can be \textit{an object category generates another adjacent object category if there is a directed edge from former to later in the graph}. 
We now need to create a Context Free Grammar from the causal graph. The CFG is characterised by non-terminal symbols that generate other symbols.  Suppose we choose a particular node in the causal graph (\ie\ an object category) and assume it is  non-terminal. By tracing the full set of directed edges in the causal graph emanating from this node we create a set of production rules.  This non-terminal and associated rules are then tested against the dataset to determine how many scenes are explained (formally ``covered'') by the rules.  A good choice of non-terminals will lead to good coverage.  Our task then is to determine an optimal set of non-terminals and associated rules to give the best coverage of the full dataset of scenes.

Note that finding such a set is a combinatorial hard problem. Therefore, we devise a greedy algorithm to select non-terminals and find approximate best coverage. Let $X_j$, an object category, be a potential non-terminal symbol and $\cR_j$ be the set of production rules derived from $X_j$ in the causal graph. Let $C_j$ be the set of terminals that $\cR_j$ covers (essentially nodes that $X_j$ leads to in $\cG$). Our greedy algorithm begins with an empty set $\cR = \emptyset$ and chooses the node $X_j$ and associated production rule set $\cR_j$ to add that maximize the \emph{gain} in coverage $\cG_{gain}(\cR_j, \cR) = \frac{1}{|\cR_j|} \sum_{I_i \in \cI \setminus \cC} |Y_i| / |I_i|$. %\\ \vspace{-0.5em}

\begin{figure}[H]
\centering \scriptsize 
\begin{tabular}{@{\hspace{0.1em}}c@{\hspace{0.2em}}c@{\hspace{0.5em}}c@{\hspace{0.5em}}c} 
 \includegraphics[width=0.22\textwidth, height=2.0cm, trim={1.5 1.5 8 1.5},clip]{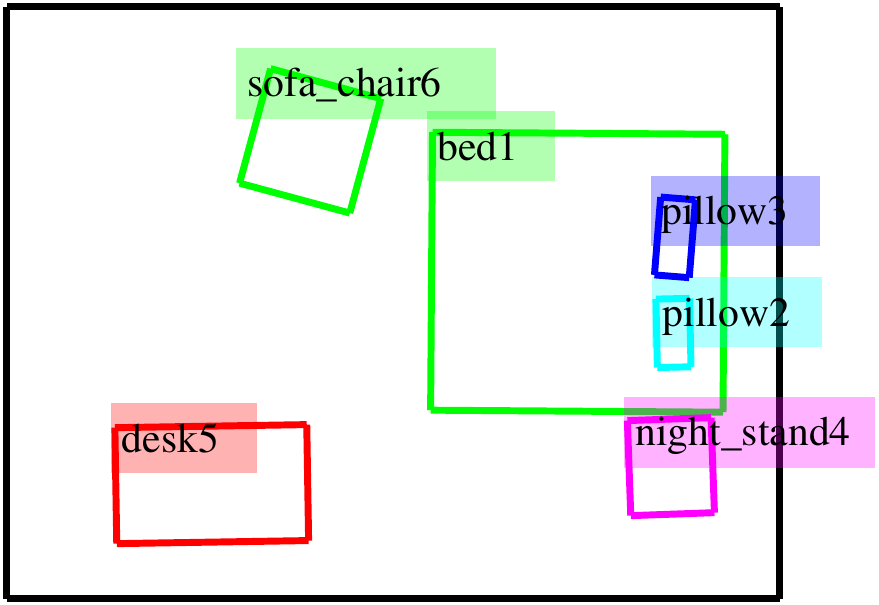} & 
\includegraphics[width=0.20\textwidth, height=2.0cm, trim={1 1 1 1},clip]{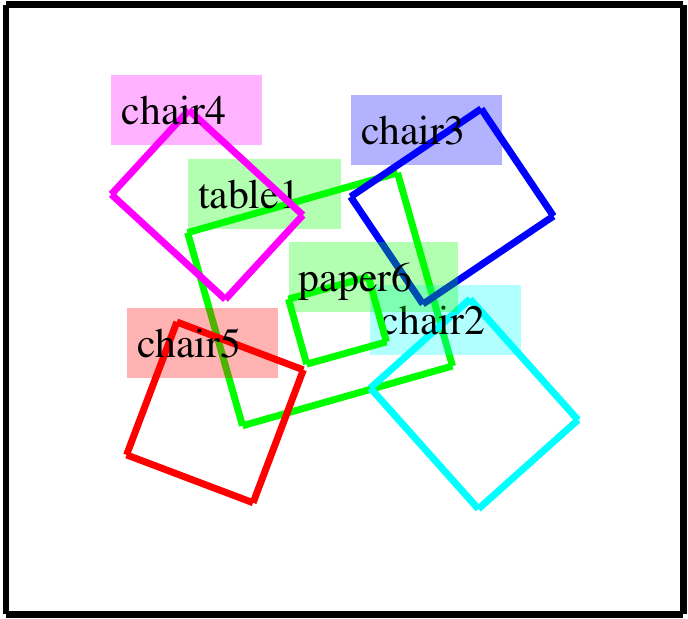} &
\includegraphics[width=0.22\textwidth, height=2.0cm, trim={1 1 8 1},clip]{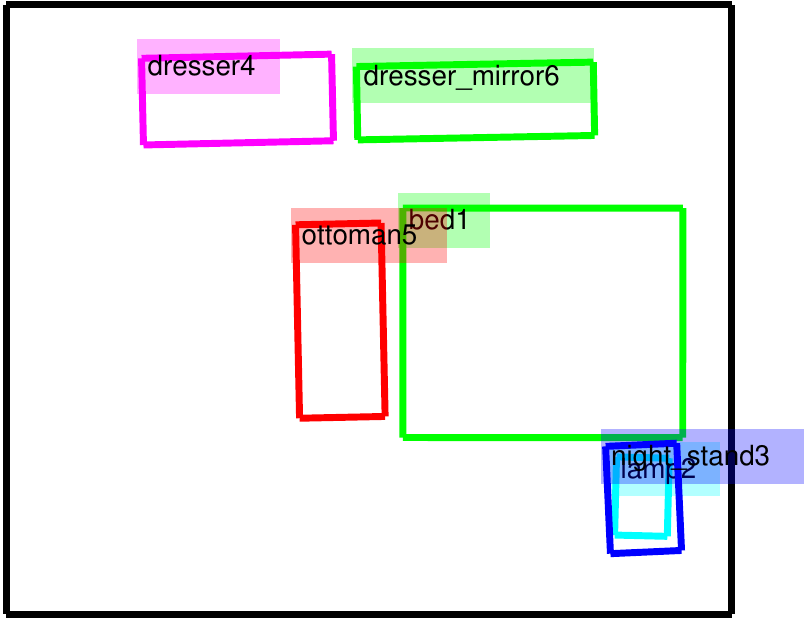} &
\includegraphics[width=0.20\textwidth, height=2.0cm, trim={1 1 1 1},clip]{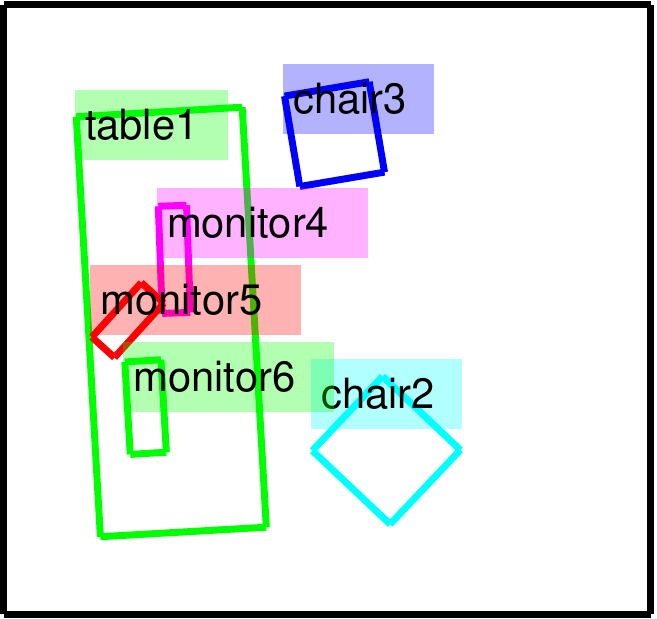} \\ 
(a) Ground-truth 1 & (b) Ground-truth 2 & (c) Ground-truth 3 & (d) Ground-truth 4 
\end{tabular}
\caption{The parsing order is displayed by a numeral concatenated with the object name. (a) the objects are sorted in the order of precedence defined by the grammar, and (b) multiple chairs are sorted in the appearance w.r.t. the table in anti-clockwise direction starting from the bottom right corner. (c)-(d) More examples of the object order in the ground-truth samples from SUN RGB-D dataset~\cite{song2015sun}. }
\label{fig:2dprojsample}
\end{figure}

\noindent  {\bf Unique parsing} Given a scene there could be multiple parse trees derived by the leftmost derivation grammar and hence produces different sequences and different representations. For example, for a scene consist of \texttt{bed}, \texttt{sofa} and \texttt{pillow}, the terminal object \texttt{pillow} could be generated by any of the non-terminals \texttt{\small BED} or \texttt{\small SOFA}. This ambiguity can confuse the parser while encoding a scene. Further, different orderings of multiple occurrences of an object lead to different parse trees.  % We  fix the ambiguity as below.  %
We consider following parsing rules to remove the ambiguity:
\begin{itemize}[itemsep=0.5ex,leftmargin=4.5mm,topsep=0.5ex,partopsep=0.5ex]
\item Fix the order of the object categories in the order of precedence defined by the grammar. %and the terminals as the proportion of outward degree and inward degree of a node in the causal graph. % The object categories in a scene are also sorted in the same order.  
\item Multiple occurrences of an object are sorted in the appearance w.r.t. the preceded object in the anti-clockwise direction starting from the object  making minimum angle to the orientation of the preceded object. One such example  is shown in Figure~\ref{fig:2dprojsample}. Note that this ordering is only required during training. 
% \item Inspect the pose dependency of the objects in a scene as described in \ref{sec:poseprior} and assign a higher precedence to the corresponding rule if a dependency was found. 
\end{itemize}

 %\vspace{-2em}

\section{Experiments}\label{sec:experiments} 
\noindent {\bf Dataset}\label{sec:datasets}
We evaluate the proposed method on SUN RGB-D Dataset~\cite{song2015sun} consisting of ${10,335}$ real scenes with ${64,595}$ 3D bounding boxes and about $800$ object categories. The dataset is a collection of multiple datasets~\cite{janoch2013category,silberman2012indoor,xiao2013sun3d}, and  is highly unbalanced: \eg\ a single object category \emph{chair} corresponds to about $31\%$ of all the bounding boxes and $38\%$ of total object categories occur just once in the entire dataset. We consider object categories appearing at least $10$ times in the dataset for evaluation. Further, very similar object categories are merged, yielding $84$ object categories and ${62,485}$ bounding boxes. %\vspace{-1em}

\begin{table}[H]\setlength{\tabcolsep}{0.5pt}
\centering 
\caption{Results of 3D bounding box reconstruction of some of the frequent objects under a valid reconstruction with IoU $> 0.25$~\cite{qi2020imvotenet}. The pose estimation results are furnished within braces (angular errors in degrees, displacement errors in meters).} \tiny  
\begin{tabular}{@{\hspace{0.01em}}c|c@{\hspace{0.01em}}c@{\hspace{0.01em}}c|c@{\hspace{0.01em}}c@{\hspace{0.01em}}c|c@{\hspace{0.01em}}c@{\hspace{0.01em}}c|c@{\hspace{0.01em}}c@{\hspace{0.01em}}c|c@{\hspace{0.01em}}c@{\hspace{0.01em}}c@{\hspace{0.01em}}}
\toprule
 Objects & \multicolumn{3}{c|}{\texttt{chair}} & \multicolumn{3}{c|}{\texttt{bed}} & \multicolumn{3}{c|}{\texttt{table}} & \multicolumn{3}{c|}{\texttt{ktchn$\_$cntr}} & \multicolumn{3}{c}{\texttt{piano}} \\ 
\midrule 
{SG-VAE}  & $\bm{92.3}$ & $({\bf 5.88^{\circ}}{,}$ & $\bm{0.13m})$ & $\bm{100.0}$ & $(\bm{2.80^{\circ}},  $ & $\bm{0.08m})$ & ${\bf 96.2}$ & $({\bf 2.84^{\circ}},  $ & $\bm{0.08m})$  & ${\bf 100.0}$ & $(\bm{1.69^{\circ}},  $ & $\bm{0.08m} )$ & ${\bf 67.1}$ & $({\bf 6.93^{\circ}},  $ & $\bm{0.11m}) $ \\
{BL1} & ${75.4}$ & ${(8.30^{\circ}},  $ & $\bm{0.14m})$ & $98.5$ & ${(5.70^{\circ}},  $ & $\bm{0.09m} )$ & $93.8$ & ${(3.70^{\circ}},  $ & $\bm{0.08m}  ) $ & $\bm{ 100.0}$ & $({\bf 1.62^{\circ}},  $ & $\bm{0.06m}) $ & $48.7$ & ${(10.3^{\circ}},  $ & $\bm{0.12m} )$\\
{BL2}~\cite{gomez2018automatic} & $      33.7$ & ${(28.1^{\circ}},  $ & ${0.32m)}$ & $75.1$ & ${(7.12^{\circ}},  $ & ${0.45m)     } $ & ${ 90.2}$ & ${(7.69^{\circ}},  $ & ${0.33m)}   $  & $83.2$ & ${(5.96^{\circ}},  $ & ${0.09m)} $ & $0.80$ & ${(45.1^{\circ}},  $ & ${0.43m) }$\\
{BL3}\cite{kusner2017grammar}+\cite{yu2011make} & $      29.2$ & ${(41.8^{\circ}},  $ & ${0.26m)}$ & $88.7$ & ${(35.7^{\circ}},  $ & ${0.58m)}      $ & $72.9$ & ${(56.7^{\circ}},  $ & ${0.34m) }  $  & ${ 100.0}$ & ${(52.5^{\circ}},  $ & ${0.78m)} $ & $26.3$ & ${(17.4^{\circ}},  $ & ${0.33m)} $
%{\bf BL3} & -- & -- & --  & -- \\ % & -- 
\\ \bottomrule
\end{tabular} %\vspace{-1em}
\label{tab:results} 
\end{table} 
%Although, the room boundaries (floor, wall, ceiling) are interesting but not considered in this work. 
\noindent We separated $10\%$ of the data at random for validation. On average there are $4.29$ objects per image and maximum number of objects in an image is considered to be $15$. This also provides the upper bound of the length of the sequence generated by the grammar.   
Note that the dataset is the intersection of the given dataset and the scene language, \ie\ the possible set of scenes generated by the CFG.  %\vspace{-2em}
\begin{table}[H]\setlength{\tabcolsep}{13pt} %{r}{0.50\textwidth}
%\vspace{-10pt} 
\caption{IoU of room layout estimation}\label{wrap-tab:1}
\centering 
\scriptsize 
\begin{tabular}{c|c|c|c|c}\toprule    
Methods & {\bf SG-VAE}  & {\bf BL1} & {\bf BL2}~\cite{gomez2018automatic} & {\bf BL3}~\cite{kusner2017grammar}+\cite{yu2011make}  \\ \midrule
Grammar & \checkmark & \checkmark & & \checkmark \\ 
Pose $\&$ Shape & \checkmark & \checkmark & \checkmark & \\  \midrule 
%Grammar & \checkmark & \checkmark & & \checkmark 
IoU & $\bm{0.6240}$ & $0.5673$ & $0.2964$ & $0.5119$ \\ \bottomrule 
%\vspace{-25pt}
\end{tabular} %\vspace{-1em}
\end{table}

\noindent {\bf Baseline methods for evaluation (ablation studies)}
To evaluate the individual effects of (i) output of the decoder structure, (ii) usage of grammar, and (iii) the pose and shape attributes, the following baselines are chosen: 
\begin{enumerate}[({\bf BL}1),itemsep=0.5ex,leftmargin=12mm,topsep=0.5ex,partopsep=0.5ex]
\item \emph{Variant of SG-VAE}: In contrast to the proposed SG-VAE where attributes of each rule are directly concatenated with 1-hot encoding of the rule, in this variant separate attributes for each rule type are predicted by the decoder and rest are filled with zeros. 
\item \emph{No Grammar VAE}~\cite{gomez2018automatic}: No grammar is considered in this baseline. The 1-hot encodings correspond to the object type is concatenated with the absolute pose of the objects (in contrast to rule-type and relative pose in SG-VAE) respectively. 
\item \emph{Grammar VAE}~\cite{kusner2017grammar} + \emph{Make home}~\cite{yu2011make}: The Grammar VAE is incorporated with our extracted grammar to sample a set of coherent objects and~\cite{yu2011make} is used to arrange them. Sampled $10$ times and solution corresponding to best IoU w.r.t.\ groundtruth is employed. The details of the above baselines are provided in the supplementary.  %({\bf BL3L}) is the same baseline but we sample $32$ times in this case. % We could not find an open source implementation and rely on our own implementation. 
\end{enumerate}
All the baselines including SG-VAE are implemented in \texttt{python 2.7 (Tensorflow)} and trained on a GTX 1080 Ti GPU.  %\vspace{-0.5em}
%The relevant source codes are shared~\footnote{\href{https://github.com/xxx/SG-VAE}{https://github.com/pulak09/SG-VAE}} for reproducibility. \\ 
~\\ 
\noindent {\bf Evaluation metrics} 
The relative poses of individual objects in a scene are accumulated to compute their absolute poses which are then combined with the shape parameters to compute the scene layout. 
The reconstructed scene layouts are then compared against the groundtruth layouts. 
\begin{itemize}[itemsep=0.5ex,leftmargin=4.5mm,topsep=0.5ex,partopsep=0.5ex]
\item {\bf 3D bounding box reconstruction} 
We employed IoU to measure the shape similarity of the bounding boxes. 
% $ S_{bb}(m, n) = {V_{ov_{mn}}}/{V_m + V_n - V_{ov_{mn}}} $ where $V_{ov_{mn}}$ is the volume of the intersection between $m$ and $n$, and $V_m$ and $V_n$ are the volumes of $m$ and $n$ respectively. 
 Reconstructed bounding boxes with IoU $>0.25$ are considered as true positives. 
The results are reported in Table~\ref{tab:results}. 
\item {\bf 3D Pose estimation}
%
%The symmetry property of individual objects is not customized, for example, a rectangular object has a 2-fold rotational symmetry about the vertical axis. 
The average pose error is considered over two separate metrices: {\bf (i)}  angular error in degrees, {\bf (ii)} displacement error in meters (see Table~\ref{tab:results}). % Only the true positive boxes are considered. 
\item {\bf Room Layout Estimation}
The evaluation is conducted as the IoU of the \emph{occupied} space between the groundtruth and the predicted layouts (see Table~\ref{wrap-tab:1}). The intersection is computed only over the true positives.  % The details of the evaluation strategy can be found in~\cite{song2015sun}.   
\end{itemize} 
\begin{figure}[H] \tiny \centering 
\begin{tabular}{@{\hspace{0.01em}}l@{\hspace{1.0em}}c@{\hspace{0.3em}}c@{\hspace{0.3em}}c@{\hspace{0.3em}}c@{\hspace{0.3em}}c@{\hspace{0.3em}}c@{\hspace{0.3em}}c} \\ %\vspace{-0.1cm} \\ %\vspace{0.01cm}
  \begin{picture}(1,25)\put(0, 5){\rotatebox{90}{($\alpha = 0.9$)}}\end{picture} & 
  \includegraphics[width=0.154\textwidth, height=1.4cm ]{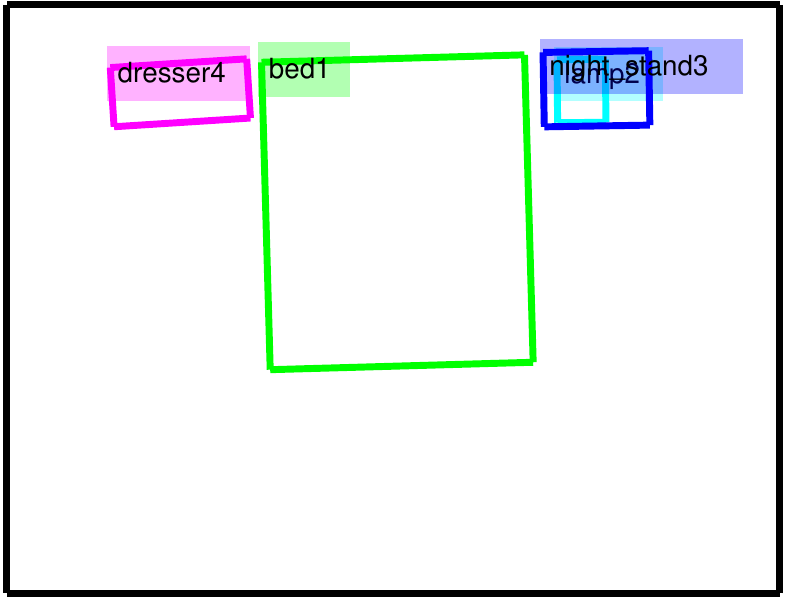}  & 
\includegraphics[width=0.154\textwidth, height=1.4cm ]{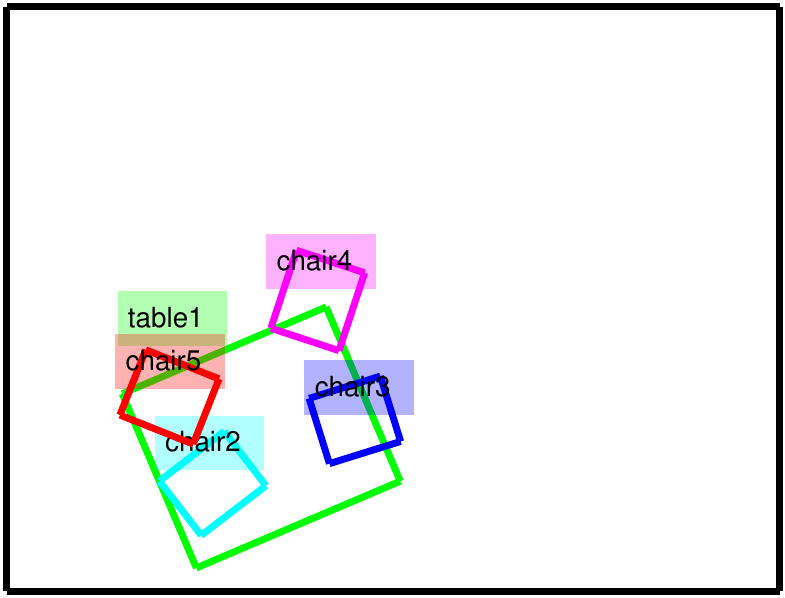}  & 
\includegraphics[width=0.154\textwidth, height=1.4cm ]{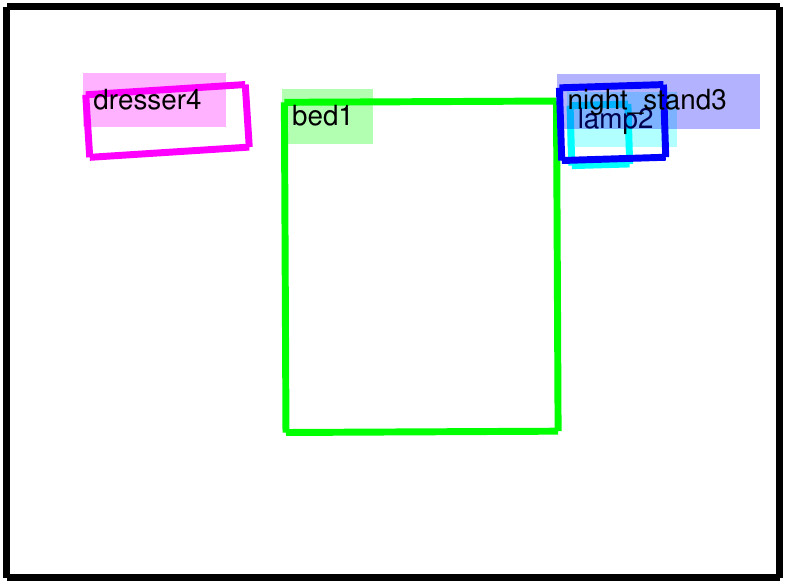}  & 
\includegraphics[width=0.154\textwidth, height=1.4cm ]{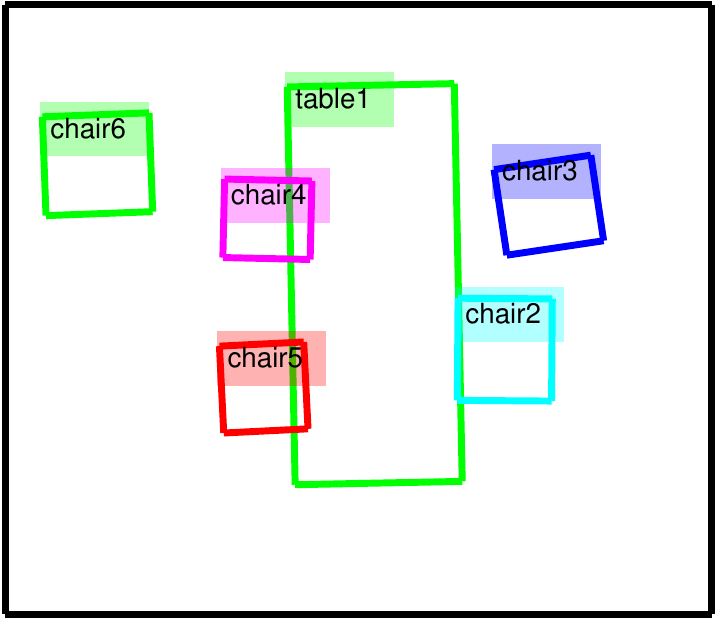}  & 
\includegraphics[width=0.154\textwidth, height=1.4cm ]{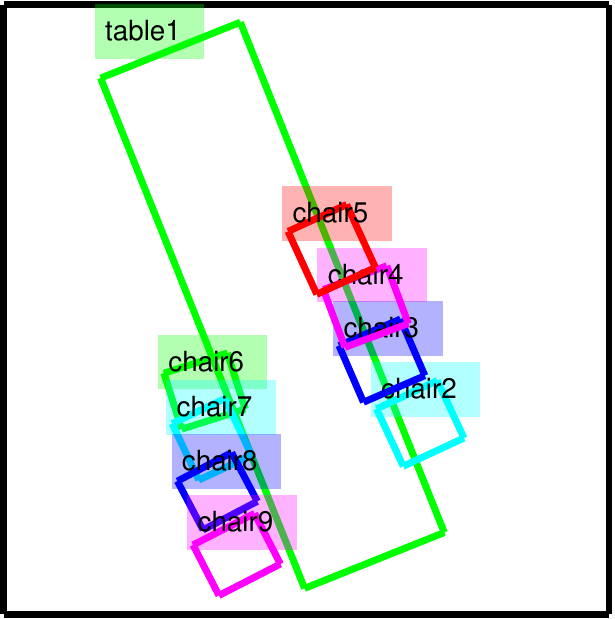}  & 
\includegraphics[width=0.154\textwidth, height=1.4cm ]{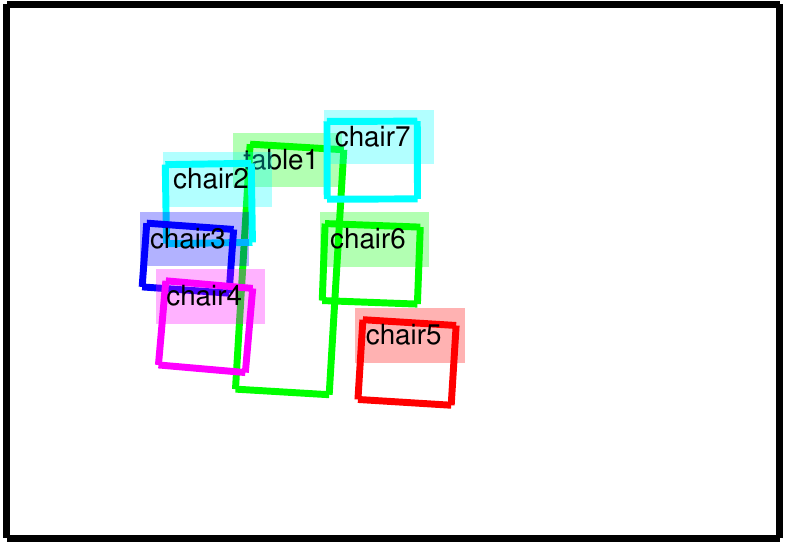}  & 
  \begin{picture}(1,25)\put(0, 5){\rotatebox{90}{Top View}}\end{picture} \\ 

  \begin{picture}(1,25)\put(0, 5){\rotatebox{90}{($\alpha = 0.9$)}}\end{picture} & 
  \includegraphics[width=0.154\textwidth, height=1.4cm ]{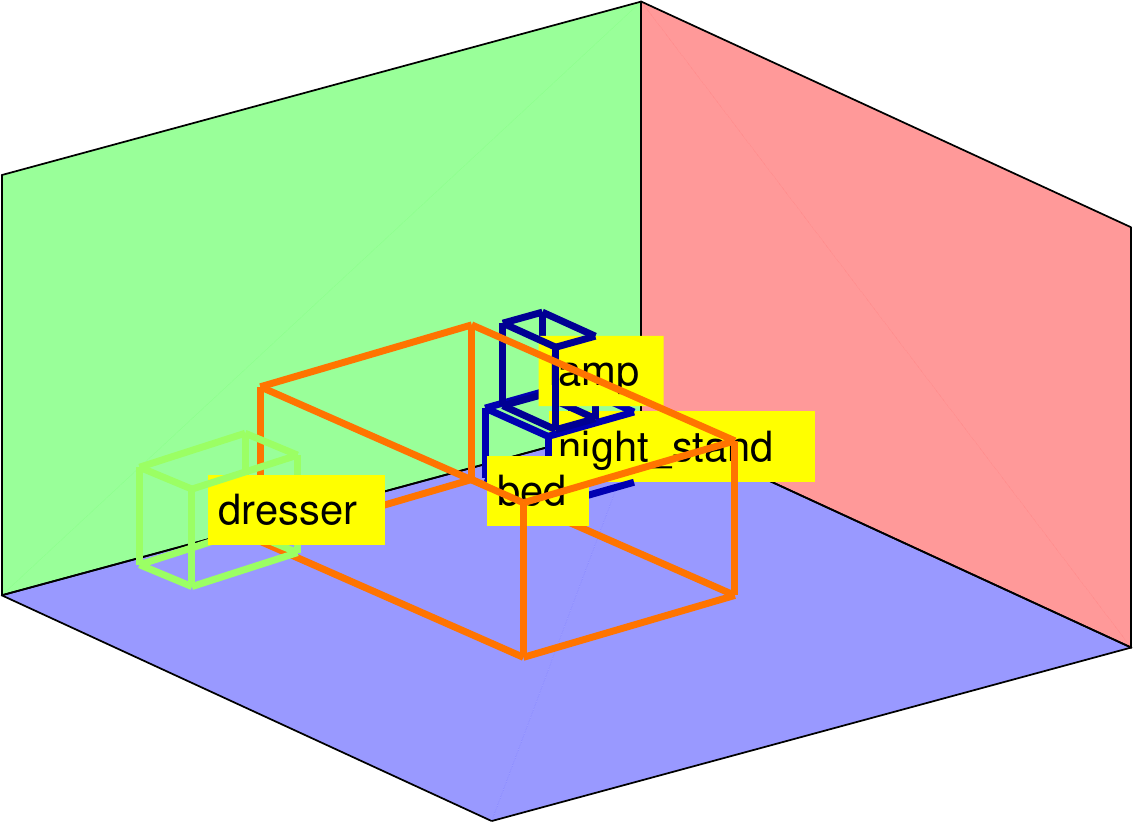}  & 
\includegraphics[width=0.154\textwidth, height=1.4cm ]{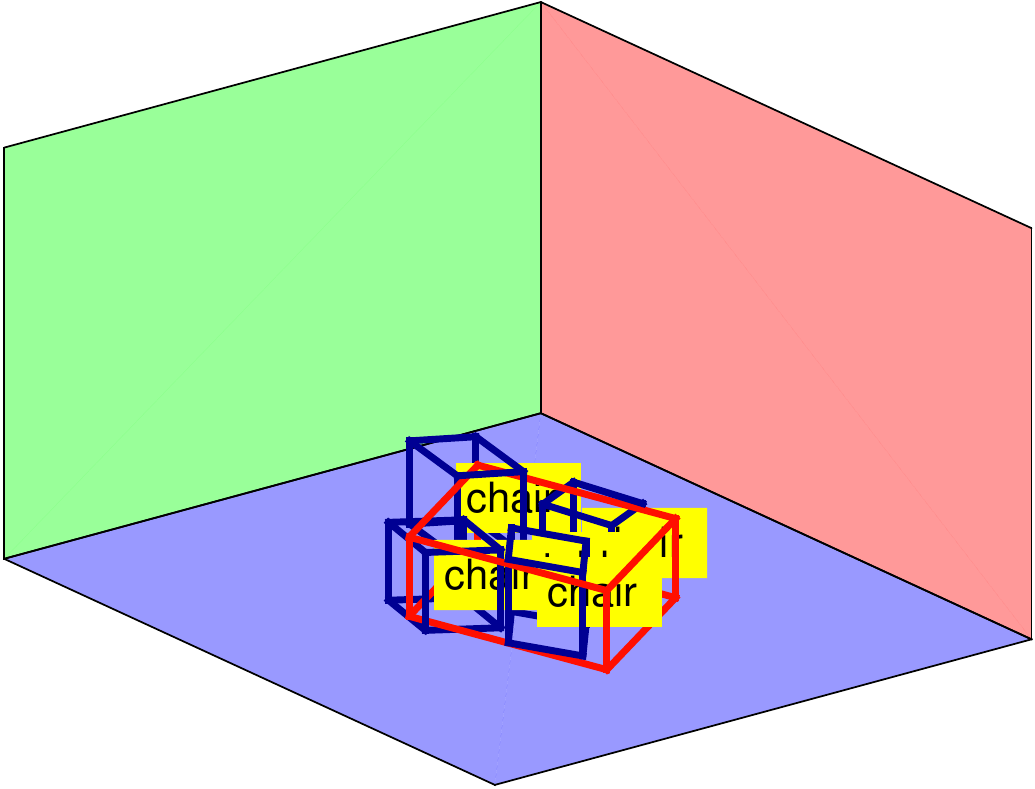}  & 
\includegraphics[width=0.154\textwidth, height=1.4cm ]{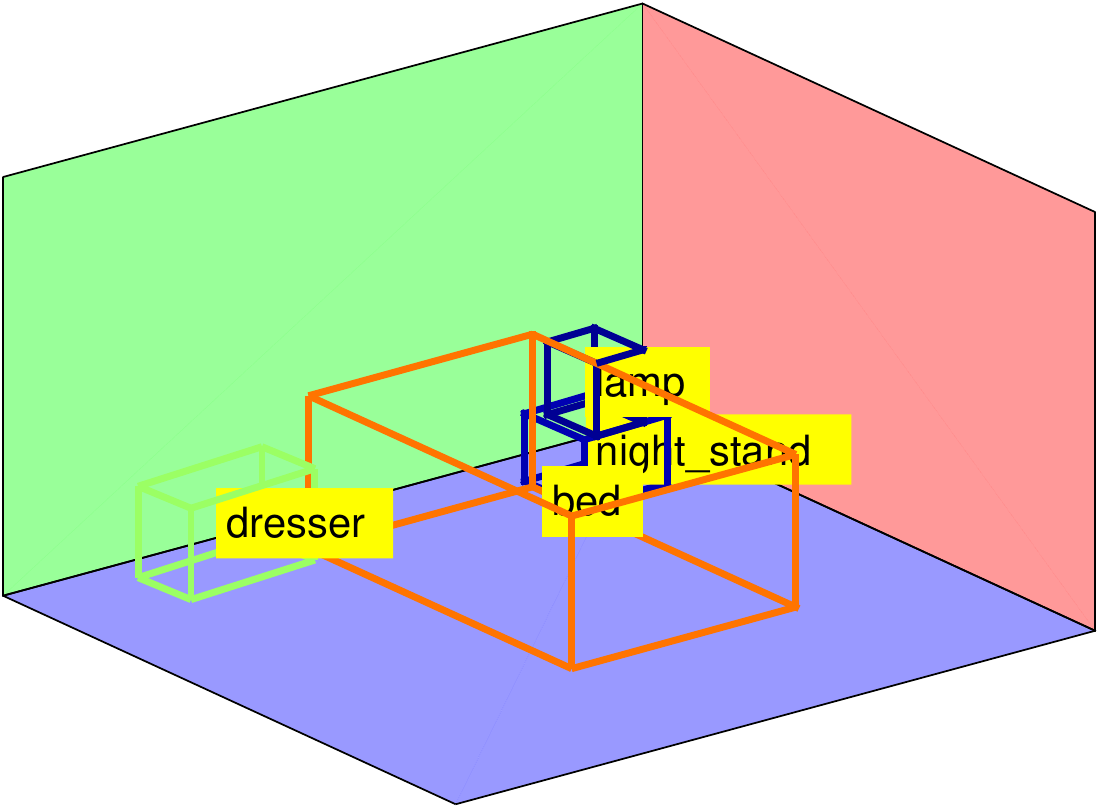}  & 
\includegraphics[width=0.154\textwidth, height=1.4cm ]{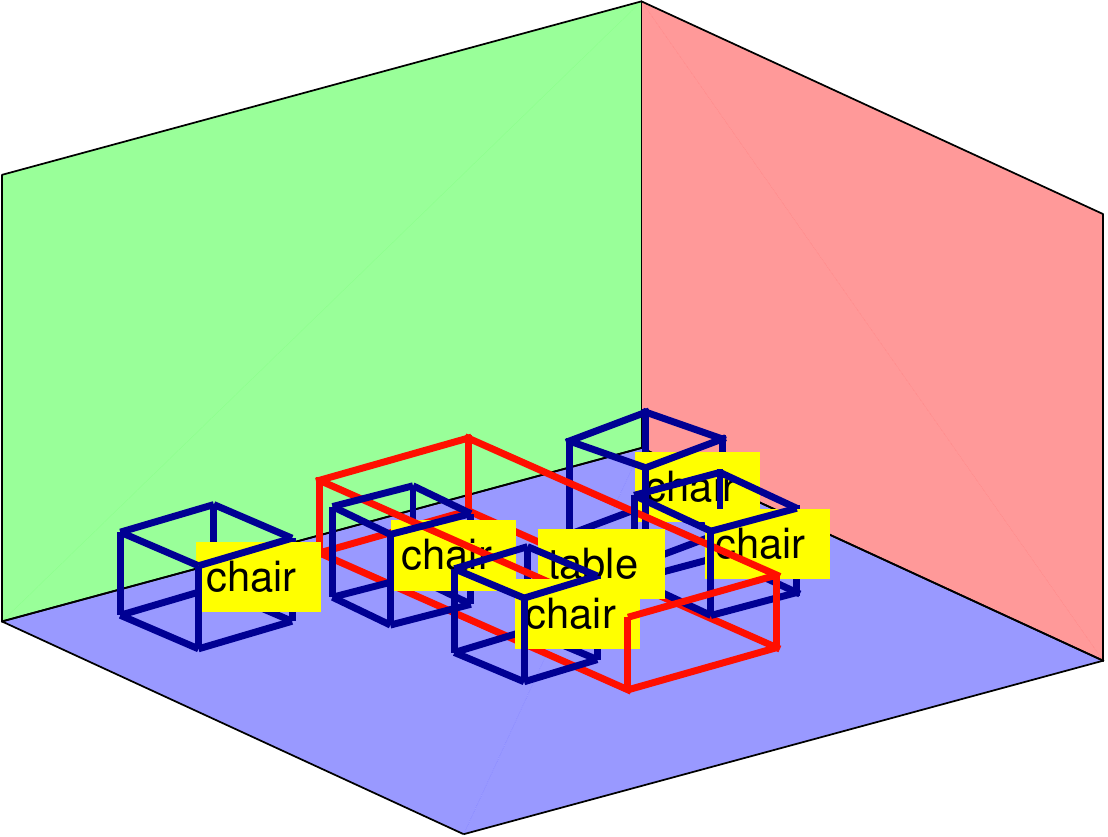}  & 
\includegraphics[width=0.154\textwidth, height=1.4cm ]{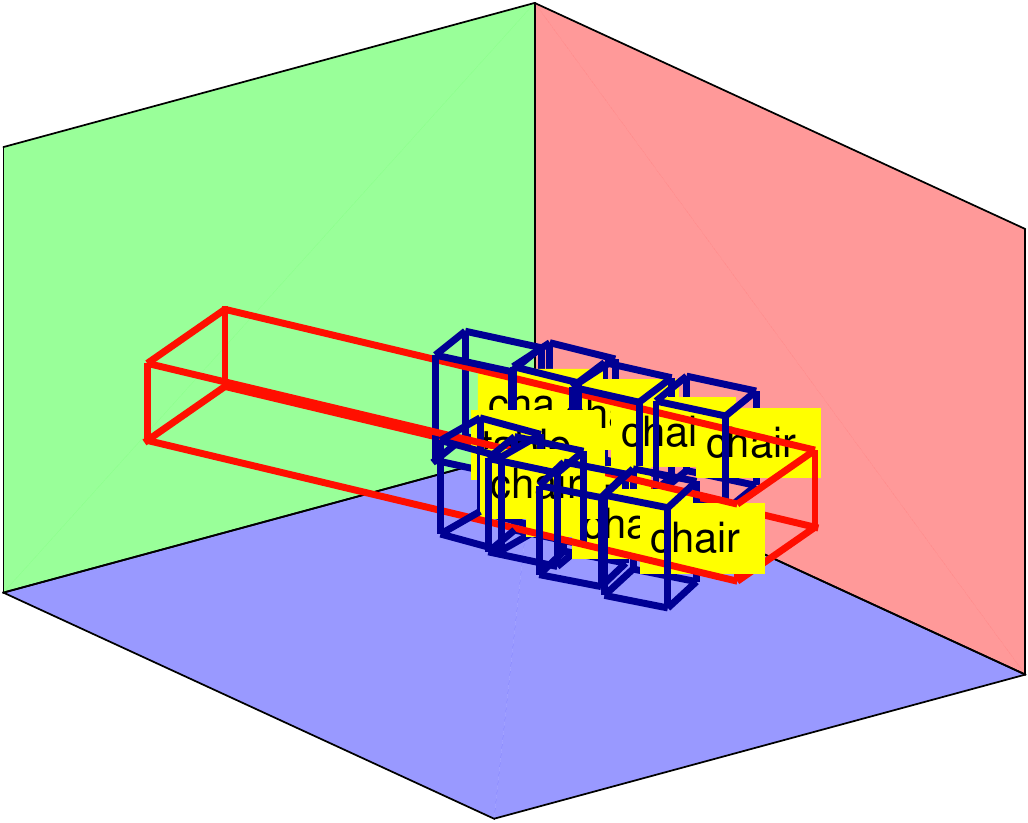}  & 
\includegraphics[width=0.154\textwidth, height=1.4cm ]{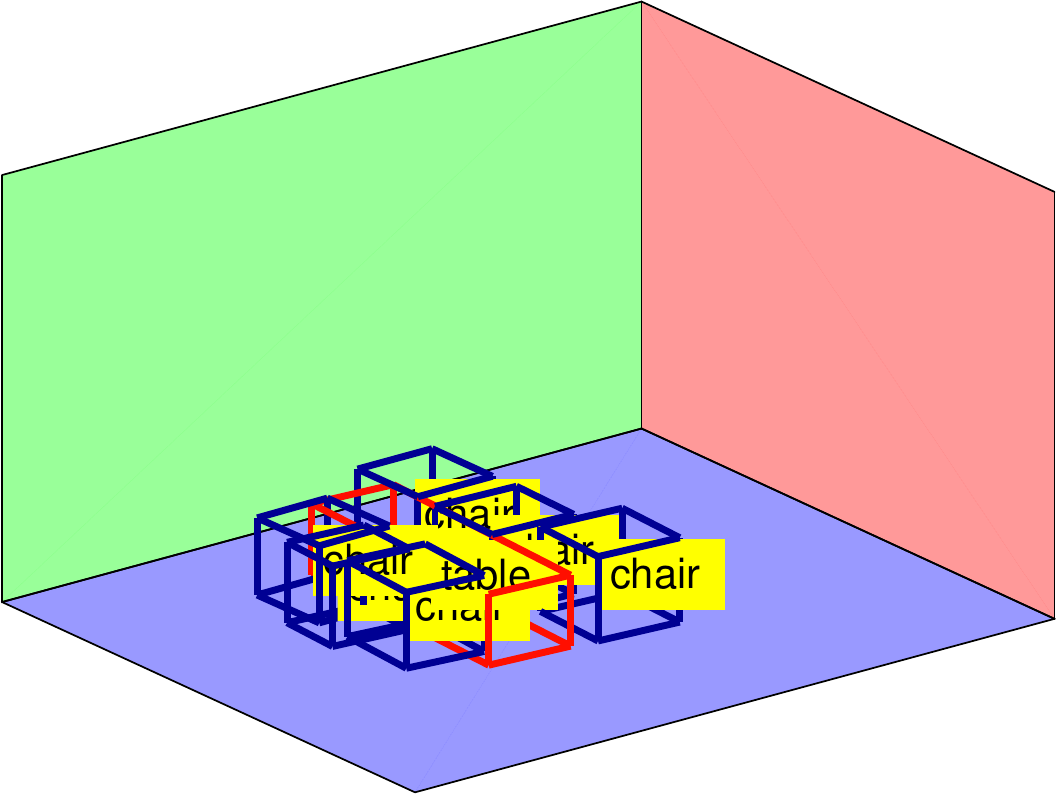}  & \\ 

  \begin{picture}(1,25)\put(0, 5){\rotatebox{90}{($\alpha = 0.6$)}}\end{picture} & 
  \includegraphics[width=0.154\textwidth, height=1.4cm ]{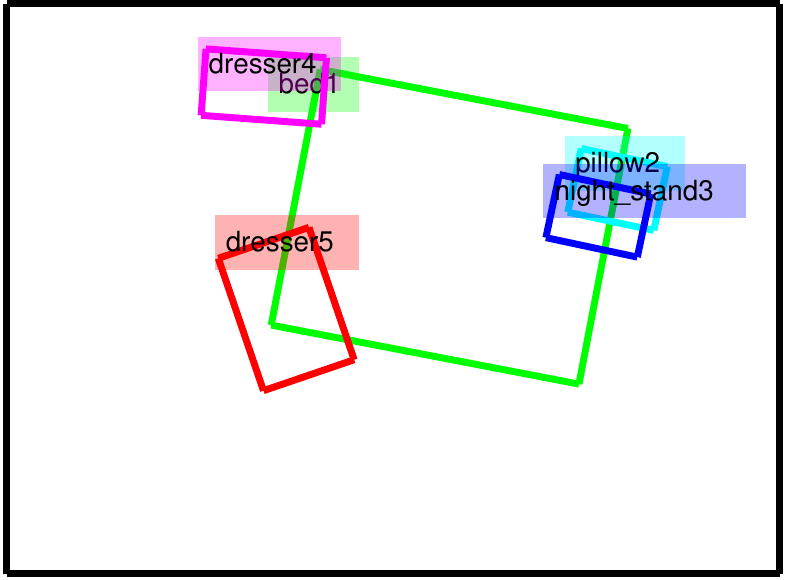}  & 
\includegraphics[width=0.154\textwidth, height=1.4cm ]{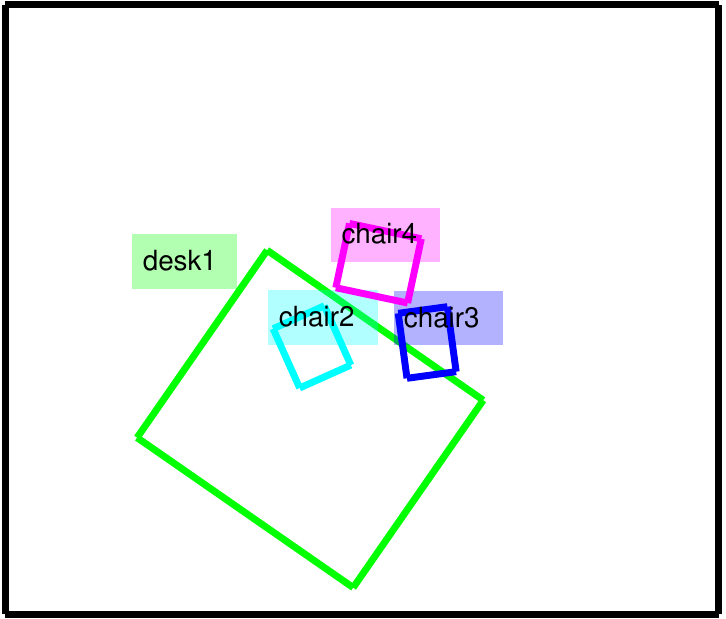}  & 
\includegraphics[width=0.154\textwidth, height=1.4cm ]{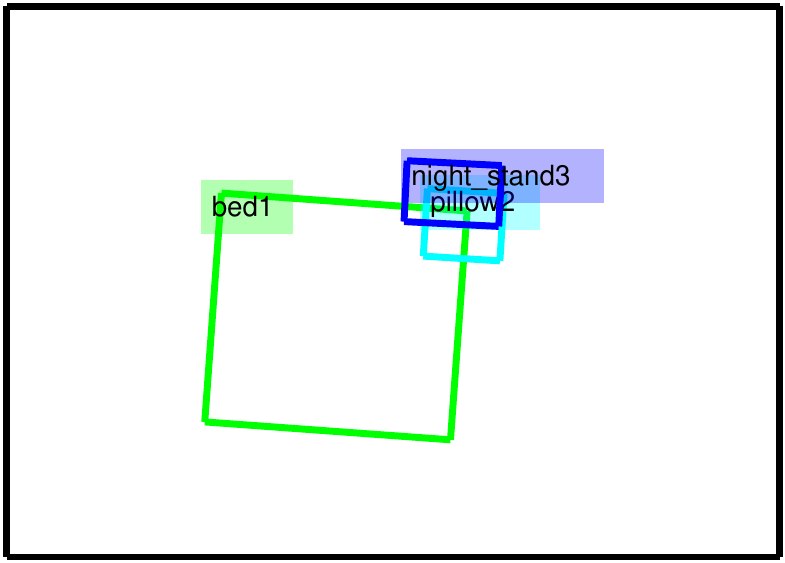}  & 
\includegraphics[width=0.154\textwidth, height=1.4cm ]{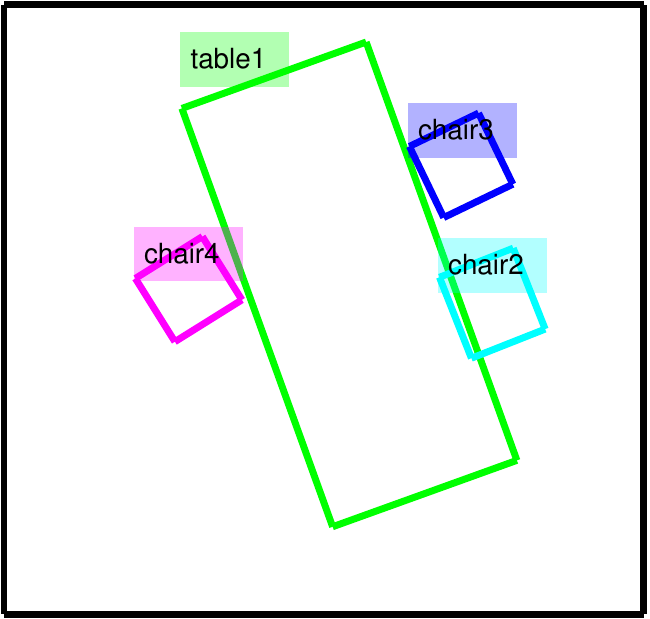}  & 
\includegraphics[width=0.154\textwidth, height=1.4cm ]{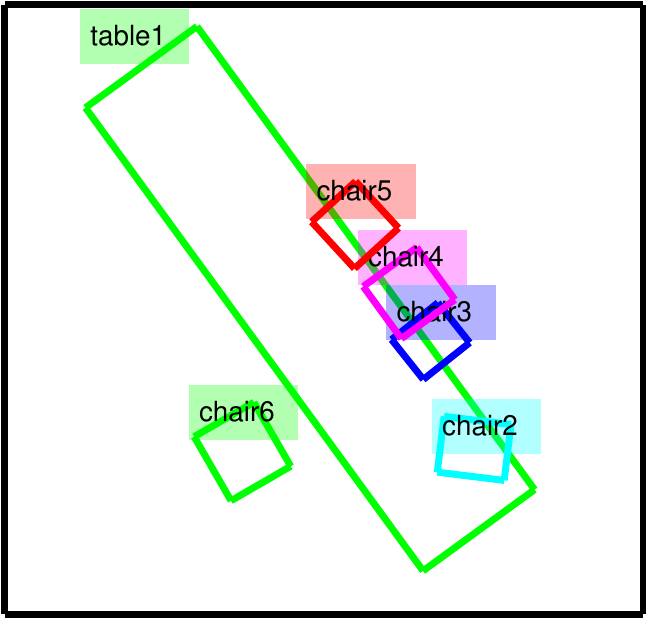}  & 
\includegraphics[width=0.154\textwidth, height=1.4cm ]{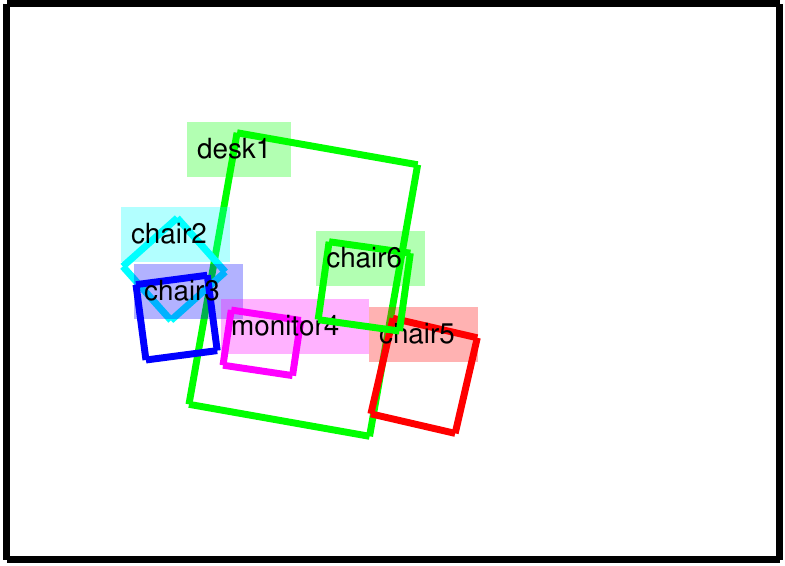} & 
  \begin{picture}(1,25)\put(0, 5){\rotatebox{90}{Top View}}\end{picture} \\ 

  \begin{picture}(1,25)\put(0, 5){\rotatebox{90}{($\alpha = 0.6$)}}\end{picture} & 
  \includegraphics[width=0.154\textwidth, height=1.4cm ]{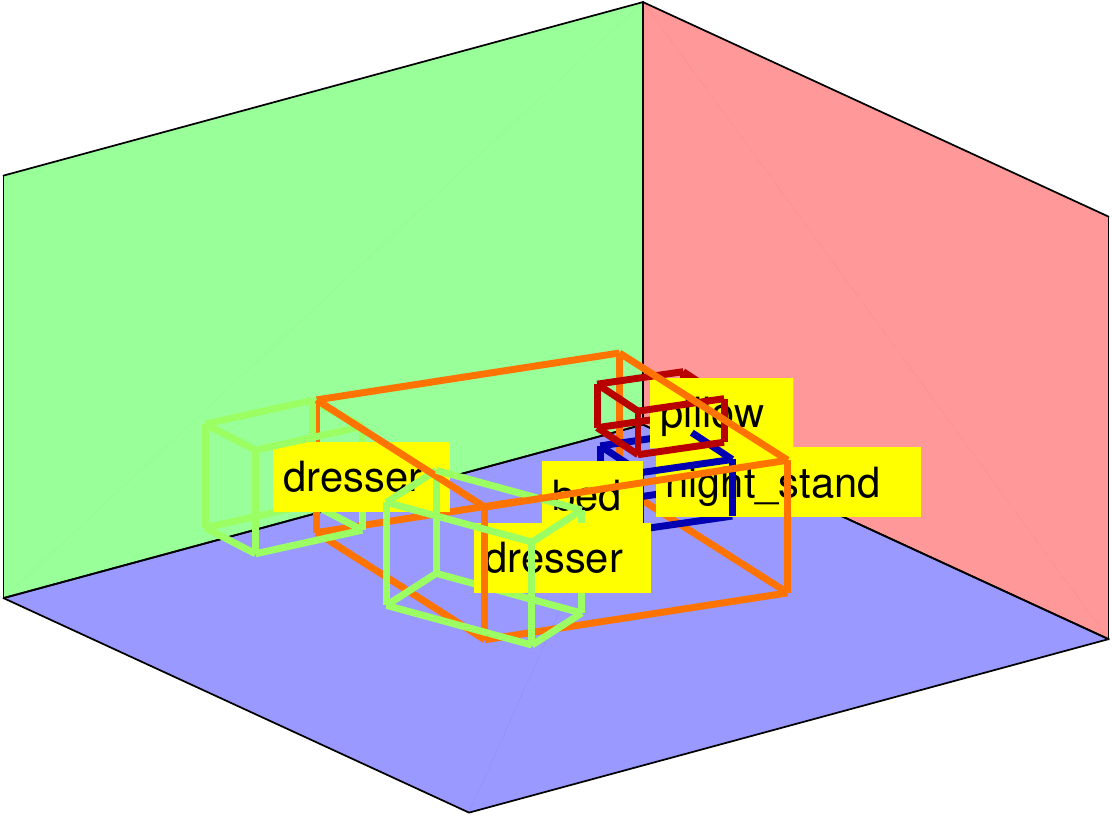}  & 
\includegraphics[width=0.154\textwidth, height=1.4cm ]{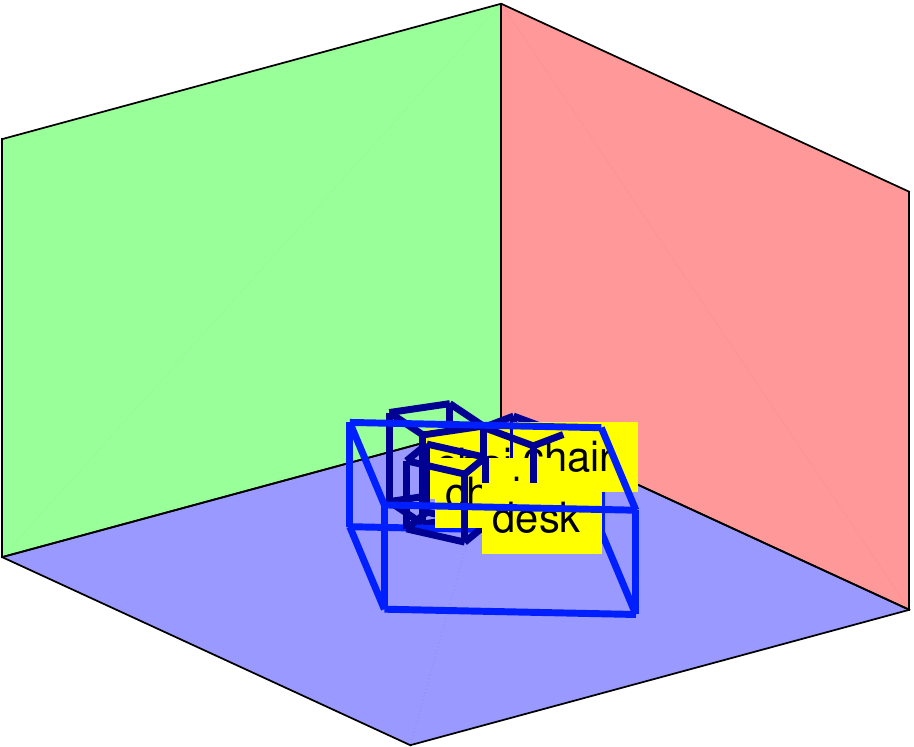}  & 
\includegraphics[width=0.154\textwidth, height=1.4cm ]{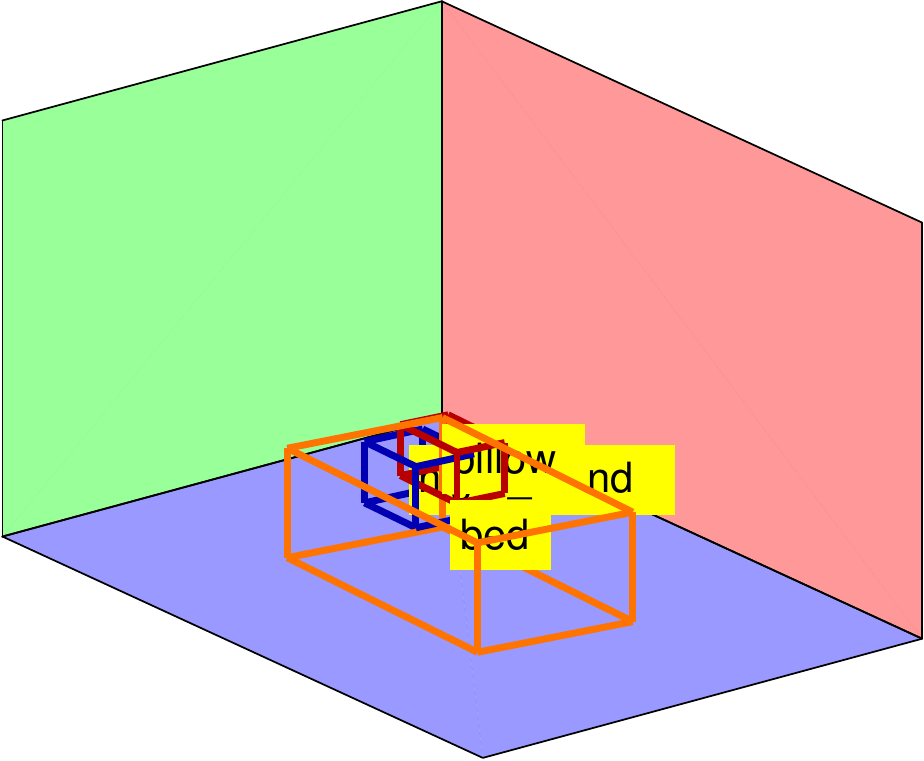}  & 
\includegraphics[width=0.154\textwidth, height=1.4cm ]{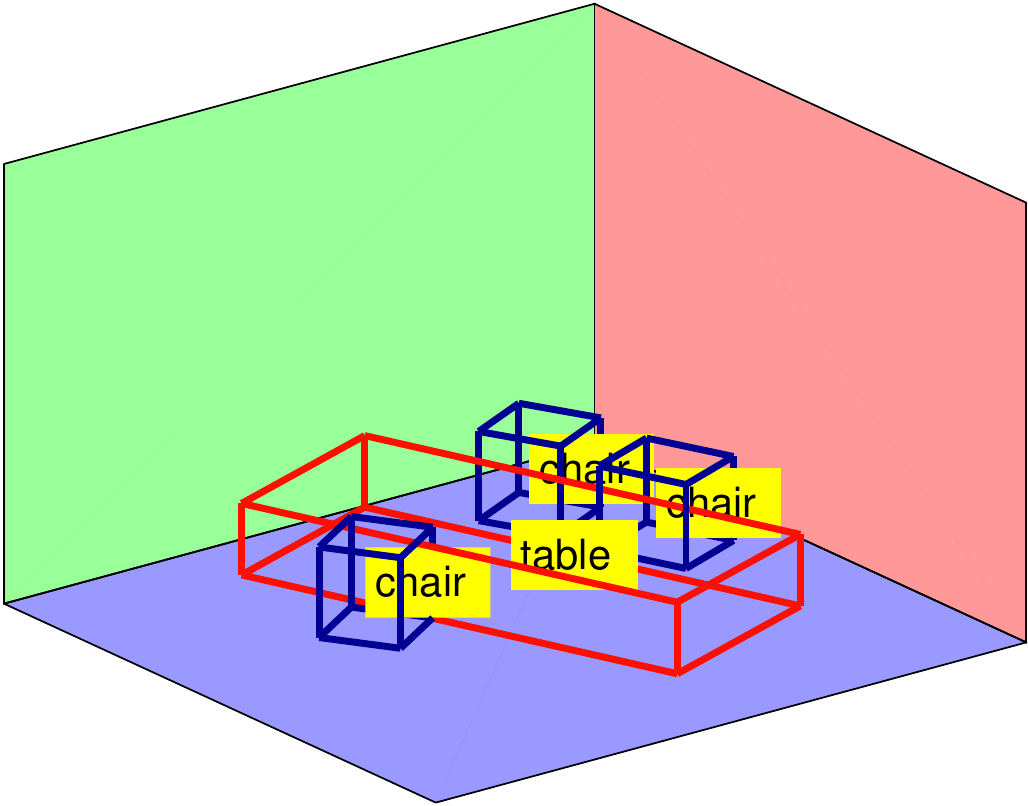}  & 
\includegraphics[width=0.154\textwidth, height=1.4cm ]{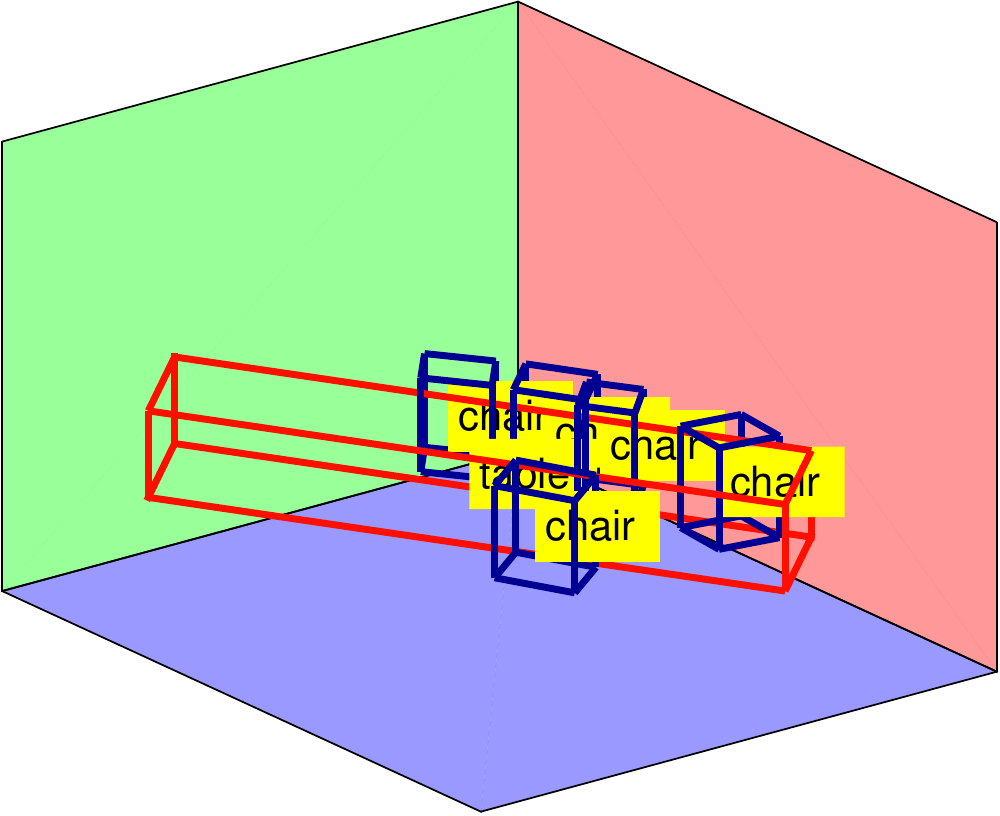}  & 
\includegraphics[width=0.154\textwidth, height=1.4cm ]{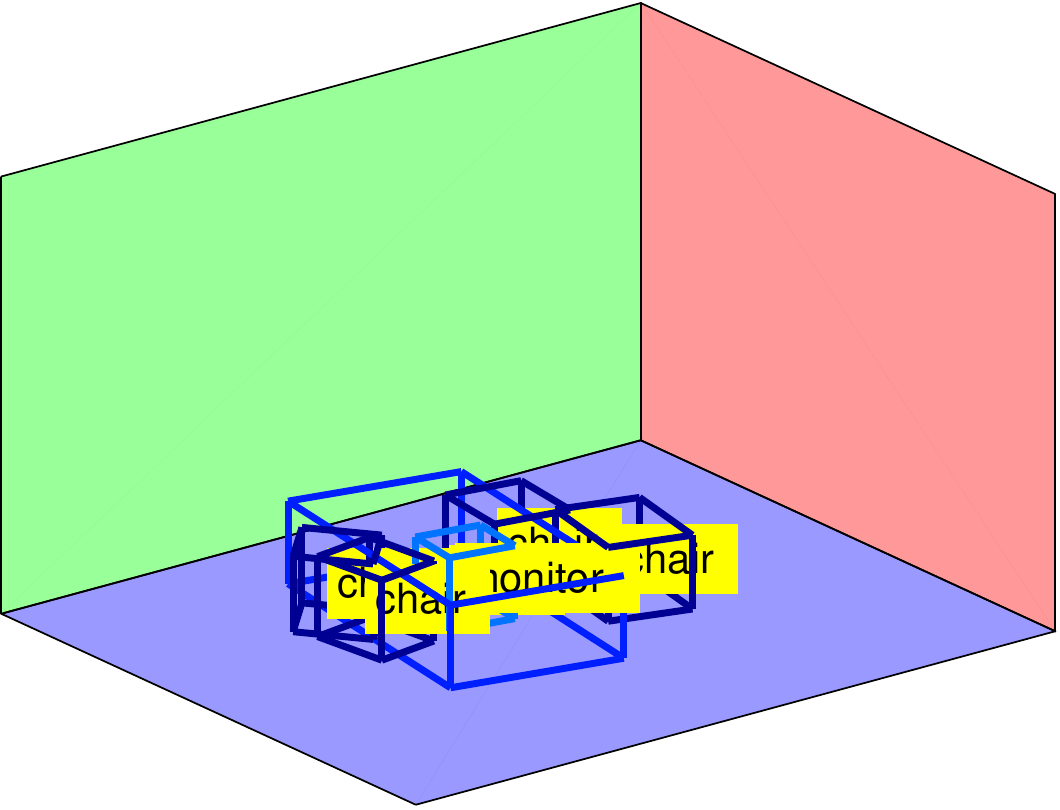} & \\ 
%\includegraphics[width=0.154\textwidth, height=1.4cm ]{slerp2d/pred_test1850-crop.pdf}  \\ 
%
%
%  \begin{picture}(1,25)\put(0, 5){\rotatebox{90}{($\alpha = 0.5$)}}\end{picture} & 
%  \includegraphics[width=0.154\textwidth, height=1.4cm ]{slerp2d/pred_test_2D_51-crop.pdf}  & 
%\includegraphics[width=0.154\textwidth, height=1.4cm ]{slerp2d/pred_test_2D_969-crop.pdf}  & 
%\includegraphics[width=0.154\textwidth, height=1.4cm ]{slerp2d/pred_test_2D_284-crop.pdf}  & 
%\includegraphics[width=0.154\textwidth, height=1.4cm ]{slerp2d/pred_test_2D_842-crop.pdf}  & 
%\includegraphics[width=0.154\textwidth, height=1.4cm ]{slerp2d/pred_test_2D_1202-crop.pdf}  & 
%\includegraphics[width=0.154\textwidth, height=1.4cm ]{slerp2d/pred_test_2D_87-crop.pdf}  \\ 
%%\includegraphics[width=0.154\textwidth, height=1.4cm ]{slerp2d/pred_test_2D_1849-crop.pdf}  \\ 
%
%  \begin{picture}(1,25)\put(0, 5){\rotatebox{90}{($\alpha = 0.5$)}}\end{picture} & 
%  \includegraphics[width=0.154\textwidth, height=1.4cm ]{slerp2d/pred_test_2D_51-crop.pdf}  & 
%\includegraphics[width=0.154\textwidth, height=1.4cm ]{slerp2d/pred_test_2D_969-crop.pdf}  & 
%\includegraphics[width=0.154\textwidth, height=1.4cm ]{slerp2d/pred_test_2D_284-crop.pdf}  & 
%\includegraphics[width=0.154\textwidth, height=1.4cm ]{slerp2d/pred_test_2D_842-crop.pdf}  & 
%\includegraphics[width=0.154\textwidth, height=1.4cm ]{slerp2d/pred_test_2D_1202-crop.pdf}  & 
%\includegraphics[width=0.154\textwidth, height=1.4cm ]{slerp2d/pred_test_2D_87-crop.pdf}  \\ 
%%\includegraphics[width=0.154\textwidth, height=1.4cm ]{slerp2d/pred_test_2D_1849-crop.pdf}  \\ 

  \begin{picture}(1,25)\put(0, 5){\rotatebox{90}{($\alpha = 0.4$)}}\end{picture} & 
  \includegraphics[width=0.154\textwidth, height=1.4cm ]{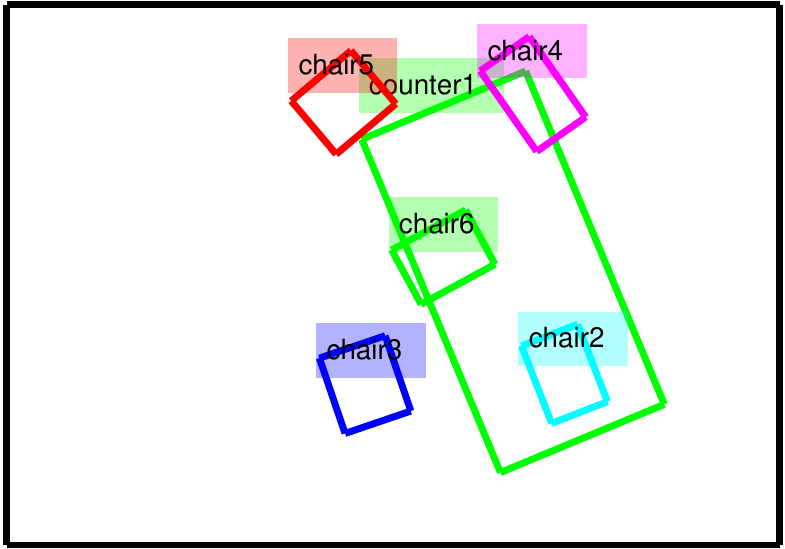}  & 
\includegraphics[width=0.154\textwidth, height=1.4cm ]{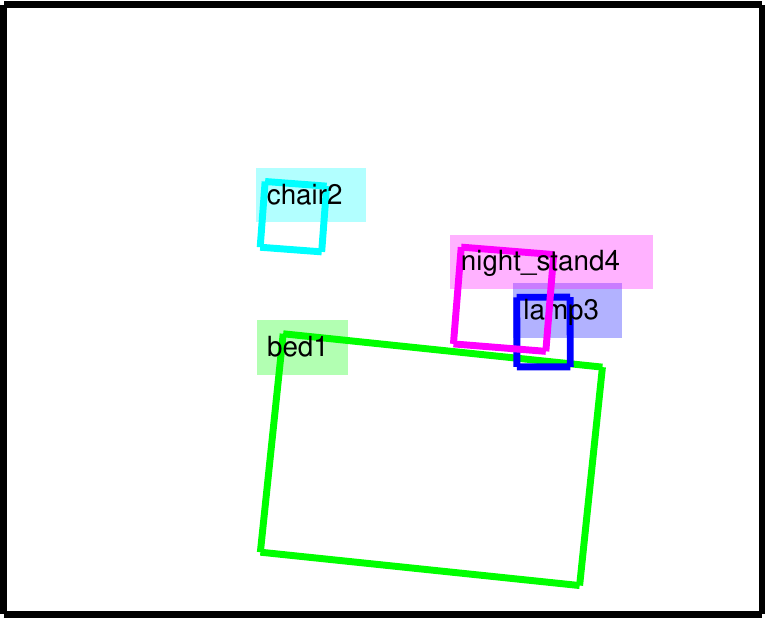}  & 
\includegraphics[width=0.154\textwidth, height=1.4cm ]{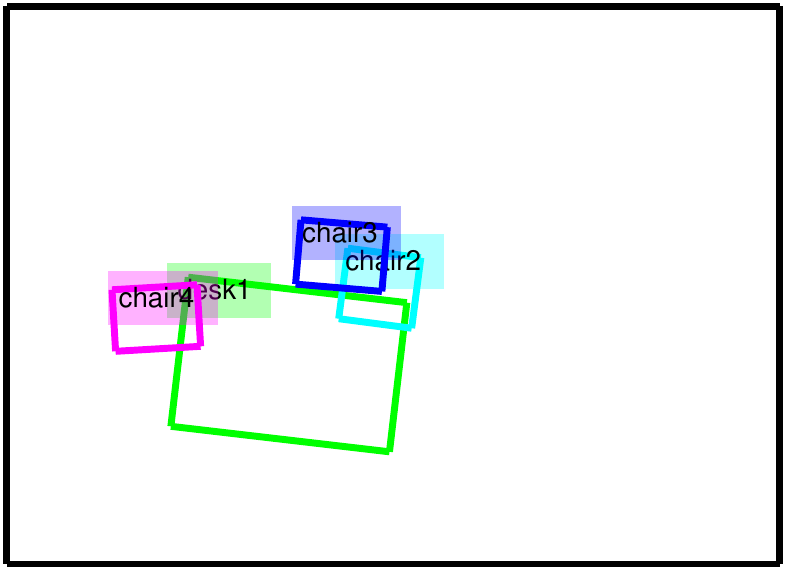}  & 
\includegraphics[width=0.154\textwidth, height=1.4cm ]{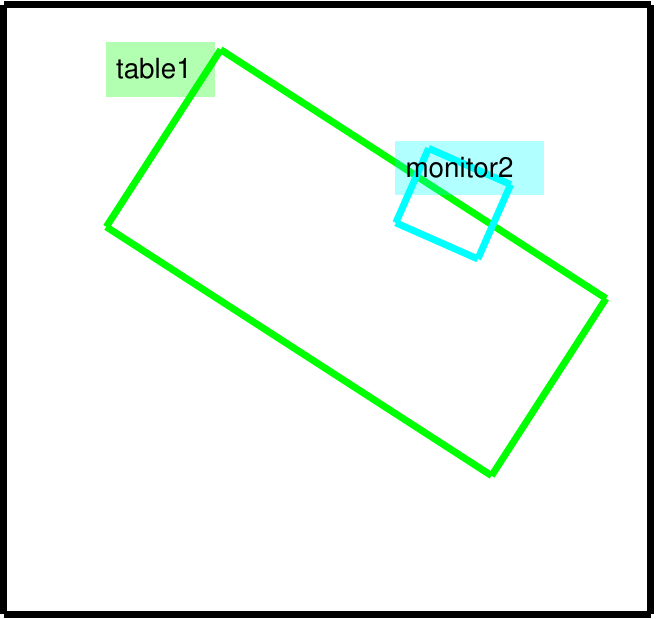}  & 
\includegraphics[width=0.154\textwidth, height=1.4cm ]{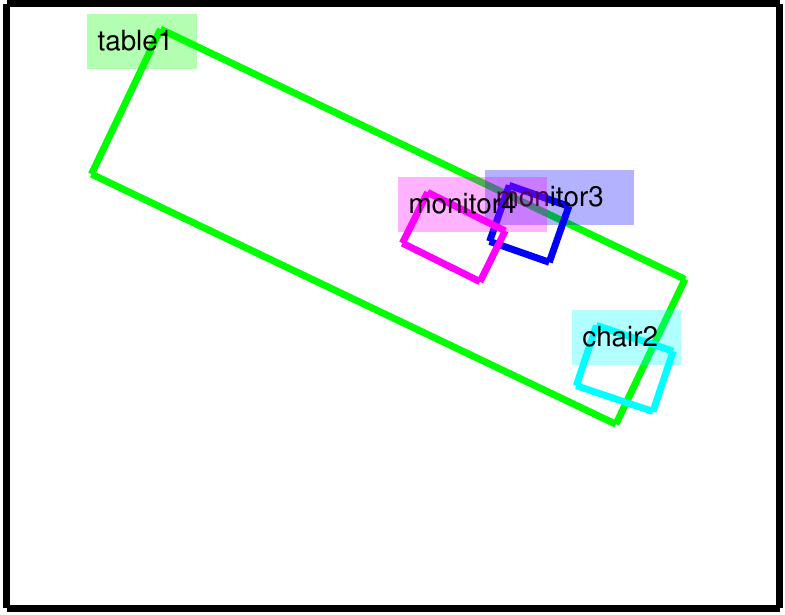}  & 
\includegraphics[width=0.154\textwidth, height=1.4cm ]{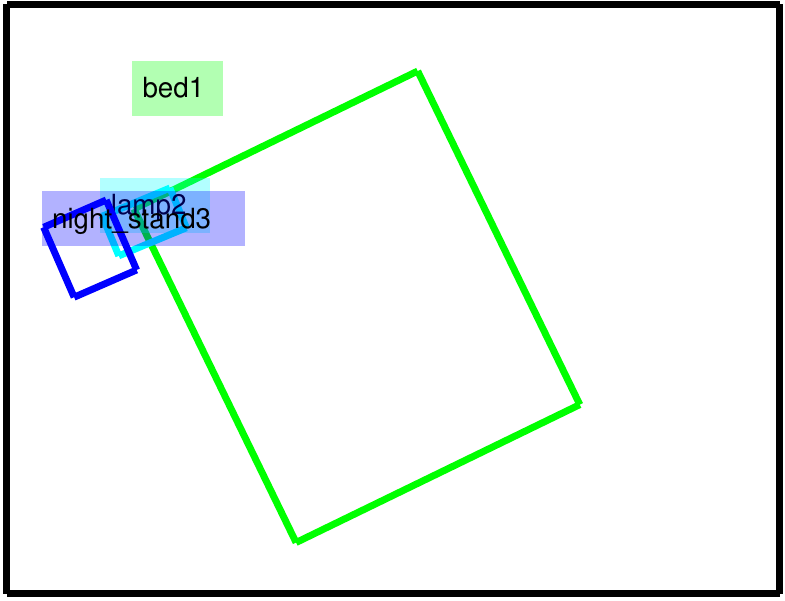} & 
  \begin{picture}(1,25)\put(0, 5){\rotatebox{90}{Top View}}\end{picture} \\ 

  \begin{picture}(1,25)\put(0, 5){\rotatebox{90}{($\alpha = 0.4$)}}\end{picture} & 
  \includegraphics[width=0.154\textwidth, height=1.4cm ]{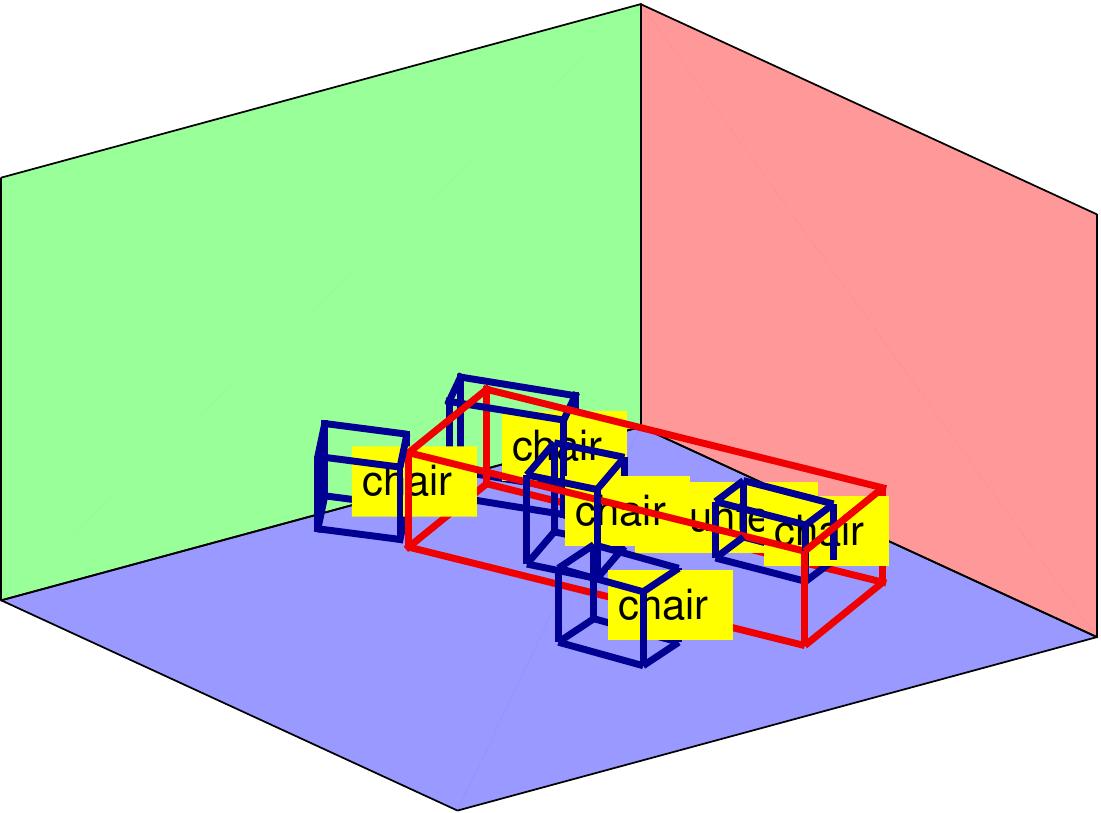}  & 
\includegraphics[width=0.154\textwidth, height=1.4cm ]{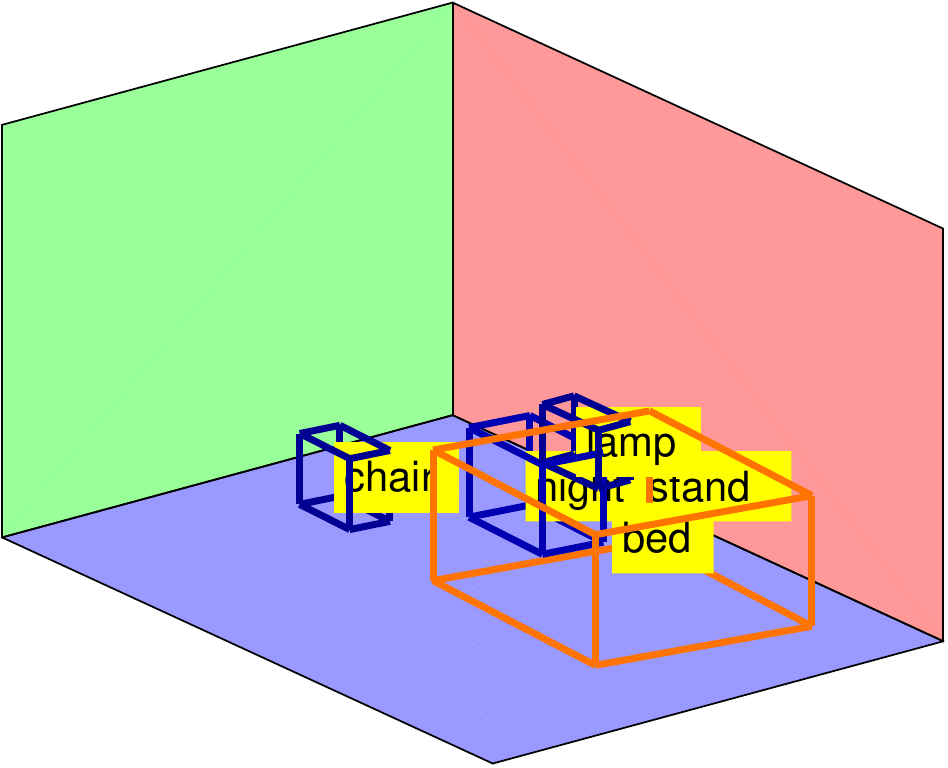}  & 
\includegraphics[width=0.154\textwidth, height=1.4cm ]{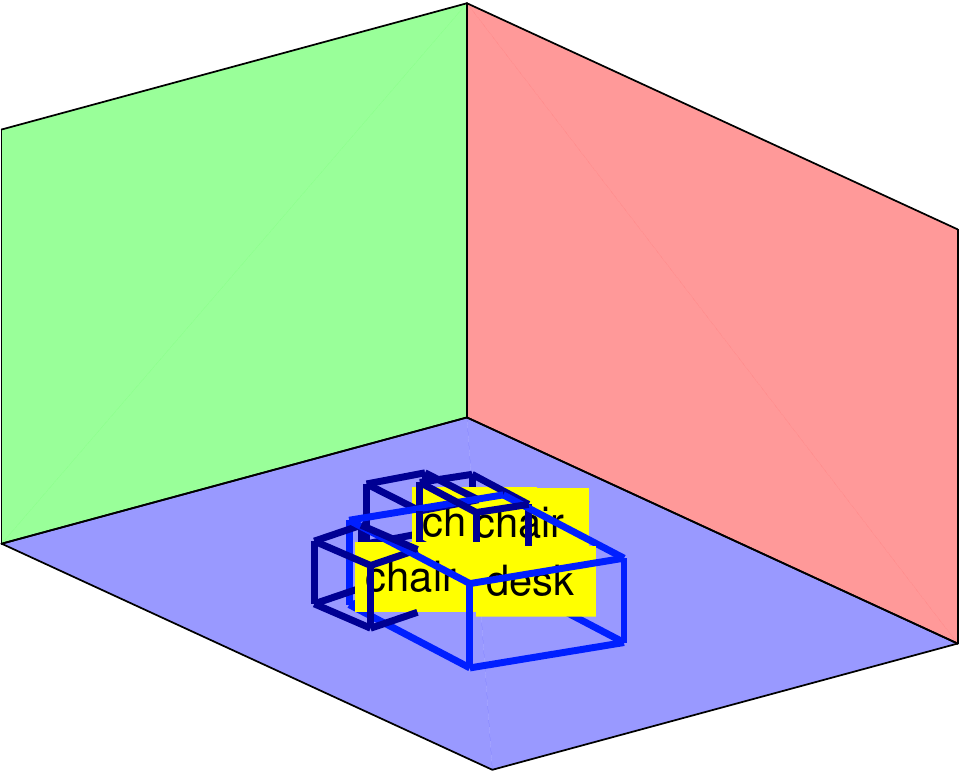}  & 
\includegraphics[width=0.154\textwidth, height=1.4cm ]{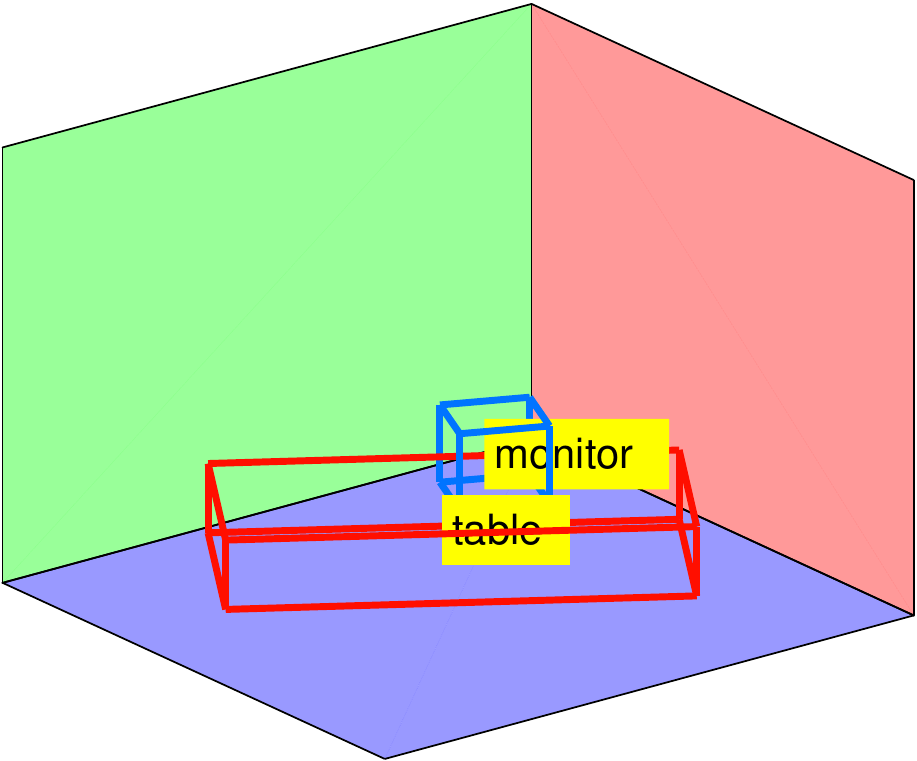}  & 
\includegraphics[width=0.154\textwidth, height=1.4cm ]{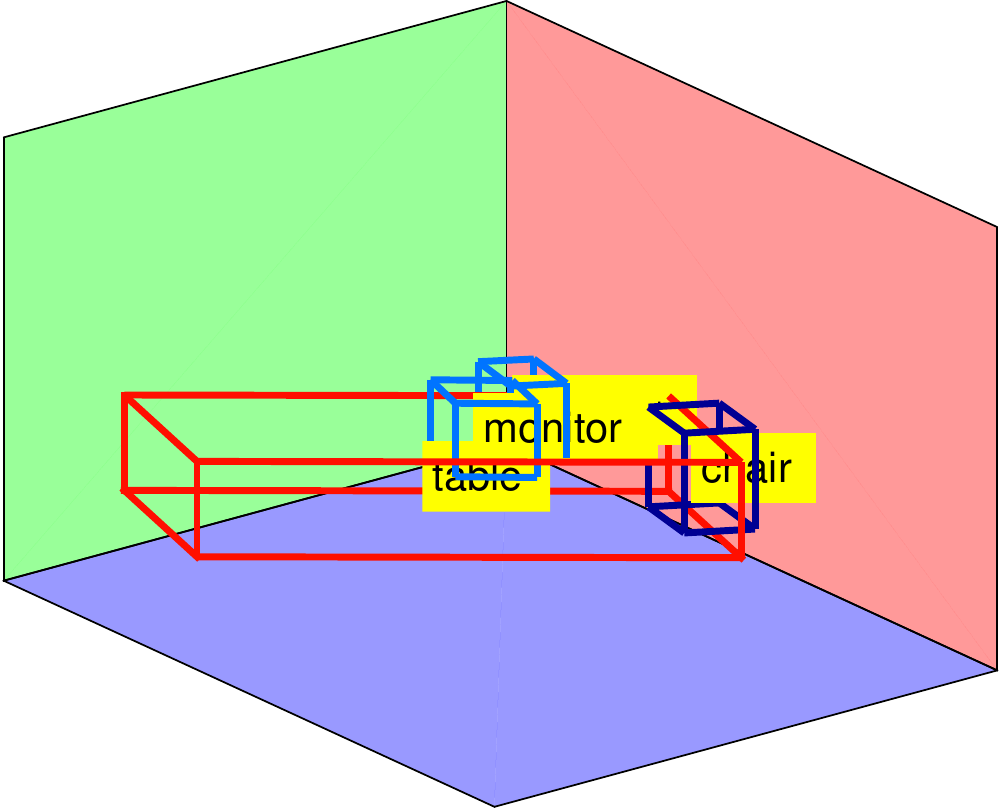}  & 
\includegraphics[width=0.154\textwidth, height=1.4cm ]{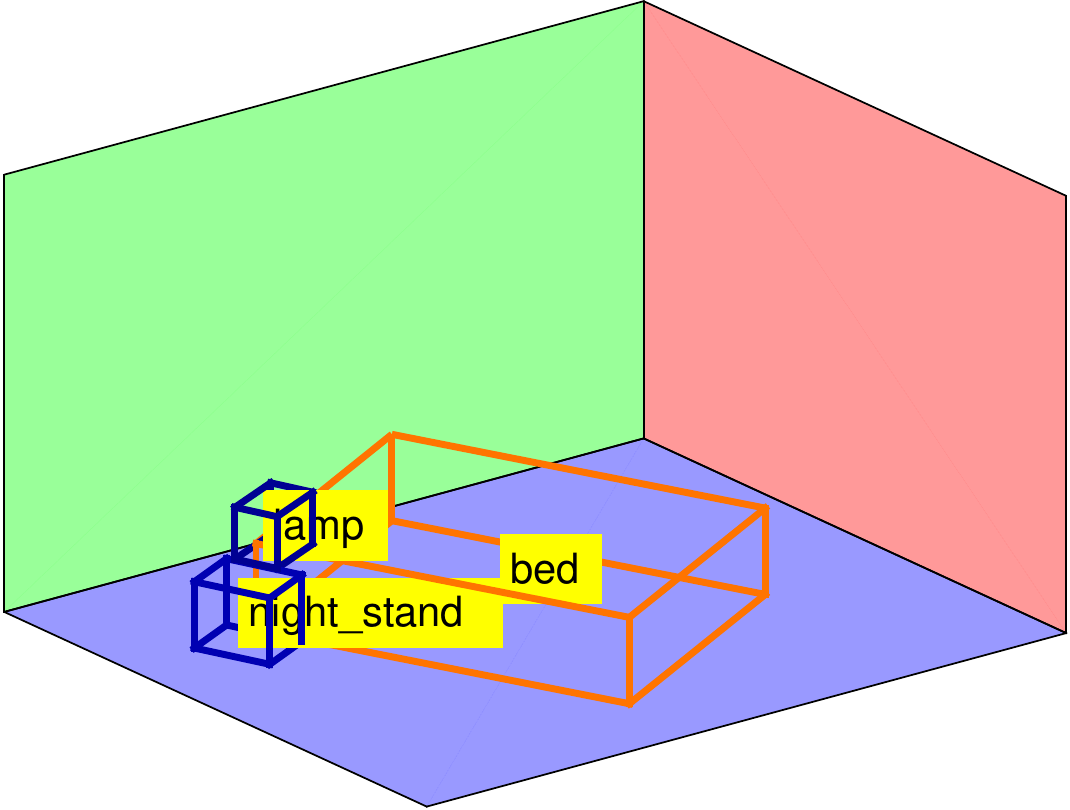}  & \\ 

  \begin{picture}(1,25)\put(0, 5){\rotatebox{90}{($\alpha = 0.1$)}}\end{picture} & 
\includegraphics[width=0.154\textwidth, height=1.4cm ]{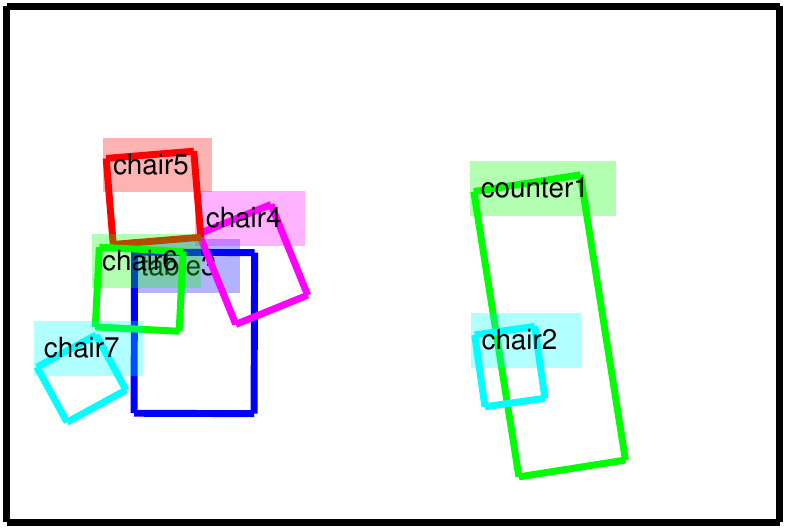}  & 
\includegraphics[width=0.154\textwidth, height=1.4cm ]{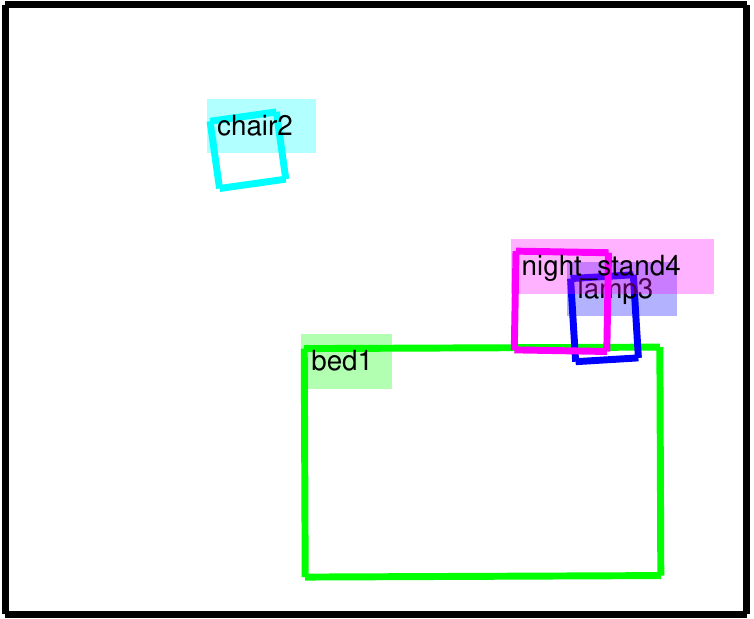}  & 
\includegraphics[width=0.154\textwidth, height=1.4cm ]{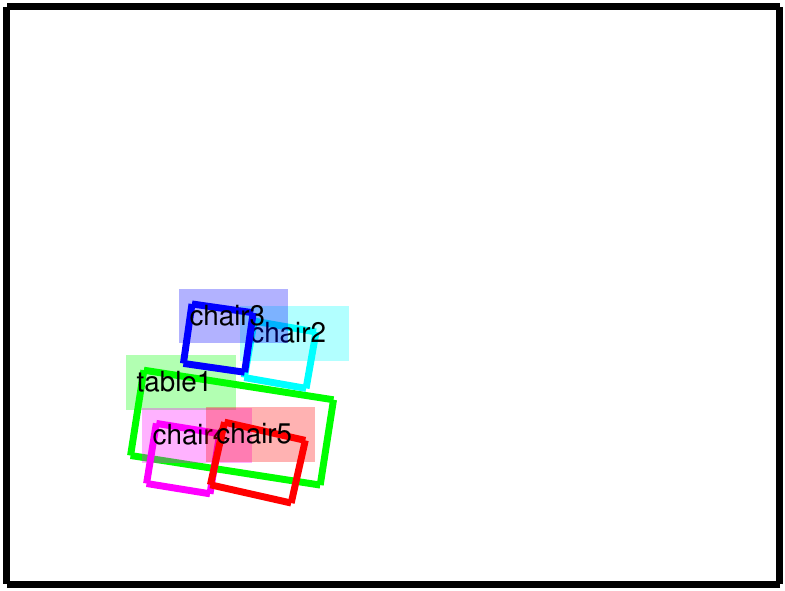}  & 
\includegraphics[width=0.154\textwidth, height=1.4cm ]{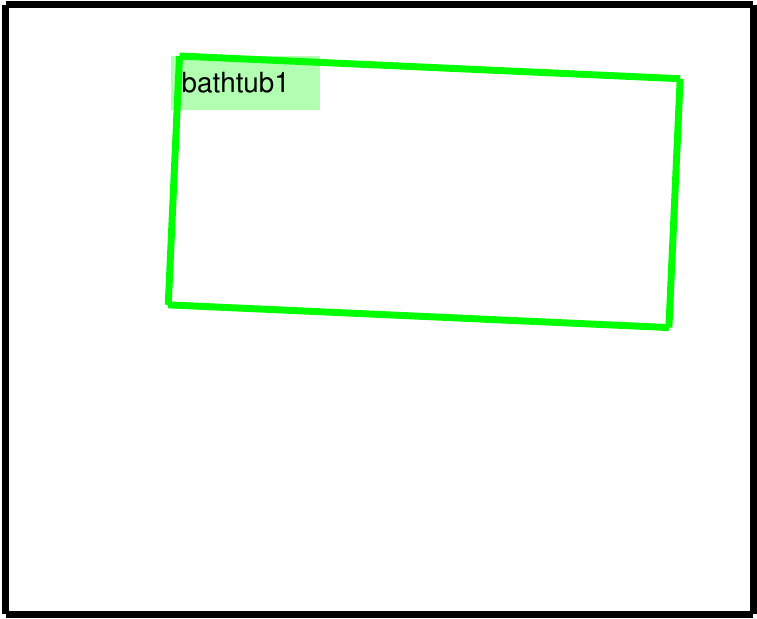}  & 
\includegraphics[width=0.154\textwidth, height=1.4cm, trim={0 0 8 0},clip]{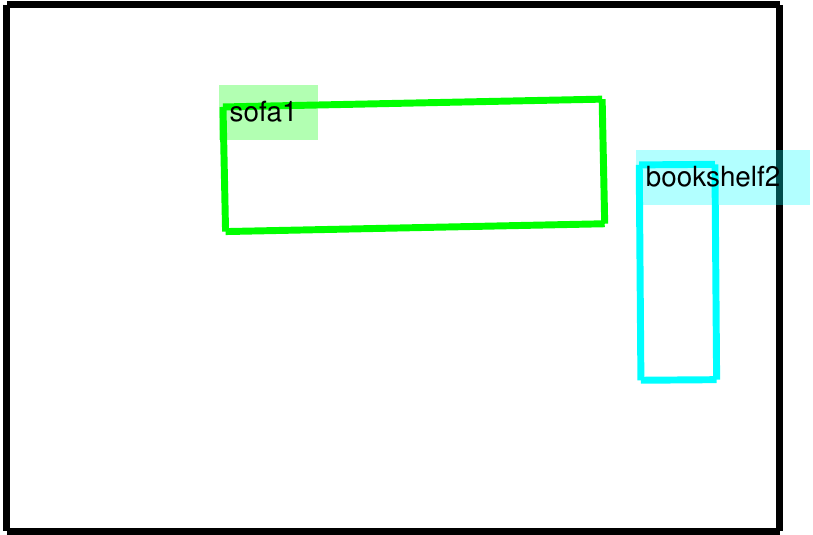}  & 
\includegraphics[width=0.154\textwidth, height=1.4cm ]{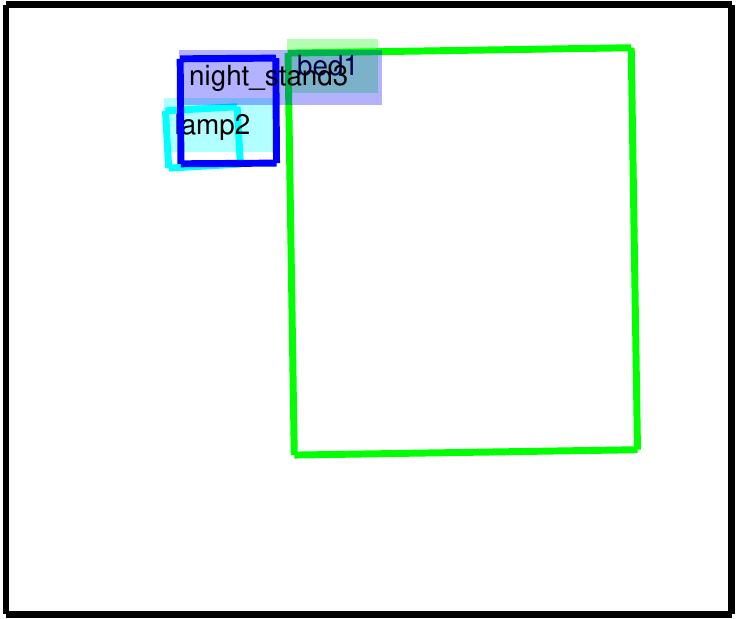} & 
  \begin{picture}(1,25)\put(0, 5){\rotatebox{90}{Top View}}\end{picture} \\ 

  \begin{picture}(1,25)\put(0, 5){\rotatebox{90}{($\alpha = 0.1$)}}\end{picture} & 
\includegraphics[width=0.154\textwidth, height=1.4cm ]{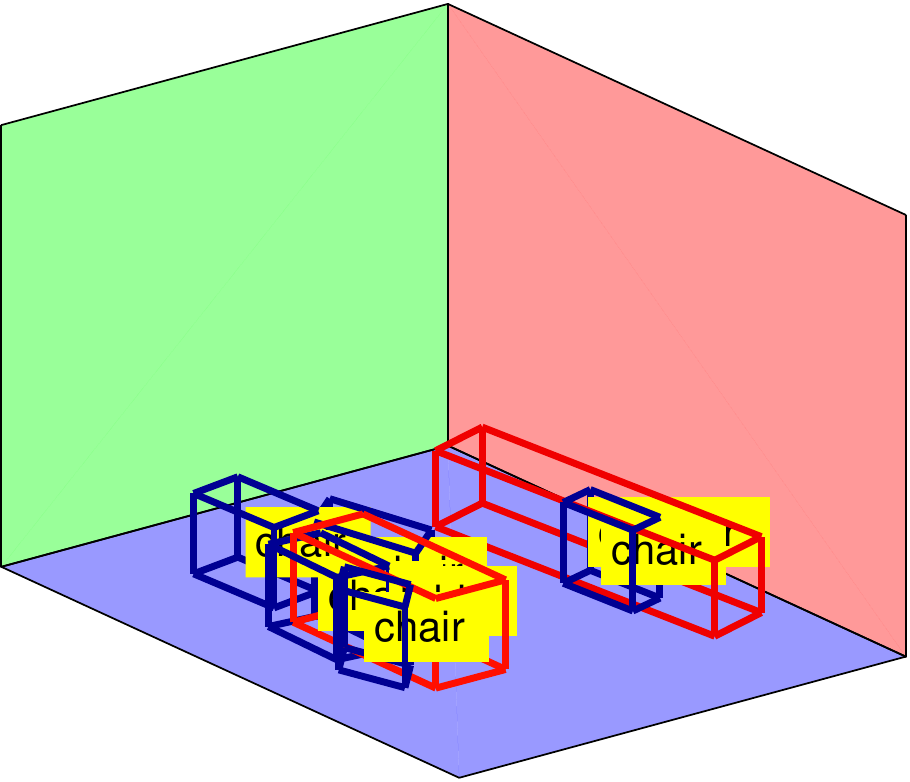}  & 
\includegraphics[width=0.154\textwidth, height=1.4cm ]{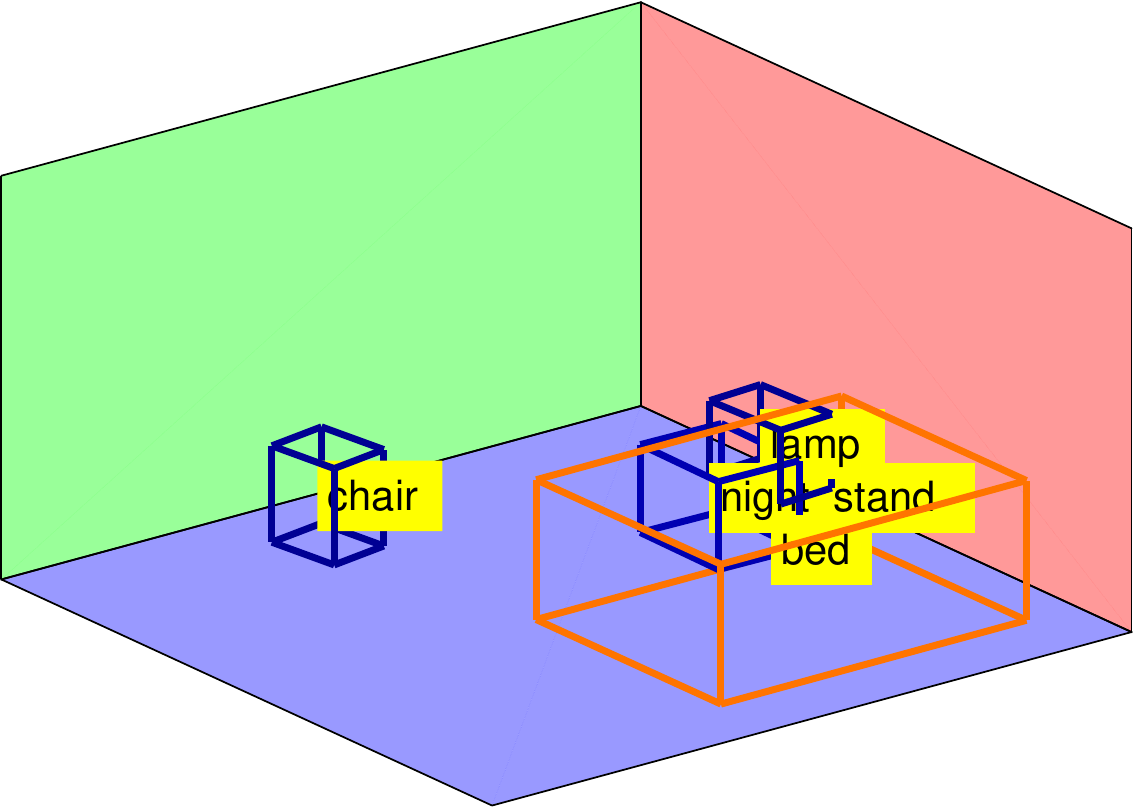}  & 
\includegraphics[width=0.154\textwidth, height=1.4cm ]{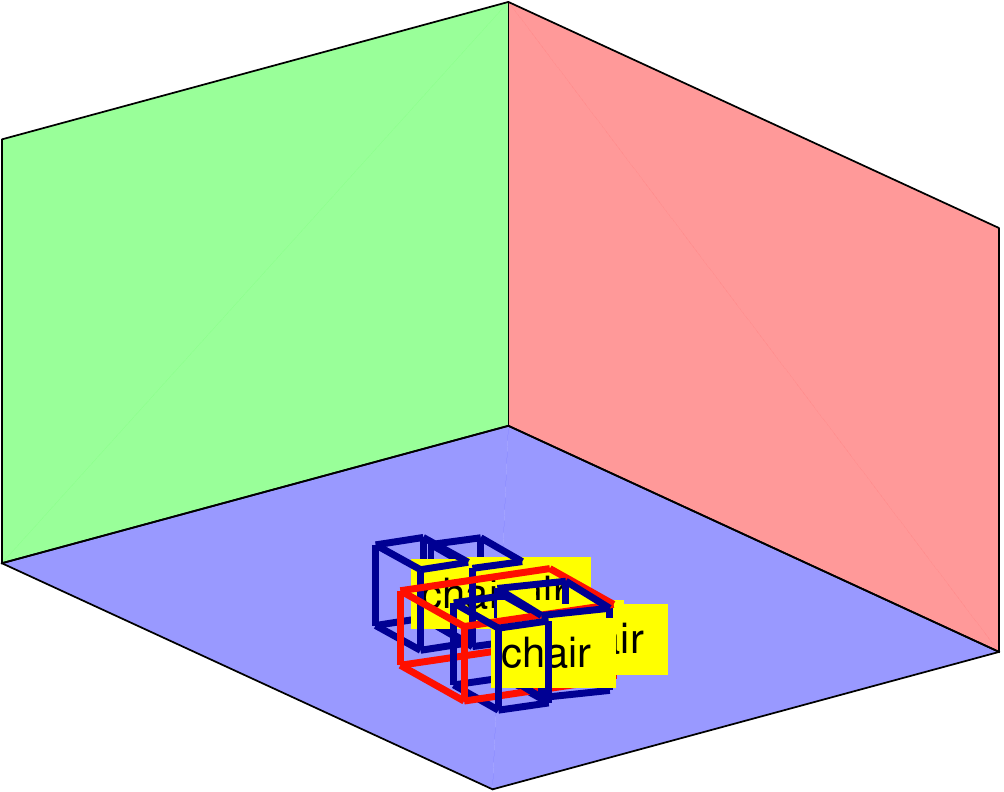}  & 
\includegraphics[width=0.154\textwidth, height=1.4cm ]{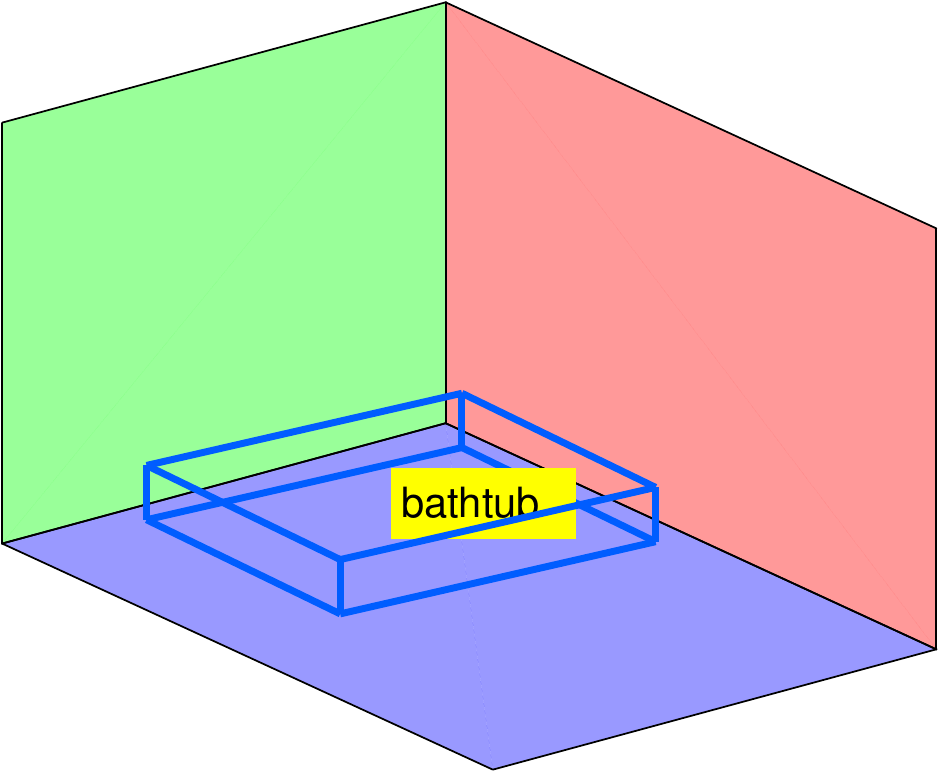}  & 
\includegraphics[width=0.154\textwidth, height=1.4cm, trim={0 0 8 0},clip]{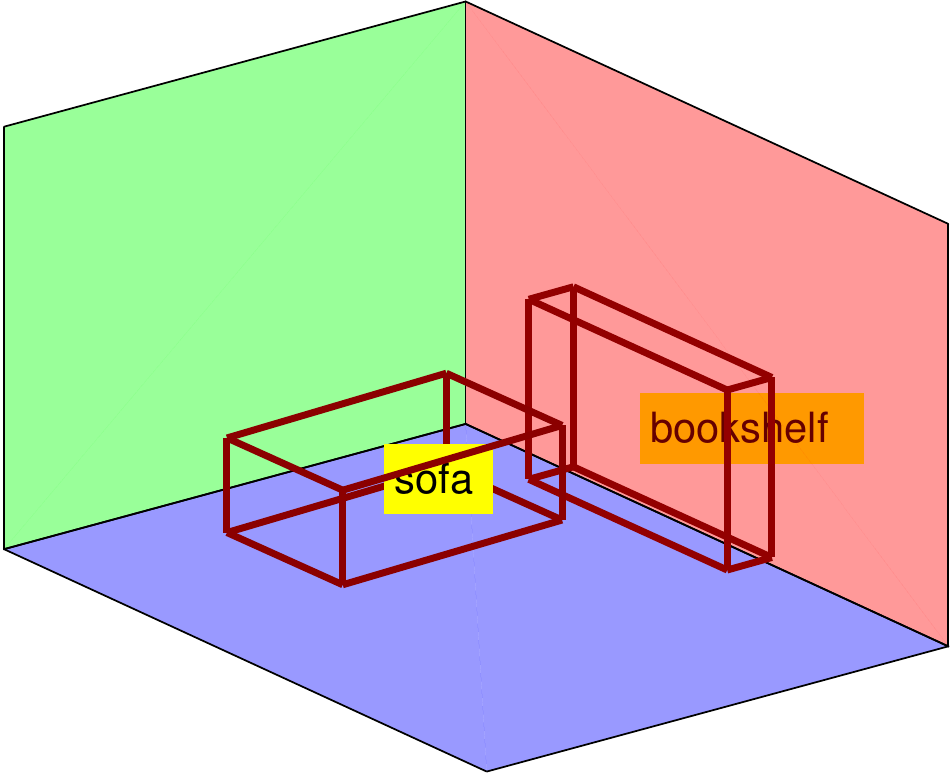}  & 
\includegraphics[width=0.154\textwidth, height=1.4cm ]{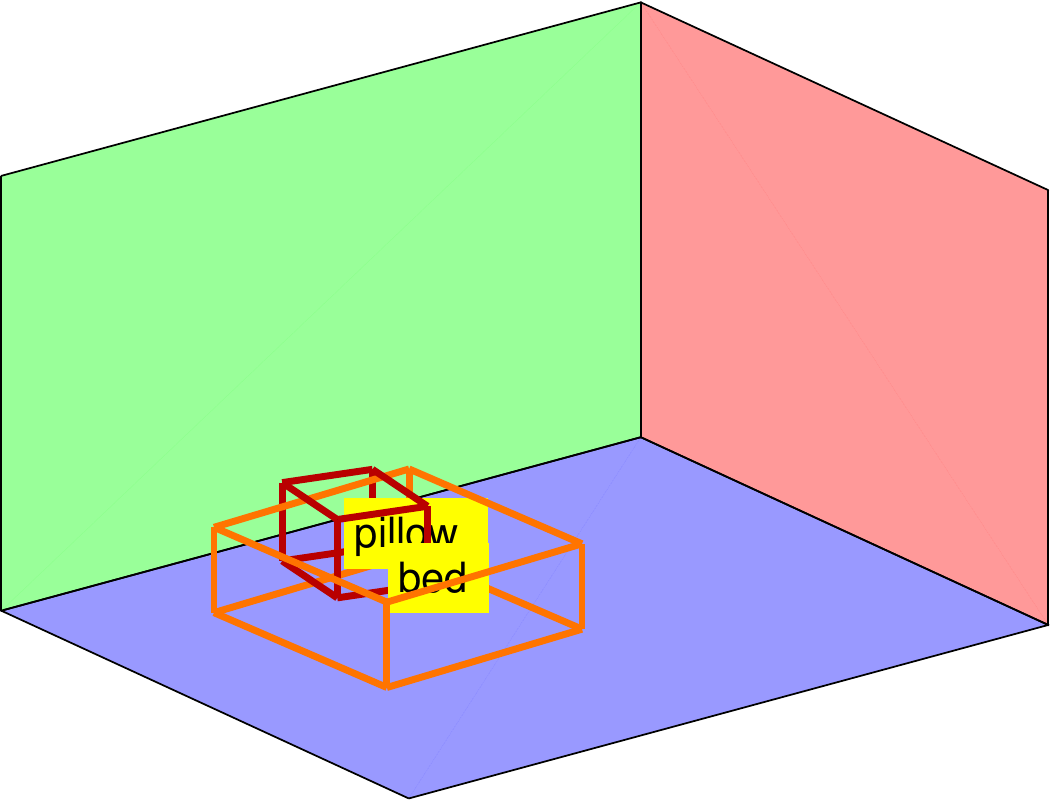}  & \\
& Interpolation 1 & Interpolation 2 & Interpolation 3 & Interpolation 4 & Interpolation 5 & Interpolation 6  
\end{tabular}
\caption{Synthetic scenes decoded from linear interpolations $\alpha \bm{\mu_1} + (1 - \alpha)\bm{\mu_2}$ of the means $\bm{\mu_1}$ and $\bm{\mu_2}$ of the latent distributions of two separate scenes. The generated scenes are valid in terms of the co-occurrences of the object categories and their shapes and poses (more examples can be found in the supplementary). The room-size and the camera view-point are fixed for better visualization. Best viewed electronically. }
\label{fig:qualititive} %\vspace{-2em}
\end{figure}
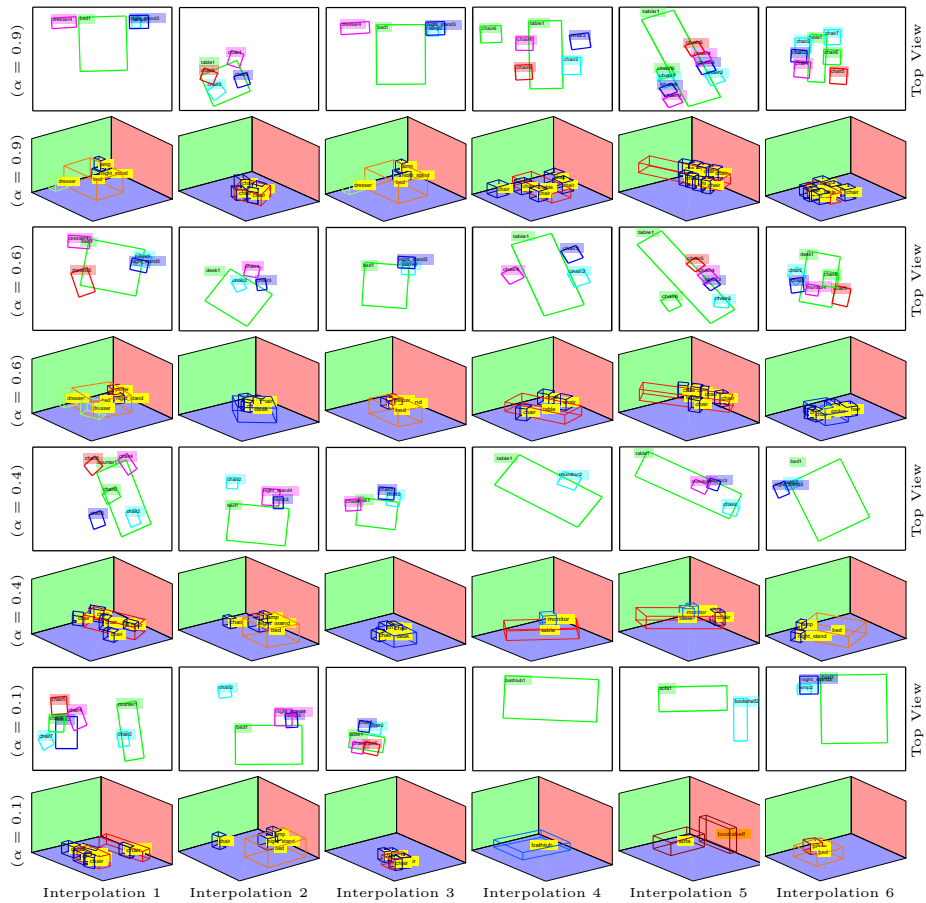 
\noindent {\bf Interpolation in latent space}
Two distinct scenes are encoded into the latent space, \eg.,  $\cN(\bm{\mu_1}, \bm{\Sigma_1})$ and $\cN(\bm{\mu_2}, \bm{\Sigma_2})$ and new scenes are then synthesized from interpolated vectors of the means, \ie~from $\alpha \bm{\mu_1} + (1 - \alpha)\bm{\mu_2}$. 
We performed the experiment on a set of random pairs chosen from the test dataset. The results are shown in Figure~\ref{fig:qualititive}. Notice that the decoder behaves gracefully w.r.t.\ perturbations of the latent code and always yields a valid and realistic scene. %\vspace{-0.5em}

\begin{figure}[H]
\centering \scriptsize
\begin{tabular}{l@{\hspace{1.01em}}c@{\hspace{0.3em}}c@{\hspace{0.2em}}@{\hspace{0.2em}}c@{\hspace{0.2em}}@{\hspace{0.2em}}c@{\hspace{0.2em}}@{\hspace{0.2em}}c@{\hspace{0.2em}}} 
%& SUN RGB-D~\cite{song2015sun} & \multicolumn{4}{c}{SUNCG~\cite{song2017semantic}} \\
%  \cmidrule[0.06em](r){2-2}\cmidrule[0.06em](lr){3-6} 
 \begin{picture}(1,25)\put(0, 8){\rotatebox{90}{Bed Room}}\end{picture} &
 \includegraphics[width=0.17\textwidth, height=1.7cm, trim={1.5 1.5 2 2},clip]{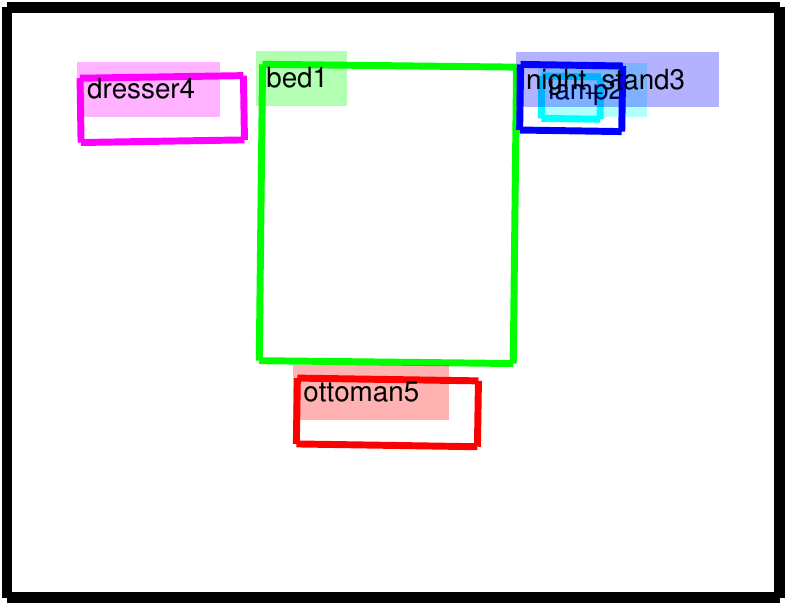} & 
\includegraphics[width=0.17\textwidth, height=1.7cm, trim={1 1 21.5 1},clip]{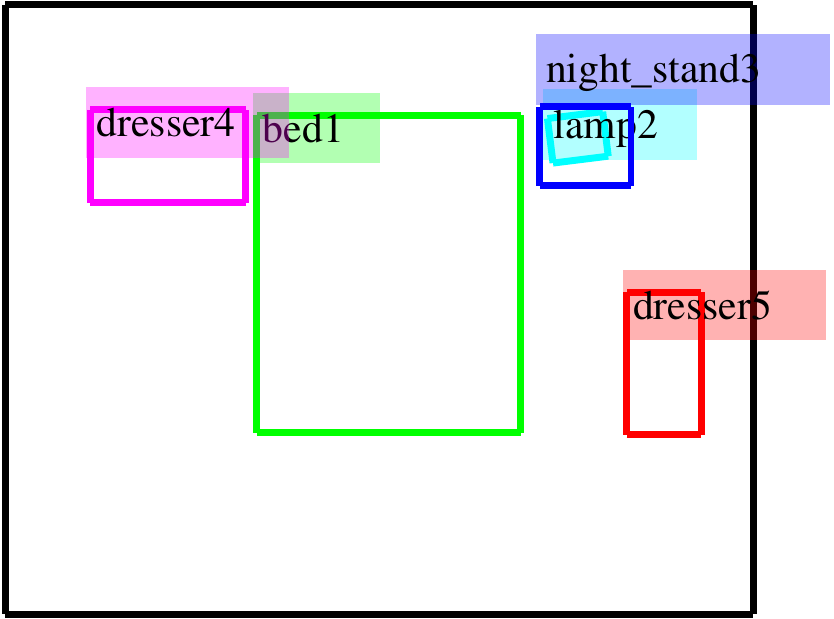} & 
\includegraphics[width=0.17\textwidth, height=1.7cm, trim={1 1 1 1},clip]{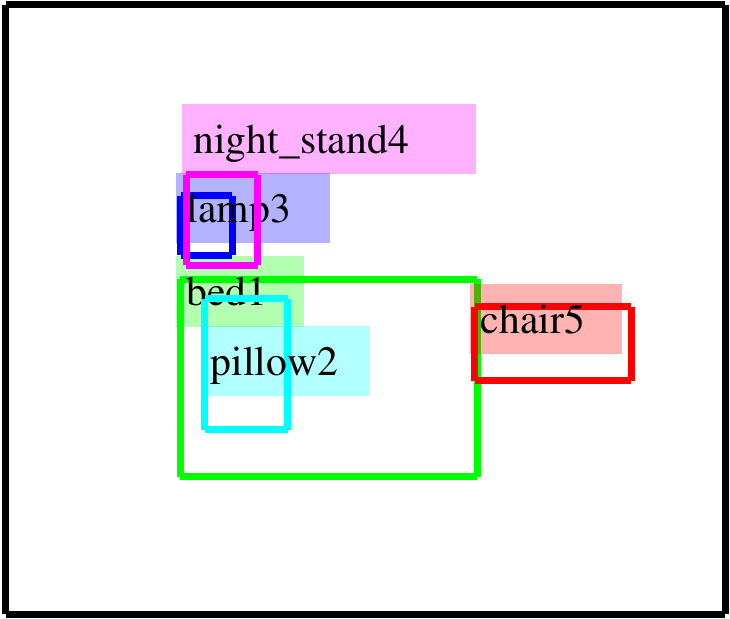} & 
\includegraphics[width=0.17\textwidth, height=1.7cm, trim={1 1 32 1},clip]{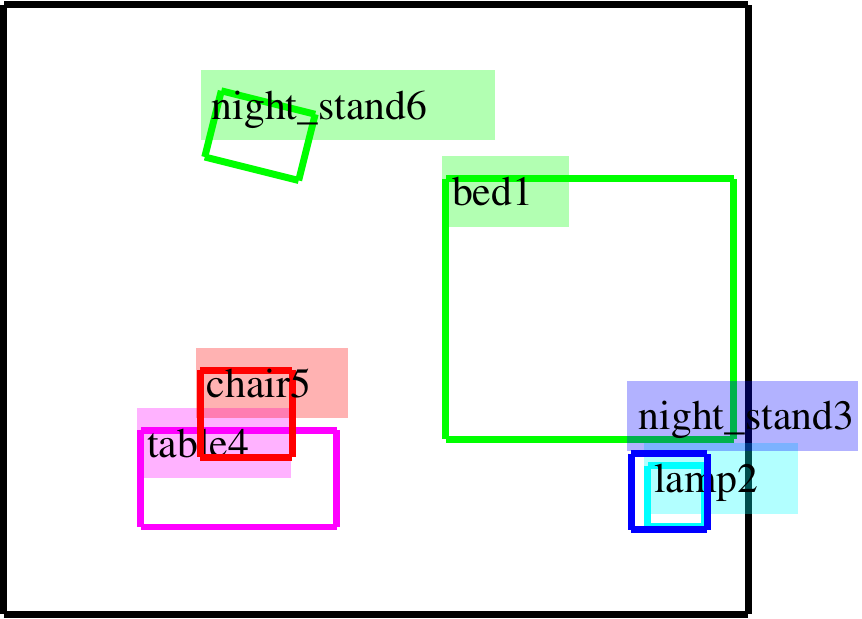} & 
\includegraphics[width=0.17\textwidth, height=1.7cm, trim={1 1 1 1},clip]{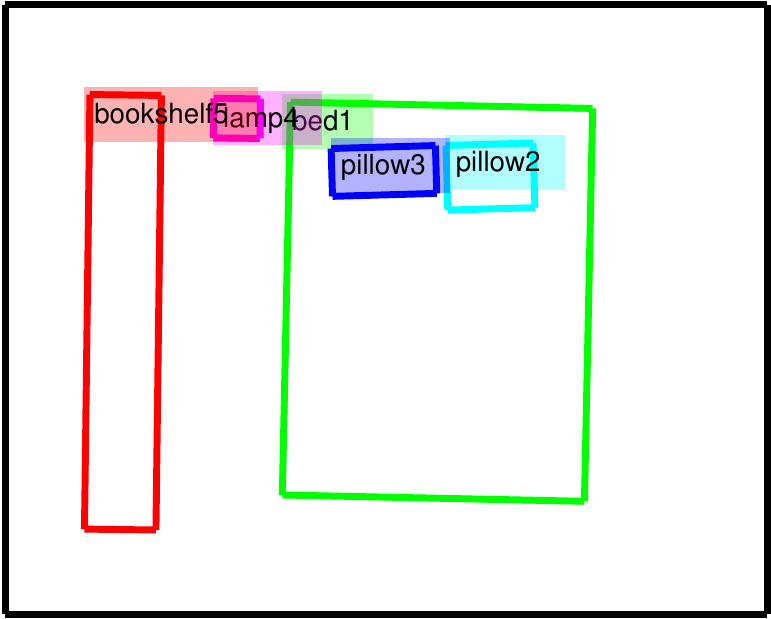} \\ %& \includegraphics[width=0.17\textwidth, height=1.7cm, trim={1.5 2 25 1.5},clip]{2dproj/pred_16-crop.pdf} \\
% \begin{picture}(1,25)\put(0, 8){\rotatebox{90}{Bed Room 2}}\end{picture} &
% \includegraphics[width=0.17\textwidth, height=1.7cm, trim={1 1 18 1},clip]{2dproj/org_45-crop.pdf} & 
%\includegraphics[width=0.17\textwidth, height=1.7cm, trim={1 1 4.5 1},clip]{2dproj/19-crop.pdf} & 
%\includegraphics[width=0.17\textwidth, height=1.7cm, trim={1 1 32.5 1},clip]{2dproj/fast_synth1-crop.pdf} & 
%\includegraphics[width=0.17\textwidth, height=1.7cm, trim={1 1 36 1},clip]{2dproj/sample_0000000061-crop.pdf} \\
 \begin{picture}(1,25)\put(0, 14){\rotatebox{90}{Office}}\end{picture} &
\includegraphics[width=0.17\textwidth, height=1.7cm, trim={1.5 1 1 1},clip]{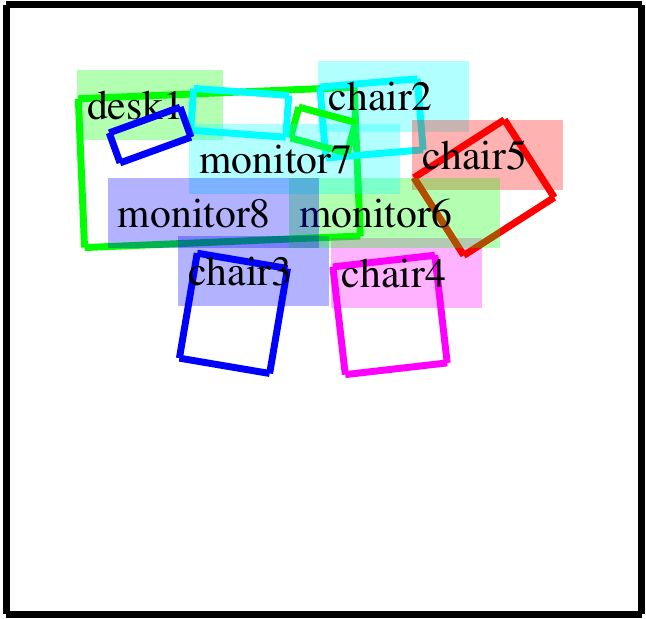} & 
\includegraphics[width=0.17\textwidth, height=1.7cm, trim={1 1 1 1},clip]{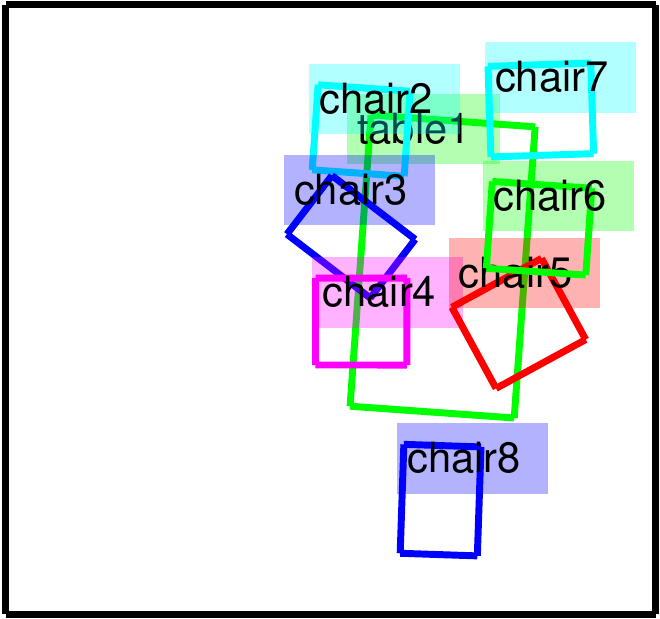} & 
\includegraphics[width=0.17\textwidth, height=1.7cm, trim={1.5 1 1.0 1.5},clip]{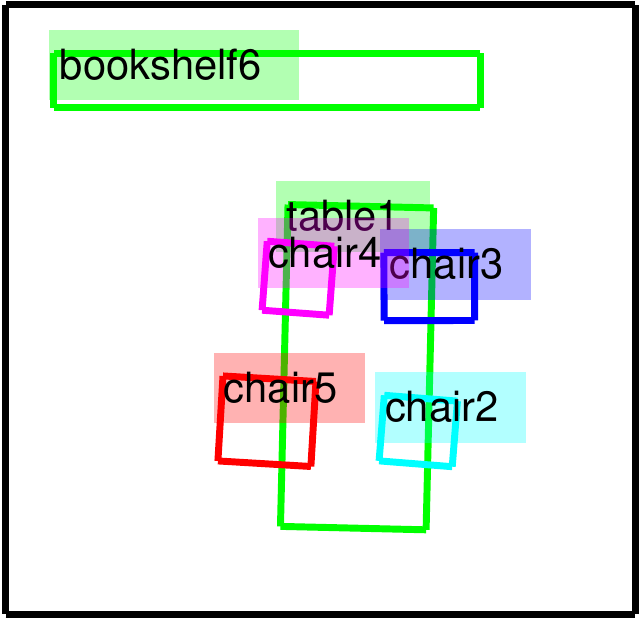} & 
\includegraphics[width=0.17\textwidth, height=1.7cm, trim={1.5 1 1.0 1.5},clip]{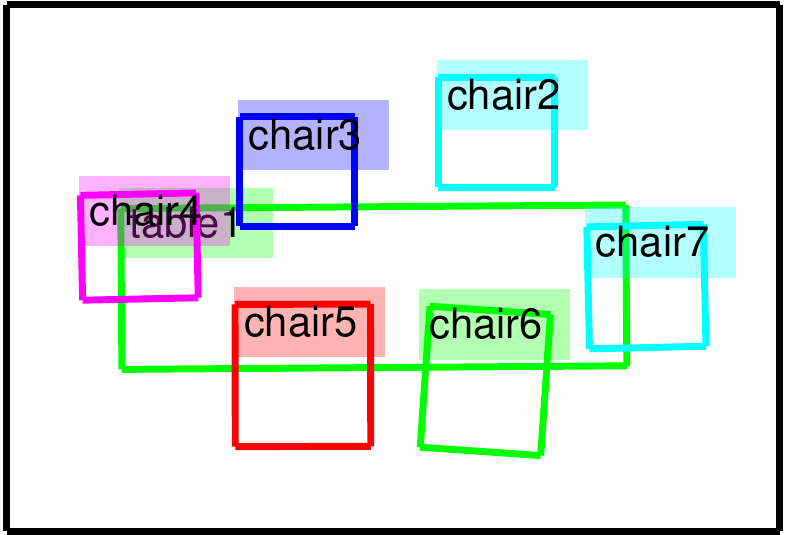} & 
\includegraphics[width=0.17\textwidth, height=1.7cm, trim={1 1 25 1},clip]{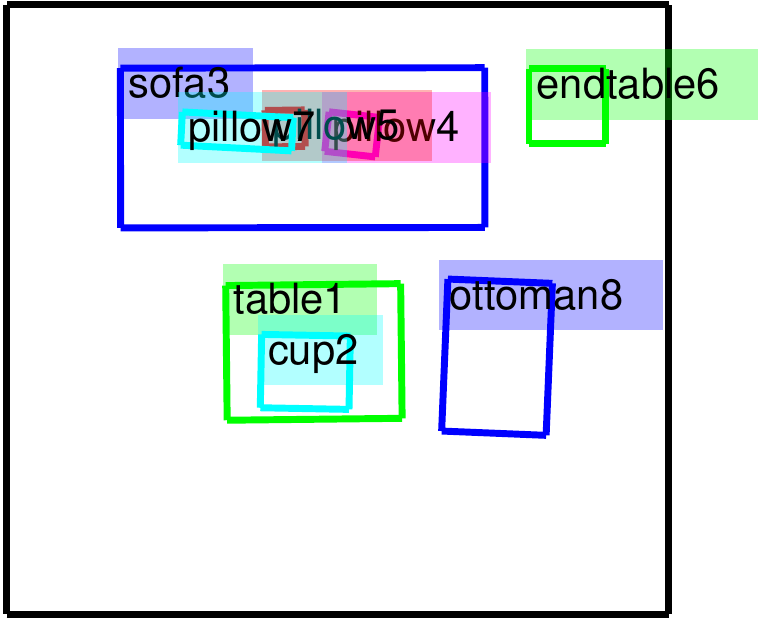} \\ 
% \begin{picture}(1,25)\put(0, 12){\rotatebox{90}{Office 2}}\end{picture} &
%\includegraphics[width=0.17\textwidth, height=1.7cm, trim={0.5 1 8 1},clip]{2dproj/org_6253-crop.pdf} & 
%\includegraphics[width=0.17\textwidth, height=1.7cm, trim={1 1 1 2},clip]{2dproj/375-crop.pdf} & 
%\includegraphics[width=0.17\textwidth, height=1.7cm, trim={1.5 1 23.0 1.5},clip]{2dproj/fast_synth2-crop.pdf} & 
%\includegraphics[width=0.17\textwidth, height=1.7cm, trim={1 1 9 1},clip]{2dproj/sample_0000000007-crop.pdf} \\ 
\begin{picture}(1,25)\put(0, 12){\rotatebox{90}{Kitchen}}\end{picture} &
\includegraphics[width=0.17\textwidth, height=1.7cm, trim={1 1 37 1},clip]{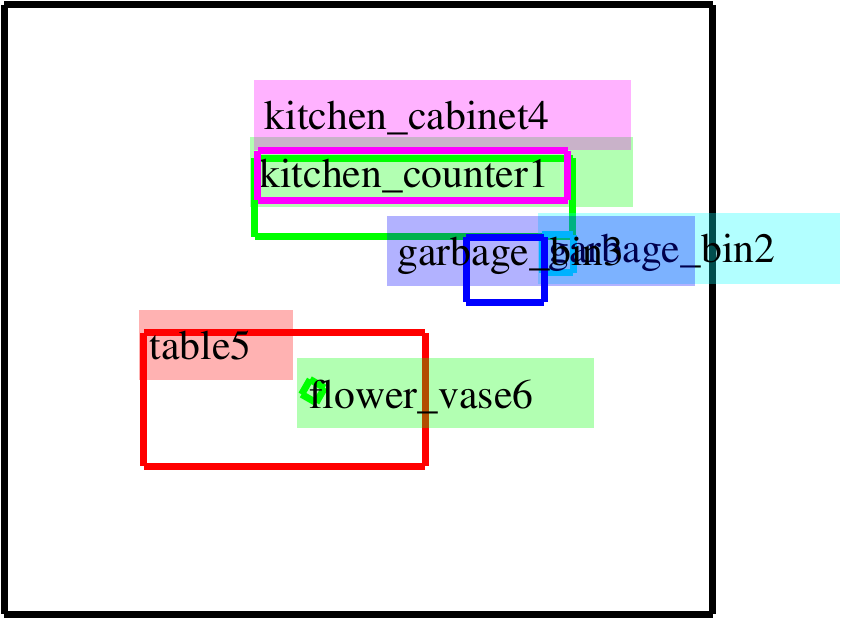} & 
\includegraphics[width=0.17\textwidth, height=1.7cm, trim={1 1 6.5 1},clip]{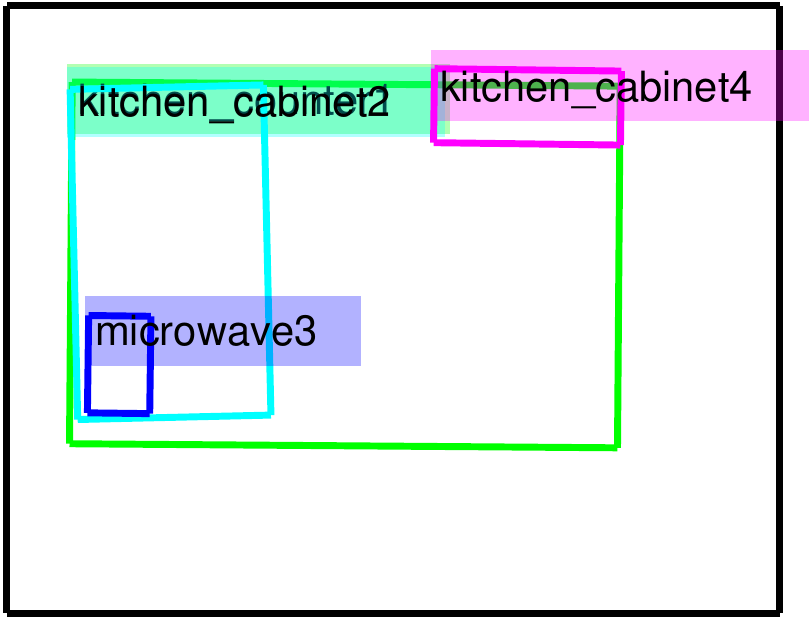} & 
\includegraphics[width=0.17\textwidth, height=1.7cm, trim={1 1 35.5 1},clip]{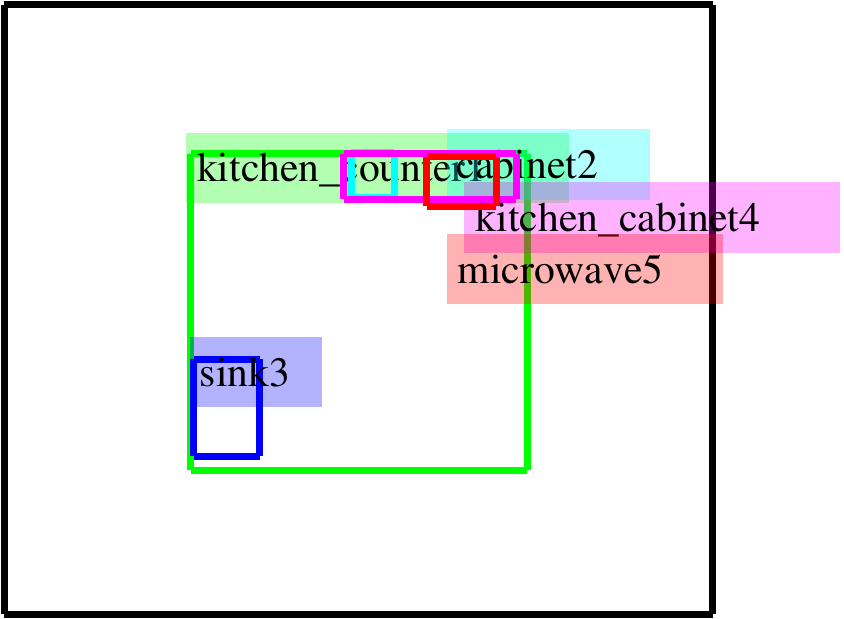} & 
\includegraphics[width=0.17\textwidth, height=1.7cm, trim={1 1 1 1},clip]{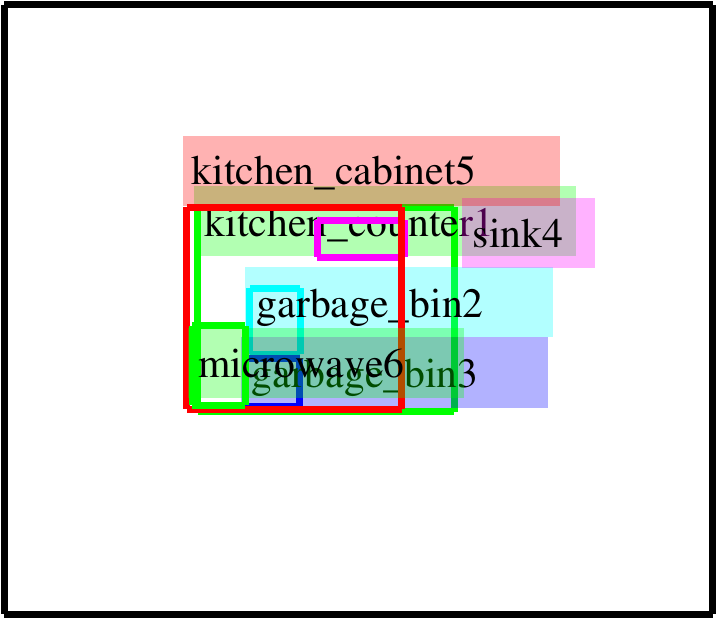} & 
\includegraphics[width=0.17\textwidth, height=1.7cm, trim={1 1 6 1},clip]{2dproj/org_2157-crop.pdf} \\
%\begin{picture}(1,25)\put(0, 10){\rotatebox{90}{Kitchen 2}}\end{picture} &
%\includegraphics[width=0.17\textwidth, height=1.7cm, trim={0.5 1 1 1},clip]{2dproj/org_3212-crop.pdf} & 
%\includegraphics[width=0.17\textwidth, height=1.7cm, trim={9 1 2.5 2.5},clip]{2dproj/125-crop.pdf} & 
%\includegraphics[width=0.17\textwidth, height=1.7cm, trim={1 1 55.5 1},clip]{2dproj/fast_synth3-crop.pdf} & 
%\includegraphics[width=0.17\textwidth, height=1.7cm, trim={1 1 1 1},clip]{2dproj/sample_0000000206-crop.pdf} \\
%~~~~~& (a) {SG-VAE} & (b) {SG-VAE} & (c) Grains~\cite{li2019grains} & (d) FS~\cite{ritchie2019fast}  & (e) HC~\cite{qi2018human} 
% & Sample 1 & Sample 2 & Sample 3 & Sample 4 & Sample 5 
\end{tabular} 
\caption{Top-views of the synthesized scenes generated by the SG-VAE on SUN RGB-D. A detailed comparison with other baselines on SUNCG can be found in supplementary.} 
\label{fig:2dproj2}
\end{figure} %\vspace{-3em}

\subsection{Comparison with baselines on other datasets}
% The proposed method and the other baselines reported in the last section are trained on the SUN RGB-D dataset~\cite{song2015sun}. However, 
%In contrast to the proposed method, some of the 
The conventional indoor scene synthesis methods (for example, Grains~\cite{li2019grains}, Human-centric~\cite{qi2018human} (HC), fast-synth~\cite{ritchie2019fast}(FS) etc.) are tailored to and trained on SUNCG dataset~\cite{song2017semantic}. The dataset consists of synthetic scenes generated by graphic designers.  Moreover, the dataset is no longer publicly available (along with the meta-files).
\footnote{Due to the legal dispute around SUNCG~\cite{song2017semantic} we include our results for SUNCG (conducted on our internal copy) in Table~\ref{tab:synthetic} only for illustrative purposes. % We may need to remove our results on SUNCG in the final version.
%, we hereby disclose that we do not want to gain any advantage exploiting the data. The SG-VAE entry for the SUNCG data (conducted on our internal copy) placed in Table~\ref{tab:synthetic} for the reviewers for a proper evaluation and will be removed in the final version.
}
Therefore the following evaluation protocols are employed to assess the performance of different methods. 

\noindent {\bf Comparison on synthetic scene quality} To conduct a qualitative evaluation, we employ a classifier (based on Pointnet~\cite{qi2017pointnet}) to predict a scene layout to be an original or generated by a scene synthesis method. If the generated scenes are very similar to the original scenes, the classifier performs poorly (lower accuracy) and indicates the efficacy of the synthesis method. The classifier takes a scene layout of multiple objects, individually represented by the concatenation of 1-hot code and the attributes, as input and predicts a binary label according to the scene-type. The classifier is trained and tested on a dataset of $2K$ original and synthetic scenes ($50\%$ training and $50\%$ testing). Note that SG-VAE is trained on the synthetic data generated by the interpolations of latent vectors (some examples are shown in Figure~\ref{fig:qualititive}) and real data of SUN RGB-D~\cite{song2015sun}. Lower accuracy of the classifier validates the superior performance of the proposed SG-VAE.  %The same task is performed by $10$ human subjects on $10\%$ scenes of the dataset. 
The average performance is plotted in Table~\ref{tab:synthetic}. 
Examples of some synthetic scenes\footnote{We thank the authors of Grains~\cite{li2019grains} and HC~\cite{qi2018human} for sharing the code.
More results and the proposed SG-VAE for SUNCG are in the supplementary material.
%The scene layouts of the baseline indoor scene synthesis methods and the proposed SG-VAE (including the learned grammar) on SUNCG dataset are placed in Supplementary.
} generated by SG-VAE are also shown in Figure~\ref{fig:2dproj2}. %\vspace{-0.5em}

\begin{table}[H]\setlength{\tabcolsep}{13pt} %{r}{0.50\textwidth} 
\caption{Original vs. synthetic classification accuracy (trained with Pointnet~\cite{qi2017pointnet}): The accuracy indicates that the synthetic scenes are indistinguishable from the original scenes and hence lower (closer to $50\%$) is better. } 
 % {\bf Performance of Human-centric~\cite{qi2018human} to be included.}}
\centering \scriptsize  
%\begin{tabular}{l|ccc|ccc}\toprule    
%& \multicolumn{3}{c|}{Pointnet~\cite{qi2017pointnet}} & \multicolumn{3}{c}{Visual Inspection} \\ 
%Methods & {SG-VAE}  & Grains~\cite{li2019grains} & HC~\cite{qi2018human} & {SG-VAE}  & Grains~\cite{li2019grains} & HC~\cite{qi2018human}\\\midrule
%Accuracy & $\bm{71.3\%}$ & $96.4\%$ & $98.1\%$ & $\bm{63.2\%}$ & $82.9\%$ & $87.4\%$ \\ \bottomrule 
%%\vspace{-25pt}
%\end{tabular} %\vspace{-1em}
%\label{tab:synthetic} 
%\end{table}
\begin{tabular}{l|c|c|c|c}\toprule    
Datasets& SUN RGB-D~\cite{song2015sun} & \multicolumn{3}{c}{SUNCG~\cite{song2017semantic}} \\ \midrule 
 % \cmidrule[0.06em](r){2-2}\cmidrule[0.06em](lr){3-5} 
%& \multicolumn{3}{c}{Visual Inspection} \\ 
Methods & {SG-VAE} & {SG-VAE} & Grains~\cite{li2019grains} & HC~\cite{qi2018human} \\ 
%& {SG-VAE}  & Grains~\cite{li2019grains} & HC~\cite{qi2018human}\\\midrule
Accuracy & $\bm{71.3\%}$ & $\bm{83.7\%}$ & $96.4\%$ & $98.1\%$ \\ \bottomrule  %& $\bm{63.2\%}$ & $82.9\%$ & $87.4\%$ \\ 
%\vspace{-25pt}
\end{tabular} %\vspace{-1em}
\label{tab:synthetic} 
\end{table}

\noindent {\bf Runtime comparison} 
 All the methods are evaluated on a single CPU and the runtime is displayed in Table~\ref{tab:runtime}. Note that the decoder of the proposed SG-VAE takes only $\sim 1$ms to generate the parse tree and the rest of the time is consumed by the renderer (generating bounding boxes). The proposed method is almost two orders of magnitude faster than the other scene synthesis methods. %\vspace{-1em}
 
\begin{table}[H]\setlength{\tabcolsep}{12pt} %{r}{0.50\textwidth}
\caption{Average time required to generate a single scene.}
\centering \scriptsize 
\begin{tabular}{c|c|c|c|c}\toprule    
Methods & {SG-VAE~}  & Grains~\cite{li2019grains} & FS~\cite{ritchie2019fast} & HC~\cite{qi2018human}  \\\midrule
Avg. runtime & $\bm{8.5ms}$ &  $1.2\times10^{2}ms$ & $1.8\times10^{3}ms$ & $2.4\times10^{5}ms$ \\ \bottomrule 
%\vspace{-25pt}
\end{tabular} %\vspace{-2em}
\label{tab:runtime} 
\end{table}

\subsection{Scene layout estimation from the RGB-D image}
The task is to predict the 3D scene layout given an RGB-D image. Typically, the state of the art methods are based on sophisticated region proposals and subsequent processing~\cite{song2016deep,qi2020imvotenet}. With this experiment, we aim to demonstrate the potential use of the latent representation learned by the proposed auto-encoder for a computer vision task, and therefore we employ a simple approach at this point. We (linearly) map deep features (extracted from images by a DNN~\cite{zhou2014learning}) to the latent space of the scene-grammar autoencoder. The decoder subsequently generates a 3D scene configuration with associated bounding boxes and object labels from the projected latent vector. Since during the deep feature extraction and the linear projection, the spatial information of the bounding boxes are lost, the predicted scene layout is then combined with a bounding box detection to produce the final output. 
\begin{figure}[H]
\centering \tiny
\begin{tabular}{@{\hspace{0.2em}}c@{\hspace{0.2em}}c@{\hspace{0.2em}}c@{\hspace{0.2em}}c@{\hspace{0.25em}}c@{\hspace{0.2em}}c@{\hspace{0.2em}}c@{\hspace{0.2em}}c@{\hspace{-12.5em}}c@{\hspace{-0.5em}}c@{\hspace{0.2em}}c} 
(a) \texttt{\bf Input} & (b) {\bf SG-VAE} & (c) {\bf BL1} & (d) {\bf BL2}~\cite{gomez2018automatic} & (e){\bf BL3}~\cite{kusner2017grammar} & (f) {\bf DSS~\cite{song2016deep}} & (g) {\bf GrndTrth}  \\ 
\includegraphics[width=0.135\textwidth, height=1.2cm]{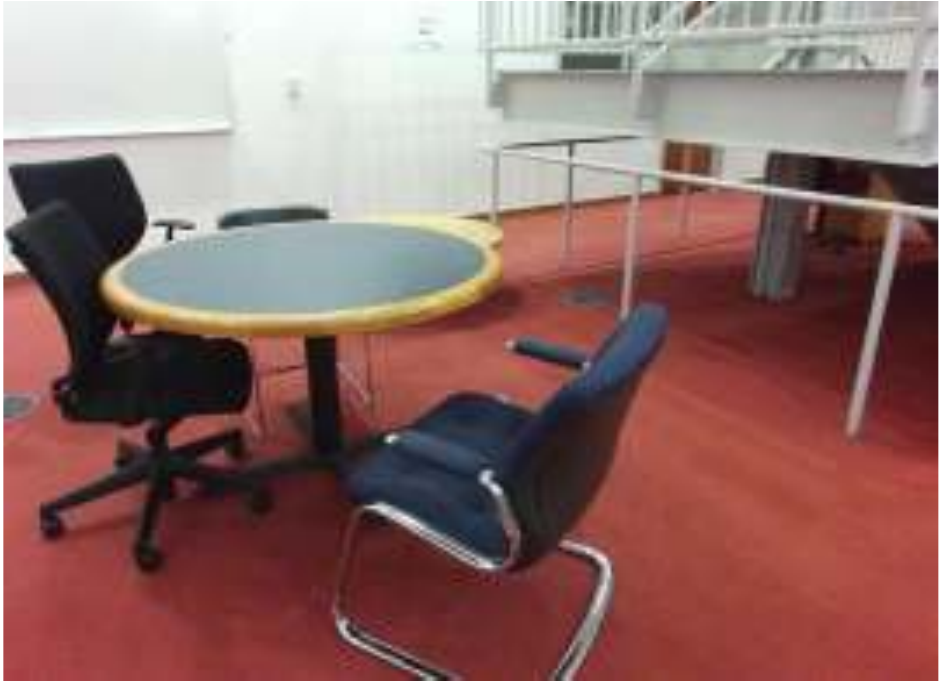} & 
\includegraphics[width=0.135\textwidth, height=1.2cm]{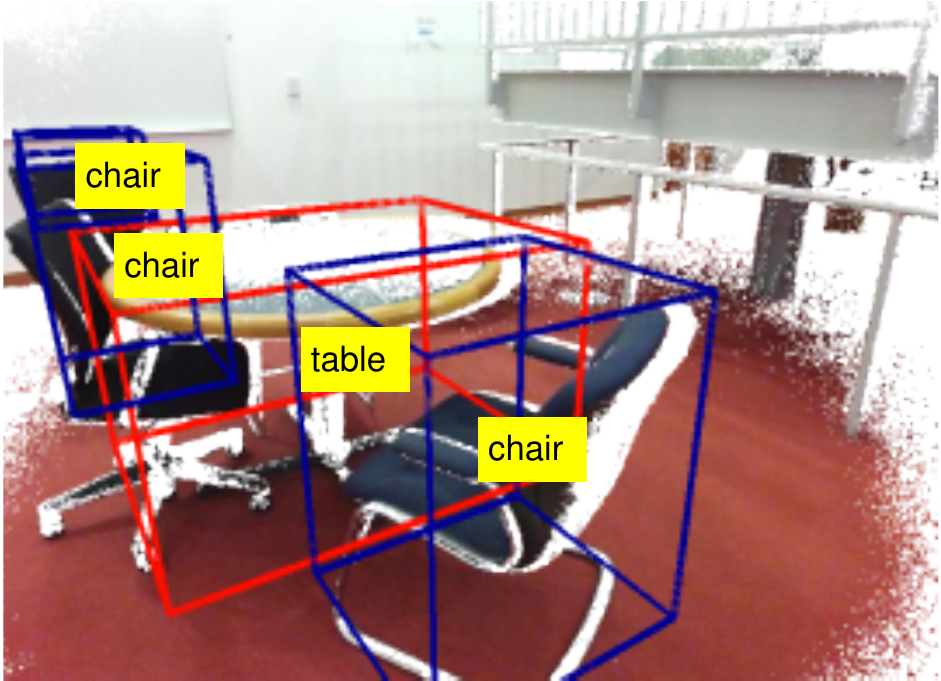} & 
\includegraphics[width=0.135\textwidth, height=1.2cm]{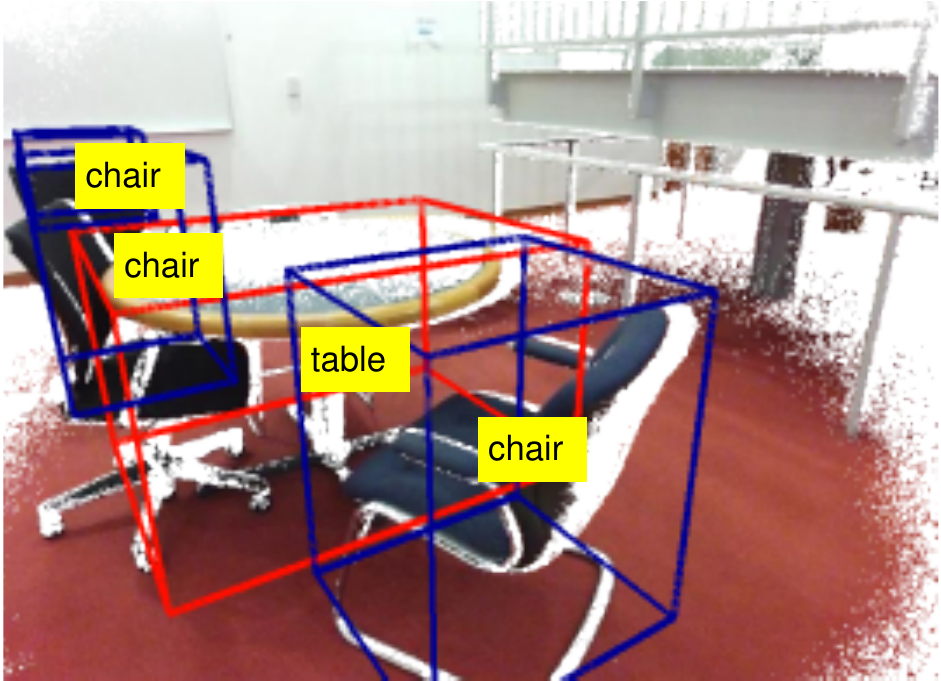} & 
\includegraphics[width=0.135\textwidth, height=1.2cm]{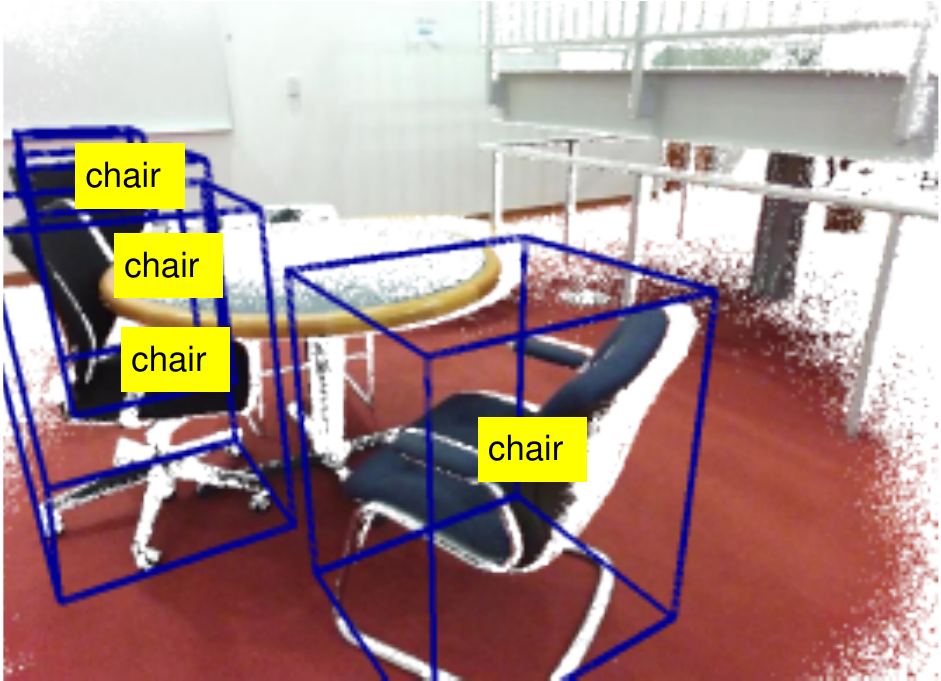} & 
\includegraphics[width=0.135\textwidth, height=1.2cm]{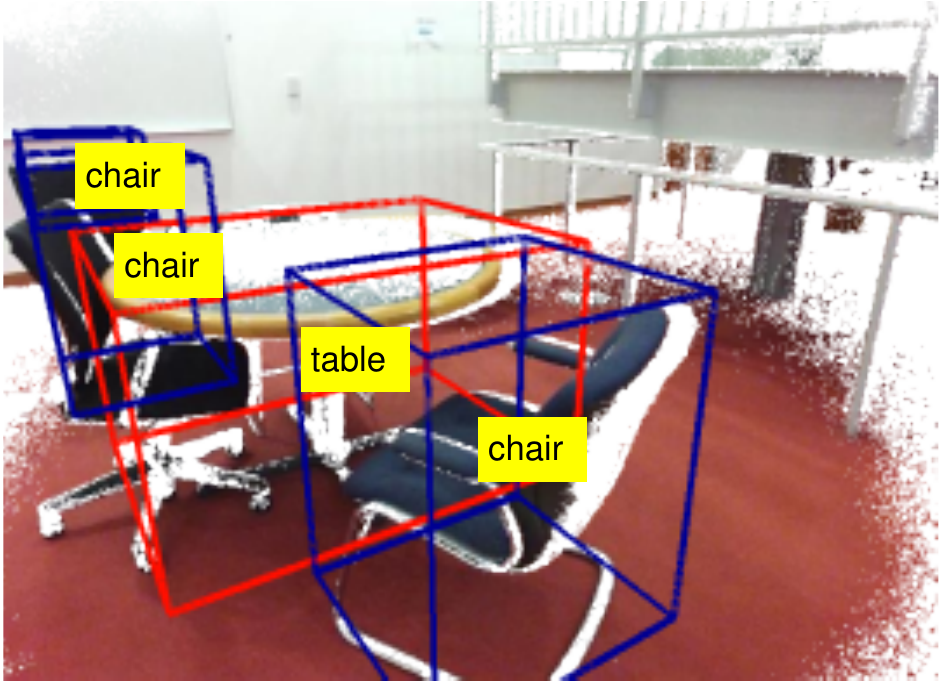} & 
\includegraphics[width=0.135\textwidth, height=1.2cm]{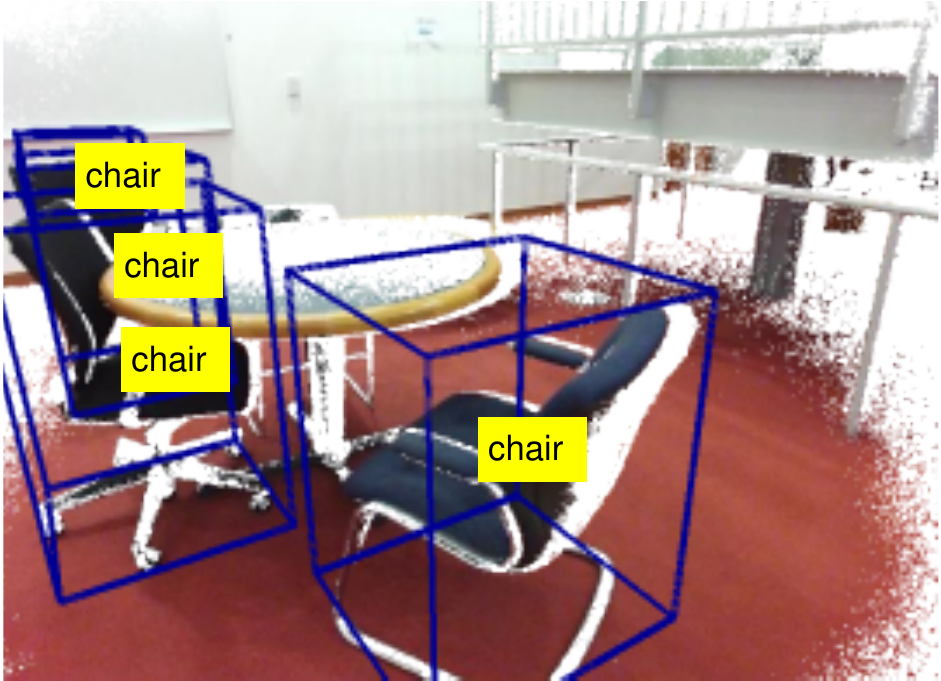} & 
\includegraphics[width=0.135\textwidth, height=1.2cm]{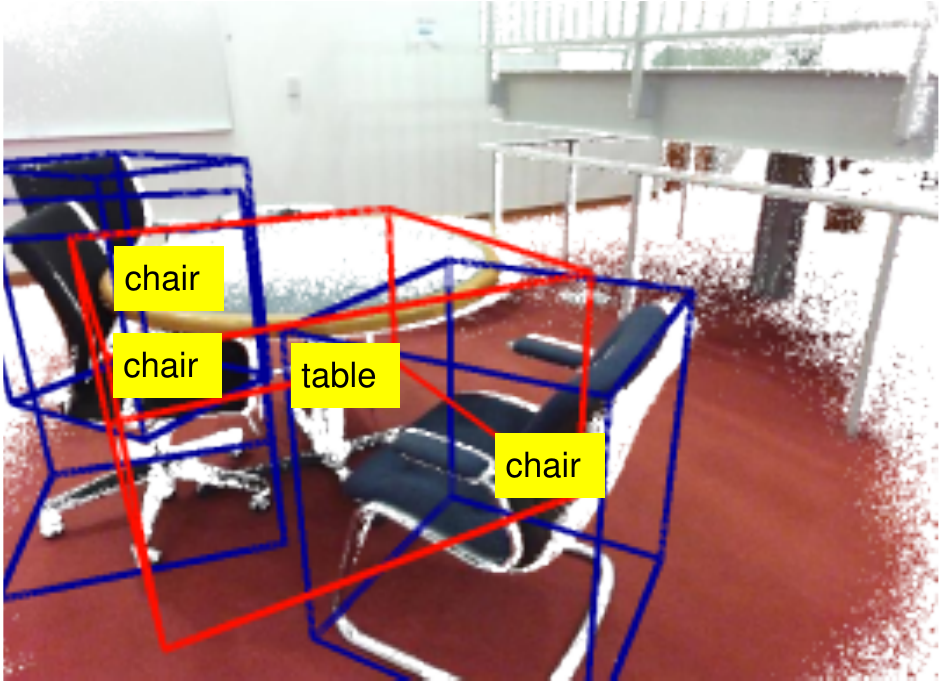}  \\ 
 %& IoU $ = 0.65$ & IoU $ = 0.65$ & IoU $ = 0.43 $ & IoU $ = 0.65$ & IoU $ = 0.43$ & \\

\includegraphics[width=0.135\textwidth, height=1.2cm]{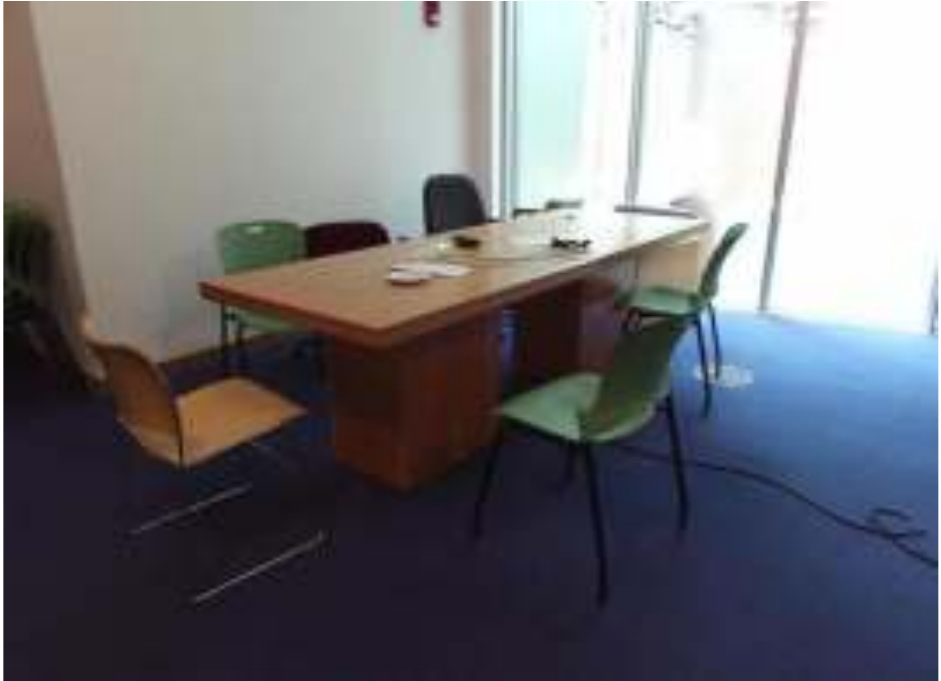} & 
\includegraphics[width=0.135\textwidth, height=1.2cm]{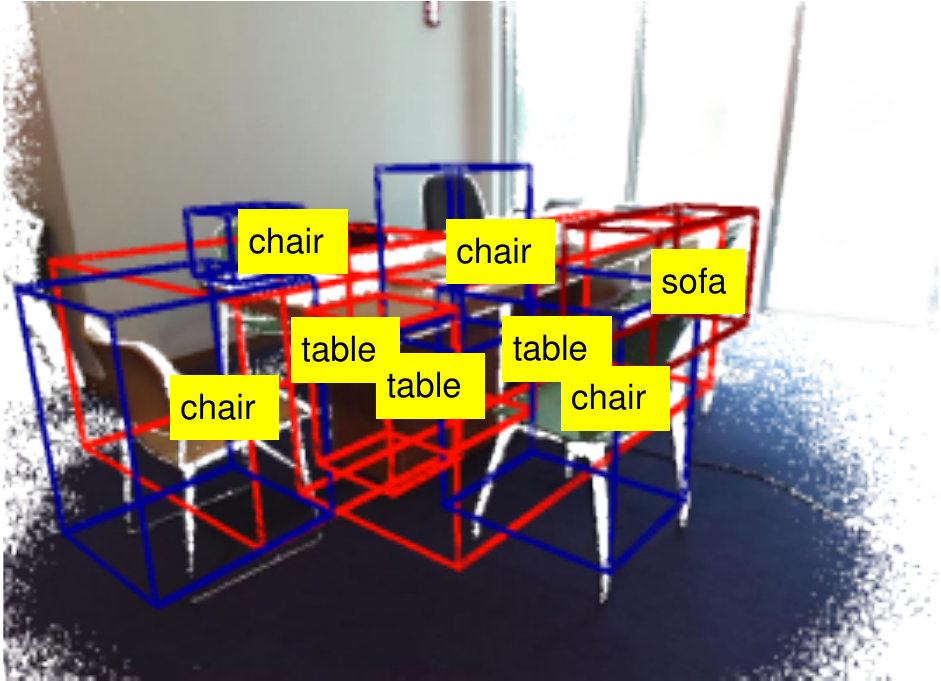} & 
\includegraphics[width=0.135\textwidth, height=1.2cm]{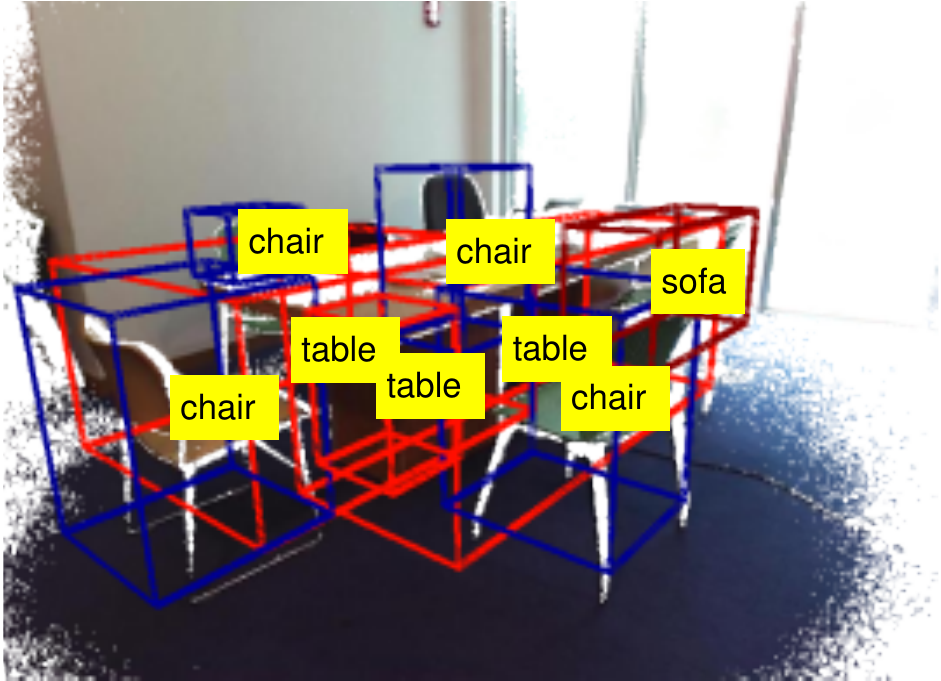} & 
\includegraphics[width=0.135\textwidth, height=1.2cm]{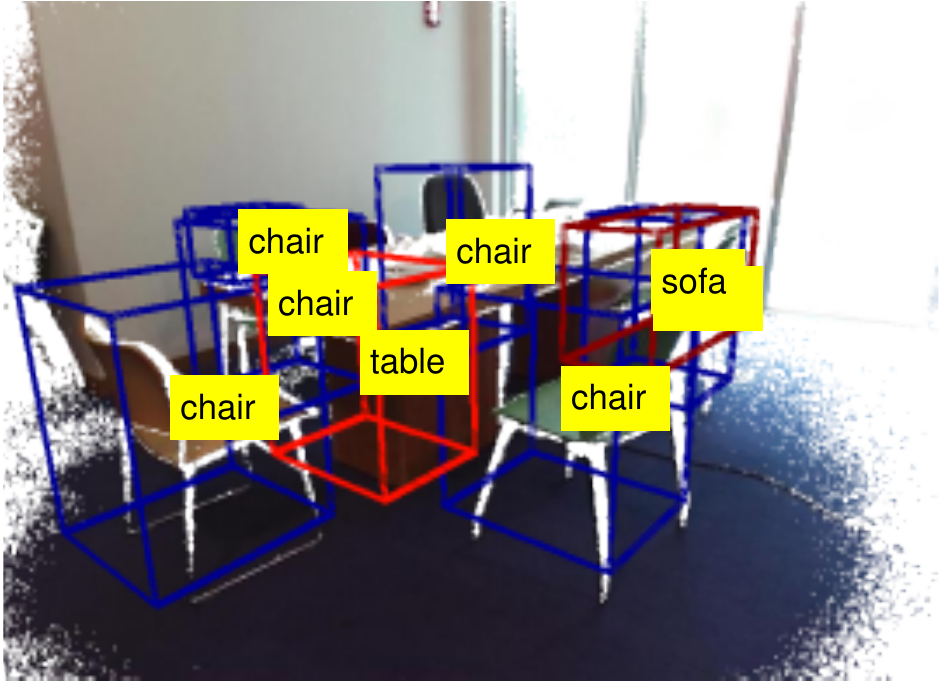} & 
\includegraphics[width=0.135\textwidth, height=1.2cm]{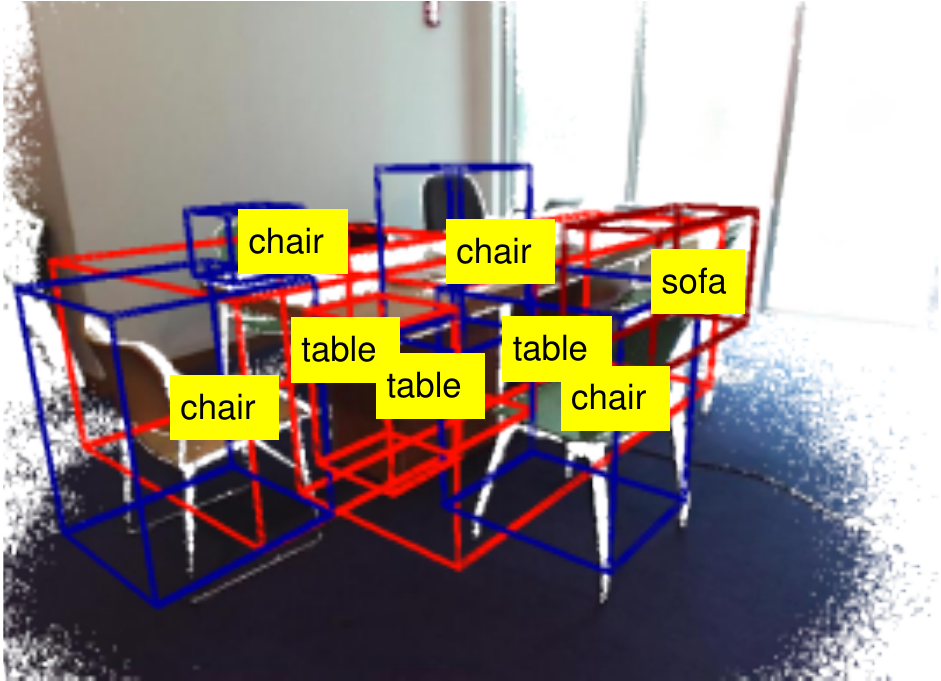} & 
\includegraphics[width=0.135\textwidth, height=1.2cm]{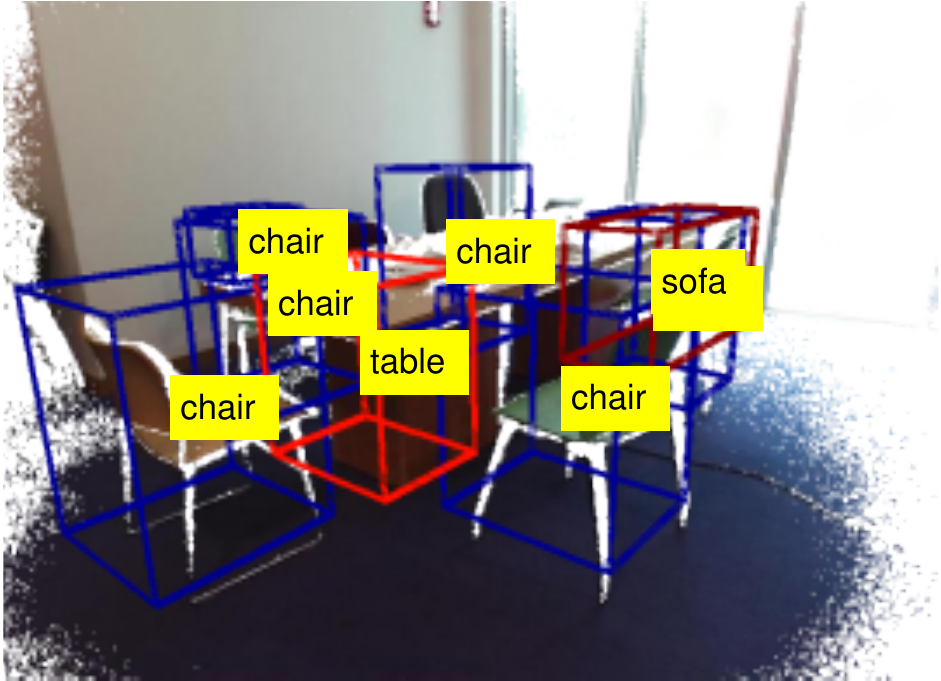} & 
\includegraphics[width=0.135\textwidth, height=1.2cm]{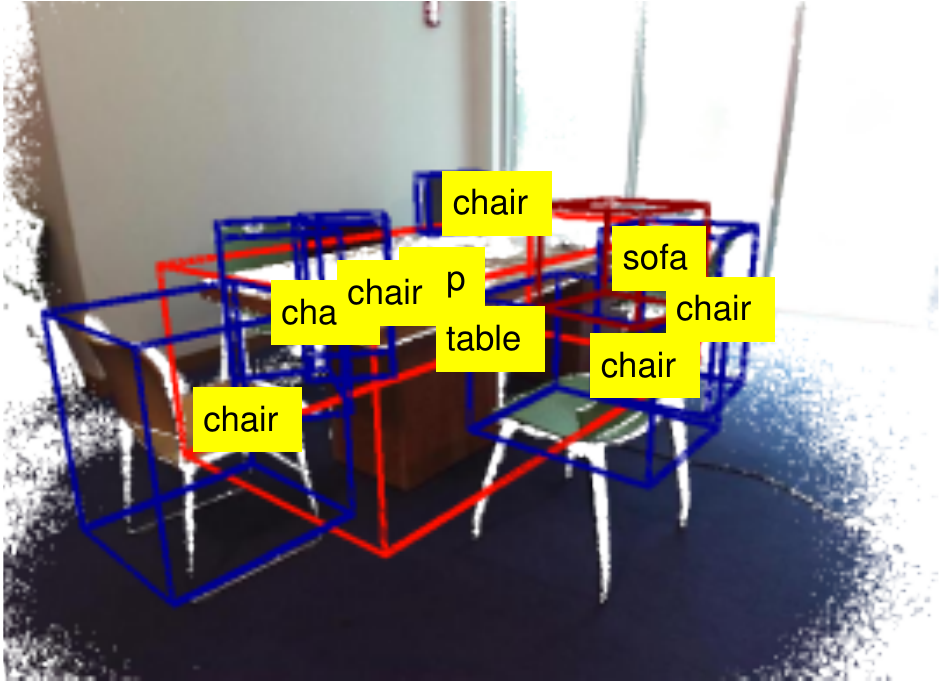}  \\  
 %& IoU $ = 0.64 $ & IoU $ = 0.64 $ & IoU $ = 0.35 $ & IoU $ = 0.64 $ & IoU $ = 0.35 $ &  \\ 
 
\includegraphics[width=0.135\textwidth, height=1.2cm]{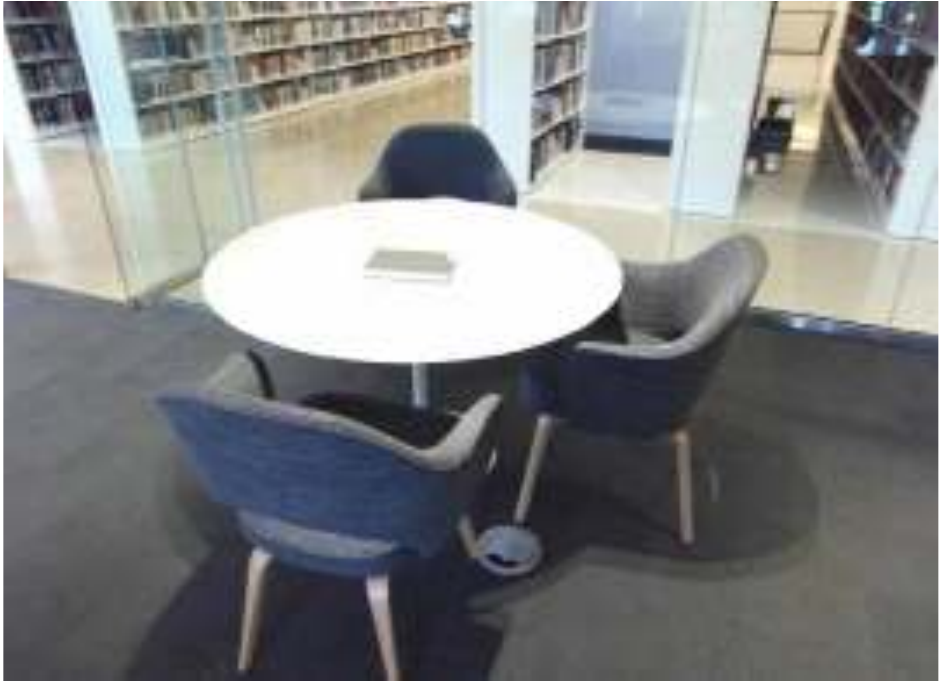} & 
\includegraphics[width=0.135\textwidth, height=1.2cm]{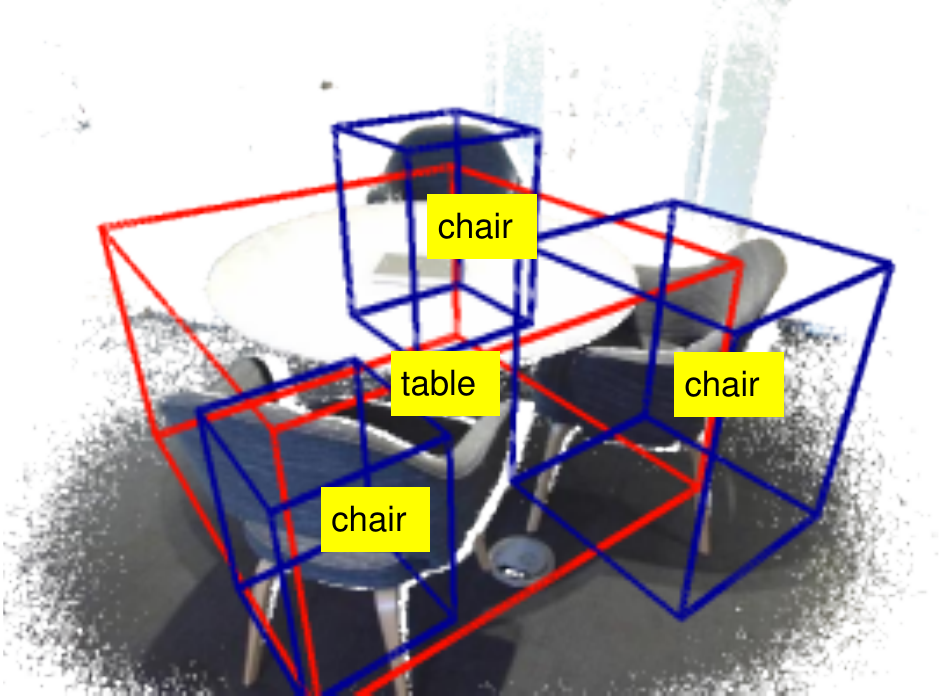} & 
\includegraphics[width=0.135\textwidth, height=1.2cm]{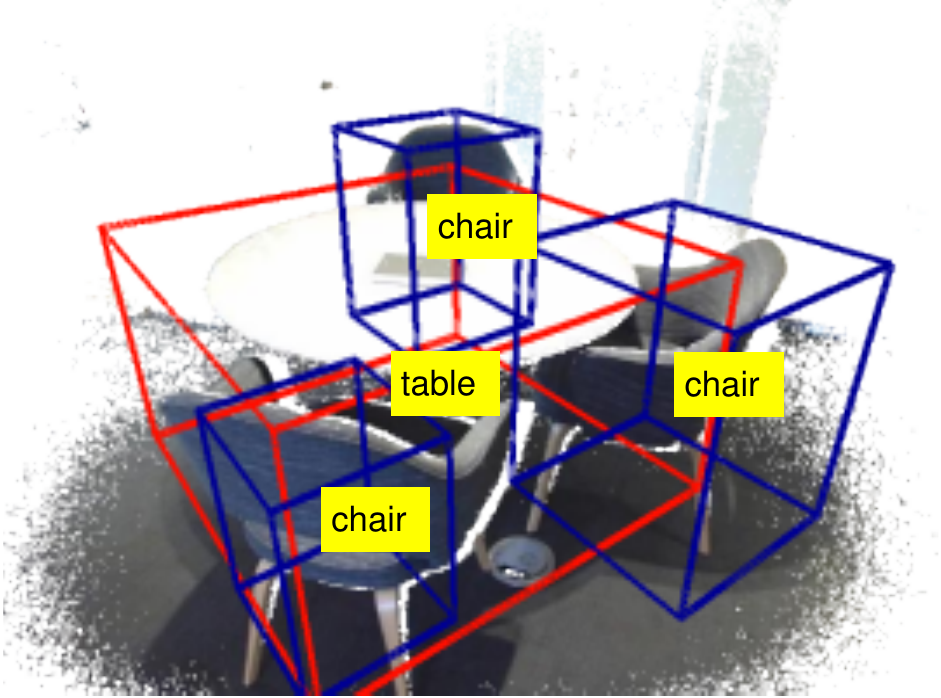} & 
\includegraphics[width=0.135\textwidth, height=1.2cm]{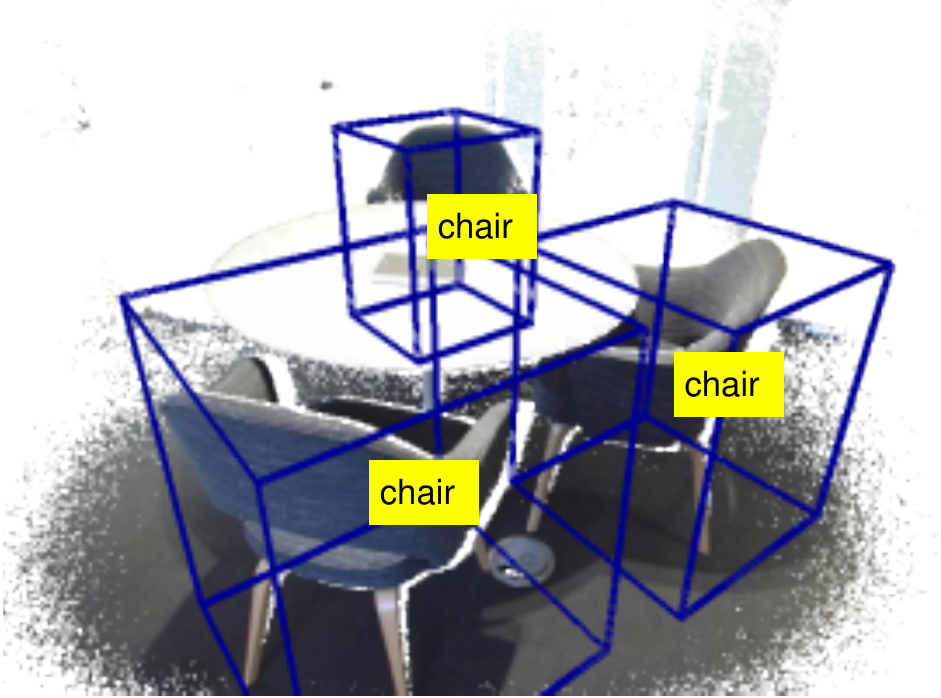} & 
\includegraphics[width=0.135\textwidth, height=1.2cm]{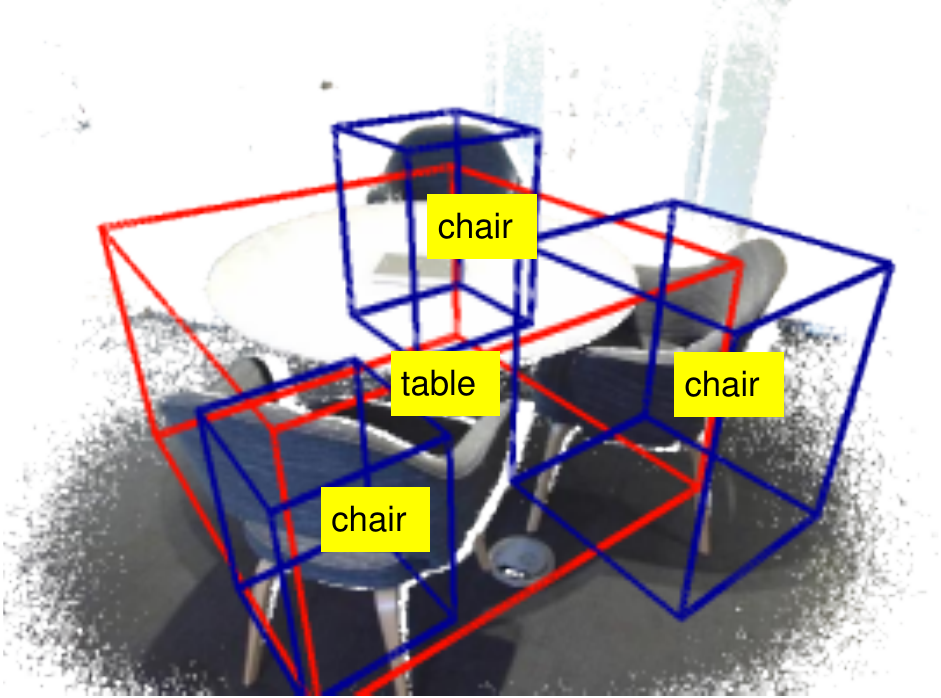} & 
\includegraphics[width=0.135\textwidth, height=1.2cm]{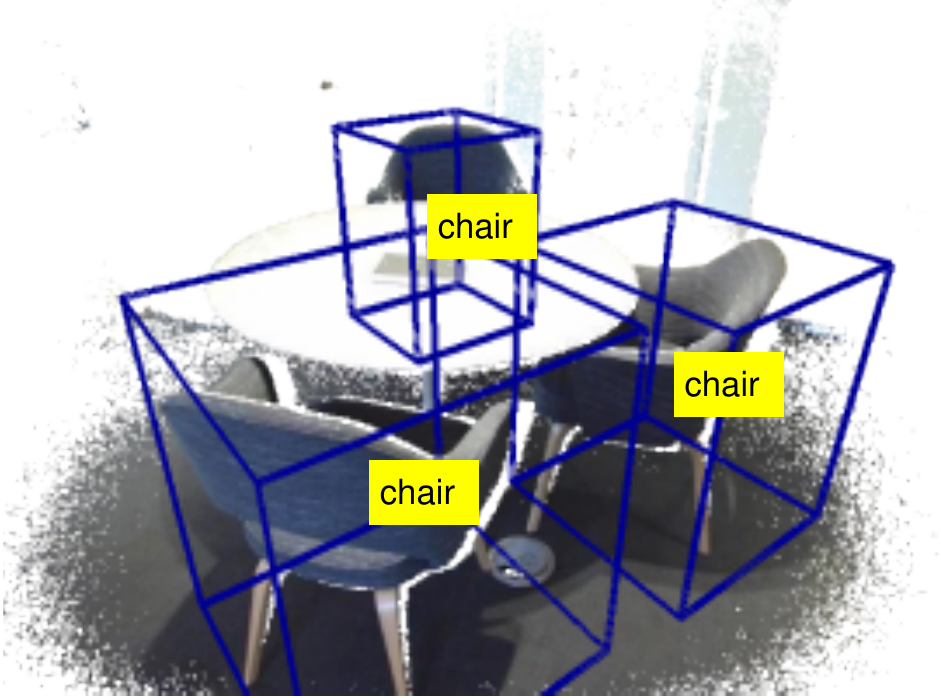} & 
\includegraphics[width=0.135\textwidth, height=1.2cm]{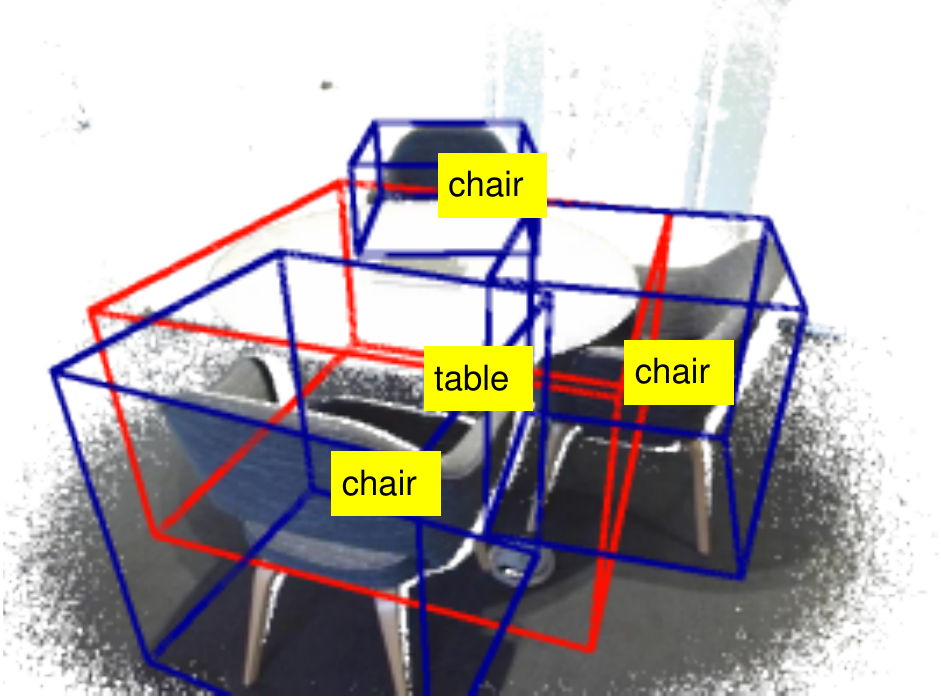}  \\
 %& IoU $ = 0.61 $ & IoU $ = 0.61 $ & IoU $ = 0.44 $ & IoU $ = 0.61 $ & IoU $ = 0.44 $ & \\

\includegraphics[width=0.135\textwidth, height=1.2cm]{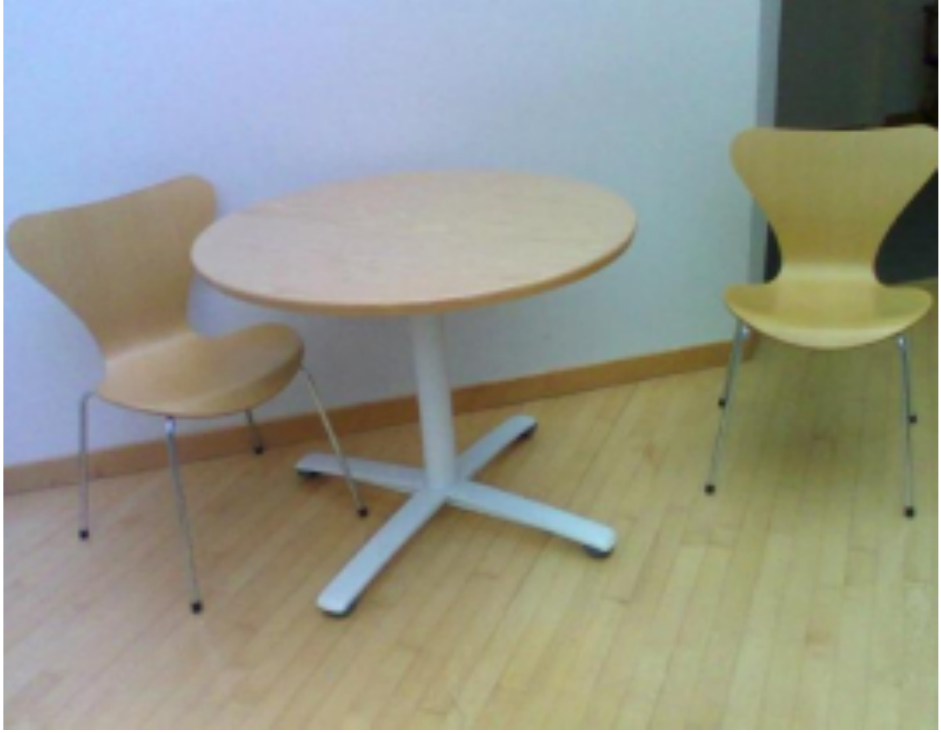} & 
\includegraphics[width=0.135\textwidth, height=1.2cm]{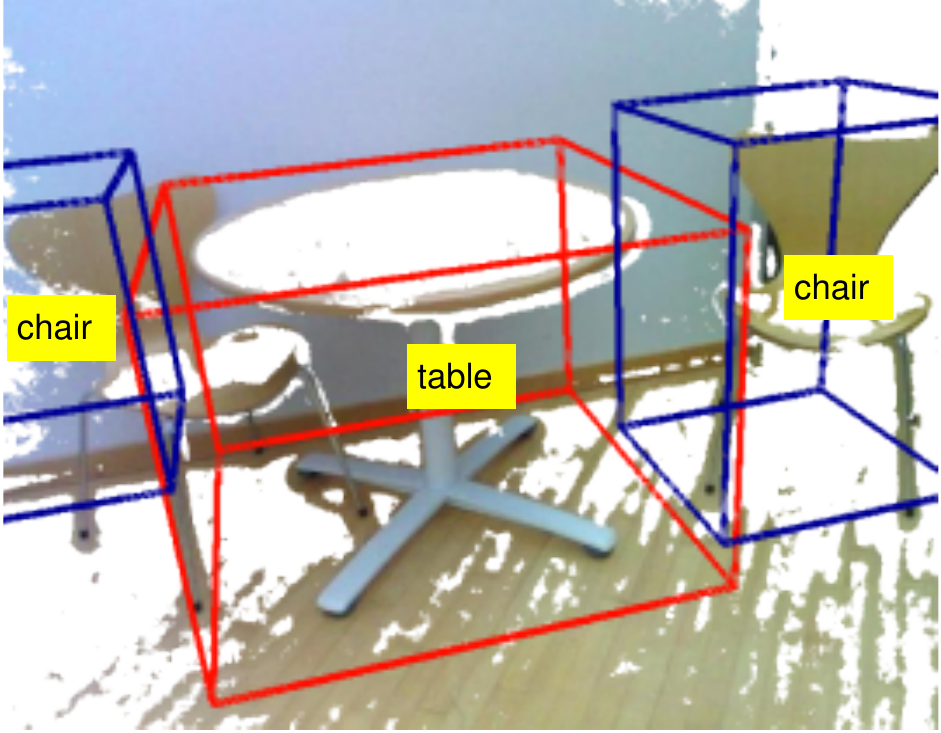} & 
\includegraphics[width=0.135\textwidth, height=1.2cm]{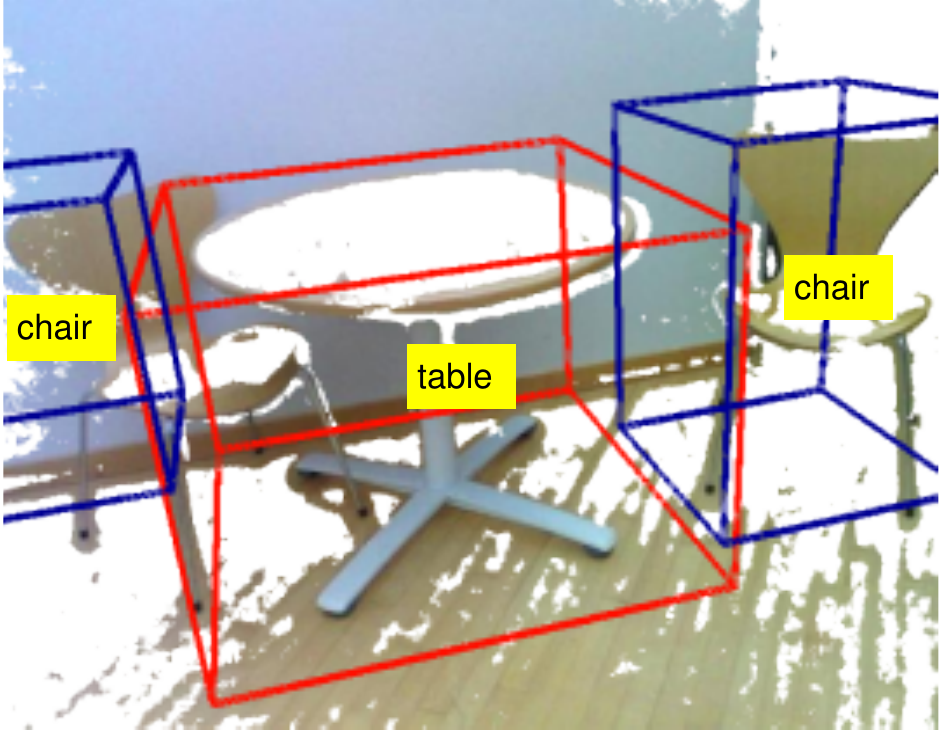} & 
\includegraphics[width=0.135\textwidth, height=1.2cm]{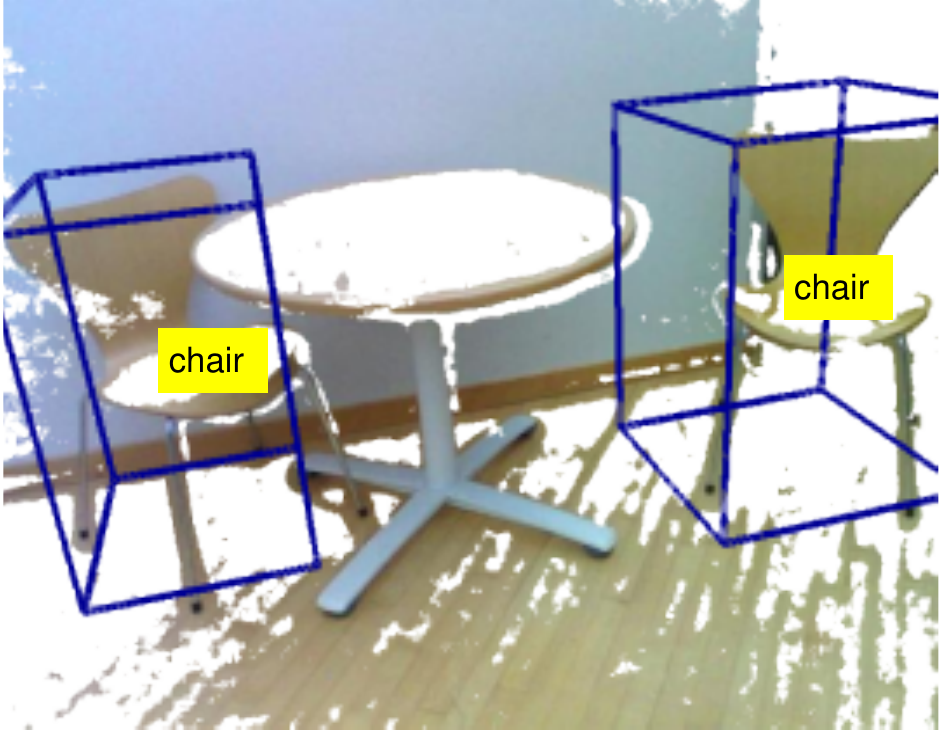} & 
\includegraphics[width=0.135\textwidth, height=1.2cm]{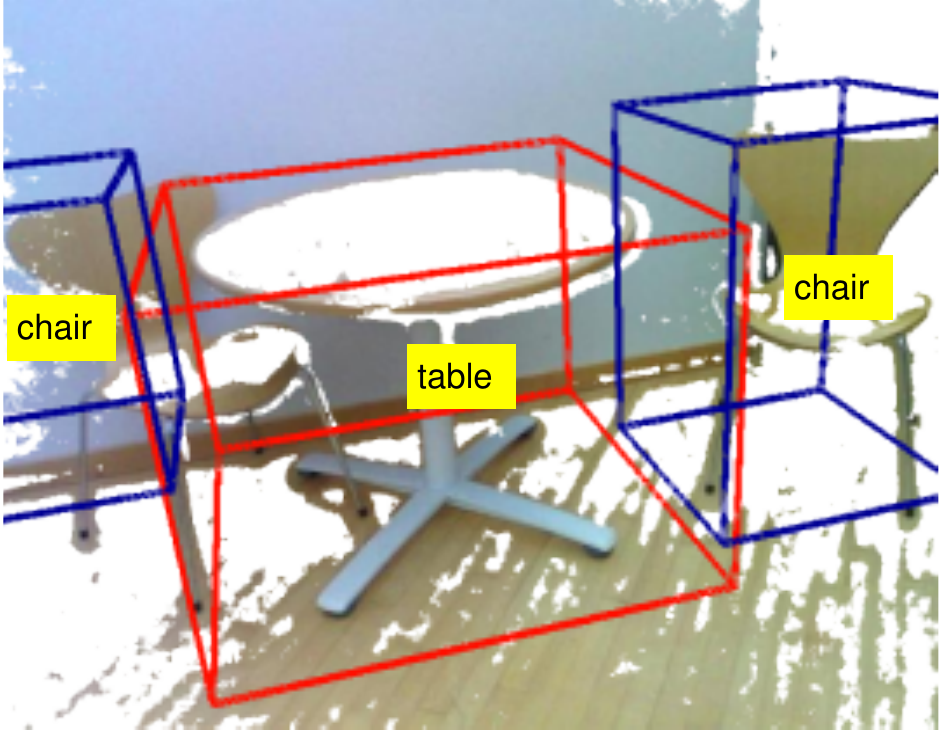} & 
\includegraphics[width=0.135\textwidth, height=1.2cm]{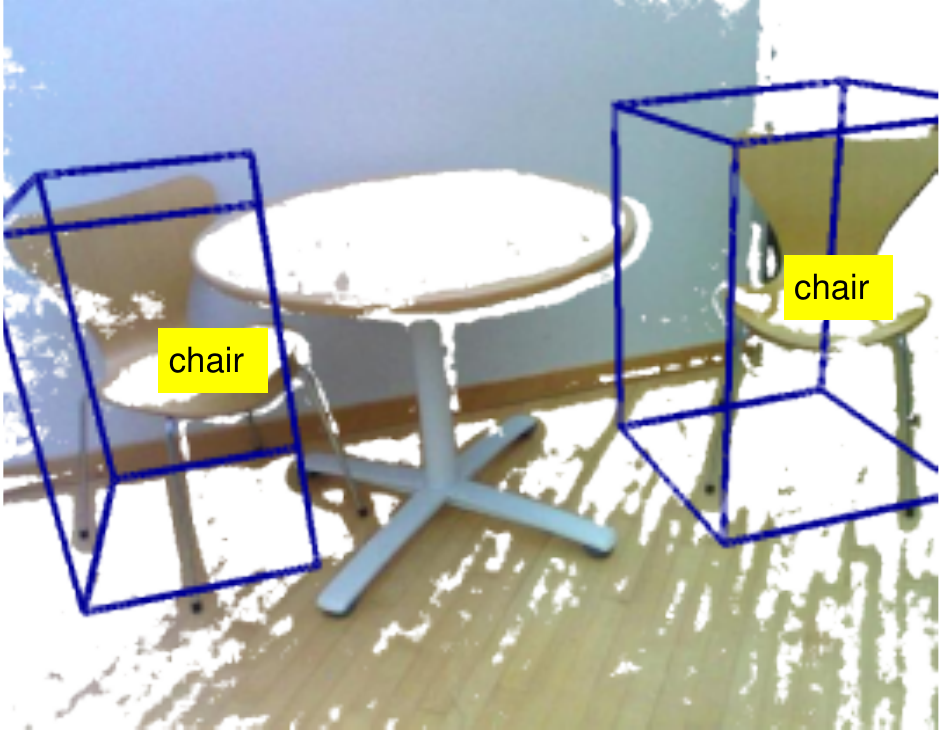} & 
\includegraphics[width=0.135\textwidth, height=1.2cm]{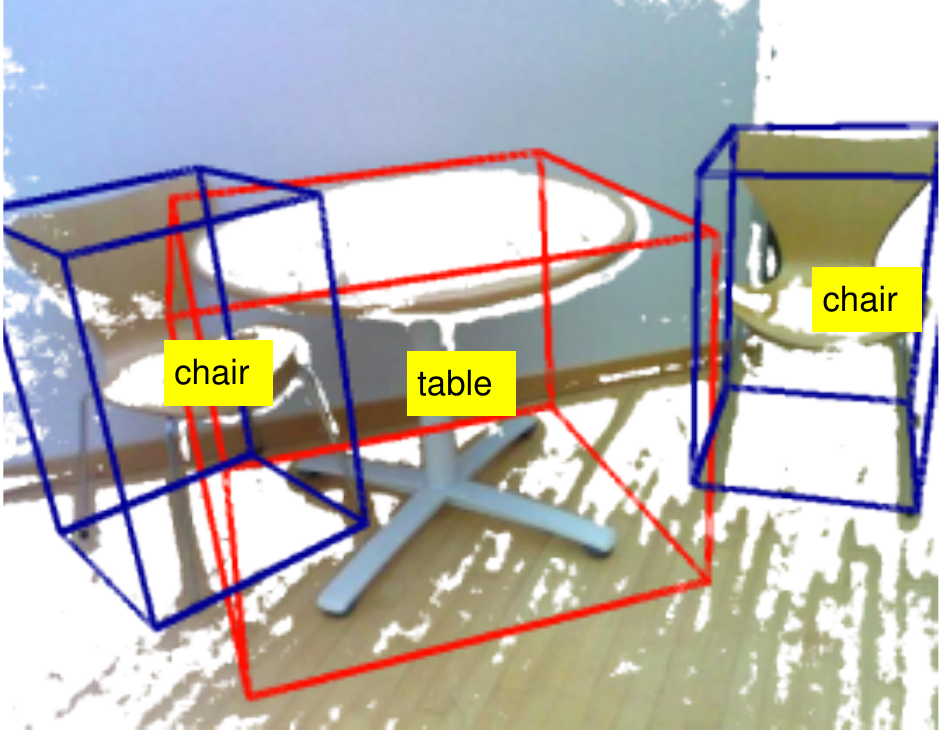}  \\  
 %& IoU $ = 0.61 $ & IoU $ = 0.61 $ & IoU $ = 0.33 $ & IoU $ = 0.61 $ & IoU $ = 0.33 $ &  \\ 
 
%  \includegraphics[width=0.135\textwidth, height=1.2cm]{vis_images/4363_rgb_0144-crop.pdf} & 
%\includegraphics[width=0.135\textwidth, height=1.2cm, trim={1 1 27.5 1},clip]{vis_images/4363_pred_0144-crop.pdf} & 
%\includegraphics[width=0.135\textwidth, height=1.2cm, trim={1 1 27.5 1},clip]{vis_images/4363_pred_pose_0144-crop.pdf} & 
%\includegraphics[width=0.135\textwidth, height=1.2cm, trim={1 1 27.5 1},clip]{vis_images/4363_pred_standrd_0144-crop.pdf} & 
%\includegraphics[width=0.135\textwidth, height=1.2cm, trim={1 1 27.5 1},clip]{vis_images/4363_pred_gvae_0144-crop.pdf} & 
%\includegraphics[width=0.135\textwidth, height=1.2cm, trim={1 1 27.5 1},clip]{vis_images/4363_all_0144-crop.pdf} & 
%\includegraphics[width=0.135\textwidth, height=1.2cm, trim={1 1 27.5 1},clip]{vis_images/4363_org_0144-crop.pdf}  \\ 
%% & IoU $ = 0.55 $ & IoU $ = 0.55 $ & IoU $ = 0.36 $ & IoU $ = 0.55 $ & IoU $ = 0.36 $ &  \\

  \includegraphics[width=0.135\textwidth, height=1.2cm]{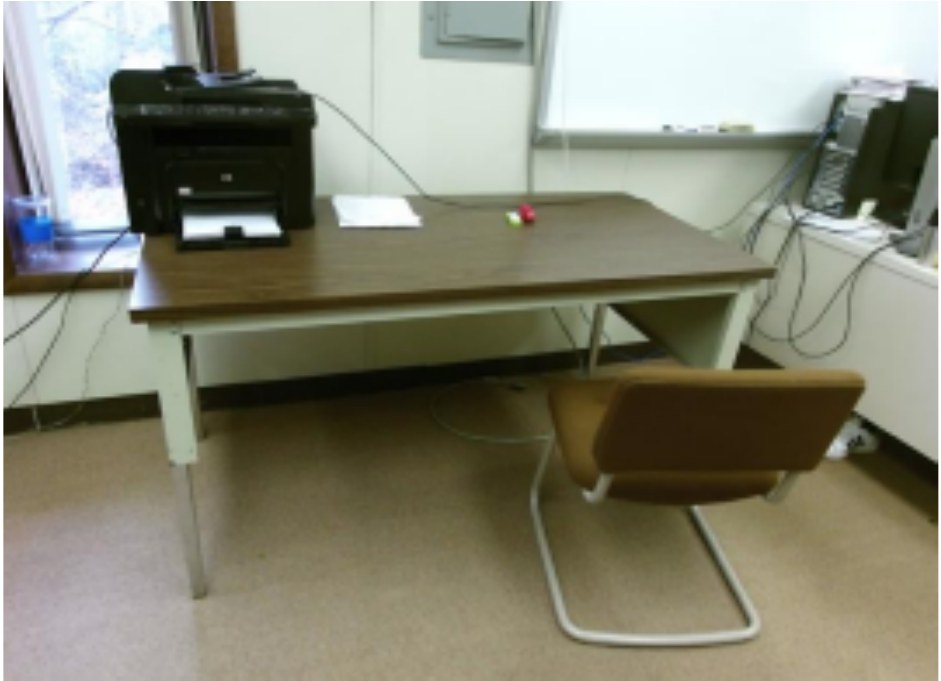} & 
\includegraphics[width=0.135\textwidth, height=1.2cm]{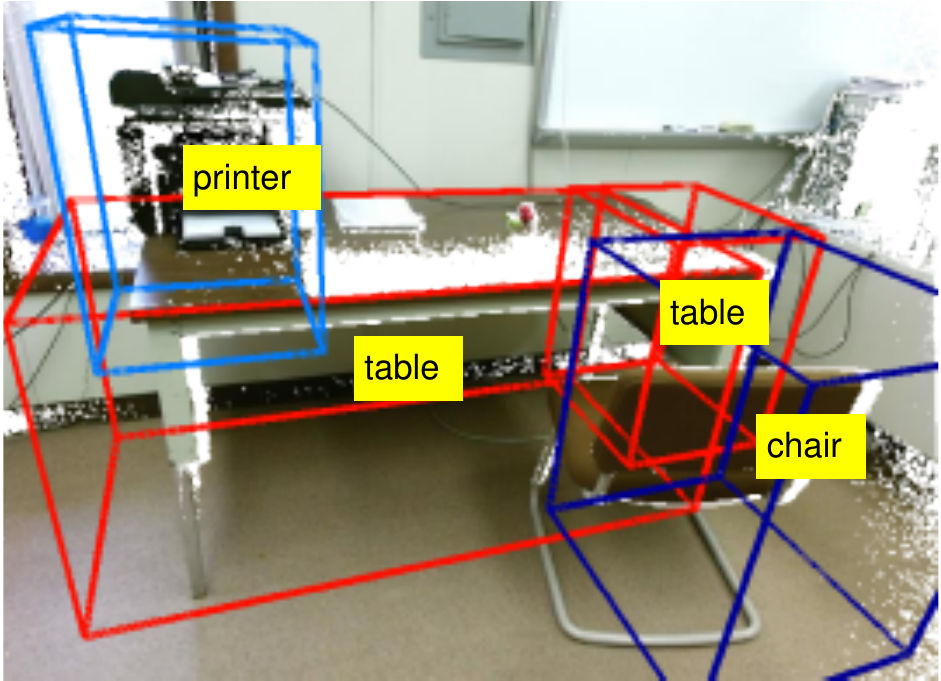} & 
\includegraphics[width=0.135\textwidth, height=1.2cm]{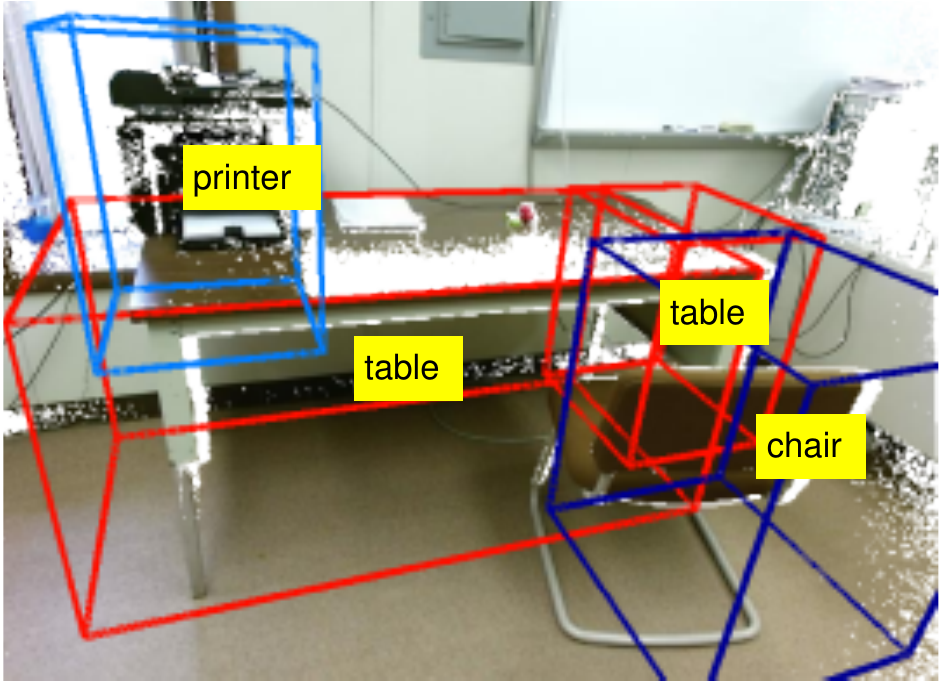} & 
\includegraphics[width=0.135\textwidth, height=1.2cm]{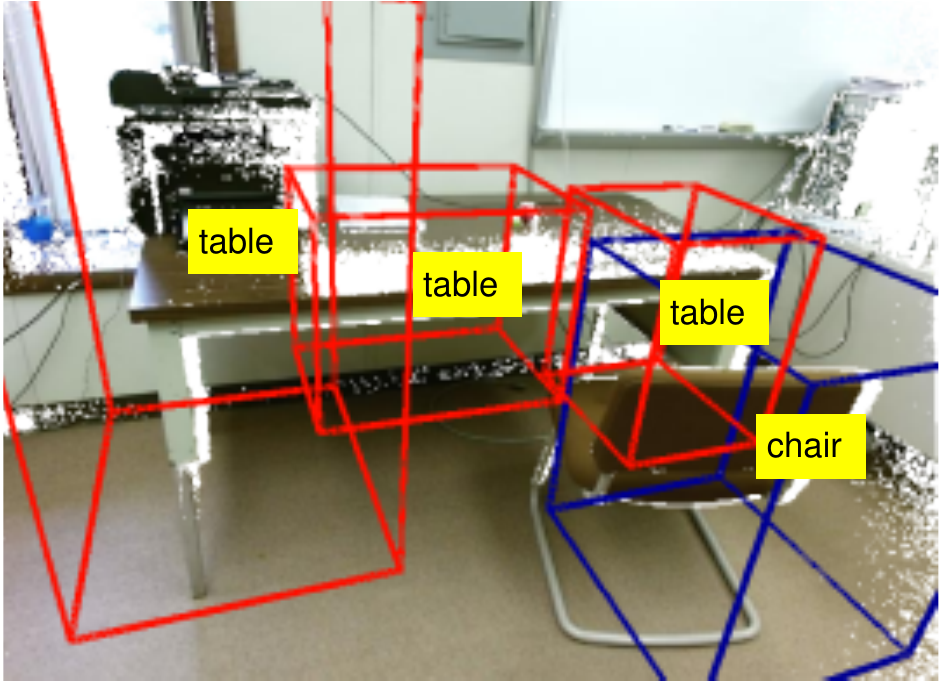} & 
\includegraphics[width=0.135\textwidth, height=1.2cm]{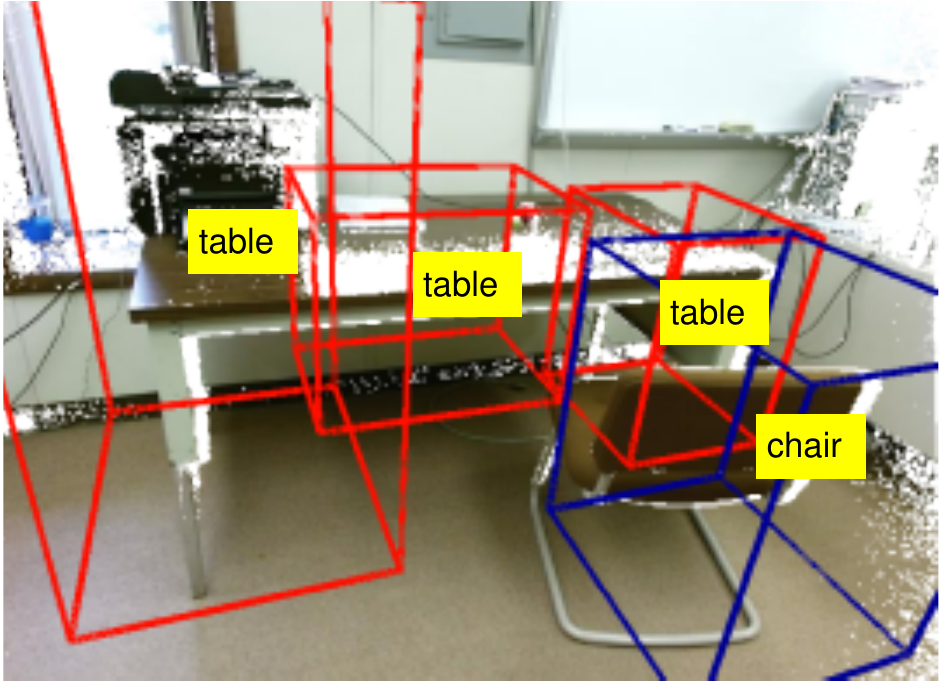} & 
\includegraphics[width=0.135\textwidth, height=1.2cm]{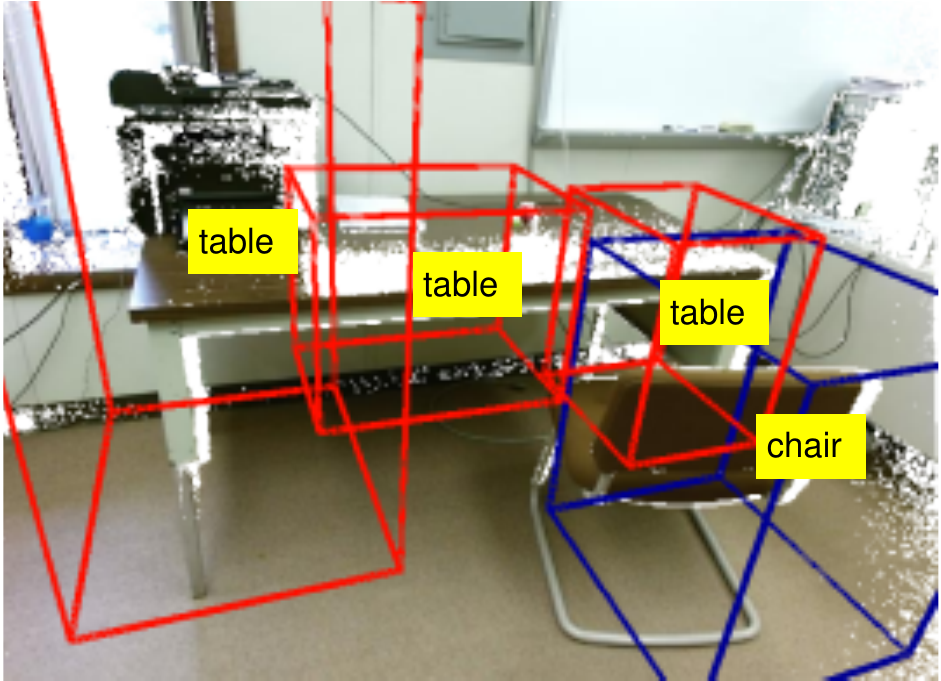} & 
\includegraphics[width=0.135\textwidth, height=1.2cm]{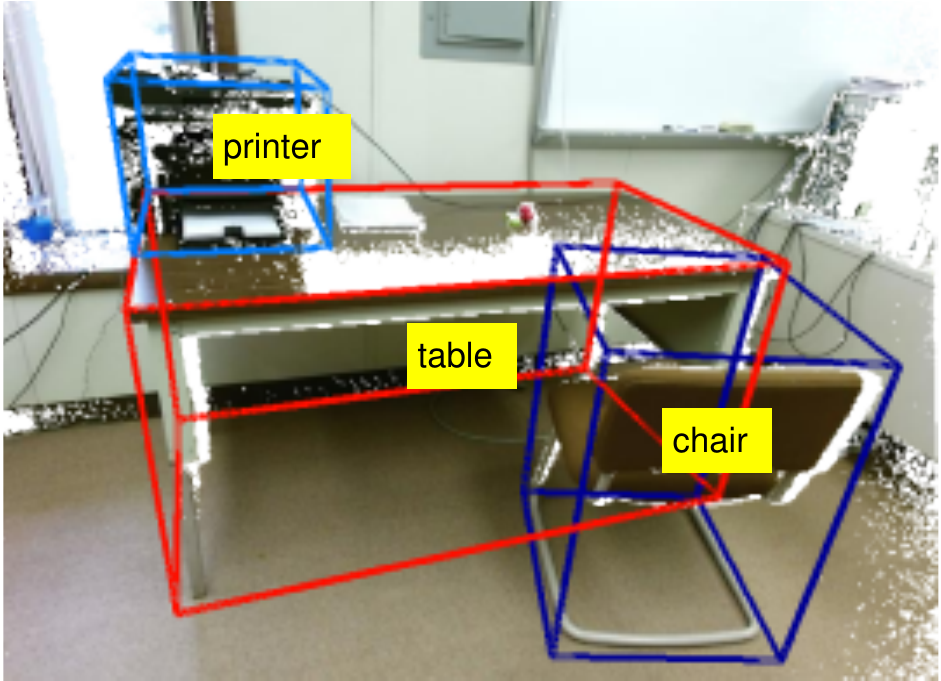}  
 %& IoU $ = 0.67 $ & IoU $ = 0.67 $ & IoU $ = 0.40 $ & IoU $ = 0.40 $ & IoU $ = 0.40 $ &  
\end{tabular} 
\caption[Results on SUNRGB inferred from RGB-D images]{A few results on SUNRGB Dataset inferred from the RGB-D images.}% \vspace{-1em}
\label{fig:rgbtoscene} 
\end{figure} 

\noindent The bounding box detector of DSS~\cite{song2016deep} is employed and the scores of the detection are updated based on our  reconstruction as follows: the score (confidence of the prediction) of a detected bounding box is doubled if a similar bounding box (in terms of shape and pose) of the same category is reconstructed by our method. A 3D non-maximum suppression is applied to the modified scores to get the final scene layout. The details can be found in the supplementary. 
 \begin{table}[H]\setlength{\tabcolsep}{11pt} % {0.52\textwidth}
%\vspace{-15pt} 
\caption{IoU for RGBD to room layout estimation}\label{wrap-tab:2}
\centering \scriptsize 
\begin{tabular}{@{\hspace{0.5em}}c@{\hspace{1.8em}}|c@{\hspace{1.6em}}|c@{\hspace{1.6em}}|c@{\hspace{1.6em}}|c@{\hspace{1.6em}}|c}\toprule    
Methods & {\bf SG-VAE}  & {\bf BL1} & {\bf BL2}~\cite{gomez2018automatic} & {\bf BL3}~\cite{kusner2017grammar}+\cite{yu2011make} & {\bf DSS~\cite{song2016deep}} \\\midrule
IoU & $\bm{0.4387}$ & ${0.4315}$ & $0.4056$ & $0.4259$ &  $0.4070$ \\ \bottomrule 
\end{tabular} %\vspace{-1em}
\end{table}

\noindent  We selected the average IoU for room layout estimation as the evaluation metric, and the results are presented in Table~\ref{wrap-tab:2}. % BL3 returns good numeric results as the solution is chosen from the best among $10$ samples. 
 The proposed method and other grammar-based baselines improve the scene layout estimation from the same by sophisticated methods such as deep sliding shapes~\cite{song2016deep}. Furthermore, the proposed method tackles the problem in a much simpler and faster way.  Thus, it can be employed to any 3D scene layout estimation method with very little overhead (\eg~ a few $ms$ in addition to $5.6s$ of~\cite{song2016deep}).
%Furthermore, our goal was to build a fast generative model for indoor scenes with an useful latent vector and the current experiment is to demonstrate the potential of this rather than benchmarking results of 3D scene layout estimation.
Results on some test images where SG-VAE produces better IoUs are displayed in Figure~\ref{fig:rgbtoscene}. %\vspace{-0.5em}
%Further applications of utilizing the learned latent space (such as multiple object detection and 3D pose estimation) are given in the supplementary material.

\section{Related Works}
The most relevant method to ours is Grains~\cite{li2019grains}. It requires training separate networks for each of the room-types---bedroom, office, kitchen etc. HC~\cite{qi2018human} is very slow  and takes a few minutes to synthesize a single layout. FS~\cite{ritchie2019fast} is fast, but still takes a couple of seconds. SceneGraphNet~\cite{Yang2019SceneGraphNet} predicts a probability distribution over object types that fits well in a target location given an incomplete layout. A similar graph-based method is proposed in~\cite{wang2019planit} and CNN-based method proposed in~\cite{wang2018deep} . A complete survey of the relavant method can be found in~\cite{zhang2019survey}. Note that all these methods are tailored to and trained on the synthetic SUNCG dataset which is currently unavailable. %and therefore can not be applied to any real-world applications. 

Koppula~\etal~\cite{koppula2011semantic} propose a graphical model that captures the local visual appearance and co-occurrences of different objects in the scene. They learn the appearance relationships among objects from the visual features that takes an RGB-D image as input and predicts 3D semantic labels of the objects as output. The pair-wise support relationships of the indoor objects are also exploited in~\cite{guo2013support,silberman2012indoor}. Learning to 3D scene synthesis from annotated RGB-D images is proposed in~\cite{kermani2016learning}. In the similar direction, an example-based synthesis of 3D object arrangements is proposed in~\cite{fisher2012example}. 

% The multiple view consistency of different objects for 3d scene-graph generation is proposed by Gay~\etal~\cite{gay2018visual}. They infer appearances (quadrics) of objects utilizing a message-passing algorithm that exchange messages, across different objects, consists of visual features of 2D bounding boxes. 
 
Grammar-based models for 3D scene reconstruction have been partially exploited before~\cite{zhao2011image,zhao2013scene}, \eg\  textured probabilistic grammar~\cite{li20183d}.  Zhao~\etal~\cite{zhao2011image} proposed handcoded grammar to its terminal symbols (line segments) and later extended to different functional groups in~\cite{zhao2013scene}. Choi~\etal~\cite{choi2013understanding} proposed a 3D geometric phrase model that estimates a scene layout with multiple object interactions. Note that all the above methods are based on hand-coded production rules, in contrast, the proposed method exploits a self-supervision to yield the production rules of the grammar.

\section{Conclusion}
We proposed a grammar-based autoencoder SG-VAE for generating natural indoor scene layouts containing multiple objects. By construction the output of SG-VAE always yields a valid configuration (w.r.t.\ the grammar) of objects, which was also experimentally confirmed. We demonstrated that the obtained latent representation of an SG-VAE has desirable properties such as the ability to interpolate between latent states in a meaningful way. The latent space of SG-VAE can also be easily adapted to computer vision problems (\eg\ 3D scene layout estimation from RGB-D images). Nevertheless, we believe that there is potential in leveraging the latent space of SG-VAEs to the other tasks, \eg\ fine-tuning the latent space for a consistent layout over multiple cameras which is part of the future work.

%involving multiple object detection and 3D pose estimation (see the supplementary material). %If our model is trained on a full 3D scenes, then given some                                                                                                                   visible objects SG-VAE should be able to infer the objects which are occluded / out of viewing frustum of the camera. 
%Further, in our current implementation, the appearance of objects is modelled using only tight 3D bounding boxes, and future work will extend our approach to utilize richer object representations based on 3D meshes or point-clouds. 
\section*{Acknowledgement}
We gratefully acknowledge the support of the Australian Research Council through the Centre of Excellence for Robotic Vision, CE140100016, and the Wallenberg AI, Autonomous Systems and Software Program (WASP) funded by the Knut and Alice Wallenberg Foundation.

\title{Supplementary Material: Scene Grammar Variational Autoencoder }

%\begin{comment}
\titlerunning{Supplementary Material: Scene Grammar Variational Autoencoder } 

\author{Pulak Purkait\inst{1} \orcidID{0000-0003-0684-1209} \and
Christopher Zach\inst{2} \orcidID{0000-0003-2840-6187} \and
Ian Reid\inst{1} \orcidID{0000-0001-7790-6423}}
\authorrunning{P. Purkait et al.}
\institute{Australian Institute of Machine Learning and School of Computer Science, \\ The University of Adelaide, Adelaide SA 5005, Australia 
%\email{\{pulak.purkait,ian.reid\}@adelaide.edu.au} 
\and Chalmers University of Technology, Goteborg 41296, Sweden 
}
%\end{comment}
%******************

\maketitle
%\tableofcontents
%\listoffigures

\section{Selection of the CFG---Algorithm details}
We aim to find suitable non-terminals and associated production rules that cover the entire dataset.  %The learned grammar is displayed in the following subsection. 
Note that finding such a set is a combinatorial hard problem. Therefore, we devise a greedy algorithm to select non-terminals and find approximate best coverage. Let $X_j$, an object category, be a potential non-terminal symbol and $\cR_j$ be the set of production rules derived from $X_j$ in the causal graph. Let $C_j$ be the set of terminals that $\cR_j$ covers (essentially nodes that $X_j$ leads to in the causal graph $\cG$). Our greedy algorithm begins with an empty set $\cR = \emptyset$ and chooses the node $X_j$ and associated production rule set $\cR_j$ to add that maximize the \emph{gain} in coverage % $\cG_{gain}(\cR_j, \cR) = \frac{1}{|\cR_j|} \sum_{I_i \in \cI \setminus \cC} |Y_i| / |I_i|$. %by adding the set of rules $\cR_j$ in the compact rule set $\cR$: 
\begin{equation}
\cG_{gain}(\cR_j, \cR) = \frac{1}{|\cR_j|} \sum_{I_i \in \cI \setminus \cC} |Y_i| / |I_i|
\label{eq:gain}
\end{equation}
where $\cC$ is the set of scenes that are already covered with a predefined fraction $p$ by the set of rules $\cR$, \ie $\cC = \{ I_i \;|\; \frac{|Y_i|}{|I_i|} > p\}$ where $Y_i$ is the set of terminal symbols occurs while parsing a scene $I_i$ by $\cR$.  The algorithm continues till no further nodes and associated rule set contribute a positive gain or until the current rule set covers the entire dataset with probability $p$ (chosen as $0.8$). We name our algorithm as \emph{p-cover} and is furnished in algorithm \ref{cover_algo}. Note that only object co-occurrences were utilized and object appearances were not incorporated in the proposed \emph{p-cover} algorithm. % In Figure~\ref{fig:concepts}, we also display a few non-terminals and associated production rules. 

To ensure the production rules to form a CFG, we select a few vertices (anchor nodes) of the causal graph and associate a number of non-terminals. A valid production rule is \emph{``an object category corresponding to an anchor node generates another object category it is adjacent to in $\cG$''.} Let us consider the set of anchor nodes forms our set of non-terminals $\Sigma$. The set of all possible objects including dummy \texttt{None} is defined as the set of terminal symbols $\cV$.  %and The rest of the section is how to pick those anchor nodes efficiently that constitutes a minimal set of rules sufficient to explain the entire scene dataset. %We utilize a greedy search method for this task. 

\begin{figure}
\centering 
\begin{tabular}{@{\hspace{-0.01cm}}c@{\hspace{0.01cm}}c@{\hspace{-0.01cm}}}
\includegraphics[width=0.49\textwidth,height=2.6cm]{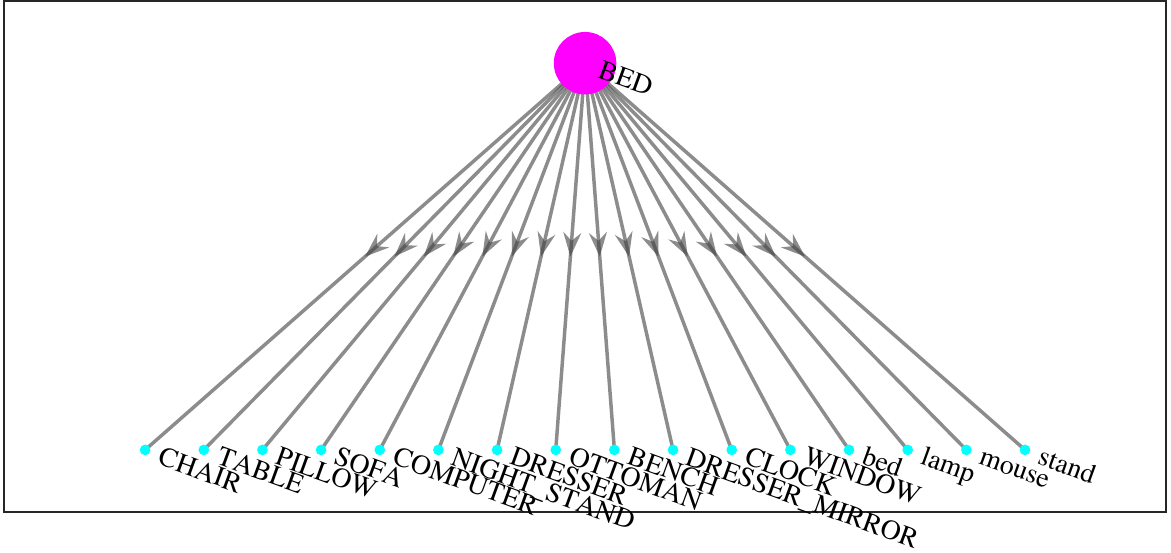} & 
\includegraphics[width=0.475\textwidth,height=2.6cm]{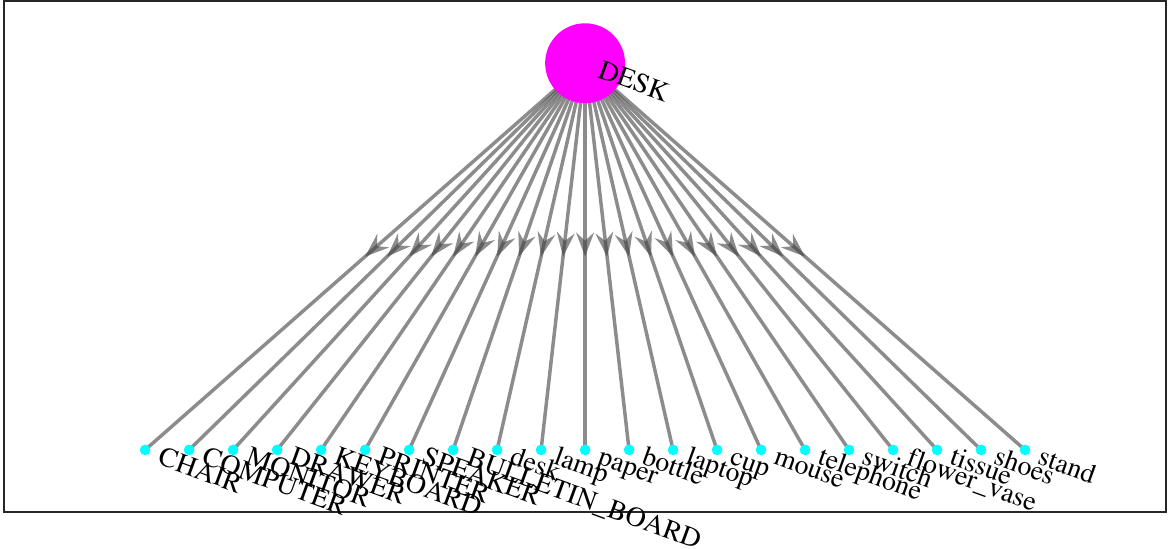} \\  
\includegraphics[width=0.49\textwidth,height=2.6cm]{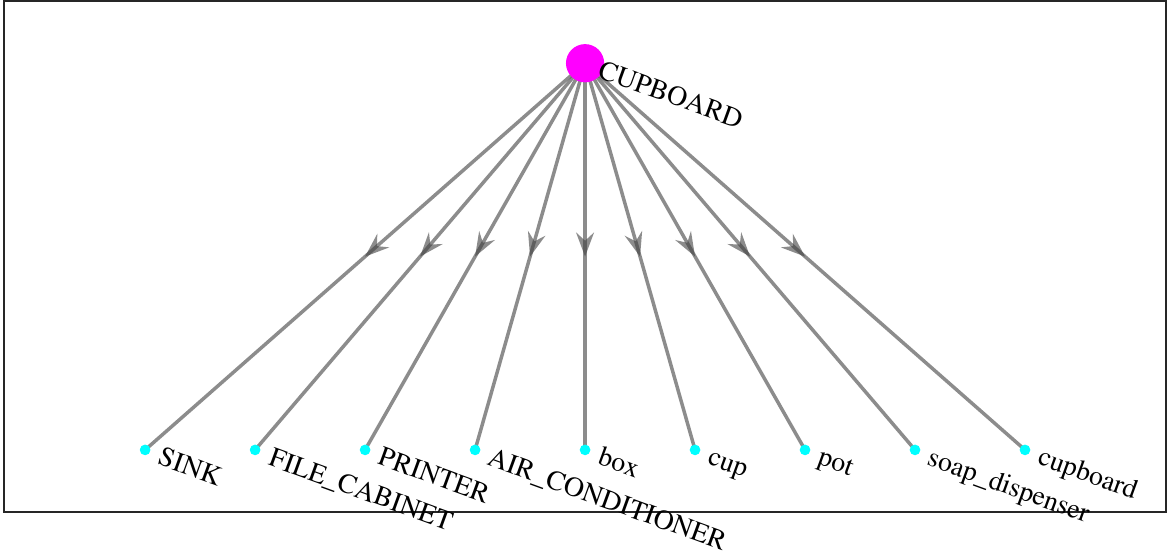} & 
\includegraphics[width=0.475\textwidth,height=2.6cm]{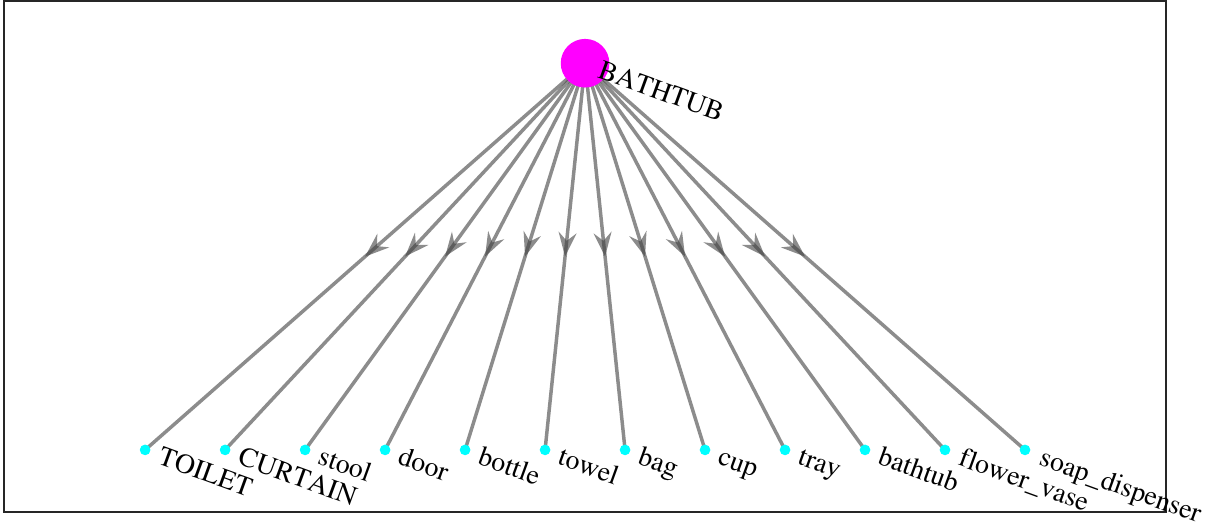} \\
\includegraphics[width=0.49\textwidth,height=2.6cm]{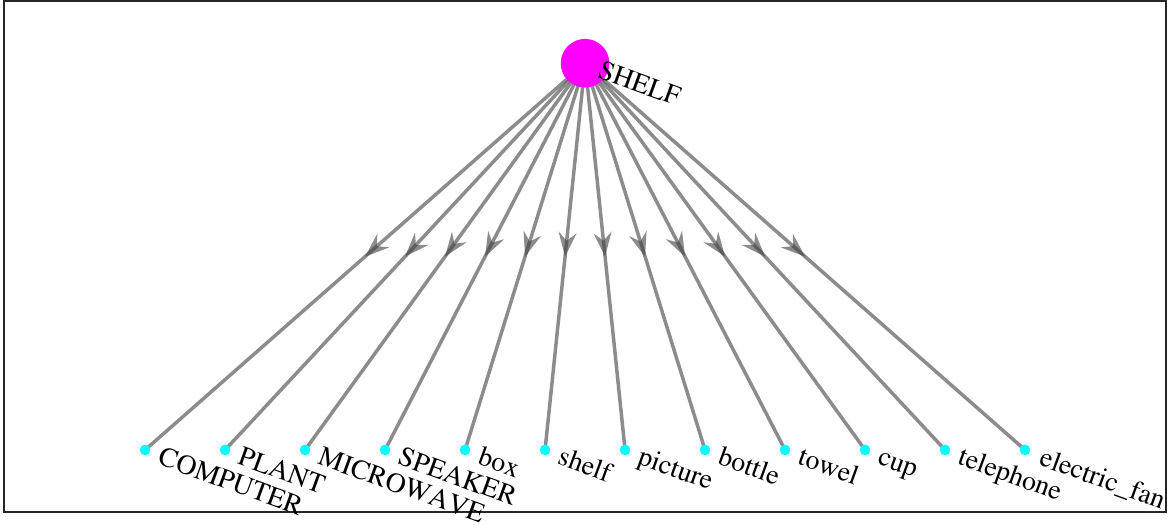} & 
\includegraphics[width=0.475\textwidth,height=2.6cm]{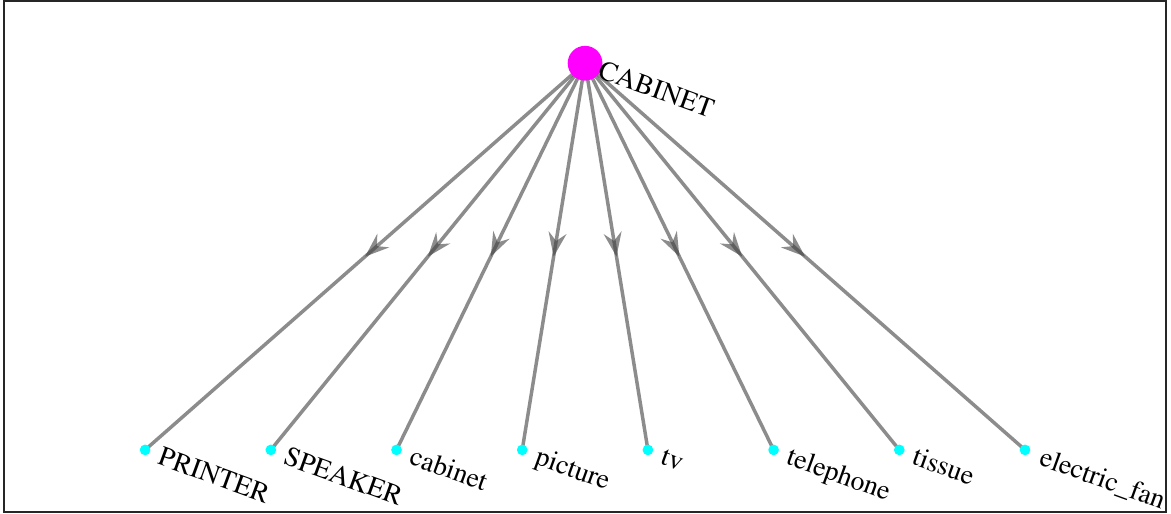} \\
\includegraphics[width=0.49\textwidth,height=2.6cm]{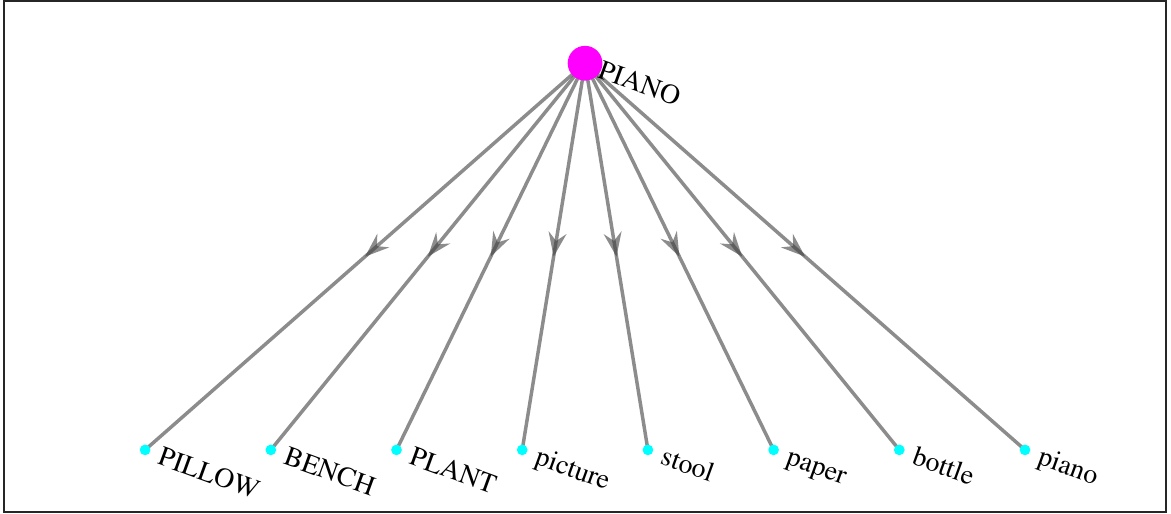} & 
\includegraphics[width=0.475\textwidth,height=2.6cm]{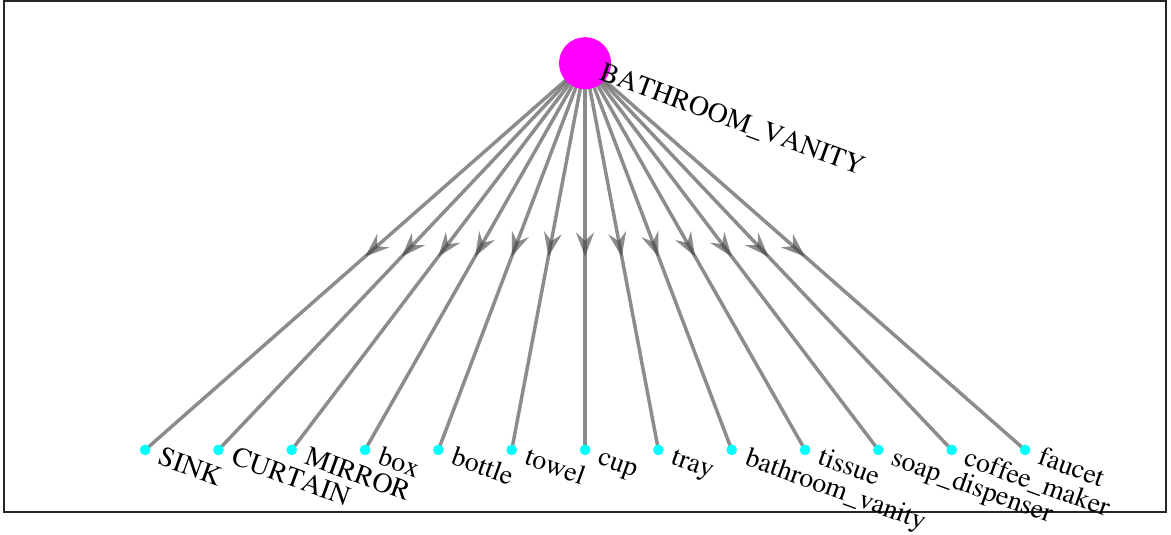} \\
\end{tabular}
\caption[Examples of the production rule-sets $R_j$]{A few examples of the production rule-sets $R_j$ which are utilized to generate the CFG. The detail algorithm is furnished in algorithm~\ref{cover_algo}. The non-terminal symbols are displayed in upper-case and terminal symbols are shown in lower-case.}
\vspace{4em}\label{fig:concepts}
\end{figure}

Let $\cR_j$ be the set of production rules derived from a non-terminal $X_j\text{\texttt{inc}}$ corresponding to an anchor object $X_j$, and $C_j$ be the set of terminals that $\cR_j$ covers (essentially nodes that $X_j$ leads to in $\cG$). Note that $\cR_j$ contains mainly four  types of production rules as described in [(R1)-(R4)] in the main text.  
A few examples of such set of production rules are displayed in Figure~\ref{fig:concepts}. 
The above strategy could lead to a large amount of production rules (ideally sum of number of the anchor points and the number of edges $|\cE|$ in the causal graph). Note that a large number of production rules increases the problem complexity. Contrarily, an arbitrary selection of few rules leads to a small number of derivable scenes (language of constituent grammar). We propose an algorithm to find a compressed (minimal size) set of rules to cover the entire dataset. Our underlying assumption is that the 
distribution of objects in the test scenes is very similar to the distribution of the same in the training scenes. 
Hence, we use the coverage of the training scenes as a proxy for the coverage of testing scenes and derive a 
probabilistic covering algorithm on the occurrences of objects in the dataset.

An example snippet for the grammar produced using this algorithm on the SUNRGBD dataset~\cite{song2015sun} is as follows: ~\\ ~\\
\begin{tabular}{cc}
\centering 
\begin{minipage}{0.5\textwidth}
%\setstretch{0.75} 
%{\scriptsize
\texttt{ \small 
\noindent \hspace{-0.3cm} S $\rightarrow$ scene SCENE \\
SCENE $\rightarrow$ bed  BED SCENE \\
BED $\rightarrow$ bed  BED \\
BED $\rightarrow$ lamp  BED \\
Bed $\rightarrow$ sofa  SOFA  BED\\
BED $\rightarrow$ pillow  PILLOW  BED\\
BED $\rightarrow$ night$\_$stand  BED\\
BED $\rightarrow$ dresser  BED\\ ...\\ 
BED $\rightarrow$ None ~~~ }
\end{minipage} 
& 
\begin{minipage}{0.5\textwidth}
%\setstretch{0.75} 
\texttt{ \small 
\noindent \hspace{-0.3cm}  SCENE $\rightarrow$ sofa sofainc SCENE\\
SOFA $\rightarrow$ sofa  SOFA\\
 SOFA $\rightarrow$ pillow PILLOW  SOFA\\
 SOFA $\rightarrow$ TOWEL  SOFA\\
 SOFA $\rightarrow$ None \\ 
...\\
SCENE $\rightarrow$ None}  
\end{minipage} %~~\\ $\cR_{bed}$ & $\cR_{sofa}$ 
\end{tabular} 
\\ ~\\
%In above, the suffix ``{\tt inc}'' is used to denote a non-terminal. 
The entire grammar is displayed in Section~\ref{sec:allrules}. In the above snippet the non-terminal \texttt{BED} generates another non-terminal \texttt{SOFA} that leads to an additional set of production rules corresponding to \texttt{SOFA}. Note that the grammar is right-recursive and not a regular grammar as some of the rules contain two non-terminals in the right hand side.  
 
Note that in this grammar non-terminal symbol \texttt{BED} generates another non-terminal \texttt{SOFA} which further leads to another set of production rules corresponding to \texttt{SOFA}. The grammar is right-recursive and not a regular grammar as some of the rules contain two non-terminals in the right hand side.

\section{Visualization of the latent space}
\begin{figure}[!ht]
\centering  
\includegraphics[width=0.8\textwidth]{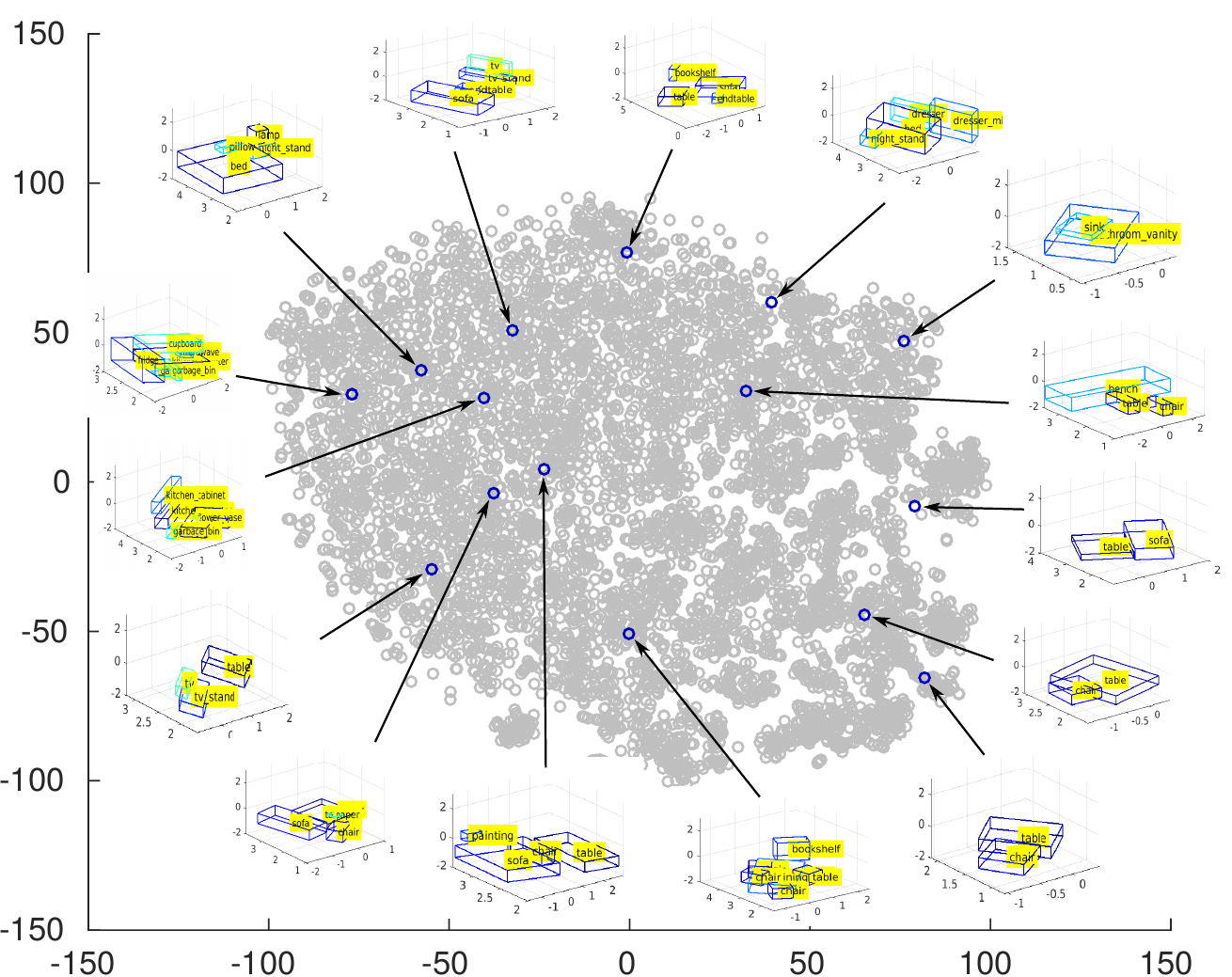}
\caption[2D projection of the mean $\bm{\mu}$ using t-SNE]{We plot the 2D projection of the mean $\bm{\mu}$ of the encoded distributions of the data (encoded with $50$ dimension) 
projected using t-SNE algorithm. $15$ chosen scenes are also displayed. Note that points with similar semantic concepts are clustered around a certain region. } \label{fig:encoded}
\end{figure}

To check the continuity of the latent space, the latent vectors (\ie, the mean $\bm{\mu}$ of the distribution $\cN(\bm{\mu}, \bm{\Sigma})$) is projected to 2D plane using data visualization algorithm t-SNE~\cite{maaten2008visualizing}. 
In Figure~\ref{fig:encoded}, we display $15$ different scenes in the latent space after t-SNE projection. We observe top left and top right regions are kitchen and bathroom scenes respectively. Whereas, top and bottom regions correspond to bedroom and dining room scenes respectively. The middle region mostly corresponds to living room and office scenes.  
Note that proposed SG-VAE not only considers the object co-occurrences, also considers object attributes (3D pose and shape parameters). For example, two scenes consists of a \emph{chair} and a \emph{table} are mapped to two nearby but distinct points. 

\section{Results on SUNCG}
 As mentioned in the paper, due to the ongoing dispute with the SUNCG dataset, we could not include these results in the main paper. The experiment is conducted only on our local copy of the dataset for the reviewing purposes. We extracted $32,765$ bedrooms, $14814$ kitchen, and $8,446$ office rooms from the dataset of $45,622$ synthetic houses. The dataset is divided into $80\%$ training, $10\%$ validation and remaining $10\%$ for testing. The bounding boxes and the relevant parameters are extracted using the scripts provided by Grains~\footnote{\href{https://github.com/ManyiLi12345/GRAINS}{https://github.com/ManyiLi12345/GRAINS}}. \vspace{1em}

\begin{figure}
\centering \scriptsize
\begin{tabular}{l@{\hspace{0.01em}}c@{\hspace{0.2em}}@{\hspace{0.2em}}c@{\hspace{0.2em}}@{\hspace{0.2em}}c@{\hspace{0.2em}}@{\hspace{0.2em}}c@{\hspace{0.2em}}} 
% & \multicolumn{4}{c}{SUNCG~\cite{song2015sun}2017semantic}} \\
%  \cmidrule[0.06em](r){2-2}\cmidrule[0.06em](lr){3-6} 
 \begin{picture}(1,25)\put(0, 8){\rotatebox{90}{~~Grains~\cite{li2019grains}}}\end{picture} ~~~&
 \includegraphics[width=0.22\textwidth, height=2.30cm, trim={1 1 1 1},clip]{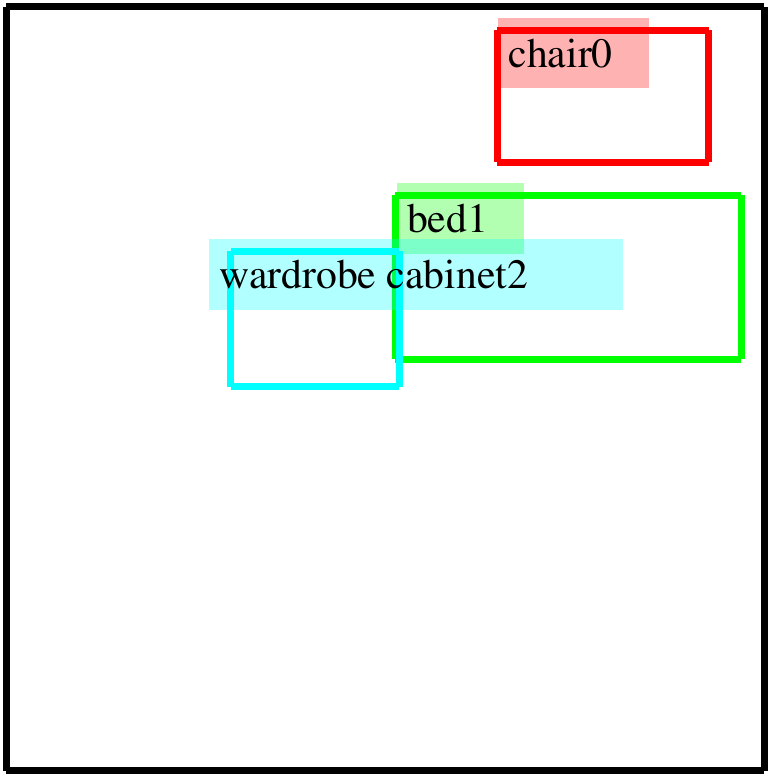} & 
\includegraphics[width=0.22\textwidth, height=2.30cm, trim={1 1 49 1},clip]{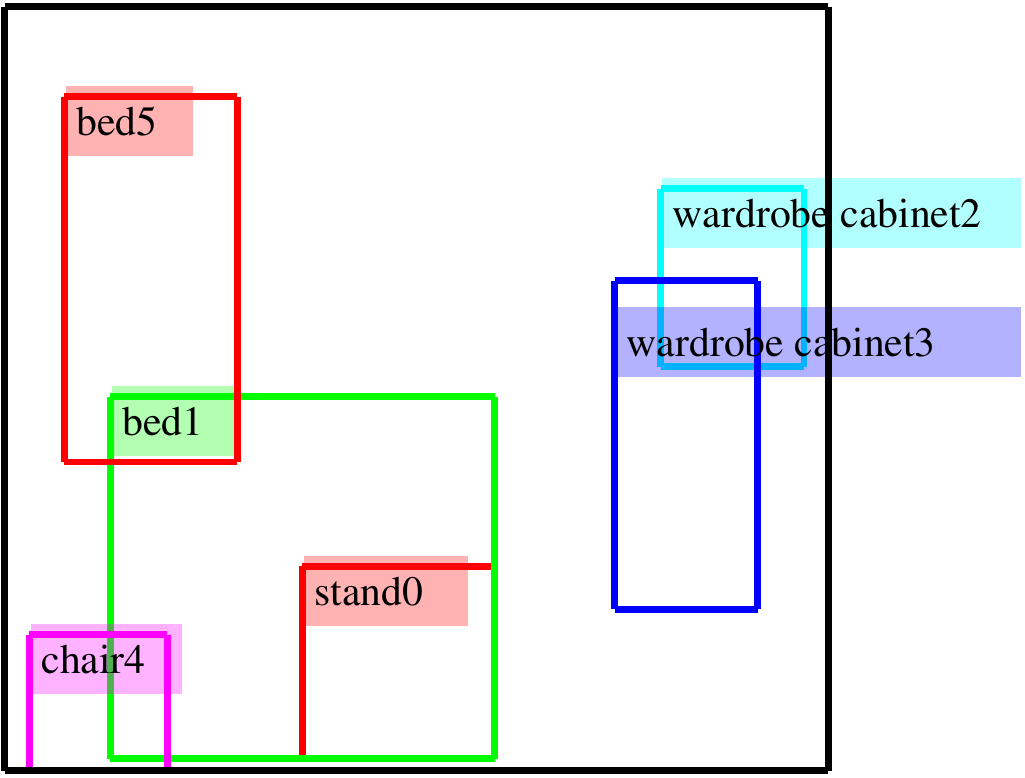} & 
\includegraphics[width=0.22\textwidth, height=2.30cm, trim={1 1 72 1},clip]{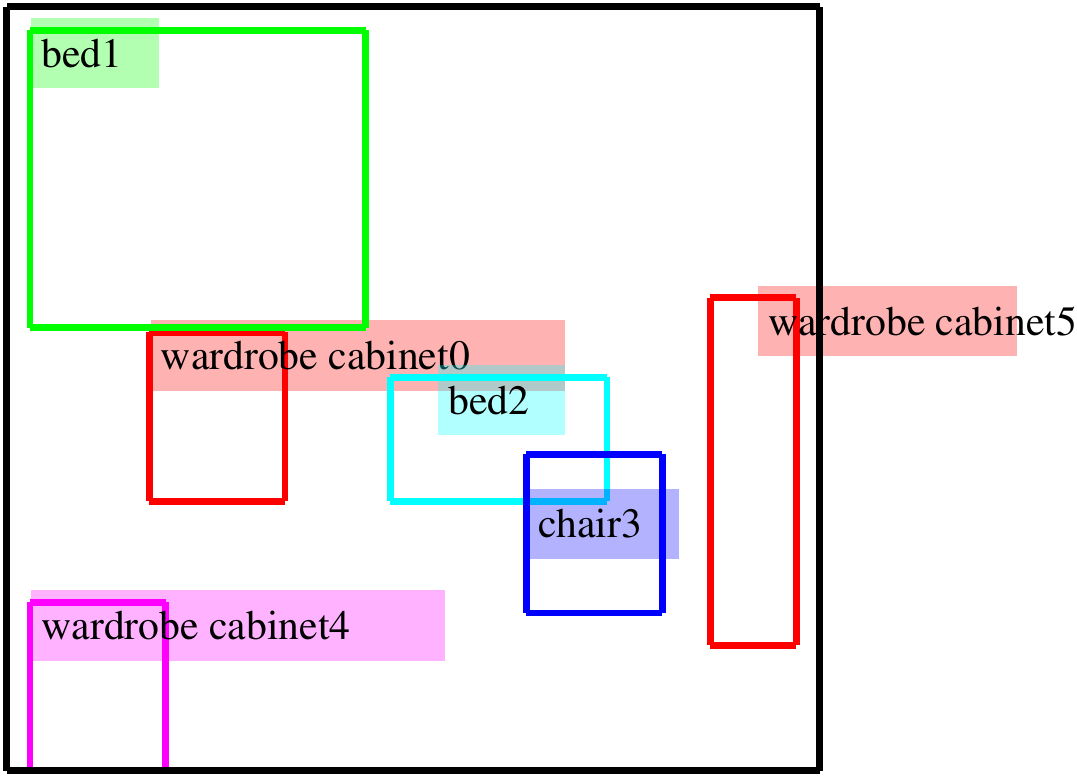} & 
\includegraphics[width=0.22\textwidth, height=2.30cm, trim={1 1 80 1},clip]{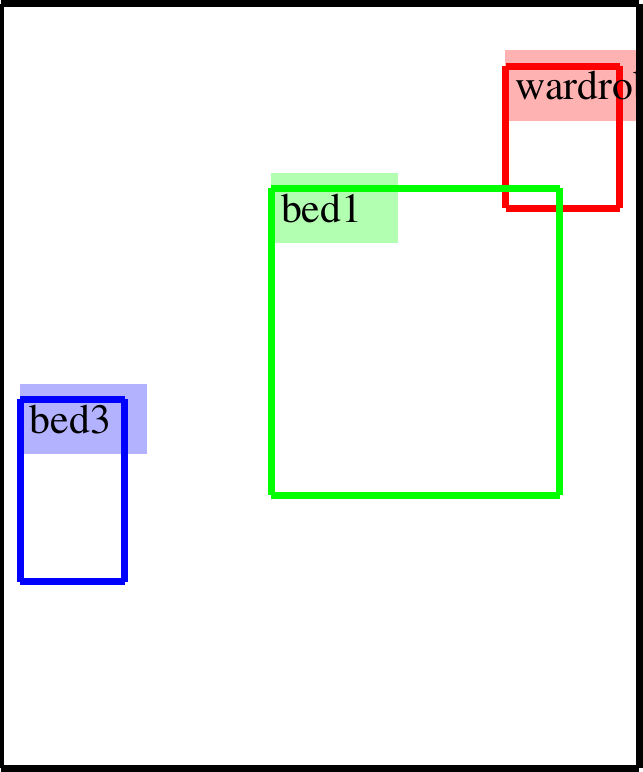} \\
% \begin{picture}(1,25)\put(0, 8){\rotatebox{90}{Bed Room 2}}\end{picture} &
% \includegraphics[width=0.22\textwidth, height=2.0cm, trim={1 1 18 1},clip]{2dproj/org_45-crop.pdf} & 
%\includegraphics[width=0.22\textwidth, height=2.0cm, trim={1 1 4.5 1},clip]{2dproj/19-crop.pdf} & 
%\includegraphics[width=0.22\textwidth, height=2.0cm, trim={1 1 32.5 1},clip]{2dproj/fast_synth1-crop.pdf} & 
%\includegraphics[width=0.22\textwidth, height=2.0cm, trim={1 1 36 1},clip]{2dproj/sample_0000000061-crop.pdf} \\
 \begin{picture}(1,25)\put(0, 14){\rotatebox{90}{SG-VAE}}\end{picture} &
 \includegraphics[width=0.22\textwidth, height=2.30cm, trim={1 1 31 5},clip]{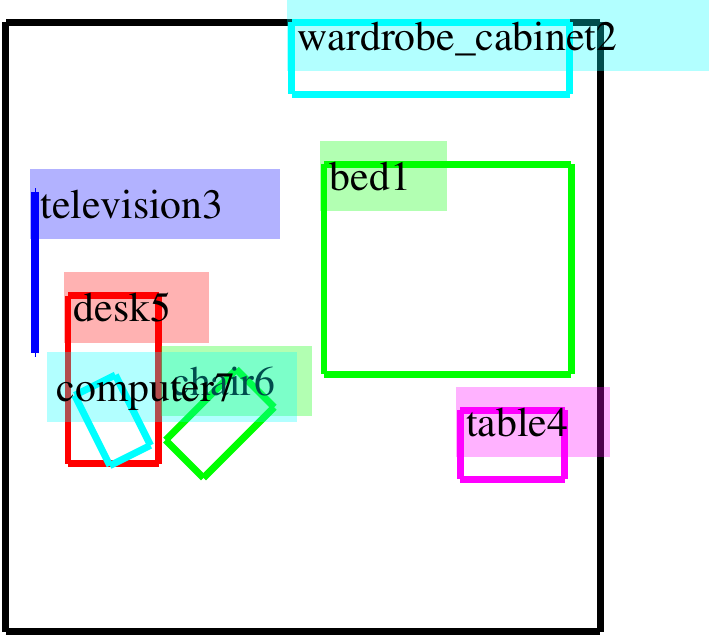} & 
\includegraphics[width=0.22\textwidth, height=2.30cm, trim={1 1 41 1},clip]{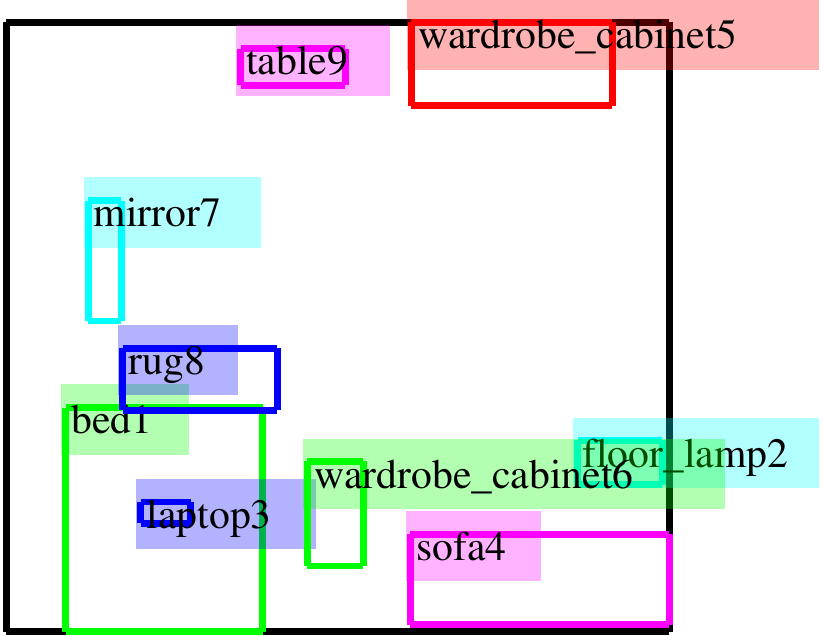} & 
\includegraphics[width=0.22\textwidth, height=2.30cm, trim={1 1 52 1},clip]{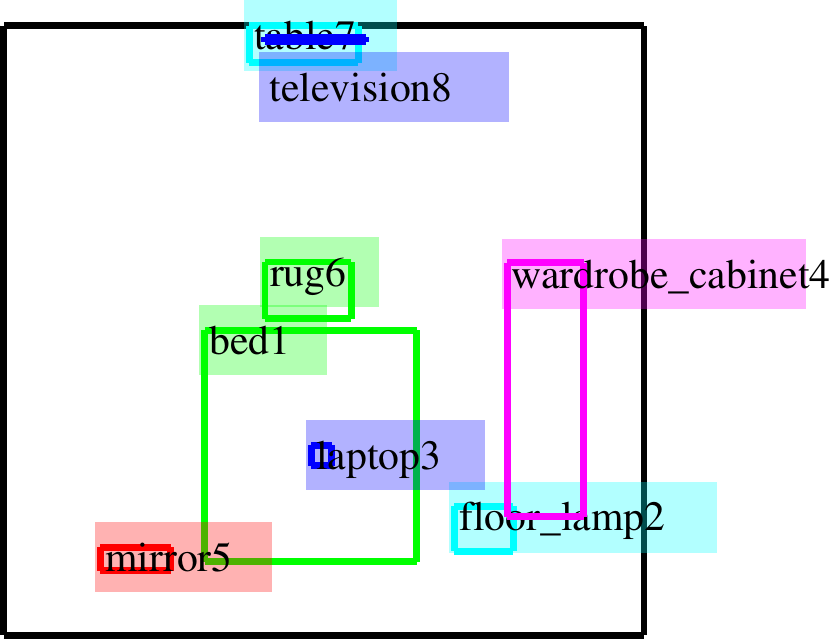} & 
\includegraphics[width=0.22\textwidth, height=2.30cm, trim={1 1 32 1},clip]{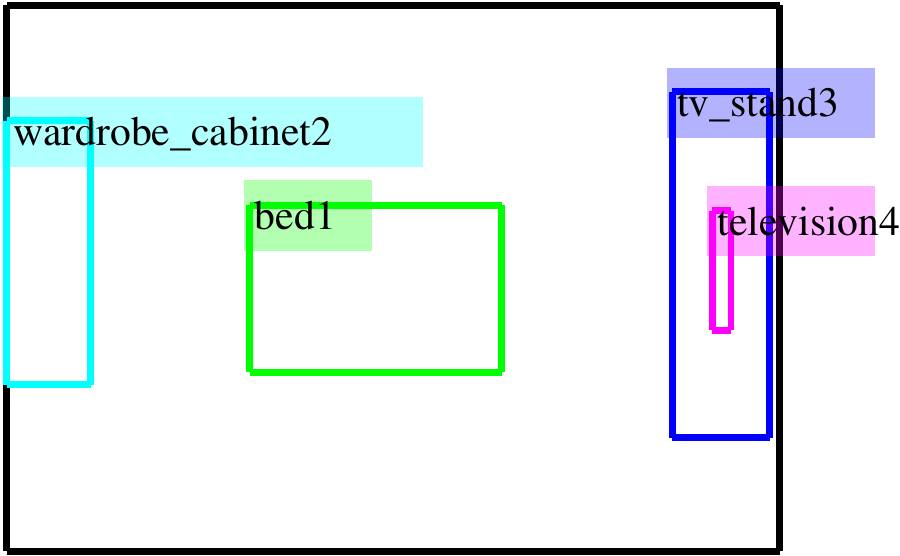}  \\ 
& Sample 1 & Sample 2 & Sample 3 & Sample 4 
\end{tabular} 
\caption[1-1 comparison with Grains on SUNCG datases]{1-1 comparison with Grains~\cite{li2019grains} on SUNCG dataset. Similar samples are chosen for a precise comparison. Most similar synthetic scenes are displayed. We observe that the proposed method SG-VAE produces more variety of objects in a scene than the baseline Grains.} 
\label{fig:grains}
\end{figure}

\noindent {\bf Baseline comparison} For visual comparison, some examples of the scene synthesized by the proposed method along with the baselines on SUNCG are shown in Figure~\ref{fig:2dproj2}.\footnote{We thank the authors of GRAINS~\cite{li2019grains} and HC~\cite{qi2018human} for sharing the code and the authors of FS~\cite{ritchie2019fast} for the results displayed in Figure~\ref{fig:2dproj2}.} Our results are similar to Grains and better than the other baselines in terms of co-occurrences and appearances (pose and shape) of different objects. A detailed 1-1 comparison with Grains is also shown in Figure~\ref{fig:grains}. The quantitative comparisons are provided in the main manuscript. \vspace{1em}

\noindent {\bf Interpolation in the latent space} Similar to the interpolation results on SUN RGB-D, shown in the main manuscript, an additional experiment is conducted on SUNCG dataset. Here, synthetic scenes are decoded from linear interpolations $\alpha \bm{\mu_1} + (1 - \alpha)\bm{\mu_2}$ of the means $\bm{\mu_1}$ and $\bm{\mu_2}$ of the latent distributions of two separate scenes. The generated scenes are valid in terms of the co-occurrences of the object categories and their shapes and poses. %The room-size and the camera view-point are fixed for better visualization. 

\vspace{2em}

\begin{figure}
\centering \scriptsize
\begin{tabular}{l@{\hspace{0.01em}}c@{\hspace{0.2em}}@{\hspace{0.2em}}c@{\hspace{0.2em}}@{\hspace{0.2em}}c@{\hspace{0.2em}}@{\hspace{0.2em}}c@{\hspace{0.2em}}} 
% & \multicolumn{4}{c}{SUNCG~\cite{song2017semantic}} \\
%  \cmidrule[0.06em](r){2-2}\cmidrule[0.06em](lr){3-6} 
 \begin{picture}(1,25)\put(0, 8){\rotatebox{90}{Bed Room}}\end{picture} &
 \includegraphics[width=0.22\textwidth, height=2.3cm, trim={1 1 1 5},clip]{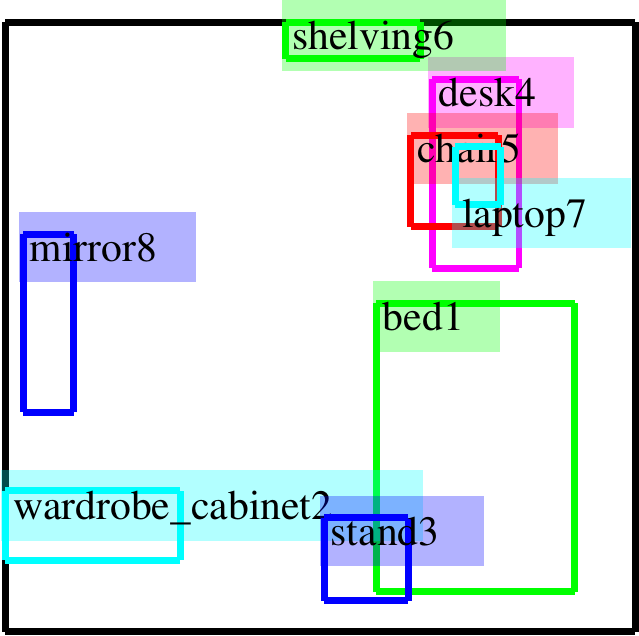} & 
\includegraphics[width=0.22\textwidth, height=2.3cm, trim={1 1 4.5 1},clip]{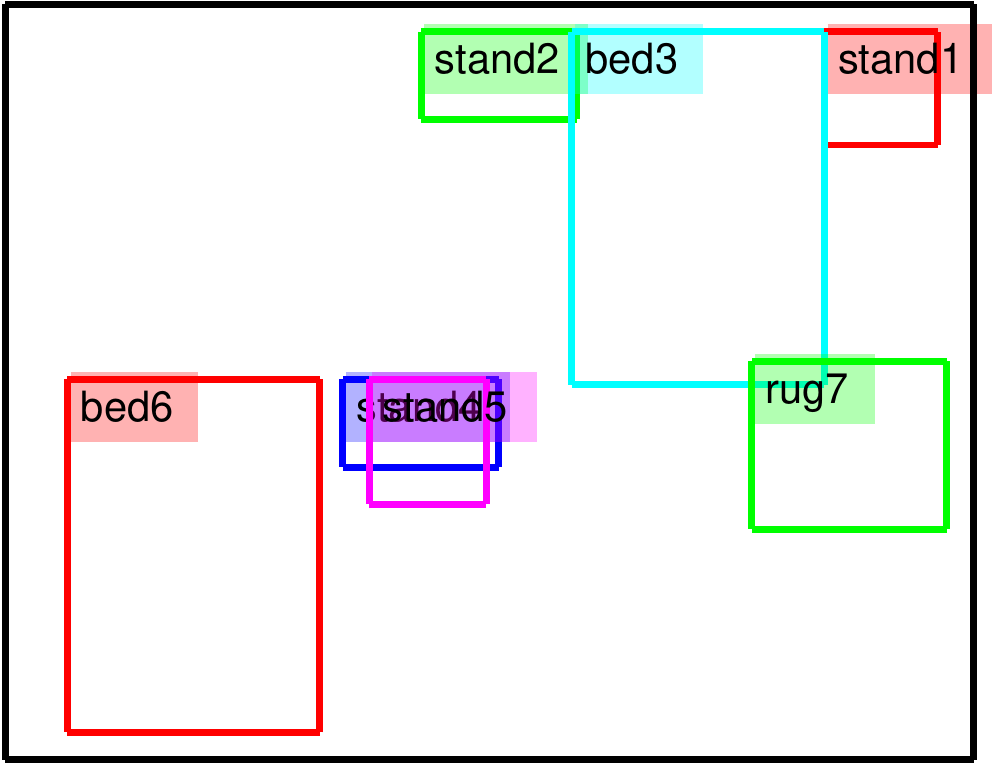} & 
\includegraphics[width=0.22\textwidth, height=2.3cm, trim={1 1 32.5 1},clip]{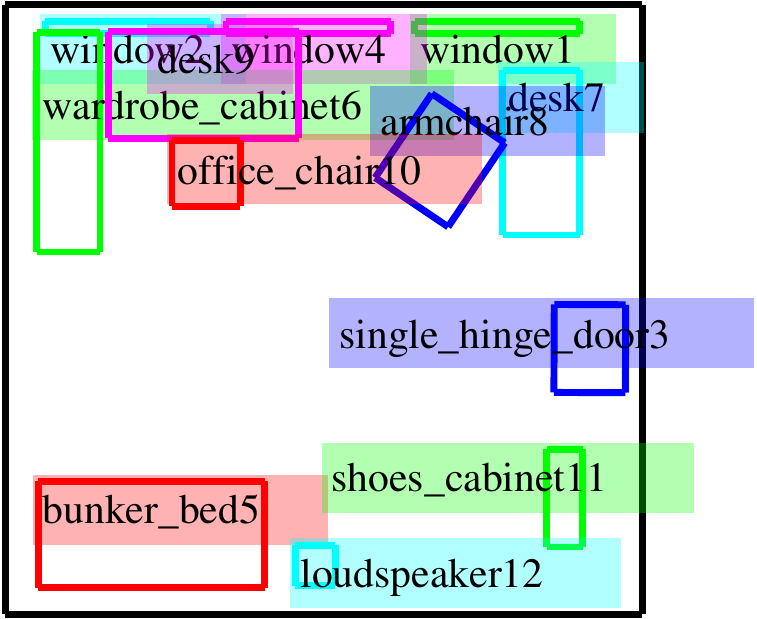} & 
\includegraphics[width=0.22\textwidth, height=2.3cm, trim={1 9 1 1},clip]{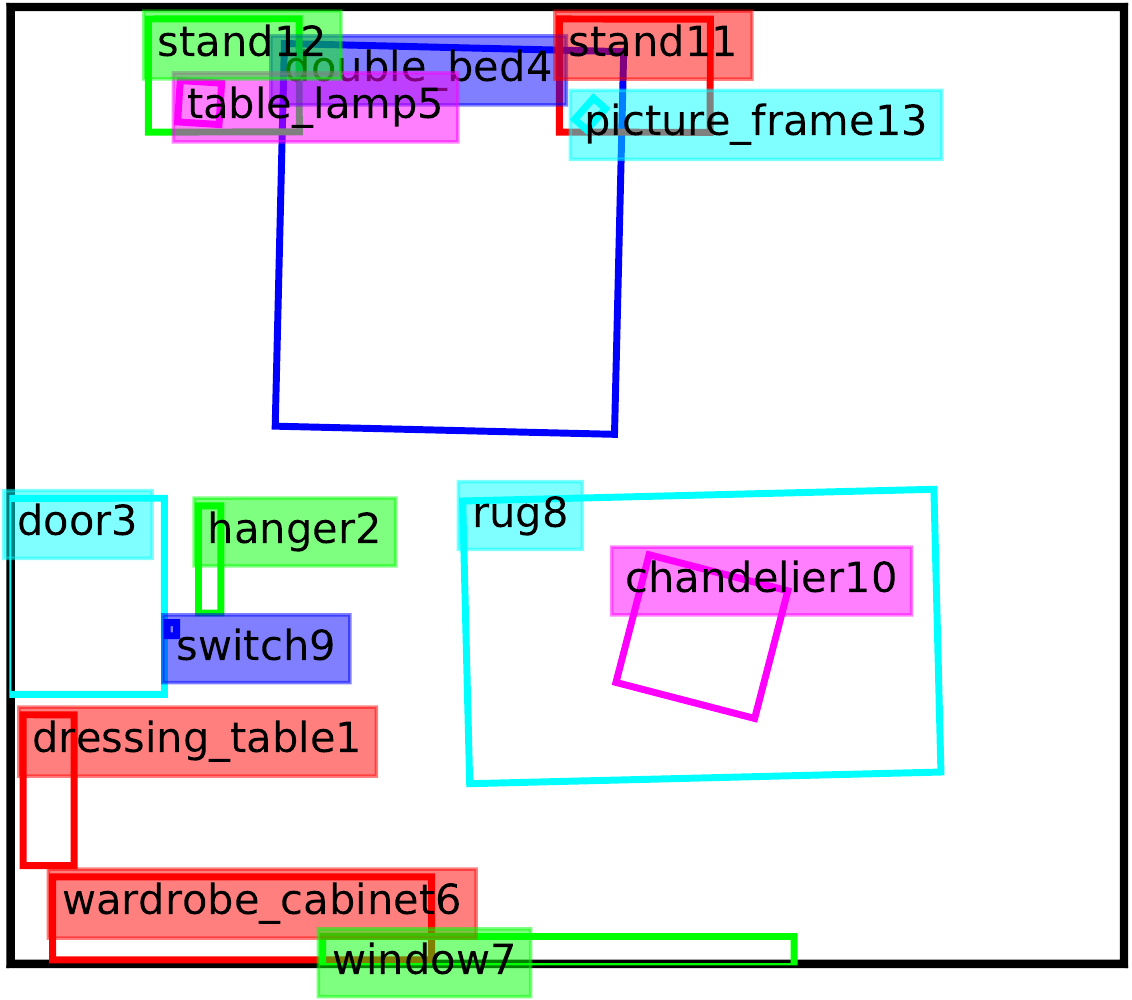} \\
% \begin{picture}(1,25)\put(0, 8){\rotatebox{90}{Bed Room 2}}\end{picture} &
% \includegraphics[width=0.22\textwidth, height=2.0cm, trim={1 1 18 1},clip]{2dproj/org_45-crop.pdf} & 
%\includegraphics[width=0.22\textwidth, height=2.0cm, trim={1 1 4.5 1},clip]{2dproj/19-crop.pdf} & 
%\includegraphics[width=0.22\textwidth, height=2.0cm, trim={1 1 32.5 1},clip]{2dproj/fast_synth1-crop.pdf} & 
%\includegraphics[width=0.22\textwidth, height=2.0cm, trim={1 1 36 1},clip]{2dproj/sample_0000000061-crop.pdf} \\
 \begin{picture}(1,25)\put(0, 14){\rotatebox{90}{Office}}\end{picture} &
\includegraphics[width=0.22\textwidth, height=2.3cm, trim={1 1 1 1},clip]{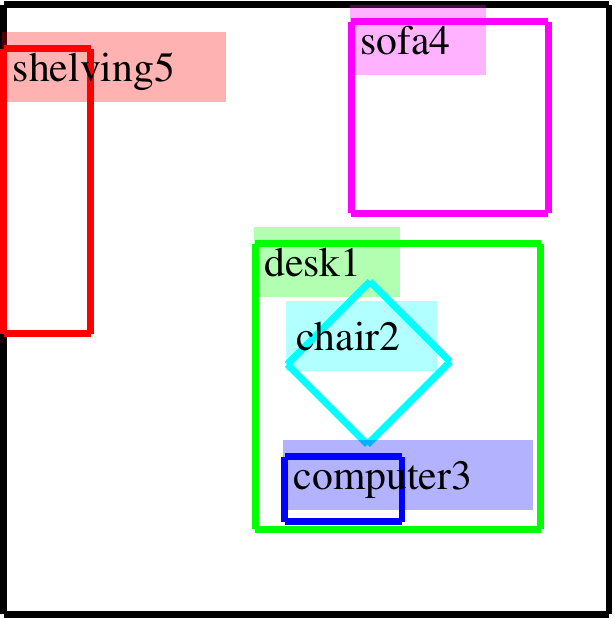} & 
\includegraphics[width=0.22\textwidth, height=2.3cm, trim={1 1 1 2},clip]{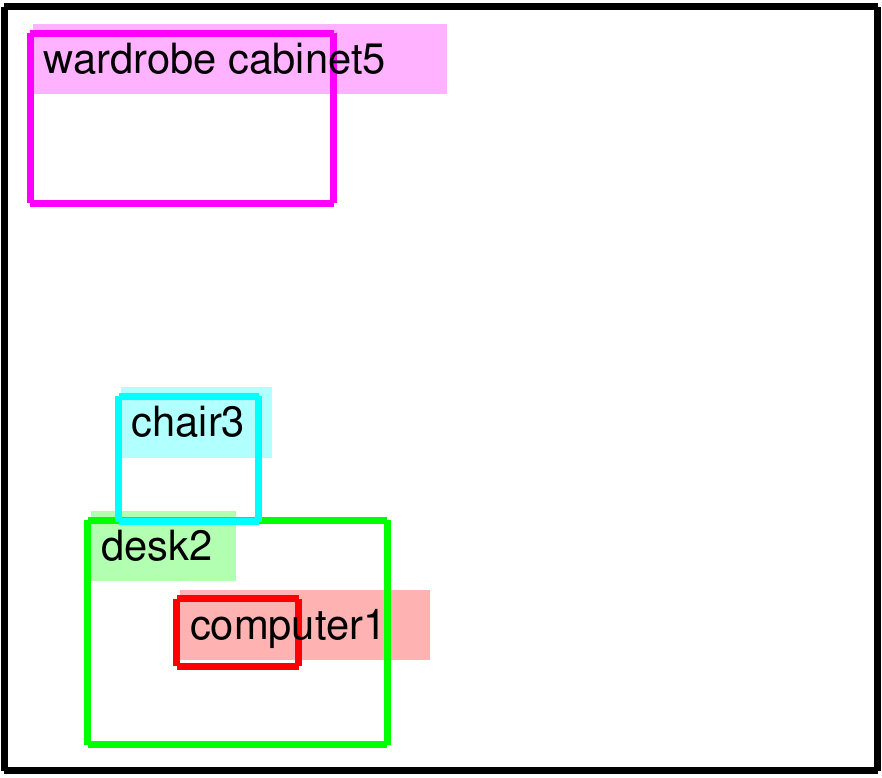} & 
\includegraphics[width=0.22\textwidth, height=2.3cm, trim={1.5 1 23.0 1.5},clip]{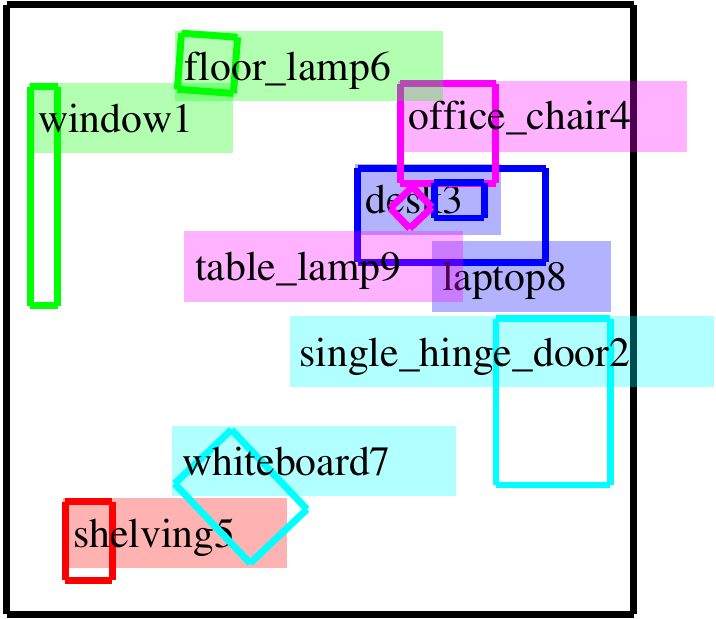} & 
\includegraphics[width=0.22\textwidth, height=2.3cm, trim={1 1 9 1},clip]{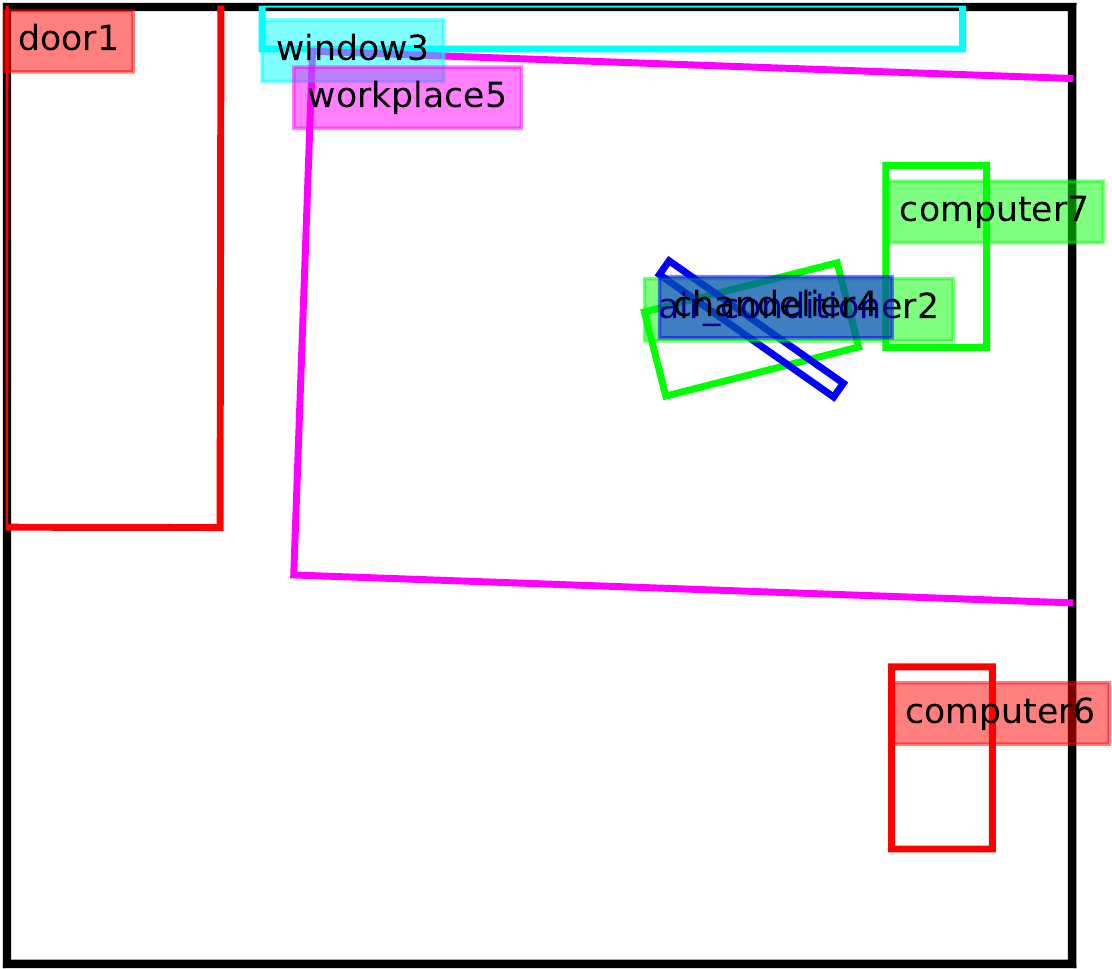} \\ 
% \begin{picture}(1,25)\put(0, 12){\rotatebox{90}{Office 2}}\end{picture} &
%\includegraphics[width=0.22\textwidth, height=2.0cm, trim={0.5 1 8 1},clip]{2dproj/org_6253-crop.pdf} & 
%\includegraphics[width=0.22\textwidth, height=2.0cm, trim={1 1 1 2},clip]{2dproj/375-crop.pdf} & 
%\includegraphics[width=0.22\textwidth, height=2.0cm, trim={1.5 1 23.0 1.5},clip]{2dproj/fast_synth2-crop.pdf} & 
%\includegraphics[width=0.22\textwidth, height=2.0cm, trim={1 1 9 1},clip]{2dproj/sample_0000000007-crop.pdf} \\ 
\begin{picture}(1,25)\put(0, 12){\rotatebox{90}{Kitchen}}\end{picture} &
\includegraphics[width=0.22\textwidth, height=2.30cm, trim={1 1 18 1},clip]{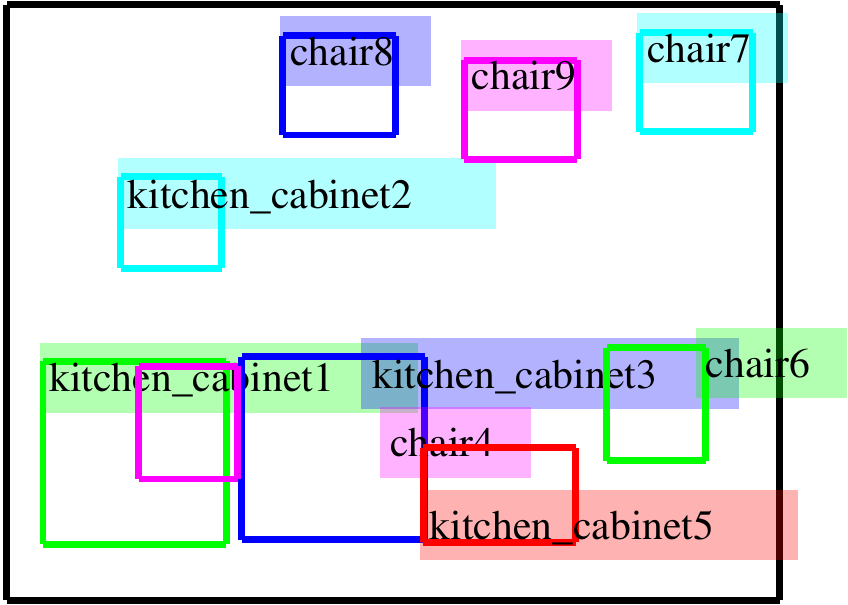} & 
\includegraphics[width=0.22\textwidth, height=2.30cm, trim={9 1 2.5 2.5},clip]{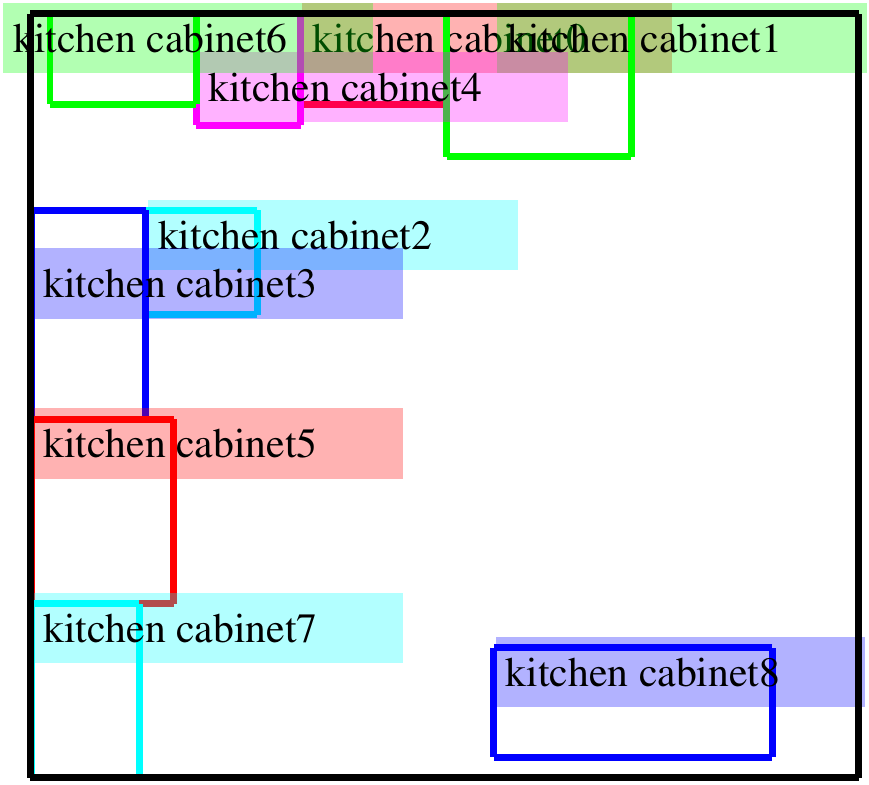} & 
\includegraphics[width=0.22\textwidth, height=2.30cm, trim={1 1 55.5 1},clip]{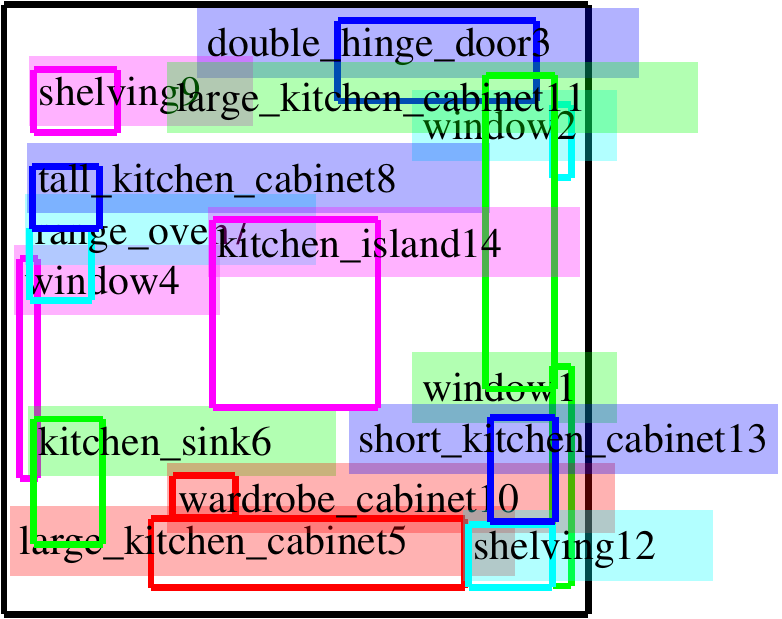} & 
\includegraphics[width=0.22\textwidth, height=2.30cm, trim={1 1 1 1},clip]{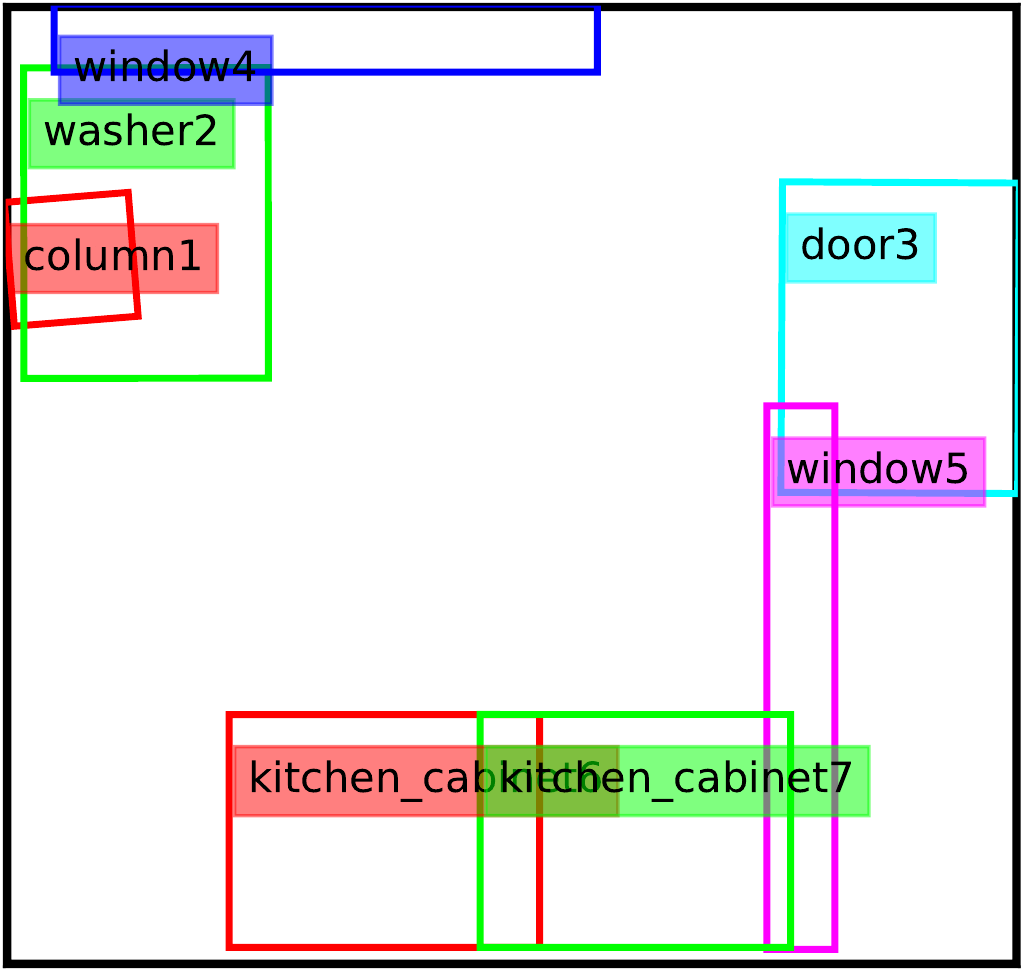} \\
%\begin{picture}(1,25)\put(0, 10){\rotatebox{90}{Kitchen 2}}\end{picture} &
%\includegraphics[width=0.22\textwidth, height=2.0cm, trim={0.5 1 1 1},clip]{2dproj/org_3212-crop.pdf} & 
%\includegraphics[width=0.22\textwidth, height=2.0cm, trim={9 1 2.5 2.5},clip]{2dproj/125-crop.pdf} & 
%\includegraphics[width=0.22\textwidth, height=2.0cm, trim={1 1 55.5 1},clip]{2dproj/fast_synth3-crop.pdf} & 
%\includegraphics[width=0.22\textwidth, height=2.0cm, trim={1 1 1 1},clip]{2dproj/sample_0000000206-crop.pdf} \\
~~~~~& (a) {SG-VAE} & (c) Grains~\cite{li2019grains} & (d) FS~\cite{ritchie2019fast}  & (e) HC~\cite{qi2018human} 
\end{tabular} 
\caption[Sampled views by baseline methods on the SUNCG datases]{Top-views of the synthesized scenes generated by the proposed and the baseline indoor scene synthesis methods on SUNCG datasets. Note that these are just some random samples taken from the generated scenes.} 
\label{fig:2dproj2}
\end{figure} %\vspace{-2em}

 %\vspace{-2em}

\begin{figure}[!ht] \tiny \centering 
\begin{tabular}{@{\hspace{0.01em}}l@{\hspace{1.0em}}c@{\hspace{0.3em}}c@{\hspace{0.3em}}c@{\hspace{0.3em}}c@{\hspace{0.3em}}c} \vspace{-0.1cm} \\ %\vspace{0.01cm}
  \begin{picture}(1,25)\put(0, 5){\rotatebox{90}{($\alpha = 0.9$)}}\end{picture} & 
\includegraphics[width=0.18\textwidth, height=1.71cm, trim={1 1 8 1},clip ]{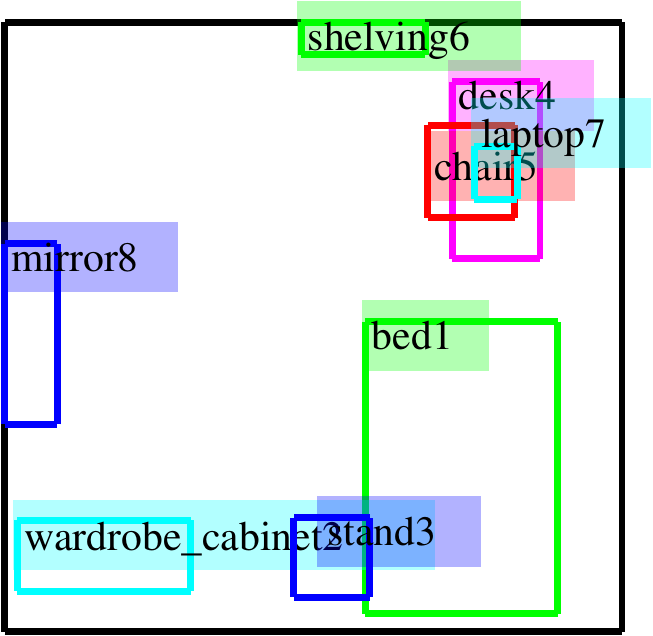} & 
  \includegraphics[width=0.18\textwidth, height=1.71cm , trim={0 0 10 0},clip]{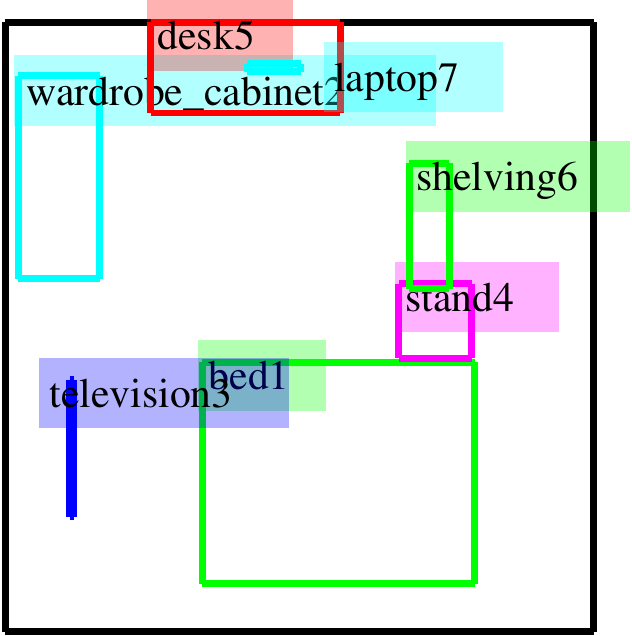}  & 
\includegraphics[width=0.18\textwidth, height=1.71cm, trim={0 0 63 8},clip ]{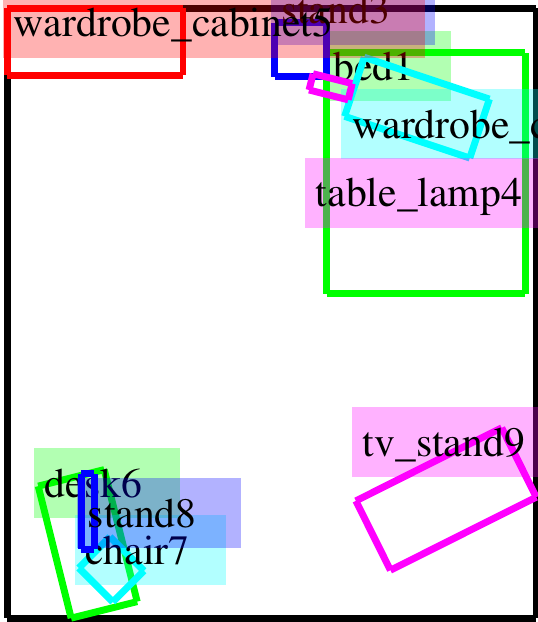}  & 
\includegraphics[width=0.18\textwidth, height=1.71cm ]{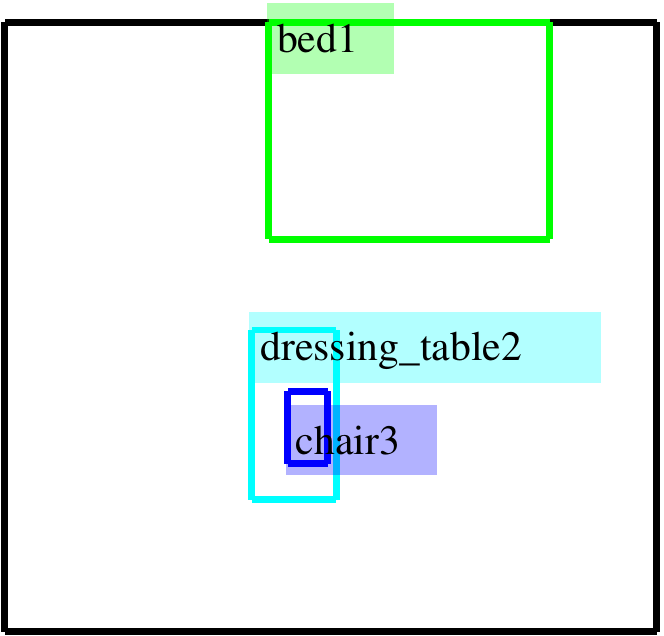}  & 
\includegraphics[width=0.18\textwidth, height=1.71cm ]{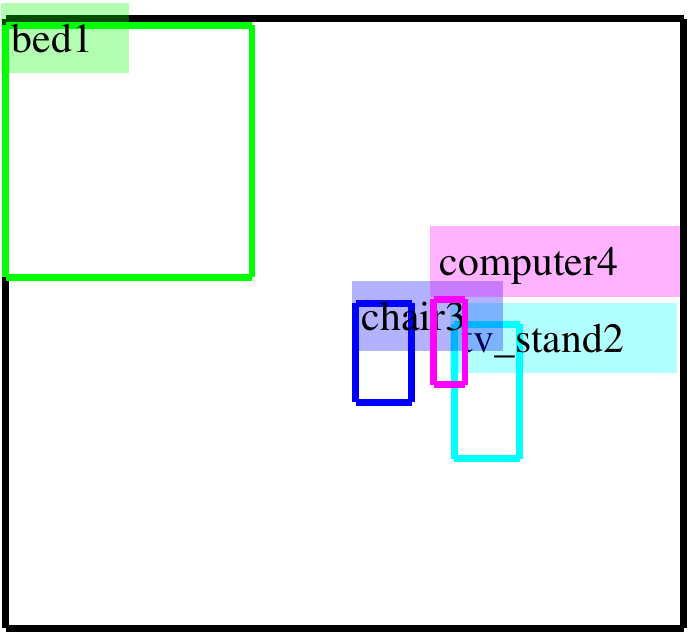}   \\ 
  \begin{picture}(1,25)\put(0, 5){\rotatebox{90}{($\alpha = 0.8$)}}\end{picture} & 
\includegraphics[width=0.18\textwidth, height=1.71cm ]{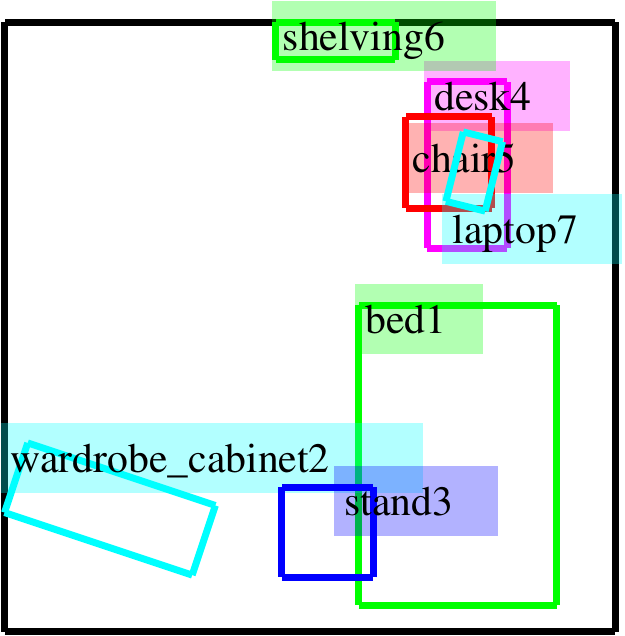} & 
  \includegraphics[width=0.18\textwidth, height=1.71cm , trim={0 0 10 0},clip]{suncg/sample2/pred_test_2D_244-crop.pdf}  & 
\includegraphics[width=0.18\textwidth, height=1.71cm, trim={0 0 63 8},clip ]{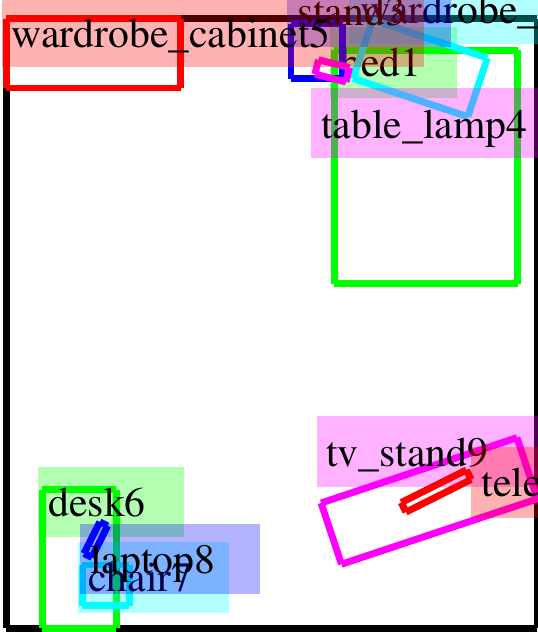}  & 
\includegraphics[width=0.18\textwidth, height=1.71cm ]{suncg/sample6/pred_test_2D_738-crop.pdf}  & 
\includegraphics[width=0.18\textwidth, height=1.71cm ]{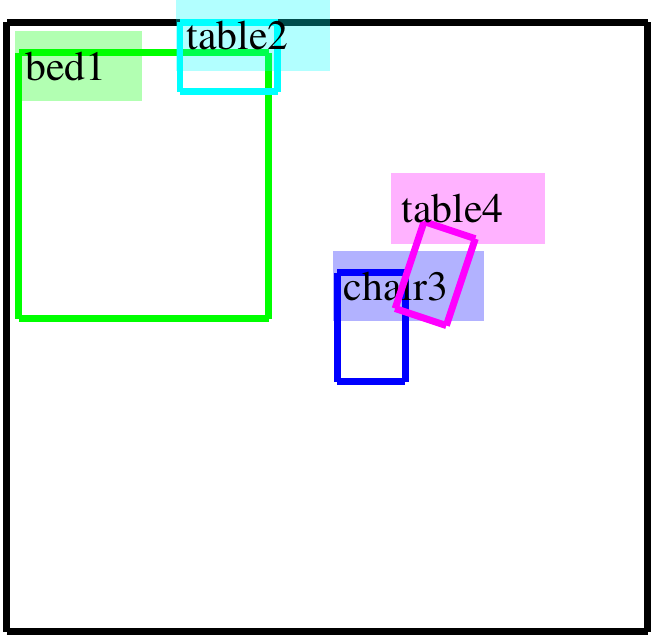}   \\ 

  \begin{picture}(1,25)\put(0, 5){\rotatebox{90}{($\alpha = 0.7$)}}\end{picture} & 
\includegraphics[width=0.18\textwidth, height=1.71cm, trim={1 1 48 1},clip]{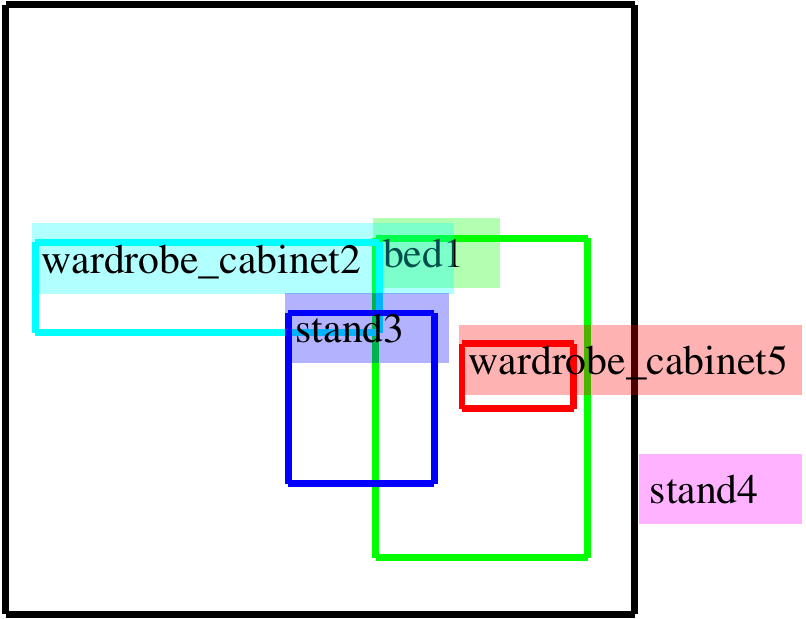} & 
  \includegraphics[width=0.18\textwidth, height=1.71cm , trim={0 0 42 0},clip]{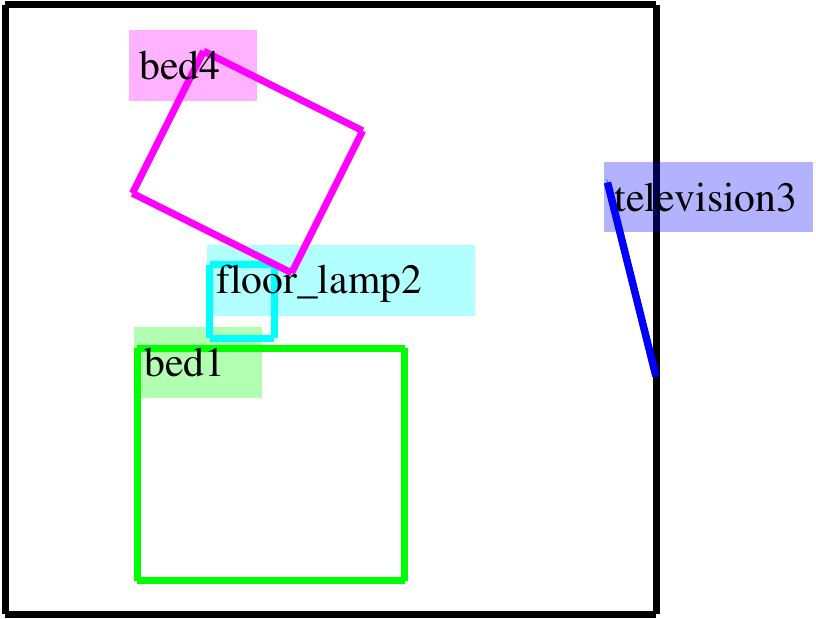}  & 
\includegraphics[width=0.18\textwidth, height=1.71cm , trim={0 0 61 8},clip]{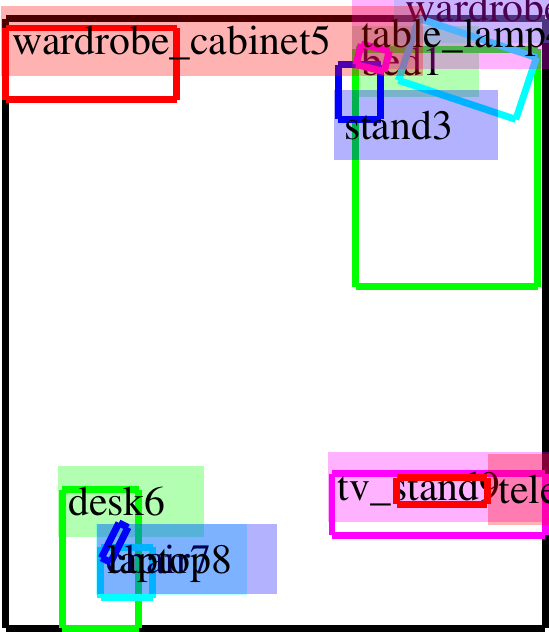}  & 
\includegraphics[width=0.18\textwidth, height=1.71cm, trim={1 1 15 1},clip ]{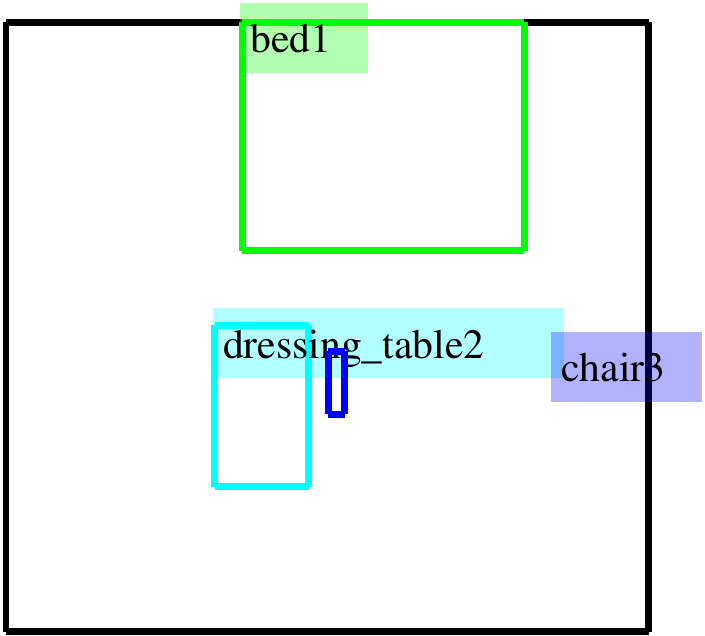}  & 
\includegraphics[width=0.18\textwidth, height=1.71cm ]{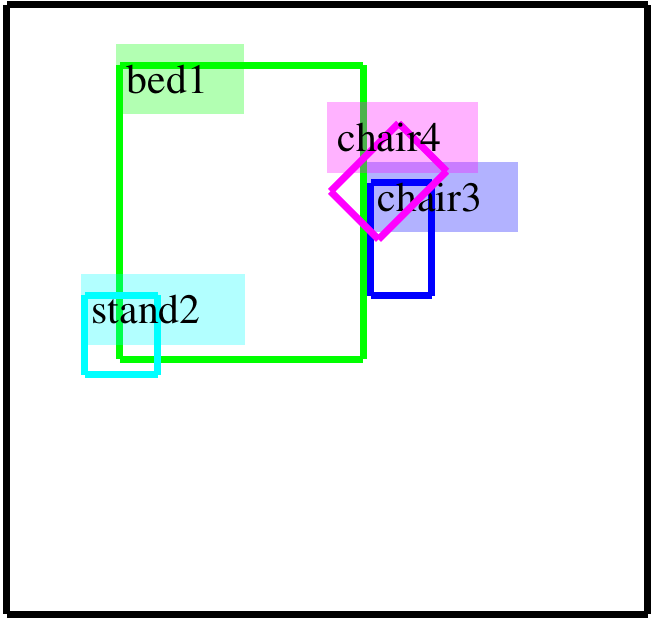}   \\ 

  \begin{picture}(1,25)\put(0, 5){\rotatebox{90}{($\alpha = 0.6$)}}\end{picture} & 
\includegraphics[width=0.18\textwidth, height=1.71cm ]{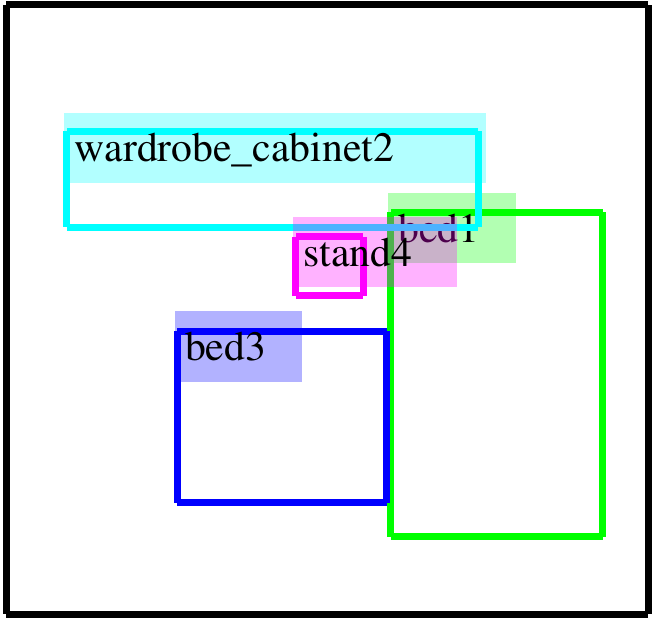} & 
  \includegraphics[width=0.18\textwidth, height=1.71cm , trim={0 0 42 0},clip]{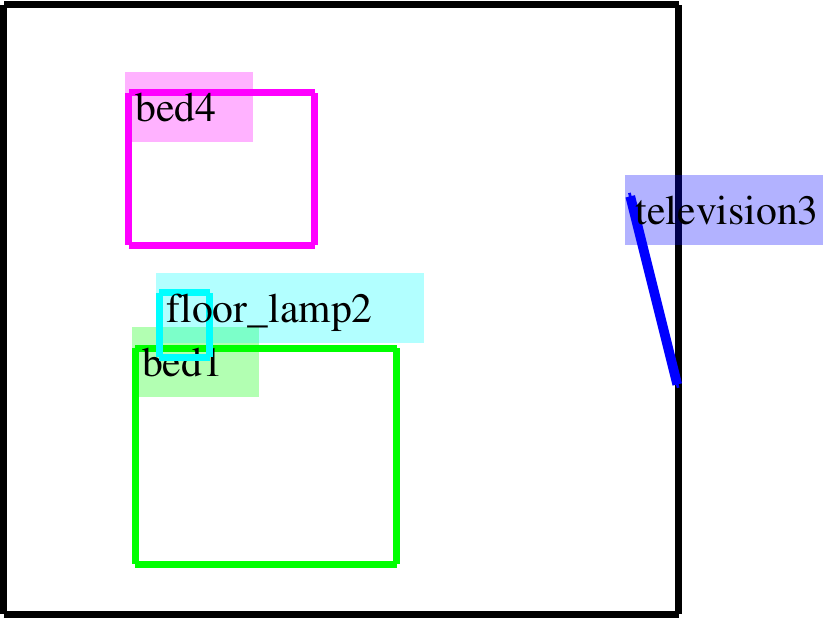}  & 
\includegraphics[width=0.18\textwidth, height=1.71cm , trim={0 0 46 8},clip]{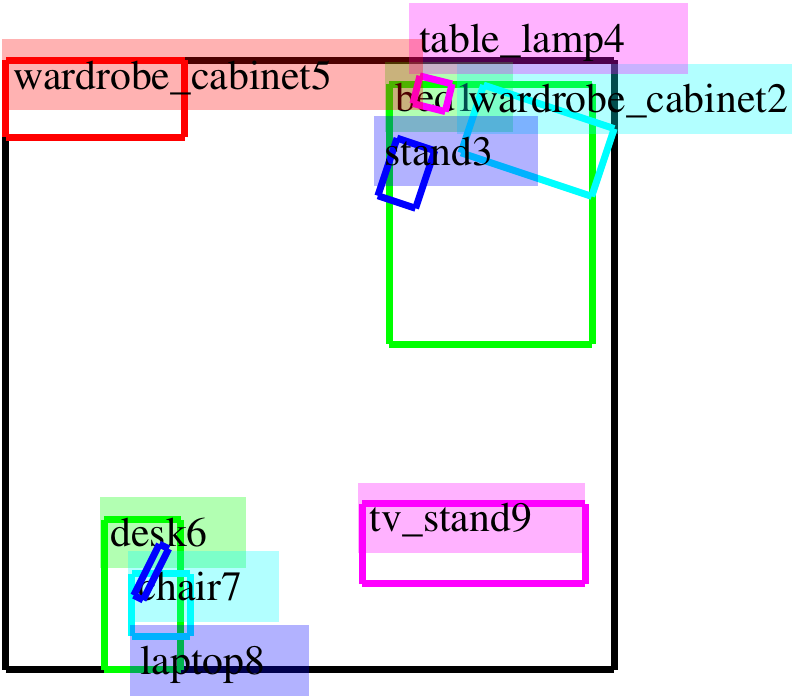}  & 
\includegraphics[width=0.18\textwidth, height=1.71cm ]{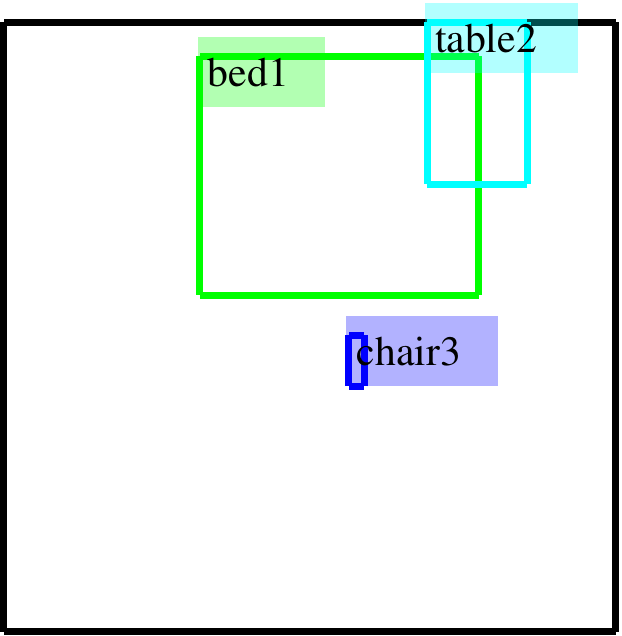}  & 
\includegraphics[width=0.18\textwidth, height=1.71cm ]{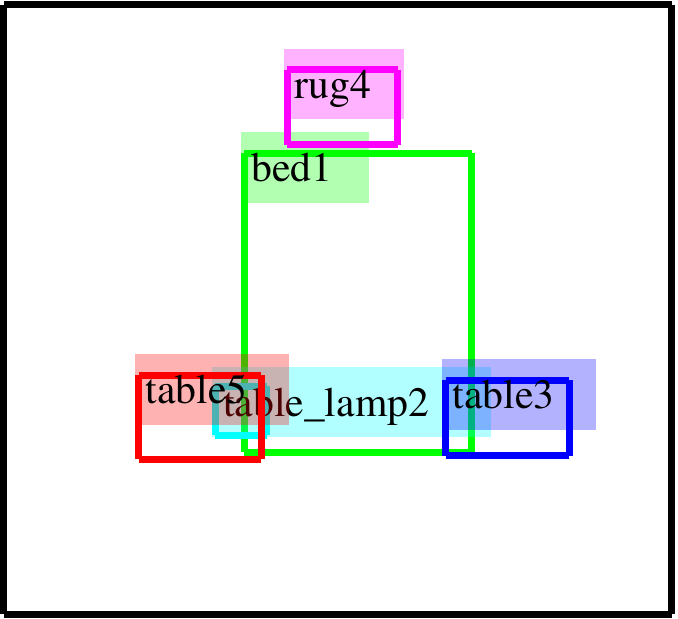}   \\ 

  \begin{picture}(1,25)\put(0, 5){\rotatebox{90}{($\alpha = 0.5$)}}\end{picture} & 
\includegraphics[width=0.18\textwidth, height=1.71cm ]{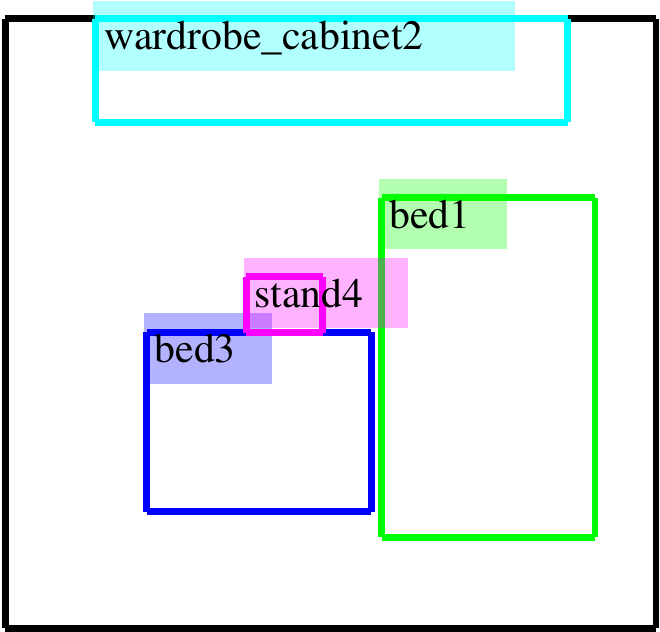} & 
  \includegraphics[width=0.18\textwidth, height=1.71cm , trim={0 0 42 0},clip]{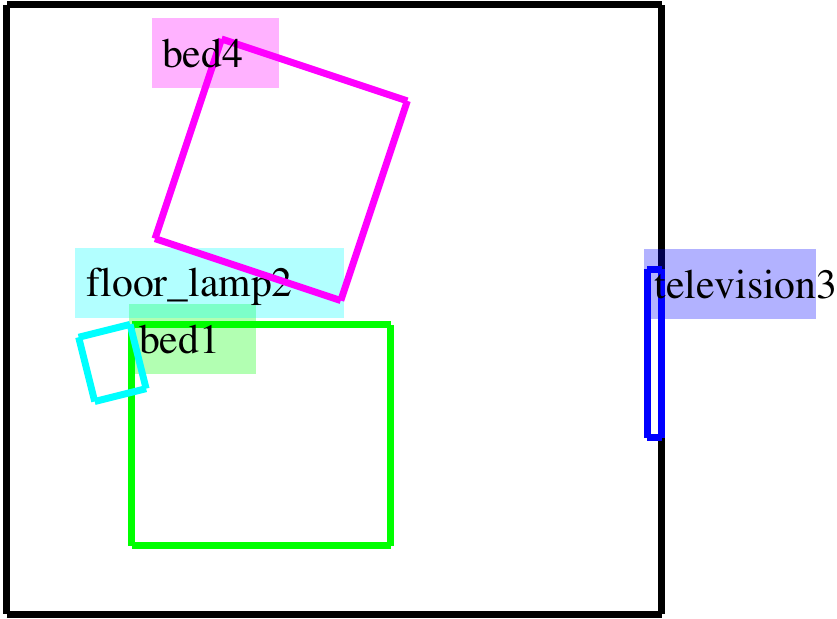}  & 
\includegraphics[width=0.18\textwidth, height=1.71cm , trim={0 0 46 3},clip]{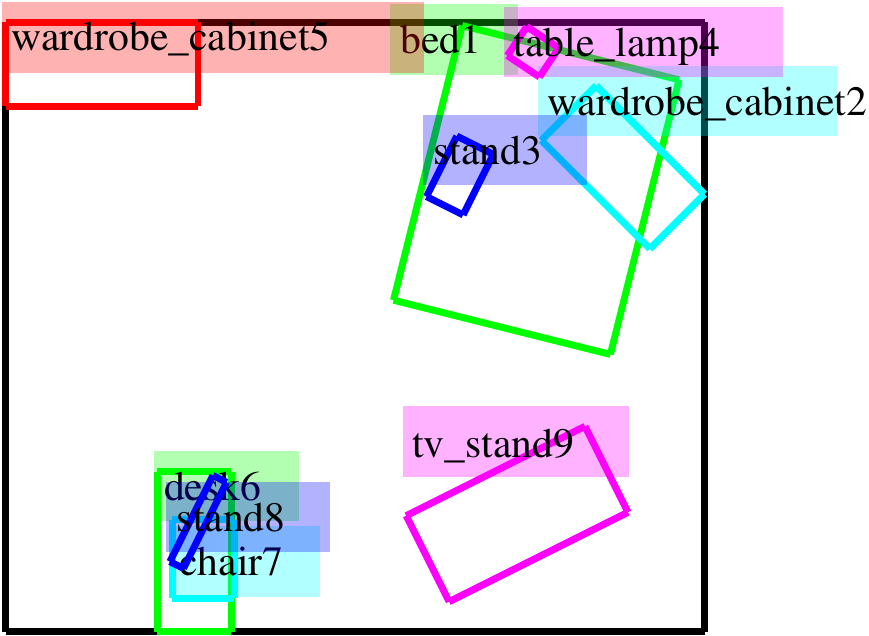}  & 
\includegraphics[width=0.18\textwidth, height=1.71cm ]{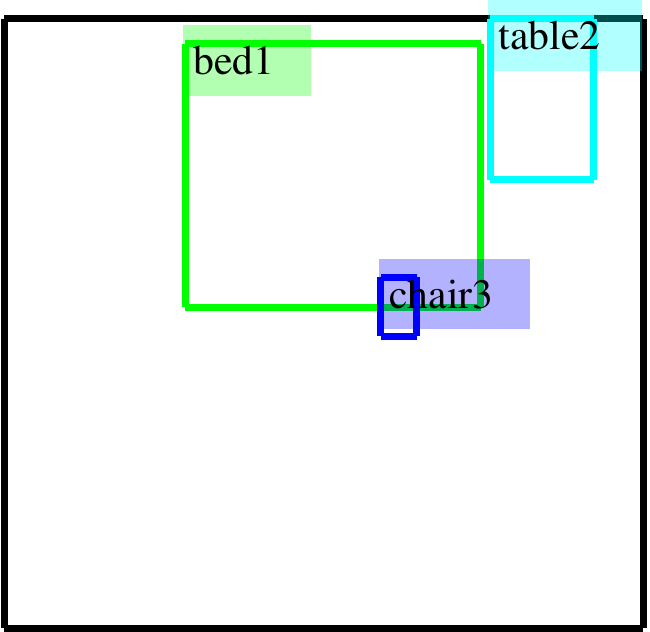}  & 
\includegraphics[width=0.18\textwidth, height=1.71cm ]{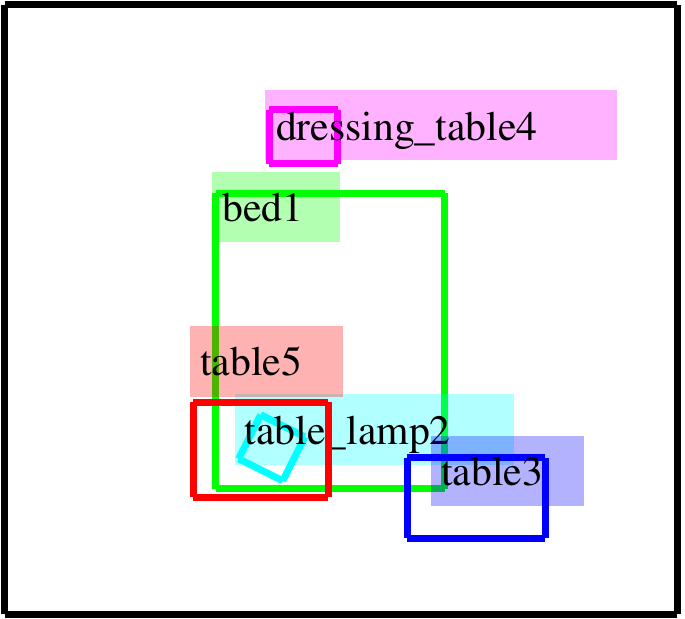}    \\ 

  \begin{picture}(1,25)\put(0, 5){\rotatebox{90}{($\alpha = 0.4$)}}\end{picture} & 
\includegraphics[width=0.18\textwidth, height=1.71cm ]{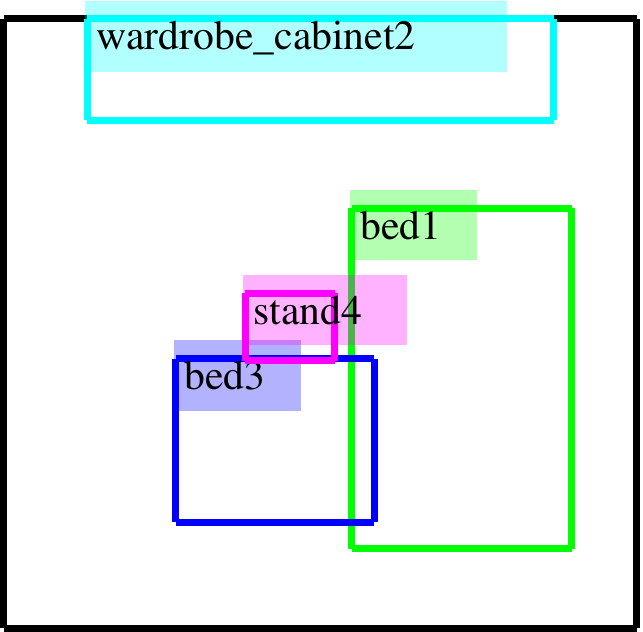} & 
  \includegraphics[width=0.18\textwidth, height=1.71cm , trim={0 0 42 0},clip]{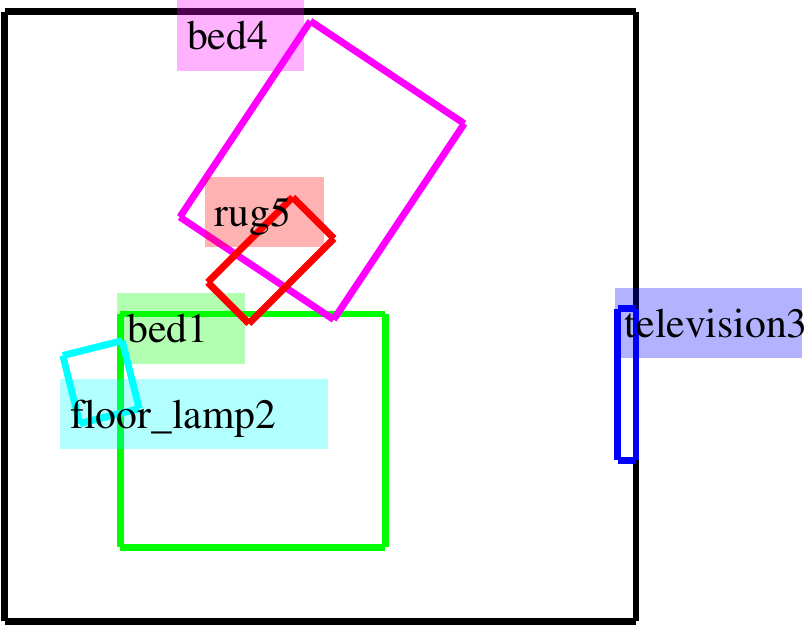}  & 
\includegraphics[width=0.18\textwidth, height=1.71cm, trim={0 0 52 0},clip ]{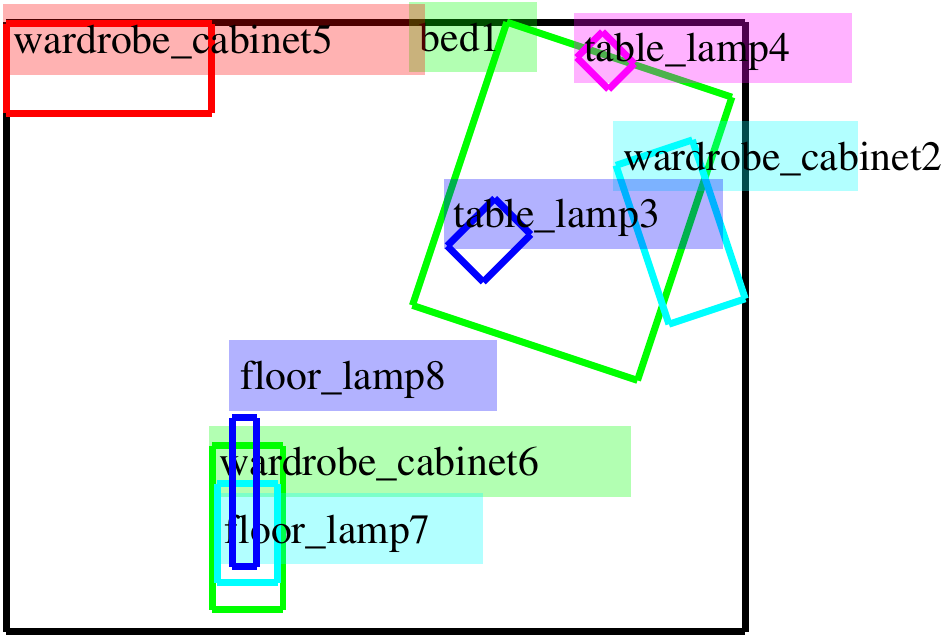}  & 
\includegraphics[width=0.18\textwidth, height=1.71cm, trim={0 0 48 0},clip ]{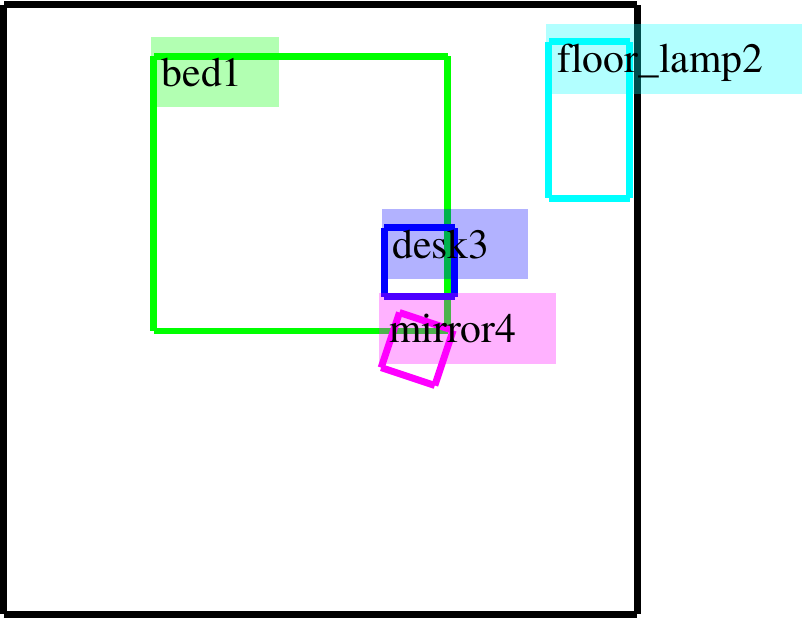}  & 
\includegraphics[width=0.18\textwidth, height=1.71cm ]{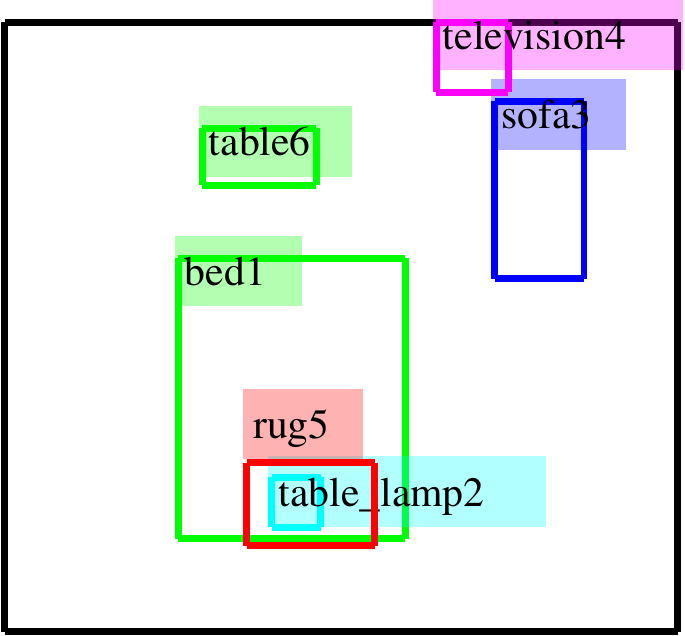}   \\ 

  \begin{picture}(1,25)\put(0, 5){\rotatebox{90}{($\alpha = 0.3$)}}\end{picture} & 
\includegraphics[width=0.18\textwidth, height=1.71cm ]{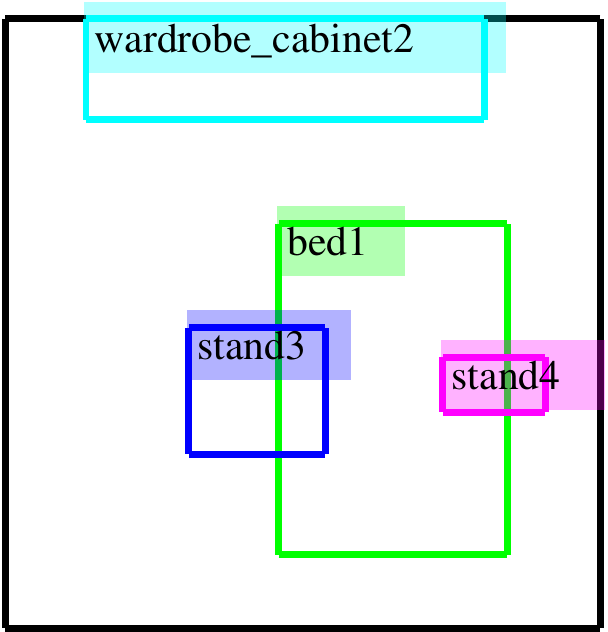} & 
  \includegraphics[width=0.18\textwidth, height=1.71cm ]{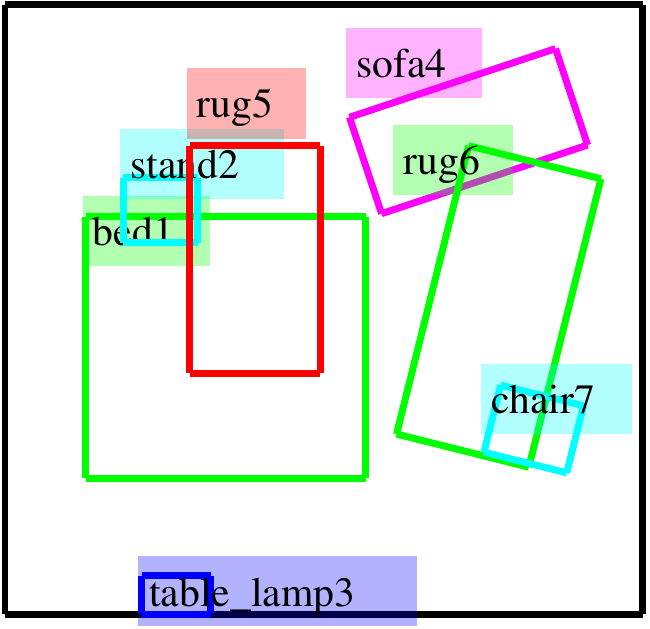}  & 
\includegraphics[width=0.18\textwidth, height=1.71cm , trim={0 0 36 0},clip]{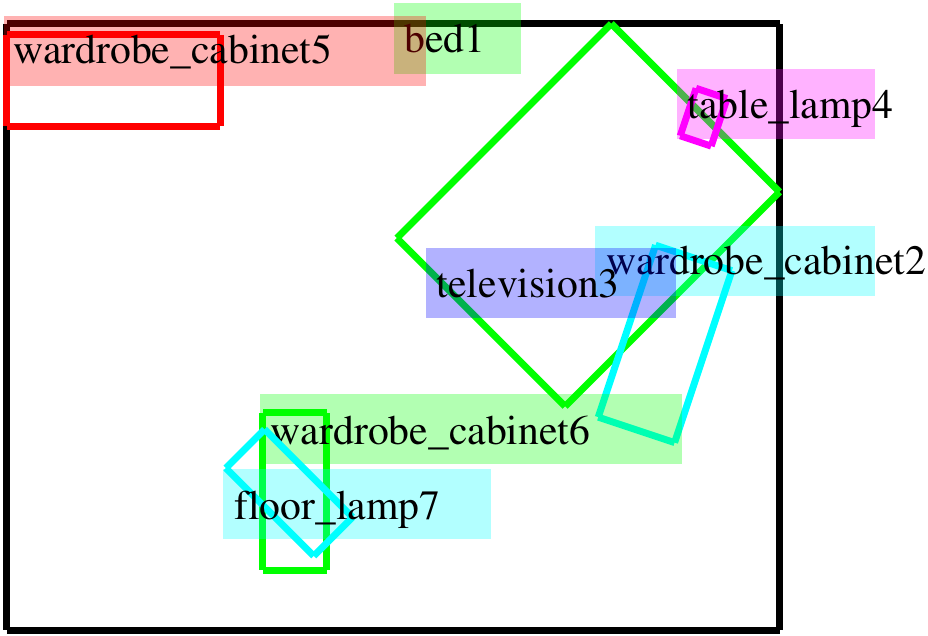}  & 
\includegraphics[width=0.18\textwidth, height=1.71cm, trim={0 0 47 0},clip  ]{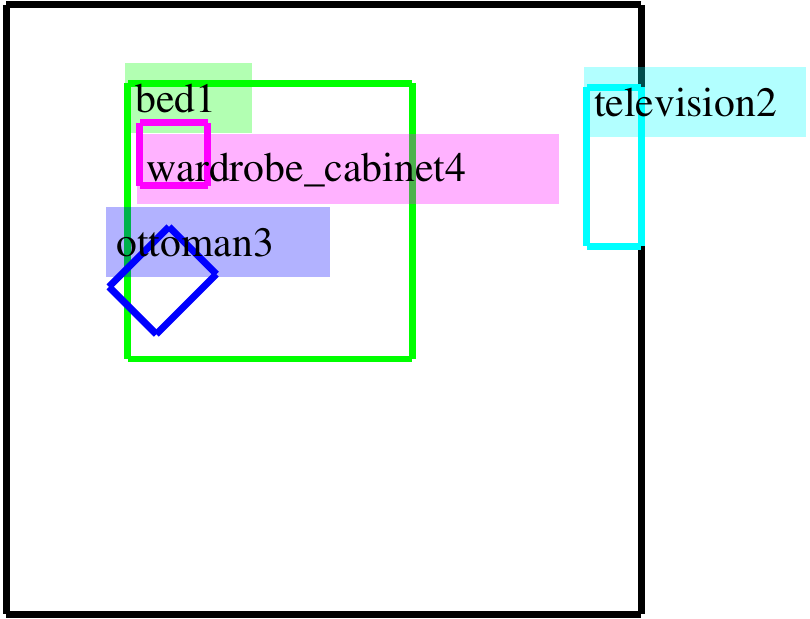}  & 
\includegraphics[width=0.18\textwidth, height=1.71cm ]{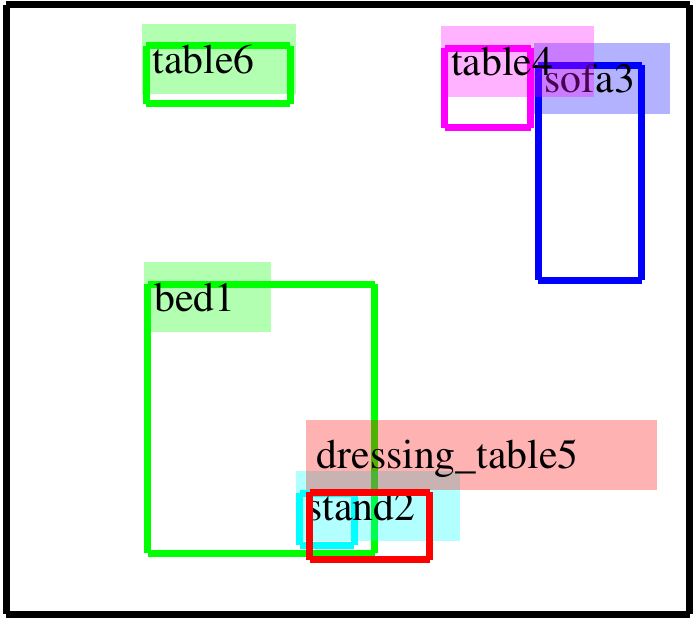}    \\ 

  \begin{picture}(1,25)\put(0, 5){\rotatebox{90}{($\alpha = 0.2$)}}\end{picture} & 
\includegraphics[width=0.18\textwidth, height=1.71cm ]{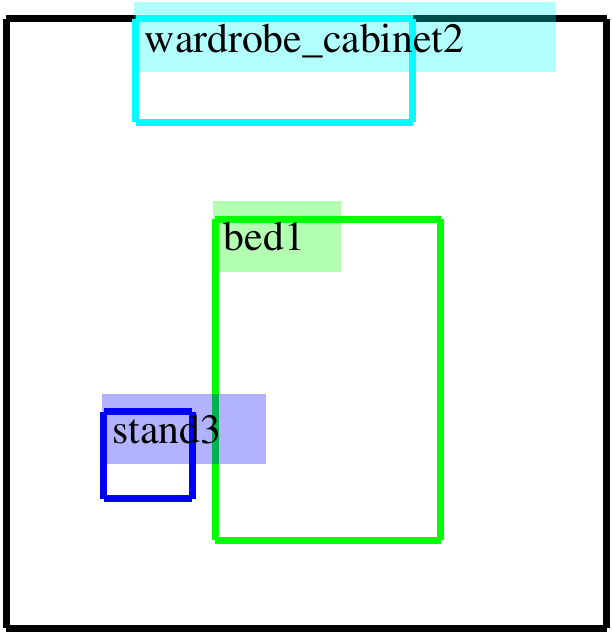} & 
  \includegraphics[width=0.18\textwidth, height=1.71cm, trim={0 0 24 0},clip ]{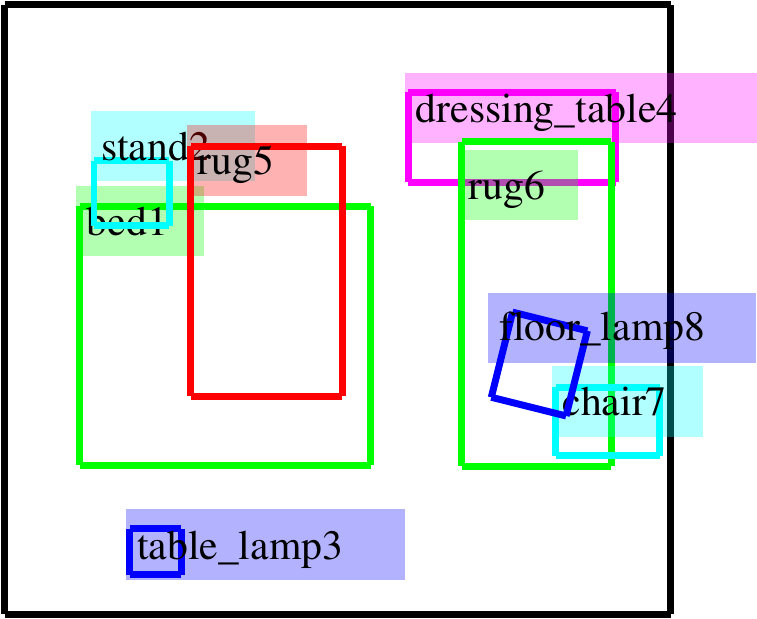}  & 
\includegraphics[width=0.18\textwidth, height=1.71cm, trim={0 0 30 0},clip ]{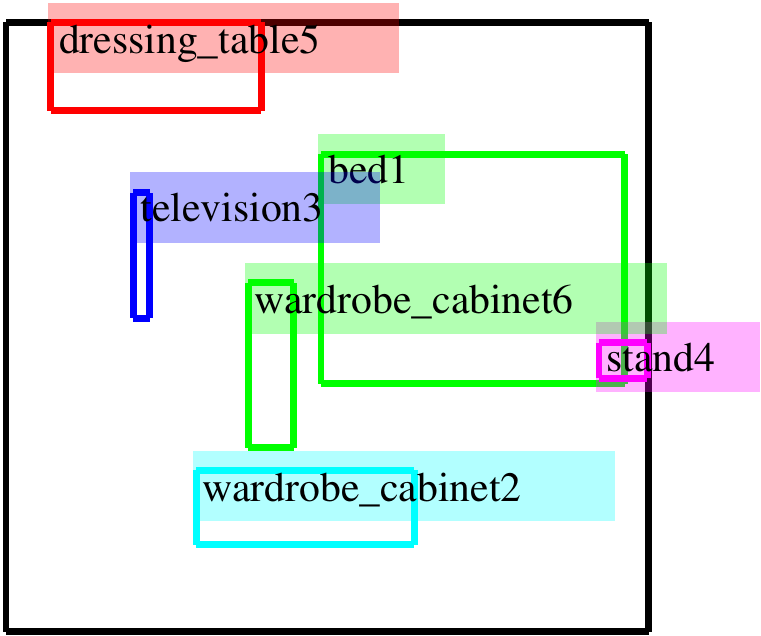}  & 
\includegraphics[width=0.18\textwidth, height=1.71cm, trim={0 0 46 0},clip  ]{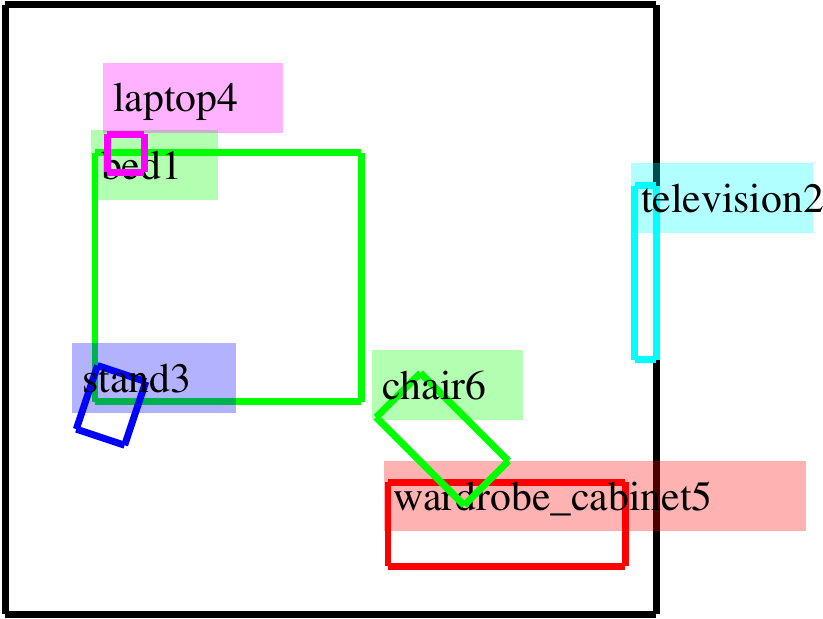}  & 
\includegraphics[width=0.18\textwidth, height=1.71cm, trim={0 0 15 0},clip ]{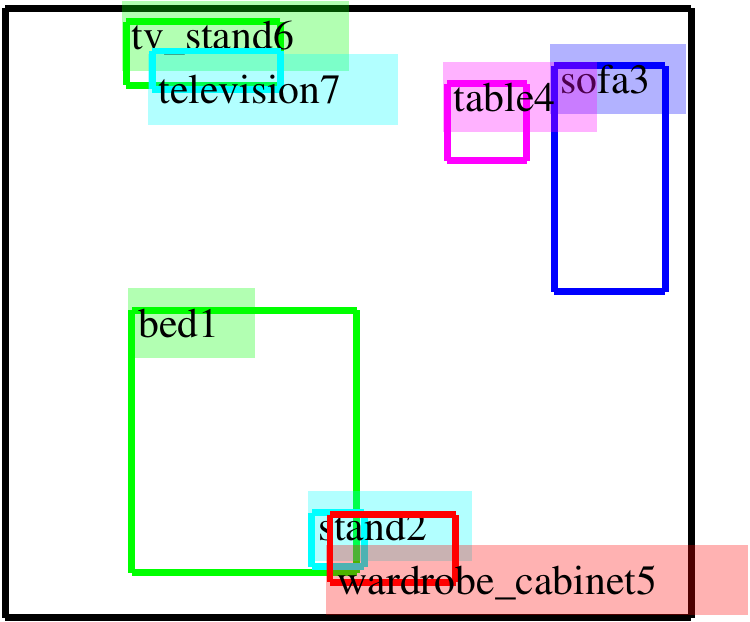}   \\ 

  \begin{picture}(1,25)\put(0, 5){\rotatebox{90}{($\alpha = 0.1$)}}\end{picture} & 
\includegraphics[width=0.18\textwidth, height=1.71cm ]{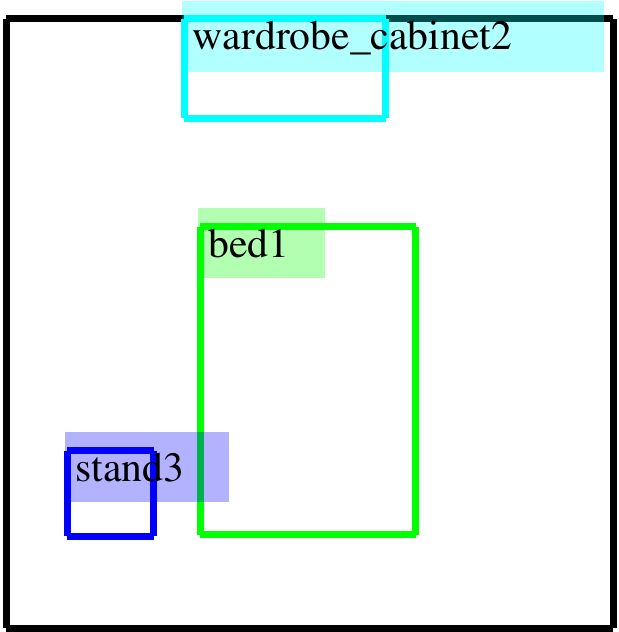} & 
  \includegraphics[width=0.18\textwidth, height=1.71cm , trim={0 0 30 0},clip]{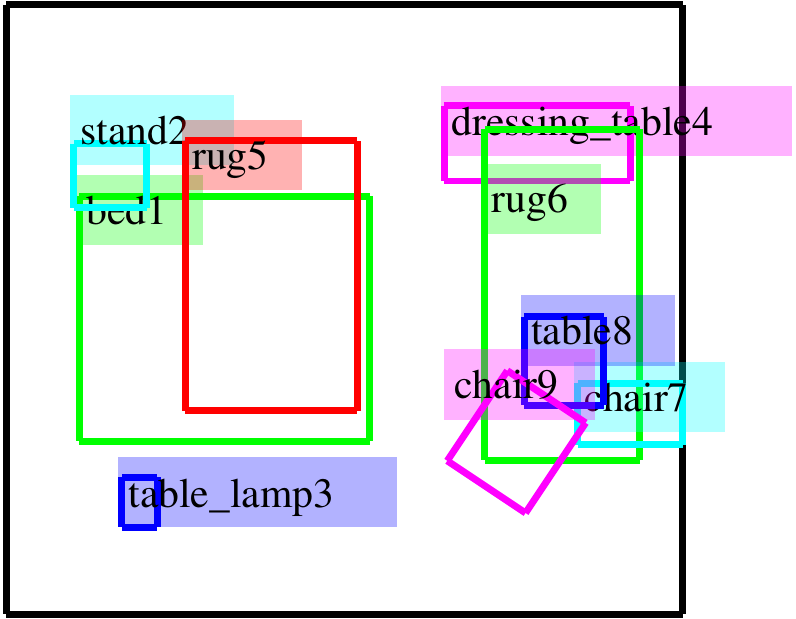}  & 
\includegraphics[width=0.18\textwidth, height=1.71cm , trim={0 0 24 0},clip]{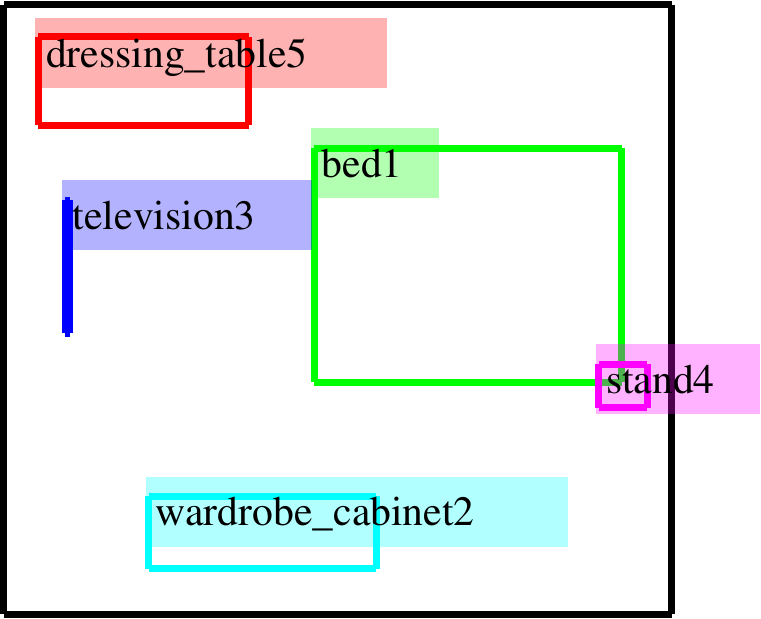}  & 
\includegraphics[width=0.18\textwidth, height=1.71cm, trim={0 0 44 0},clip  ]{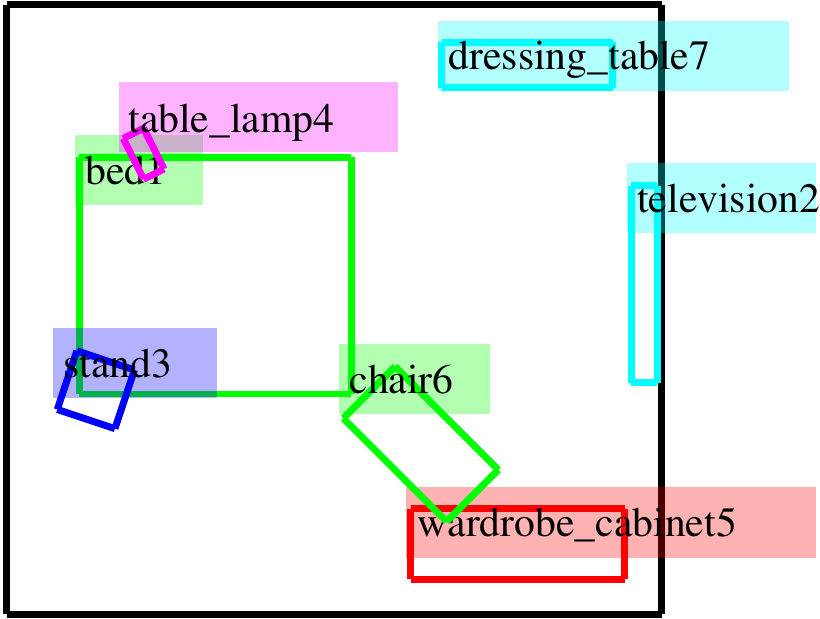}  & 
\includegraphics[width=0.18\textwidth, height=1.71cm, trim={0 0 30 0},clip ]{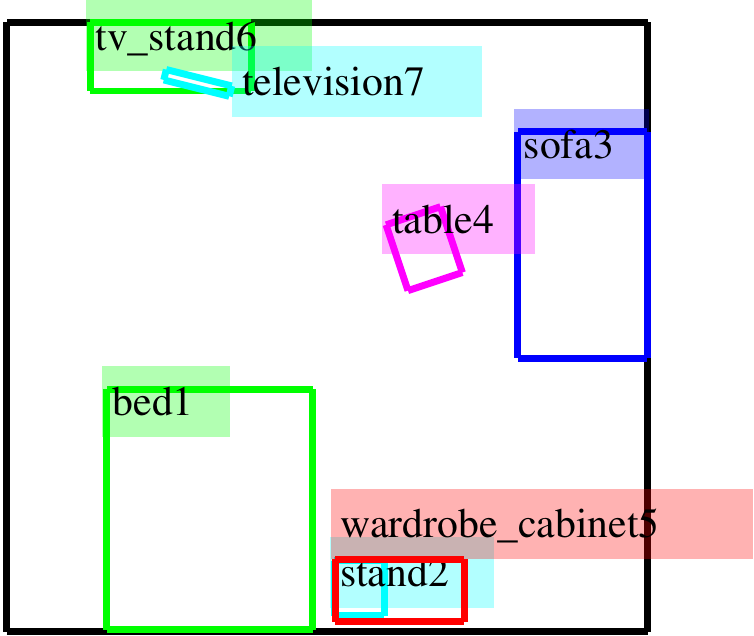}   \\ 
& Interpolation 1 & Interpolation 2 & Interpolation 3 & Interpolation 4 & Interpolation 5 % & Interpolation 6  
\end{tabular}
\caption[Latent code interpolation on SUNCG]{Latent code interpolation on SUNCG: Synthetic scenes decoded from linear interpolations $\alpha \bm{\mu_1} + (1 - \alpha)\bm{\mu_2}$ of the means $\bm{\mu_1}$ and $\bm{\mu_2}$ of the latent distributions of two separate scenes. The generated scenes are valid in terms of the co-occurrences of the object categories and their shapes and poses. The room-size and the camera view-point are fixed for better visualization. Best viewed electronically. }
\label{fig:qualititive} 
\end{figure} 

\section{Additional experiments on SUN RGBD}

\subsection{Scene layout from the RGB-D image---in detail}
The task is to predict the 3D scene layout given an RGB-D image. We (linearly) map deep features (extracted from images by a DNN~\cite{zhou2014learning}) to the latent space of the scene-grammar autoencoder. The decoder subsequently generates a 3D scene configuration with associated bounding boxes and object labels from the projected latent vector. Since during the deep feature extraction and the linear projection, the spatial information of the bounding boxes are lost, the predicted scene layout is then combined with a bounding box detection to produce the final output. \\
% \begin{table}\setlength{\tabcolsep}{6pt} % {0.52\textwidth}
%%\vspace{-15pt} 
%\caption{IoU for RGBD to room layout estimation}\label{wrap-tab:2}
%\centering \scriptsize 
%\begin{tabular}[ht]{@{\hspace{0.5em}}c@{\hspace{1.8em}}c@{\hspace{1.6em}}c@{\hspace{1.6em}}c@{\hspace{1.6em}}c@{\hspace{1.6em}}c}\toprule    
%Methods & {\bf SG-VAE}  & {\bf BL1} & {\bf BL2}~\cite{gomez2018automatic} & {\bf BL3}~\cite{kusner2017grammar}+\cite{yu2011make} & {\bf DSS~\cite{song2016deep}} \\\midrule
%IoU & $\bm{0.4387}$ & ${0.4315}$ & $0.4056$ & $0.4259$ &  $0.4070$ \\ \bottomrule 
%\end{tabular} 
%\end{table}

\noindent {\bf Training} Let $\cF_i$ be the (in our case 8192-dimensional) deep feature vector extracted from the image $I_i$, and let $\cN(\bm{\mu}_i, \bm{\Sigma_i})$ be the (e.g.\ 50-dimensional) latent representation obtained by encoding the corresponding parse tree. We align the feature vector $\cF_i$ with the latent distribution using a linear mapping $\psi(\cF_i) = A \cF_i$, where $A$ is a matrix to be learned from a training set $\mathcal T := \{I_i:= (\cF_i, \bm{\mu}_i, \bm{\Sigma_i})\}$. We minimize the cross-entropy between the predicted (deterministic) latent representation $\psi(\cF_i)$ and the target distribution $\cN(\bm{\mu}_i, \bm{\Sigma_i})$, therefore the optimal matrix $A$ is determined as  $\hat{A} = \arg \min_A \sum_{I_i \in \mathcal T} \big( A \cF_i - \bm{\mu}_i \big)^T \bm{\Sigma_i}^{-1} \big( A \cF_i - \bm{\mu}_i \big)$ $+ \lambda \| A \|_2^2$.   
The features $\cF_i$ of dimension $8192$ are then projected into the mean of the encoded vector $\bm{\mu}_i$ (typically dimension $50$).
Let $\phi: \cF_i \rightarrow \bm{\mu}_i$ be the mapping that project the feature vectors $\cF_i$ to the latent space $\bm{\mu}_i$. A neural network could be used to learn the mapping $\phi$, however, a simple linear projection is employed here, \ie $\psi(\cF_i) = A \cF_i$. %One could employ more sophisticated mapping for a better performance. % The choice of a simple linear projection is made due to limitation of the training data ($7794$ valid examples). 
The mapping $\phi$ is learned from the training examples $Tr = \{I_i:= (\cF_i, \bm{\mu}_i, \bm{\Sigma_i})\}$ as follows: 
\begin{align}
\hat{A} &= \arg \min_A \sum_{I_i \in \mathcal T} \Big( A \cF_i - \bm{\mu}_i \Big)^T \bm{\Sigma_i}^{-1} \Big( A \cF_i - \bm{\mu}_i \Big)  + \lambda \| A \|_2^2
\end{align} 
where we also added a regularization term with weight $\lambda$ (chosen as $\lambda = 100$). 
Differentiating the objective to zero, we get $\sum_{I_i \in \mathcal T} \bm{\Sigma_i}^{-1} \hat{A} (\cF_i \cF_i^T) + 2\lambda\hat{A} $ \\ = $ \sum_{I_i \in \mathcal T} \bm{\Sigma_i}^{-1}\bm{\mu}_i\cF_i^T$. Therefore, 
\begin{align}
\hat{A} & =  \Big(\sum_{I_i \in Tr}\cF_i ^T \bm{\Sigma_i}^{-1} \cF_i + 2\lambda I\Big)^{-1} \sum_{I_i \in Tr} \cF_i \bm{\Sigma_i}^{-1}\bm{\mu}_i 
\label{eq:mapping}
\end{align} 
Note that the covariance matrix $\bm{\Sigma_i}$ is chosen to be diagonal, and thus $\hat{A}$ can be solved efficiently. 
The above is a system of linear equations solved by vectorizing the matrix $\hat{A}$. \\ 

\noindent {\bf Testing} For test data image features~\cite{zhou2014learning} are extracted and then mapped to the latent space using the trained mapping $ \hat{\phi}(\cF_i) := \cF_i \hat{A}$. The scenes are then decoded from the latent vectors $\hat{\bm{\mu}}:=\cF_i \hat{A}$ using the decoder part of the SG-VAE. 
The bounding box detector of DSS~\cite{song2016deep} is employed and the scores of the detection are updated based on our  reconstruction as follows: the score (confidence of the prediction) of a detected bounding box is doubled if a similar bounding box (in terms of shape and pose) of the same category is reconstructed by our method. A 3D non-maximum suppression is applied to the modified scores to get the final scene layout. % The details can be found in the supplementary. 

 %We selected the average IoU for room layout estimation as the evaluation metric, and the results are presented in Table~\ref{wrap-tab:2}. % BL3 returns good numeric results as the solution is chosen from the best among $10$ samples. 
% The proposed method and other grammar-based baselines improve (in terms of average IoU) the scene layout estimation from the same by sophisticated methods such as deep sliding shapes~\cite{song2016deep}. Furthermore, the proposed method tackles the problem in a much simpler and faster way.  Thus, it can be employed to any 3D scene layout estimation method with very little overhead (\eg a few ms in addition to $5.6$s of~\cite{song2016deep}).
%Furthermore, our goal was to build a fast generative model for indoor scenes with an useful latent vector and the current experiment is to demonstrate the potential of this rather than benchmarking results of 3D scene layout estimation.
%Results on some test images where SG-VAE produces better IoUs are displayed in Figure~\ref{fig:rgbtoscene}. 
%Further applications of utilizing the learned latent space (such as multiple object detection and 3D pose estimation) are given in the supplementary material.

\subsection{Quality assessment of the autoencoder}
 The scenes and the bounding boxes of the test examples are first encoded to the latent representations $\cN(\bm{\mu}_i, \bm{\Sigma_i})$. The mean of the distributions $\bm{\mu}_i$ are then decoded to the scene with object bounding boxes and labels. The results are displayed below. Ideally, the decoder should produce a scene which is very similar to input test scene. IoU (computed over the occupied space) of the decoded scene and the original input scene is also shown in the main paper. 
 \noindent { Baseline methods for evaluation as follows:}
%The proposed method is evaluated against the following baselines: 
\begin{enumerate}[({\bf BL}1),leftmargin=10mm,topsep=2.25ex,partopsep=3.25ex]
\item \emph{Variant of SG-VAE}: In contrast to the proposed SG-VAE where attributes of each rule are directly concatenated with 1-hot encoding of the rule, in this variant separate attributes for each rule type are predicted by the decoder and rest are filled with zeros. \ie, the 1-hot encoding of the production rules is same as SG-VAE but the attributes are represented by a $|\cR|*\theta$ dimensional vector where $|\cR|$ is the number of production rules and $\theta$ is the size of the attributes. For example, the pose and shape attributes $\Theta^{j \rightarrow k} = ({\cP}_i^{j \rightarrow k}, \cS_i^{k})$, associated with a production rule (say $p$th rule) in which a non-terminal $X_j$ yields a terminal $X_k$, are placed in  $(p-1)*\theta + 1:p*\theta$ dimensions of the attribute vector and rest of the positions are kept as zeros.  Note that in case of SG-VAE the attributes are represented by a $\theta$ dimensional vector only. 
\item \emph{No Grammar VAE}~\cite{gomez2018automatic}: No grammar is considered in this baseline. The 1-hot encodings correspond to the object type is concatenated with the absolute pose of the objects (in contrast to rule-type and relative pose in SG-VAE) respectively. \ie each object is represented by $|\cV|$ dimensional 1-hot vector and a $\theta$ dimensional attribute vector. The objects are ordered in the same way as SG-VAE to avoid ambiguity in the representation. The same network as SG-VAE is incorporated except no grammar is employed (\ie no masking) while decoding a latent vector to a scene layout.  
\item \emph{Grammar VAE}~\cite{kusner2017grammar} + \emph{Make home}~\cite{yu2011make}: The Grammar VAE is incorporated with our extracted grammar to sample a set of coherent objects and~\cite{yu2011make} is used to arrange them. Sampled $10$ times and solution corresponding to best IoU w.r.t.\ groundtruth is employed. Here no pose and shape attributes are incorporated while training the autoencoder. The attributes are estimated by \emph{Make home}~\cite{yu2011make} and the best (in terms of IoU) is chosen comparing the ground-truth. 
\end{enumerate}
%The results are shown in Figure~\ref{fig:autoencoder}. Note that in all the examples shown, \emph{No Grammar VAE}, \ie {\bf BL2} can not reconstruct the input scene whereas the SG-VAE performs the best. 
The detailed quantitative numbers are presented in the main manuscript.  \\ \vspace{6cm}

\IncMargin{1.5em} \setlength{\textfloatsep}{0.1cm}  \setlength{\floatsep}{0.1cm} 
\begin{algorithm} \small 
\caption[Modified Inductive Causation (IC)~\cite{pearl1995theory}]{Structure learning: Inductive Causation (IC)~\cite{pearl1995theory}}\label{graph_algo}
\Indm 
    \KwIn {a dataset $\cD$ of natural scenes formed by a set of objects $\cV = \{X_1,\ldots, X_n\}$ }
    \KwOut{ a graph $\cG = \{\cV, \cE\}$ representing the causal relationships between the variables }
%\begin{algorithmic}[1]
\Indp 
 \text{ Initialize } $\cG := \{\cV;\;\cE = \cE_0\} $ \tcc*{Initialize with prior edges} 
\For {every pair of objects $(X_j\in \cV;\;X_{j^\prime}\in \cV)$ } {
\For {every conditioning variable $X_k \in \cV \setminus \{X_i, X_{j^\prime}\}$} { 
 hypothesis test $X_j \indep X_{j^\prime} \; |\; X_k$ in $\cD$  \tcc*{Using Conditional Algo~1} 
} %\tcc*{Only frequent objects were considered} 
\If {no independence was found} {
add an undirected edge $(X_j,X_j^\prime)$ in $\cE$, \ie, $\cE = \cE\cup (X_j, X_j^\prime)$.  
}
}
\For {every pair of objects $(X_j\in \cV;\;X_{j^\prime}\in \cV)$ with a common neighbor $X_k$} {
\If {$(X_j,X_j^\prime) \notin \cE$} {
\If {one of $(X_k,X_j)$ and $(X_k, X_j^\prime)$ is directed and the other is undirected {\normalfont \bf \large or} \\ both are undirected}{
turn the triplet into a common parent structure, \ie,  $X_j \reversecausal X_k \causal X_{j^\prime}$ 
}}
}
Propagate the arrow orientation for all undirected edges (modify the set $\cE$ accordingly) without introducing a directed cycle.
\tcc*{Following Dor and Tarsi~\cite{dor1992simple}}
\Return $\cG = \{\cV, \cE\}$
\end{algorithm}
\DecMargin{1.5em} \vspace{-1.5em}

%\input{file_list_images4.tex} 
%\input{file_list_images.tex} 
%\input{file_list_images5.tex} 
%\input{file_list_images6.tex} 
%\input{file_list_images3.tex} 

%\newpage{}  
\vspace{4em}
\section{Production rules extracted from SUN RGB-D}\label{sec:allrules}
Here we describe the production rules of the CFG extracted from SUN RGB-D dataset in detail. The total number of rules generated by the algorithm described in section~3 of the main draft is $399$, number of non-terminals is $49$ and number of terminal objects is $84$.  In the following we display the entire learned grammar. Note again that the non-terminal symbols are displayed in upper case, \emph{S} is the start symbol and \emph{None} is the empty object. The rules are separated by semicolons `\;;\;' symbol.  \\

\IncMargin{1.5em}
\begin{algorithm} \small 
\caption[A greedy algorithm for p-cover]{{\bf p-Cover} : A greedy algorithm for p-cover}\label{cover_algo}
\Indm
    \KwIn {a dataset $\cD$ of indoor scenes $\cI$ formed by a set of objects $\cV = \{X_1,\ldots, X_n\}$, the causal graph $\cG=(\cV, \cE)$ obtained by Algorithm~\ref{graph_algo} and a probability $p$ (chosen as $0.8$)}
    \KwOut{a set of rules $\cR$ that explains the occurrences of objects in the scenes $\cI$ } 
\Indp 
Choose the set of potential non-terminals as $\cV^\prime = \{X_i \in \cV: deg_{out}(X_i)/(deg_{in}(X_i)+\epsilon) > 1 \}$
 \tcc*{Proportion of the outward degree and inward degree; $\epsilon < 1$} 
 {Generate set of concepts and associated rules $\{\cR_j\}_{j\in \cV^\prime} $ by choosing adjacent objects} \tcc*{Some examples are displayed in Figure~\ref{fig:concepts}} 
 \text{Initialize } $\cR \gets \emptyset $, $\cV^\star = \emptyset$, and $C = \emptyset$ \tcc*{Initialize by empty set;} 
 \While {\text{the cover set $\cC$ covers the dataset with probability $p$}} { 
%  \text{Choose the next objects $\cV^\prime \subset \cV$ in the dataset; } \\
\For {\text{every non-terminal and associated set of rules $\cR_j$, $\forall j \in \cV^\prime \setminus \cV^\star$}} {
\text{Compute the gain $\cG_{gain}(\cR_j, \cR)$ as referred in Eq.~2 of the main draft} 
}
 compute next anchor node $X_{\bar{j}} = \argmax_{X_j \in \cV^\prime \setminus \cV^\star} \cG_{gain}(\cR_j, \cR)$ and $ \cV^\star = \cV^\star \cup {\bar{X_j}} $   \\
 $\cR = \cR \cup \cR_{\bar{j}} $ \\ 
 $ C = C \cup C_{\bar{j}} $ \tcc*{Update the rule set and the cover set;} 
%$\cR = \cR \cup [\bar{X_j}\text{\texttt{inc}} \rightarrow \text{\texttt{'None'}}]$
} 
$\cR = \cR \cup [S \rightarrow \text{\texttt{'None'}}]$ \\
\Return $\cR $
%\end{algorithmic}
\end{algorithm}
\DecMargin{1.5em}

{\small
\bibliographystyle{splncs04}
\bibliography{egbib}
}

%\clearpage 
%\title{Learning to generate new indoor scenes}
%\maketitle
%\appendix 
%\addcontentsline{toc}{chapter}{APPENDICES}

\end{document}